# Causal Machine Learning Methods for Estimating Personalised Treatment Effects - Insights on validity from two large trials

**Authors**

Hongruyu Chen[*], Helena Aebersold[*], Milo Alan Puhan[*], Miquel Serra-Burriel[*]

[*] Epidemiology, Biostatistics and Prevention Institute, University of Zurich, Zurich, Switzerland

**Correspondence**

Miquel Serra-Burriel,

Epidemiology, Biostatistics & Prevention Institute (EBPI),

University of Zurich,

Hirschengraben 84,

8001 Zurich, Switzerland,

miquel.serraburriel@uzh.ch

**Abstract**

Causal machine learning (ML) methods hold great promise for advancing precision medicine by estimating personalized treatment effects. However, their reliability remains largely unvalidated in empirical settings. In this study, we assessed the internal and external validity of 17 mainstream causal heterogeneity ML methods—including metalearners, tree-based methods, and deep learning methods—using data from two large randomized controlled trials: the International Stroke Trial (N=19,435) and the Chinese Acute Stroke Trial (N=21,106). Our findings reveal that none of the ML methods reliably validated their performance, neither internal or external, showing significant discrepancies between training and test data on the proposed evaluation metrics. The individualized treatment effects estimated from training data failed to generalize to the test data, even in the absence of distribution shifts. These results raise concerns about the current applicability of causal ML models in precision medicine, and highlight the need for more robust validation techniques to ensure generalizability.

# 1. Introduction

Personalized medicine requires identifying which patients will benefit from a given treatment and which will not[1,2]. Traditionally, evidence-based treatments are built upon Randomized Controlled Trials (RCTs). RCTs are widely considered as the highest standard of evidence due to having the highest degree of internal validity. The main results of RCTs investigating therapeutic options are average treatment effects (ATEs), the estimated average benefit or harm in the trial population based on the primary outcome measure.

Although ATEs provide a straightforward, easy-to-interpret summary of the overall effect, reliance on ATEs for decision-making introduces an inherent limitation[3]. ATEs do not account for individual patient differences, limiting their applicability to subsets of the trial population or when extrapolating results to a broader population of interest[4,5].

To address this limitation in personalized treatment research, numerous causal Machine Learning (ML) methods have emerged in recent years. They focus on individualized or subgroup treatment effects, leveraging patient characteristics to provide personalized treatment recommendations[6–9]. There is a substantial body of literature developing causal ML methods for learning heterogeneous treatment effects, comparing these methods using synthetic, RCT, or observational data, and applying them in practical contexts[10–16].

Validation plays an essential role in machine learning[17,18]. Models that lack proper validation are unreliable and lead to biased predictions. Unfortunately, the importance of validating these causal ML models is often overlooked in the current literature. Most studies make use of synthetic data to test the model performance[10,19], or when using RCT data, stopped at providing seemly correct density of individualized treatment effect estimates, treating this as sufficient evidence of the usefulness of given causal ML methods[12]. There also exists research testing the estimations between different models on an observational study[14]. However, both synthetic data and observational studies have fundamental drawbacks, synthetic data makes the true data generating process observable by nature, while observational data presents inherent unobservable confounding.

This study seeks to address this literature gap by investigating to what extent estimates of individualized and subgroup treatment effects based on causal ML methods withstand systematic validation. Given that RCTs offer distinct advantages over observational studies, particularly the lack of confounding[20], and that ML methods rely on large datasets, we utilized two large, parallel-group RCTs designed for parallel analysis. First, we estimated individualized treatment effects (ITEs) using a wide range of causal ML models that are used in medical studies. Second, we validated these estimates through internal training-test data splits and external cross-validation, employing established and novel metrics including c-for-benefit[21], mbcb[22], calibration-based pseudo R-squared[23], outcome-ITE plots, ATE-ITE subgroup analysis, and benefit-harm ATE density plots. We assess model performance in internal validation scenarios to test model performance on data with similar distributions, and also external validation under different experimental settings, known as distribution shift scenarios. Furthermore, we conducted simulation studies, with and without unobservables to explore the fundamental limitations of causal ML methods.

## 2. Methods

### 2.1 Data

The International Stroke Trial (IST)[24] and the Chinese Acute Stroke Trial (CAST)[25] were two large, parallel-group randomized controlled trials that investigated the effects of early aspirin therapy in patients with suspected acute ischemic stroke. IST enrolled 19,435 patients from 16 countries across 467 hospitals. In this trial, patients were randomly allocated to receive either 300 mg of aspirin daily for 14 days or no aspirin in a factorial design. Similarly, CAST enrolled 21,106 patients from 413 hospitals in China, where participants were randomly assigned to receive either 160 mg of aspirin daily for four weeks or a placebo. The primary outcomes measured in IST were death within 14 days and death or dependency at 6 months. In contrast, CAST focused on death within four weeks and death or dependency at discharge. Both trials meticulously collected baseline patient characteristics at the time of randomization, including variables such as time from stroke onset, age, sex, level of consciousness, systolic blood pressure, etc. This comprehensive data collection was intended to facilitate an integrated analysis of the two trials, and the characteristics of patients in the two trials were summarized[26]. The individual patient data from IST has been made publicly available by the International Stroke Trial Collaborative Group[27]. And the patient entry and discharge forms from CAST were provided by the University of Oxford upon request. In this study, we focused on the intention-to-treat contrast, including all patients with consistent data irrespective of their final diagnosis of ischemic stroke. Considering a relatively small fraction of missing data and large trial sizes, we only used the complete cases for analysis. All the data cleaning and causal ML analyses in this paper were conducted using R statistical software version 4.3.3[28].

### 2.2 Heterogenous treatment effect modelling

We made use of Rubin's potential outcome framework[29]. In the context of binary treatment and binary observable outcome, let $T$ denote treatment assignment (where $T = 0$ for control and $T = 1$ for treatment) and $Y$ denote the observed outcome ($Y = 0$ or $Y = 1$). The potential outcomes framework defines $Y^{(0)}$ as the potential observable outcome for a patient without treatment and $Y^{(1)}$ as the potential observable outcome for a patient with treatment. For a unique individual $i$, the individual treatment effect is defined as the difference between two potential outcomes of that individual: $Y_i^{(1)} - Y_i^{(0)}$. Average treatment effect (ATE) is then given by the average of individual treatment effect in a subset sample S: $ATE_S = \frac{1}{n}\sum_{i \in S}(Y_i^{(1)} - Y_i^{(0)})$. In the literature, it is common to transform the ATE into metrics[30] such as the odds ratio: $ATE_S^{OR} = \frac{P(Y^{(1)}=1)}{P(Y^{(1)}=0)} \Big/ \frac{P(Y^{(0)}=1)}{P(Y^{(0)}=0)}$, or risk ratio: $ATE_S^{RR} = \frac{P(Y^{(1)}=1)}{P(Y^{(0)}=1)}$, which provide additional interpretability and flexibility.

The term "potential outcome" emphasizes that we can only observe one outcome for any individual patient, while the other remains a counterfactual outcome, rendering the individual treatment effect non-identifiable. Acknowledging this fact, treatment effects can still be described as "individual" granularity. Let $X$ denote patient baseline characterization covariates. The individualized treatment effect (ITE) of patient $i$ with baseline covariates $x_i$ can be defined as:

$$\text{ITE}_i(x_i) = P\left(Y_i^{(1)} = 1 \big| X = x_i\right) - P\left(Y_i^{(0)} = 1 \big| X = x_i\right). \quad (1)$$

Under identifiability assumptions on consistency, exchangeability, positivity, and no interference[10], we can rewrite the definition of individualized treatment effect in terms of observational data:

$$ITE_i(x_i) = P(Y_i = 1|T = 1, X = x_i) - P(Y_i = 1|T = 0, X = x_i). \quad (2)$$

With treatment effects defined independently of counterfactual potential outcomes, causal ML models can be effectively applied to real data. More specifically, we applied causal ML methods on trial data to learn $P(Y_i = 1|T = 1, X = x_i)$ and $P(Y_i = 1|T = 0, X = x_i)$.

Current causal ML methods for learning ITE can be categorized into three groups[9,16]: metalearners, tree-based methods and deep learning methods. In this study, we evaluated 17 causal machine learning methods, spanning all three categories, that have been used in medical research for identifying heterogeneous treatment effects.

*2.2.1 Metalearners*

Metalearners are flexible, model-agnostic algorithms designed to estimate ITE by decomposing the estimation process into multiple prediction tasks, each solvable by machine learning models[9,15]. Notable examples include the T-learner, S-learner, X-learner and DR-learner. T-learner trains two prediction models separately for the treatment group ($P(Y_i = 1|T = 1, X = x_i)$) and control group ($P(Y_i = 1|T = 0, X = x_i)$). It allows for flexibility in selecting different models for the treated and control groups. However, to facilitate efficient model comparison, researchers typically constrain both arms to use the same type of model. S-learners incorporate the treatment assignment variable into the covariates list and train a single model. To account for potential variations in treatment effects among individuals, one way is to introduce interaction terms between treatment and covariates for the linear model or directly apply nonlinear models. Additionally, spline functions and penalization techniques are applied to address potential model misspecifications. X-learner first models outcomes for each group, imputes counterfactual outcomes to estimate treatment effects, and then refines these estimates through weighted averaging[15]. DR-learner combines pseudo-outcome regression and propensity score models using a double robust approach to correct for selection bias between treatment groups[11,31]. We used T-learners based on logistic regression[10], penalized logistic regression[32], random forest[33], support vector machine[34], and XGBoost[35]; S-learners based on logistic regression, penalized logistic regression, XGBoost and BART[36,37]; X-learners based on random forest and BART; DR-learners based on random forest.

*2.2.2 Tree-based methods*

Tree-based methods, such as causal tree[38], causal forest[39], model-based recursive partitioning[40,41], and Bayesian causal forest[42], are designed to partition the covariate space in such a way that the resulting subgroups exhibit heterogeneous treatment effects. Causal forest extends random forest and integrates an honesty mechanism in the training process to overcome potential over-fitting issues. Model-based recursive partitioning automatically identifies patient subgroups with differential treatment effects by segmenting a model of the overall treatment effect, and linking subgroups to predictive factors using a decision tree approach. Bayesian causal forest incorporates an estimate of the propensity function into the response model and separately regularizes treatment effect heterogeneity. It addresses biases from confounding and improves control over effect estimation. We used causal forest, model-based recursive partitioning, and Bayesian causal forest in this study.

*2.2.3 Deep learning methods*

Deep learning methods model the relationship between covariates and outcomes using neural networks. Conditional Variational Autoencoders (CVAE) learns a latent variable representation of treatment and covariates to estimate individualized treatment effects[43]. Generative Adversarial Nets for inference of Individualized Treatment Effects (GANITE)[44] uses adversarial learning to capture the uncertainty in counterfactual outcomes. We used CVAE and GANITE as two representative deep learning methods.

## 2.3 Evaluation metrics

Traditionally, ITE density plots and outcome prediction accuracy have been used to assess model validity[10,12,14]. Additionally, we utilized both visual-based and quantitative metrics to directly assess model reliability and the accuracy of ITE estimates. Visual metrics include outcome-ITE plots, subgroup ATE-ITE plots, and benefit-harm density plots, while quantitative validation metrics include c-for-benefit[21], mbcb[22], and calibration-based pseudo R-squared[23].

*2.3.1 Visual-based metrics*

We used outcome-ITE plots (see Figure 2a as an example) to examine the relationship between estimated ITE values and observed outcomes under different treatment conditions. Logistic regression lines with 95% confidence intervals were used to illustrate the predicted probability of outcomes as a function of ITE, stratified by treatment groups. Additionally, we conducted subgroup treatment effects trend analyses and designed benefit-harm effects density plots (see Figure 2a and 2b as an example), where we plotted ATEs across subgroups delineated by the estimated ITEs. Similar ideas using strata defined by clinical cut-offs for predicted treatment benefits have been used in previous research[10,14,23,45]. Ideally, the trend in ATEs should mirror that of ITEs, indicating consistency in treatment effect estimation. Besides, calibration curves and receiver operating characteristic (ROC) curves were used to evaluate the accuracy of outcome predictions.

*2.3.2 Quantitative metrics*

To complement these visual tools, we incorporated quantitative metrics to assess discrimination, calibration, and overall model performance. C-for-benefit and mbcb that measure model discrimination were used for treatment effects. c-for-benefit evaluates whether the model accurately identifies who is likely to benefit more from a treatment[21]. mbcb extends c-for-benefit by using probabilistic modelling instead of matching[22]. Calibration-based pseudo R-squared provided a statistical assessment of how well the model captures the heterogeneous treatment effects comparing to a constant model based on ITE partitions[23]. For all these metrics, a quantitative score closer to 1 indicates a better fit.

## 2.4 Validation strategy

Similar to the standard sample splitting validation approach[46], our validation framework made use of standard internal-external validation parts to test the robustness and generalizability of the models (see Figure 1). Internal validation uses a train-test split within the same dataset, allowing us to evaluate model performance on unseen data from the same source. External validation applies the models to a new dataset, providing an additional layer and evaluating the models' applicability to distribution shift scenarios.

*2.4.1 Internal validation*

We first conducted internal validation within IST dataset and CAST dataset separately (Figure 1, Panel a). Patients were randomly allocated to either training or test data using a 2:1 ratio. The training data was utilized for fine-tuning the causal ML models, and we subsequently evaluated the trained models on the test data. Besides random allocation, we performed geographical split within IST dataset. Furthermore, we merged the data from the two trials into a unified dataset, which was then split into training and test data for analysis akin to the previous approach (see Figure 1, Panel b). To eliminate the impact of a single random split, we further validated the model by aggregating results across 100 or 1000 random training-test splits.

*2.4.2 External validation*

We conducted external validation using both the IST and CAST datasets, leveraging the similarities and differences in their populations. More specifically, we trained the models on the IST dataset and validated them on the CAST dataset, followed by the reverse process to ensure a comprehensive evaluation (see Figure 1, Panel c).

## 2.5 Outcomes and covariates

For internal validation within the two trials, we considered dependency or death at 6 months for the IST dataset and dependency or death at 4 weeks for the CAST dataset as the primary outcome variables. To train the causal ML models, we used 16 baseline covariates common to both datasets: age, sex, delay in hours from onset to randomization, conscious state, CT scan results, presence of atrial fibrillation, systolic blood pressure, and neurological deficits (including 8 types of deficits). For the IST dataset, we included three additional variables: geographical information, aspirin history within three days prior to randomization, and stroke type. Internal validation on the combined dataset and external validation requires aligning the outcome and covariates between the two trial datasets. Poststroke period is typically categorized into hyperacute (0-24 hours), acute (1-7 days), subacute, and chronic stages[47]. Despite the lack of consensus in the literature regarding the distinction between subacute and chronic periods, durations of 14 days and 4 weeks are commonly grouped together. Therefore, we selected death within 14 days as the outcome for IST and death within 4 weeks as the outcome for CAST. For baseline covariates, we utilized the interaction set comprising the 16 variables identified in the internal validation of the IST and CAST datasets.

## 2.6 Simulation study

To further investigate the strengths and limitations of causal ML models in estimating heterogeneous treatment effects, we simulated data based on different mechanisms. More specifically, we assumed binary outcome and treatment, where the log-odds of the outcome depended on covariates and treatment, incorporating both a global treatment effect and covariate-interactive treatment effects. A sample size of 20,000 was chosen to reflect the scale of the stroke trials. We explored scenarios involving continuous covariates alone, a mix of continuous and discrete covariates, as well as both high- and low-dimensional covariate settings. Besides, we evaluated the causal ML methods that were tested above. The internal validation process was repeated for both arm-separated and global estimation methods using the simulated data. Additionally, to assess the impact of variable selection, we validated the models using only variables with near-zero treatment effects (i.e., non-significant variables).

## 3. Results

The IST dataset comprised 19,435 patients with 9,720 treated with aspirin and 9,715 untreated. 5.97% of individuals had incomplete data for the 6-month death or dependency outcome and 5.19% of individuals were incomplete for the 14-day death outcome. Following data cleaning and resolution of dataset inconsistencies, the CAST trial included 20,614 patients with 10,313 treated with aspirin and 10,301 untreated. Incomplete data rates were 3.57% for death or dependency outcome and 3.29% for death outcome. Given the RCT design, both treated and control groups exhibited balanced distributions across training and test data.

### 3.1 ITE estimation

The causal ML estimation in both the IST and CAST datasets revealed heterogeneous treatment effects among patients (see Appendix, Figure S2.1-2.3, Figure S3.1-3.3). The effect of aspirin treatment varied among patients, with some benefitting from the treatment (i.e., ITE < 0) and others not benefiting or even being harmed (i.e., ITE ≥ 0). For most models, we observed that this heterogeneity of ITE was consistent across training and internal test datasets, as well as between treatment and control groups. This is also supported by discrimination and calibration plots of predicted and observed outcomes (see Appendix, Figure S2.6-2.9, Figure S3.6-3.9).

### 3.2 Internal validation

Internal validation uncovered significant disparities in model performance. Figure 2 presents the results of T-learner logistic regression model for 6-month death or dependency outcome in IST dataset. Figure 2a indicates that, in the training data, individuals predicted to experience harmful treatment effects have a higher risk of death in the aspirin group compared to the control group. In contrast, for those predicted to benefit from the treatment, the risk is lower in the aspirin group than in the control group. However, this observed pattern does not hold in the internal test data. This is also evident from the subgroup ITE analysis as shown in Figure 2b. ATE in risk ratio increases at higher ITEs in the training data but not in the test data. Repeated experiments shown in Figure 2c confirm these discrepancies across multiple splits when stratified by benefit and harm groups. Penalized logistic regression lowered the discriminatory accuracy on the training data between treatment and control groups but did not lead to consistent results on the test data. Alternatively, using more complex models such as random forest, XGBoost, and causal forest drew a more pronounced separation between the treatment and control group predictions in the training data, but also failed to replicate in the test data. Table 1 summarizes the quantitative validation metrics for the training and test data across the 17 causal ML models evaluated on the IST dataset. None of the models consistently achieved metrics close to 1 across training and test data in both datasets. Additionally, geographical validation within the IST dataset, internal validation using the CAST dataset, and sensitivity analyses highlighted similar discrepancies in model performance. Detailed visual and quantitative validation results are provided in Appendix Figure S1–S3.10 and Table S1.

In the internal validation on the merged dataset, we observed generally lower risks in the outcomes (see Appendix, Figure S4.1-4.10). Besides, the estimated individualized treatment effects had narrower scales compared to long-term outcomes and inclusion of dependency data. Similarly, discrepancies in model behaviour between training and test groups persisted across all causal ML models (see also Appendix Table S2).

### 3.3 External validation

Cross-dataset external validation reiterated these findings (see Appendix, Figure S5.1-6.9, Table S3, S4). Regardless of which trial dataset served as the training or test set, we consistently observed the same disparity in model behaviour between the training and test groups across all the causal ML models we applied.

### 3.4 Simulation study

Simulation results (see Appendix, Figure S7.1-7.3) indicated that, under the correct logistic model specification, logistic regression methods—including T-learner logistic regression, T-learner penalized logistic regression, and S-learner penalized logistic regression—as well as T-learner support vector machine, accurately modelled ITEs. In contrast, causal forest and T-learner random forest performed poorly. All of our evaluation plots revealed discrepancies between training and test data for poorly performing methods but not for well-performing ones. These findings were consistent across both low- and high-dimensional settings and regardless of whether the models included a mix of discrete and continuous variables or only continuous variables. However, when models were constructed using only non-significant variables, causal ML models failed to accurately capture ITEs.

### 4. Discussion

Our study has found that mainstream causal ML methods for heterogeneous treatment effects fail to replicate at the internal validation stage in a non-simulated scenario. Through comprehensive experiments on two large randomized controlled trials, we have demonstrated that widely used causal ML models fail to generalize well, both internally and externally.

Causal ML methods are valued for their ability to predict counterfactual treatment outcomes and further estimate treatment effects at the individual level. Currently, ITE density plots are often used to visualize the distribution of ITEs and assess the model's ability to capture heterogeneity in treatment effects across patients[10,12,14]. Outcome prediction accuracy is also used to justify the ITE estimator[9,12]. From our ITE estimation results, all tested causal ML methods managed to capture well-aligned heterogeneous treatment effects, apparently achieving the goal of individual-specific estimation (see Appendix, Figure S2.1-2.3, S2.6-2.9, Figure S3.1-3.3, S3.6-3.9). And our experiments on IST and CAST datasets showed that even simple methods, such as T-learner logistic regression, precisely predicted outcomes in binary classification. These findings were consistent with previous research results[12,14].

However, it is crucial to emphasize that while predictive performance metrics are considered, the primary goal remains to validate ITE estimates rather than precise outcome prediction[48]. It is inappropriate and insufficient to overemphasize the importance of the outcome prediction accuracy. And density analysis presents only estimated quantities, lacks validation through observed values, and can be misleading in the presence of outliers. Therefore, evaluation metrics—including visual outcome-ITE plots, subgroup ATE-ITE plots, benefit-harm density plots and quantitative measures such as c-for-benefit, mbcb and calibration-based pseudo R-squared—are crucial for rigorously assessing estimates of individualized and subgroup treatment effects derived from causal ML methods. We assessed ITEs on both absolute and relative scales, as recommended by the PATH statement, which can accommodate diverse research purposes[16,49]. There are other evaluation metrics proposed in the literature, such as the precision in estimation of heterogeneous effect (PEHE)[36]. It is used to evaluate the accuracy of ITE estimates by calculating the mean squared error between the predicted and true ITE. However, in real-world data, true ITE values are not directly observable, making PEHE impractical for direct assessment in our study. Some recent work has explored the use of influence functions to efficiently estimate the loss function of causal inference models without requiring access to true causal effects[50]. While these methods offer promising alternatives, we did not include them in our study due to computational complexity and left them for future research. Additionally, our validation strategy, including both internal and external validation, tests the robustness, generalizability, and reliability of these estimates.

Internal validation assesses ITEs within the same population. A significant challenge in real-world RCTs is the unknown underlying data generating mechanism, making the true individualized treatment effect unrecoverable. While ITEs cannot be directly validated, it is expected of causal ML models to perform consistently across training and test data under random splits. Our internal validation results indicated that all tested causal ML models exhibit significant discrepancies between training and test data. Compared to regression-based methods, tree-based methods displayed nearly perfect discriminative prediction in the training data but failed completely on the test data. Penalization methods reduced the deviation between training and test data but at a discriminatory power cost between treatment and control groups. These findings indicate that current causal ML models struggle to reliably estimate ITEs within the same populations they were trained on, suggesting issues with overfitting or inadequacies in the models' design.

External validation is crucial because an ideal causal ML model should remain robust and resist distribution shifts. Patients enrolled in the IST and CAST trials exhibited distinct demographic characteristics, encompassing variations in geographical backgrounds, age distributions, and stroke severity profiles[24,25]. Additionally, clinical subtypes of stroke prevalence differed between Asian and Western populations. Our external validation experiments showed that none of the models generalize well to the test data, regardless of which trial dataset is used for training or testing.

Simulation studies demonstrated the effectiveness of our proposed visual evaluation metrics—outcome-ITE plots, subgroup ATE-ITE plots, and benefit-harm density plots—in distinguishing between models that accurately estimate ITEs and those that do not. However, it is necessary to be cautious, as good performance on these metrics does not always guarantee accurate ITE estimates, especially under different data distributions or when important covariates are missing. Our experiments showed that causal ML models failed to capture ITEs accurately when the true model was unknown or when non-significant variables were used. This highlights the need for careful model specification and consideration of the data context.

There are still several limitations in our study. First, the baseline patient characteristics used to train the models were limited by the stroke trials designs. Including currently unavailable information that moderates the treatment effects could provide a more comprehensive patient profile in which the set of ML models tested performed better. Second, the scope of our models and validation metrics cannot cover the vast set of potentially available ones. Models that consider time-varying effects are also worth investigating. However, we believe to have included the most important and widely used causal ML methods in medical research, as shown in a recent scoping review on ML methods for ITE in randomized controlled trials[16]. Finally, our estimates have focused on the intention-to-treat causal contrast, while other contrasts might be of more clinical interest. We chose the contrast with the highest degree of internal validity.

Overall, our research exemplifies that blindly applying current heterogeneous causal ML methods to individual patients does not produce reliable results and may lead to undesired outcomes. While validation is essential for randomized controlled trials, it is even more critical for observational studies, where confounding, data quality and identifiability problems further complicate the analysis. Future work is necessary to advance the credibility and evaluation guidelines for counterfactual models that are suitable for broad studies. Considering the findings of this study, the gap between standard ATE treatment approaches and individualized treatment decision driven by machine learning methods remains large at present.

**Data availability**

The code used for the analyses conducted in this study is available at GitLab (https://gitlab.uzh.ch/hongruyu.chen/causal_ml-for-ite). The individual patient data from IST has been made publicly available by the International Stroke Trial Collaborative Group (https://trialsjournal.biomedcentral.com/articles/10.1186/1745-6215-12-101#Sec8). And the patient entry and discharge forms from CAST were provided by the University of Oxford upon request.

**Acknowledgements**

This study was funded by a grant of the SBFI (Swiss State Secretariat for Education and Innovation) No. 24.00027, and partially funded by grant 216636 of the SNSF (Swiss National Science Foundation).

**Ethics declarations**

Competing interests

The authors declare no competing interests.

# Figures and Tables

**Table 1**. Summary of quantitative validation metrics on IST training and test dataset of 17 causal machine learning methods with 6 months' death or dependency outcome.

|  |  | c-for-benefit (training) | c-for-benefit (test) | mbcb (training) | mbcb (test) | Calibration-based pseudo R-squared (training) | Calibration-based pseudo R-squared (test) |
|---|---|---|---|---|---|---|---|
| **T-learner** | Logistic Regression | 0.541 | 0.502 | 0.553 | 0.551 | 0.685 | -1.701 |
|  | Random Forest | 0.972 | 0.491 | 0.718 | 0.616 | 0.245 | -5.855 |
|  | Support Vector Machine | 0.537 | 0.503 | 0.552 | 0.552 | 0.574 | -2.761 |
|  | XGBoost | 0.657 | 0.491 | 0.615 | 0.613 | 0.590 | -3.348 |
|  | Penalized Logistic Regression | 0.540 | 0.500 | 0.551 | 0.550 | 0.516 | -2.564 |
| **S-learner** | Logistic Regression | 0.507 | 0.496 | 0.512 | 0.511 | 0.144 | -0.641 |
|  | Penalized Logistic Regression | 0.531 | 0.497 | 0.520 | 0.520 | 0.508 | -1.697 |
|  | XGBoost | 0.682 | 0.493 | 0.571 | 0.566 | 0.315 | -3.066 |
|  | BART | 0.533 | 0.500 | 0.531 | 0.531 | 0.419 | -1.980 |
| **X-learner** | Random Forest | 0.890 | 0.496 | - | - | 0.204 | -4.563 |
|  | BART | 0.551 | 0.522 | - | - | 0.430 | 0.426 |
| **DR-learner** | Random Forest | 0.849 | 0.496 | - | - | 0.538 | -5.362 |
| **Tree-based methods** | Causal Forest | 0.593 | 0.507 | - | - | 0.083 | 0.053 |
|  | Bayesian Causal Forest | 0.524 | 0.524 | - | - | 0.062 | -0.014 |
|  | model-based recursive partitioning | 0.794 | 0.502 | - | - | - | - |
| **Deep Learning** | CVAE | 0.522 | 0.491 | 0.531 | 0.532 | 0.622 | -0.879 |
|  | GANITE | 0.519 | 0.512 | 0.539 | 0.540 | -0.647 | -1.870 |

Notes: Consistent values across training and test data that are closer to 1 indicate a better model fit. The dash symbol indicates instances where: The validation metric could not be assessed due to the model's mechanism, or, the metric was not applicable to the model. *Abbreviations*: IST: the International Stroke Trial.

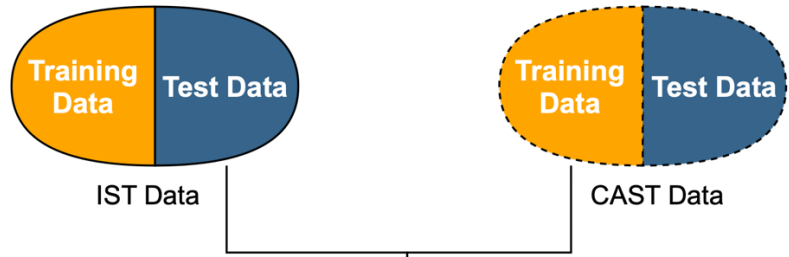
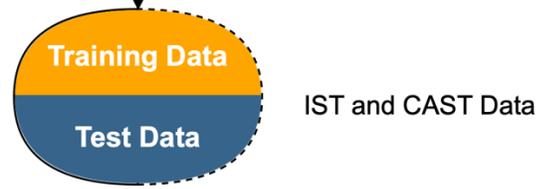
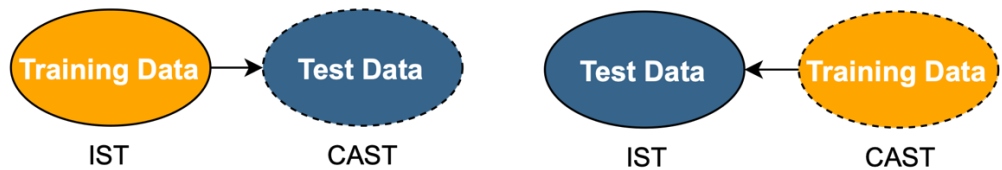

**Figure 1**. Illustration of internal and external validation strategies using IST and CAST datasets. **a**, Internal validation based on the train-test split, where the IST and CAST datasets are each divided into training and test data for model development and evaluation respectively. **b**, Internal validation on the combined IST and CAST dataset, where the combined dataset is divided into training and test data. **c**, External validation across datasets, where both the IST and CAST datasets are alternately used as training and test data. Abbreviations: IST: the International Stroke Trial, CAST: the Chinese Acute Stroke Trial.

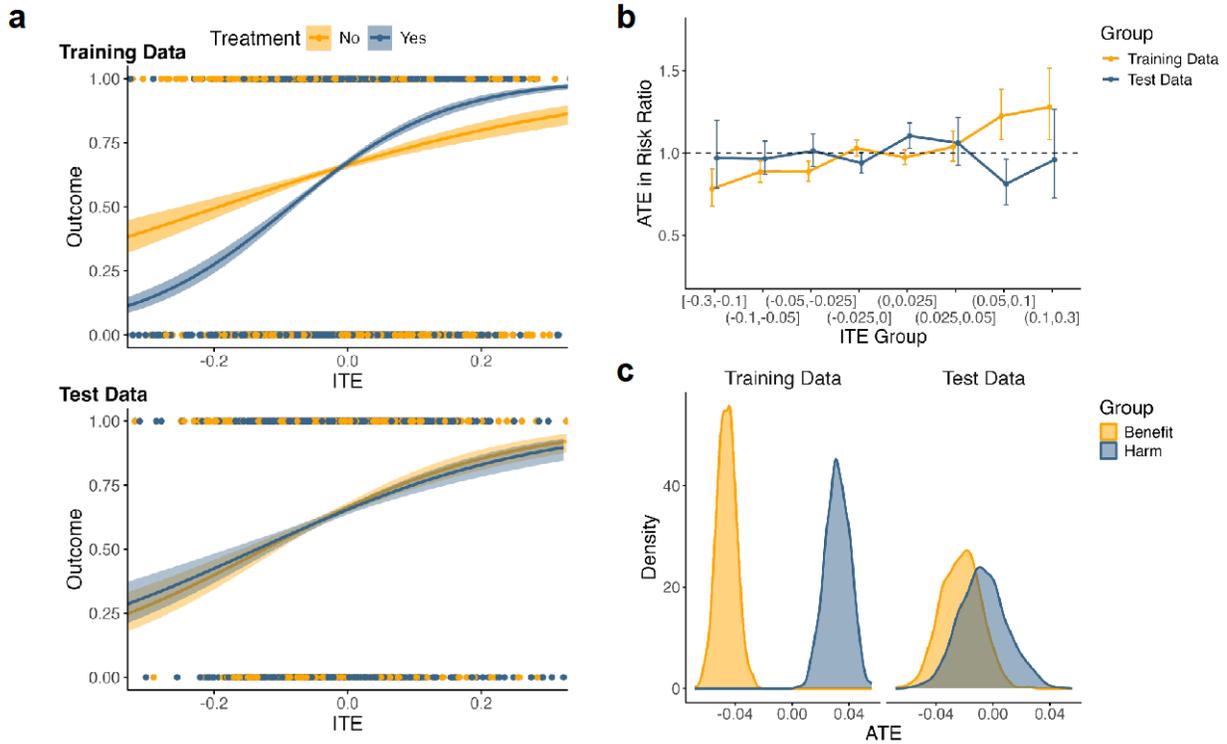

**Figure 2**. Comparative analysis of logistic regression-based individualized treatment effects in IST. **a**, Comparing patient outcomes against estimated ITE values between training data and test data. Orange dots and lines represent the true and fitted outcome probabilities for the control group, while blue dots and lines indicate the true and fitted outcome probabilities for the treatment group. **b**, ATE in risk ratio within different ITE subgroups, confidence intervals at 95% level. Orange represents the training data, and blue indicates the test data. The horizontal dashed line at 1.0 means no treatment effects. **c**, Density distributions of ATE in training data and test data stratified by benefit (orange) and harm (blue) groups. The distribution of ATE comes from 1000 random train-test split experiments. Benefit means negative ITE values and harm means positive ITE values. The outcome variable used in **a-c** is death or dependency at 6 months. Abbreviations: IST: the International Stroke Trial, ITE: individualized treatment effect, ATE: average treatment effect.

**Supplementary Materials**

**Sensitivity analysis**

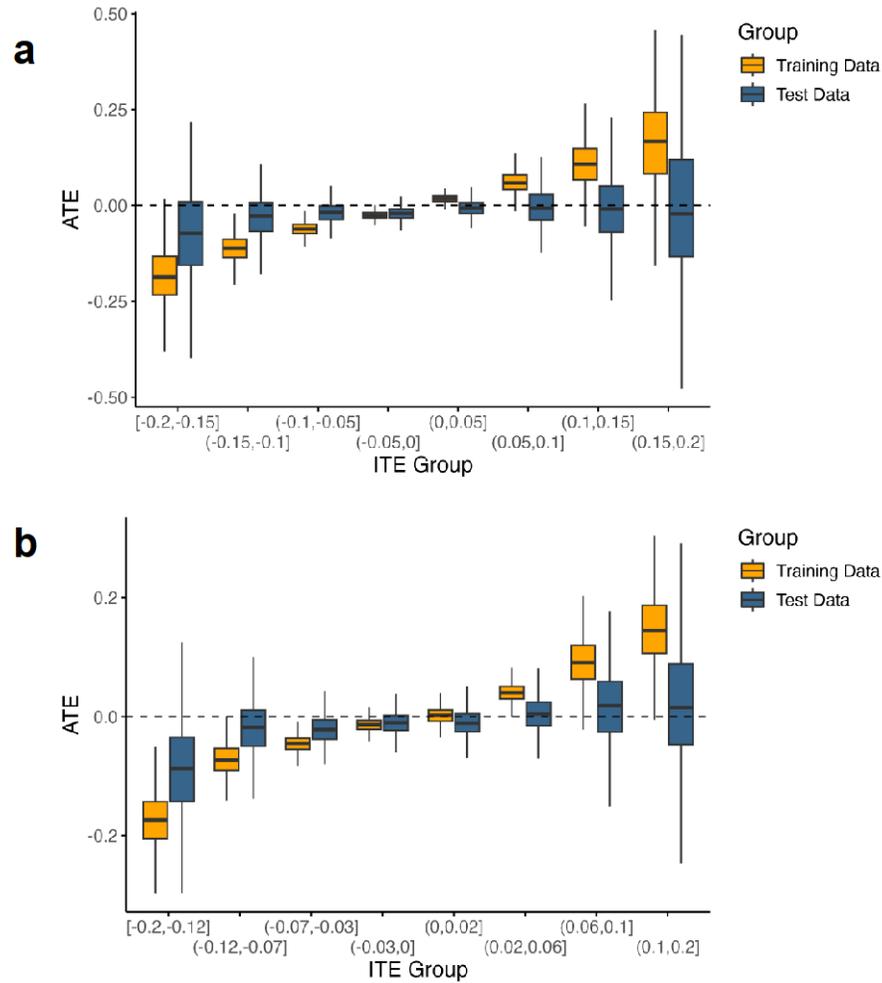

Figure S1. Box plots of ATE in risk ratio within different ITE subgroups from 1000 random train-test split experiments. It compares the trends of ATE between training data (orange bars) and test data (blue bars) across various ITE ranges. The horizontal dashed line at 0 indicates no treatment effects. **a**, ITE estimated by logistic regression model on the IST dataset with the outcome variable being death or dependency at 6 months. **b**, ITE estimated by logistic regression model on the CAST dataset with the outcome variable being death or dependency at 4 weeks. Abbreviations: IST: the International Stroke Trial, CAST: the Chinese Acute Stroke Trial, ITE: individualized treatment effect, ATE: average treatment effect.

**Results of internal validation on the IST dataset**

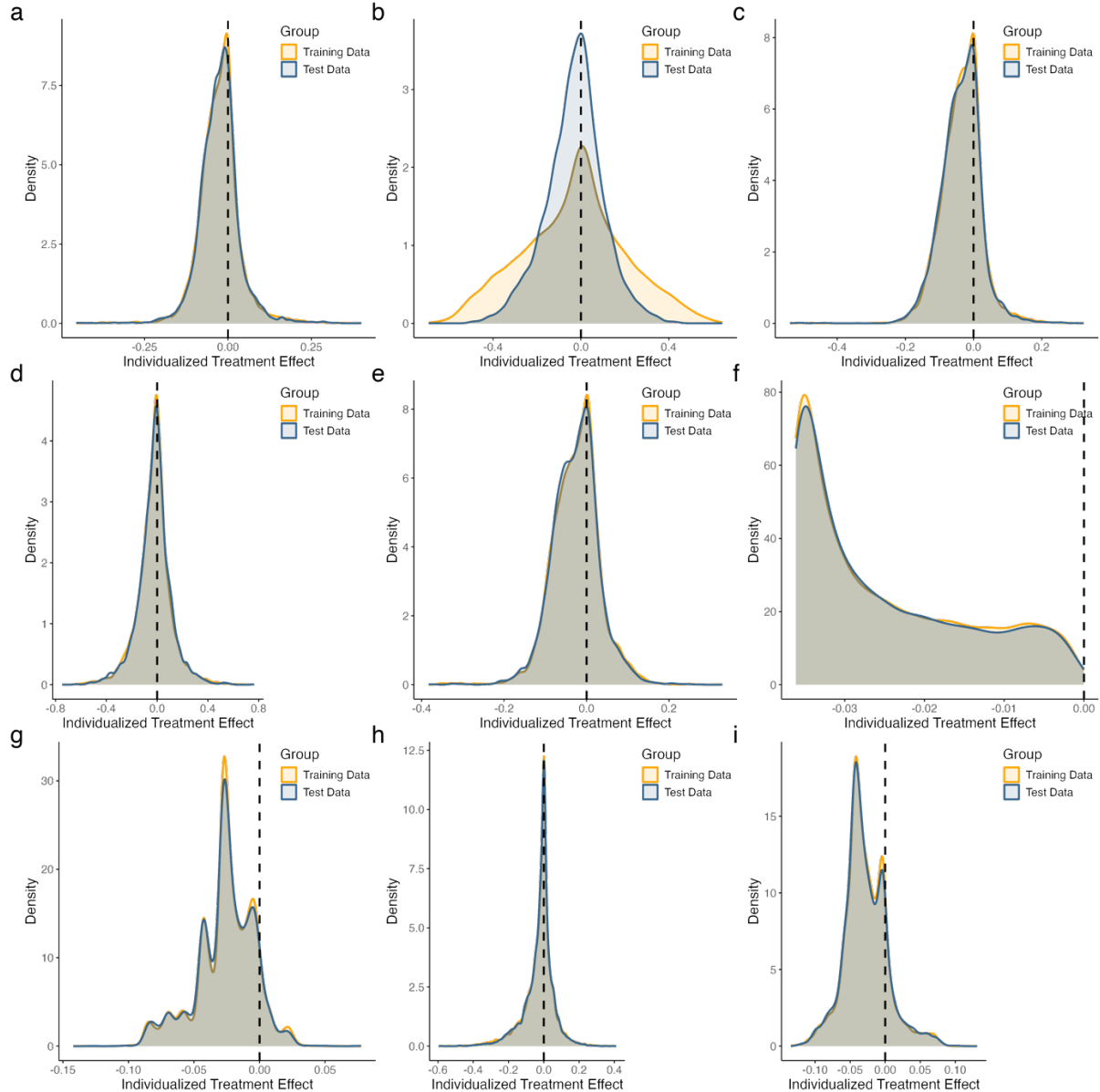

Figure S2.1.1. Internal validation on the IST dataset: density plots of causal machine learning-based individualized treatment effects. Orange represents the training data, and blue indicates the test data. The outcome variable is death or dependency at 6 months. **a**, T-learner Logistic Regression. **b**, T-learner Random Forest. **c**, T-learner Support Vector Machine. **d**, T-learner XGBoost. **e**, T-learner Penalized Logistic Regression. **f**, S-learner Logistic Regression. **g**, S-learner Penalized Logistic Regression. **h**, S-learner XGBoost. **i**, S-learner BART. Abbreviations: IST: the International Stroke Trial, ITE: individualized treatment effect.

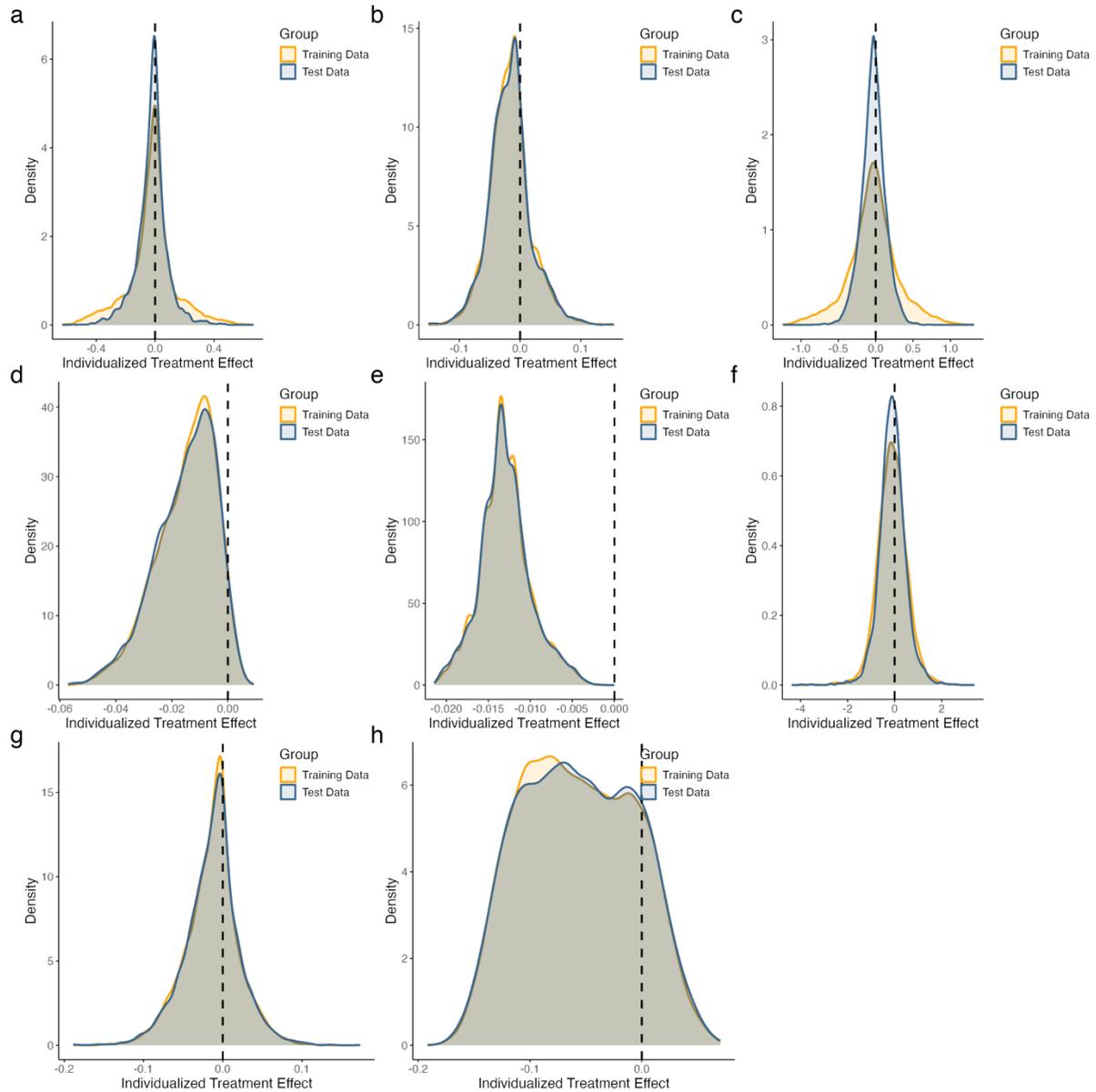

Figure S2.1.2. Internal validation on the IST dataset: density plots of causal machine learning-based individualized treatment effects. Orange represents the training data, and blue indicates the test data. The outcome variable is death or dependency at 6 months. **a**, X-learner Random Forest. **b**, X-learner BART. **c**, DR-learner Random Forest. **d**, Causal Forest. **e**, Bayesian Causal Forest. **f**, Model-based Recursive Partitioning. **g**, CVAE. **h**, GANITE. Abbreviations: IST: the International Stroke Trial, ITE: individualized treatment effect.

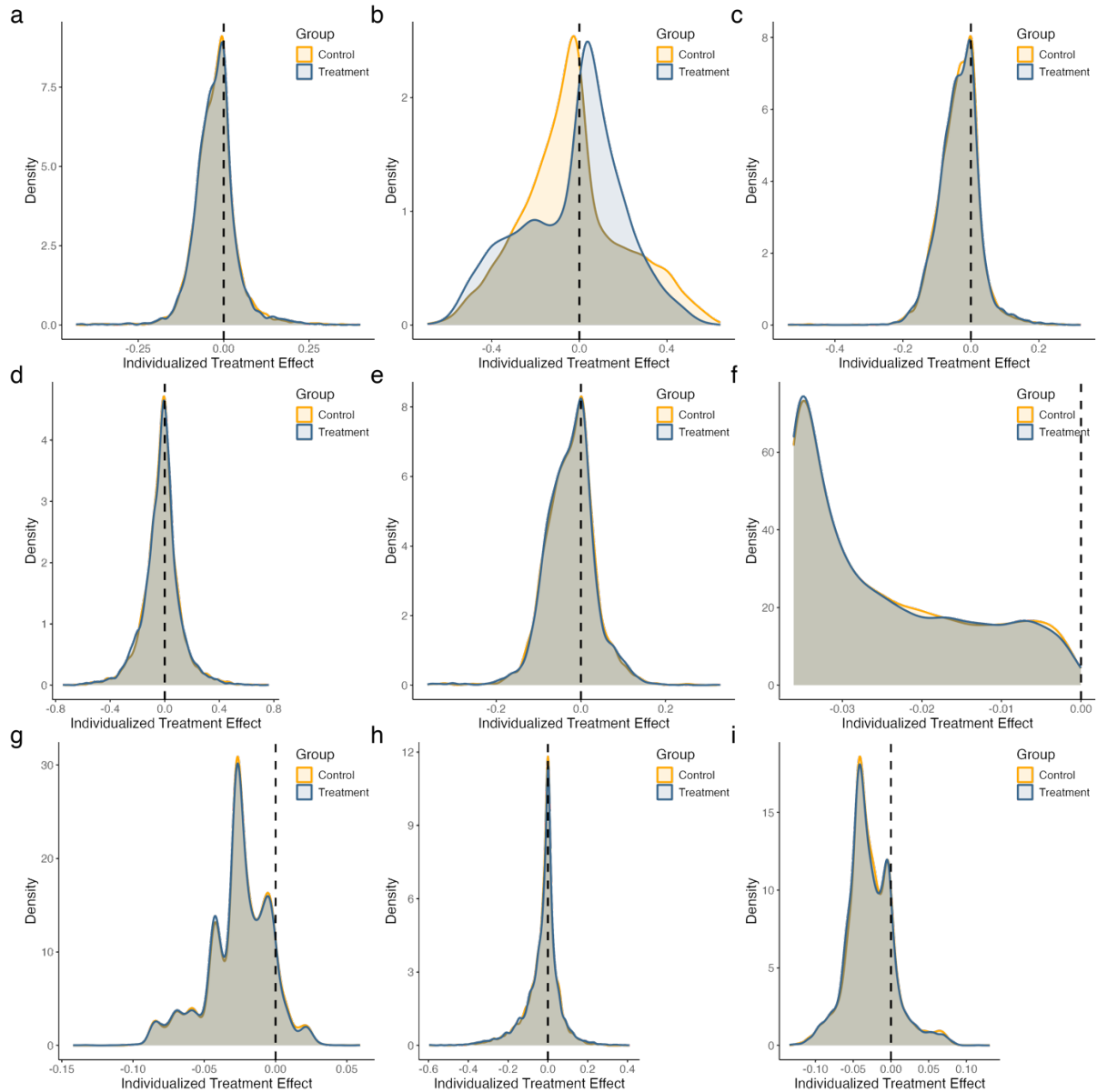

Figure S2.2.1. Internal validation on the IST training dataset: density plots of causal machine learning-based individualized treatment effects. Orange represents the control group, and blue indicates the treatment group. The outcome variable is death or dependency at 6 months. **a**, T-learner Logistic Regression. **b**, T-learner Random Forest. **c**, T-learner Support Vector Machine. **d**, T-learner XGBoost. **e**, T-learner Penalized Logistic Regression. **f**, S-learner Logistic Regression. **g**, S-learner Penalized Logistic Regression. **h**, S-learner XGBoost. **i**, S-learner BART. Abbreviations: IST: the International Stroke Trial, ITE: individualized treatment effect.

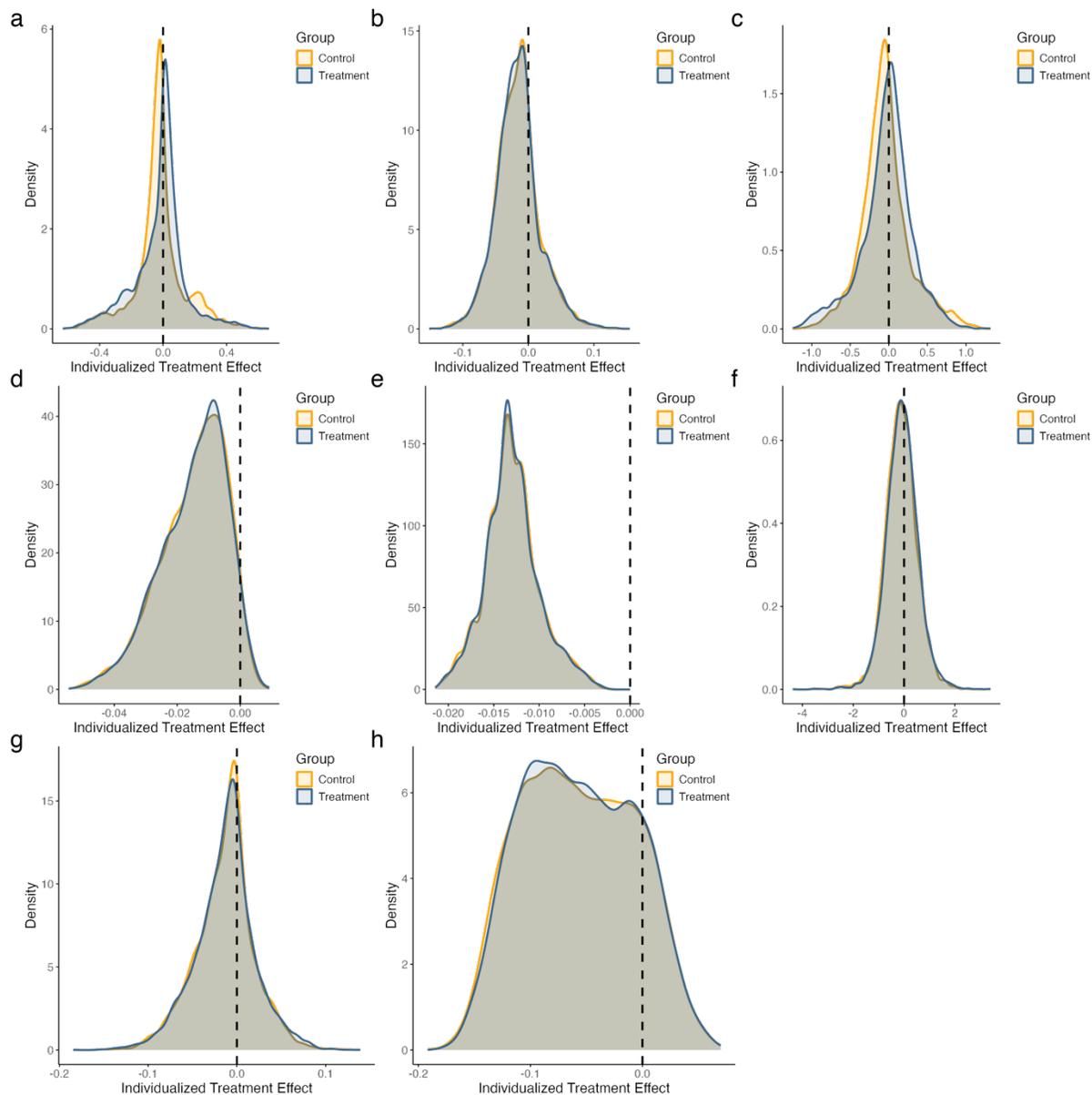

Figure S2.2.2. Internal validation on the IST training dataset: density plots of causal machine learning-based individualized treatment effects. Orange represents the control group, and blue indicates the treatment group. The outcome variable is death or dependency at 6 months. **a**, X-learner Random Forest. **b**, X-learner BART. **c**, DR-learner Random Forest. **d**, Causal Forest. **e**, Bayesian Causal Forest. **f**, Model-based Recursive Partitioning. **g**, CVAE. **h**, GANITE. Abbreviations: IST: the International Stroke Trial, ITE: individualized treatment effect.

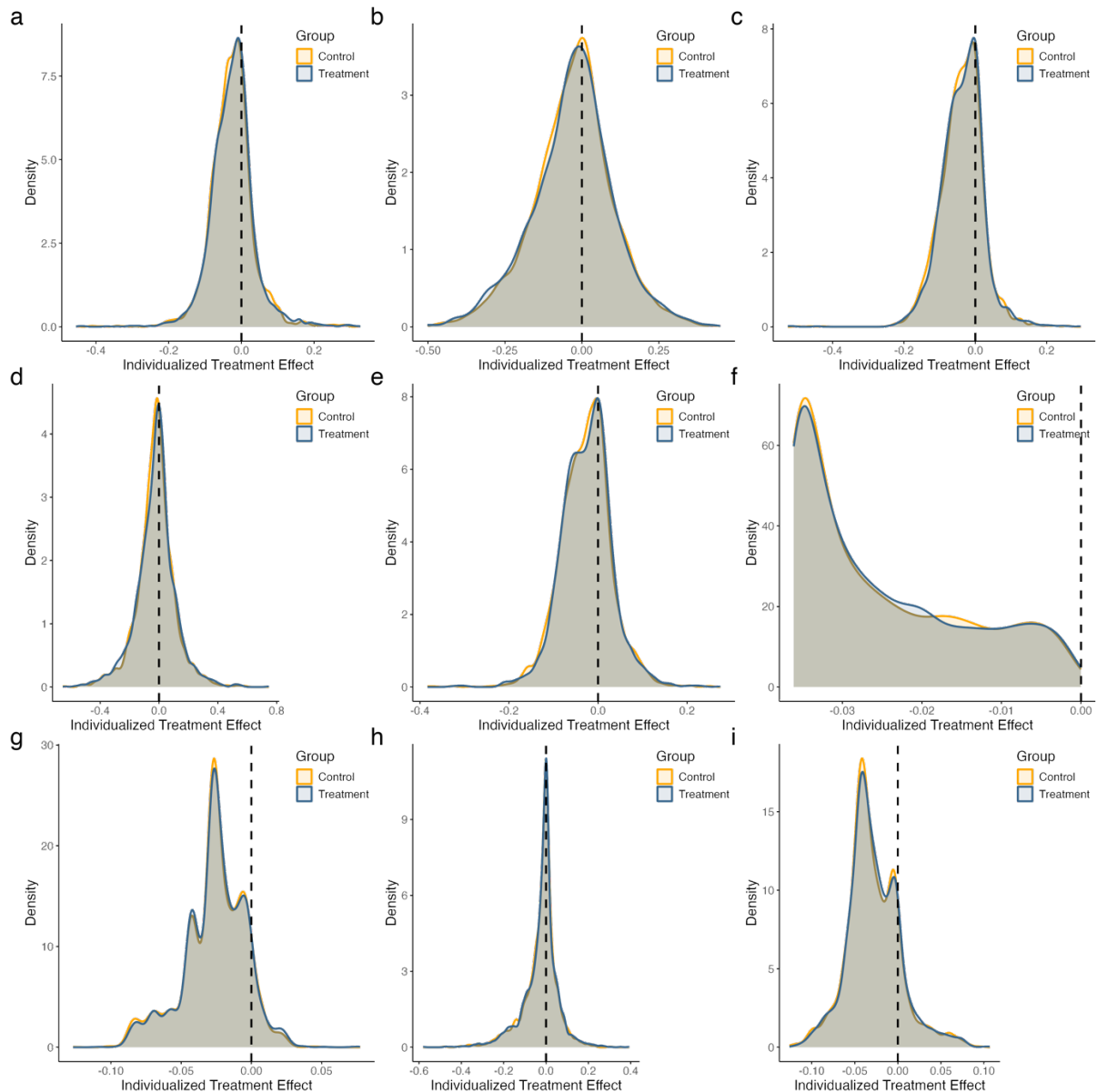

Figure S2.3.1. Internal validation on the IST test dataset: density plots of causal machine learning-based individualized treatment effects. Orange represents the control group, and blue indicates the treatment group. The outcome variable is death or dependency at 6 months. **a**, T-learner Logistic Regression. **b**, T-learner Random Forest. **c**, T-learner Support Vector Machine. **d**, T-learner XGBoost. **e**, T-learner Penalized Logistic Regression. **f**, S-learner Logistic Regression. **g**, S-learner Penalized Logistic Regression. **h**, S-learner XGBoost. **i**, S-learner BART. Abbreviations: IST: the International Stroke Trial, ITE: individualized treatment effect.

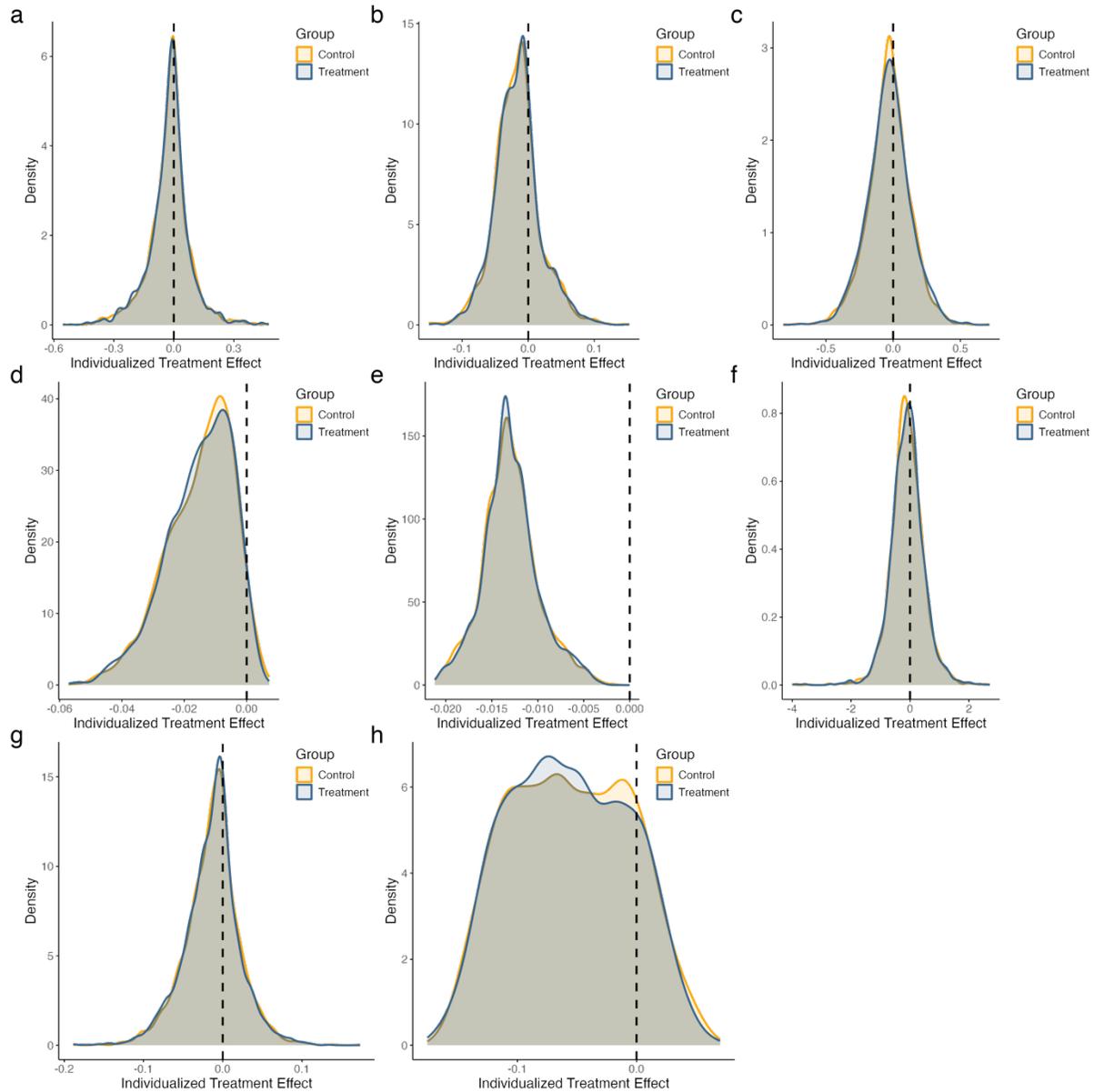

Figure S2.3.2. Internal validation on the IST test dataset: density plots of causal machine learning-based individualized treatment effects. Orange represents the control group, and blue indicates the treatment group. The outcome variable is death or dependency at 6 months. **a**, X-learner Random Forest. **b**, X-learner BART. **c**, DR-learner Random Forest. **d**, Causal Forest. **e**, Bayesian Causal Forest. **f**, Model-based Recursive Partitioning. **g**, CVAE. **h**, GANITE. Abbreviations: IST: the International Stroke Trial, ITE: individualized treatment effect.

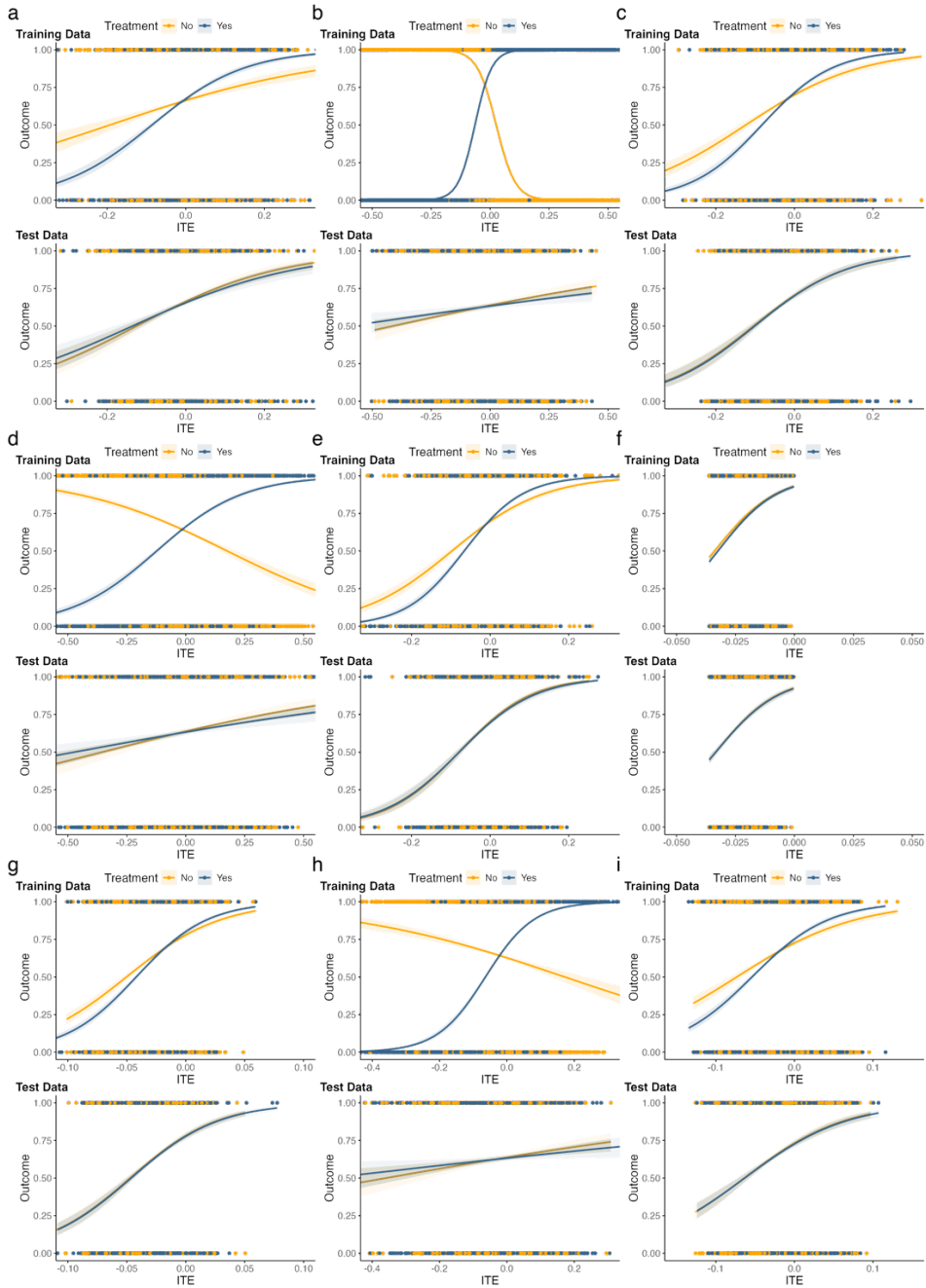

Figure S2.4.1. Internal validation on the IST dataset: outcome-ITE comparative analysis. Dot plots and line plots depict the true and fitted patient outcomes against estimated ITE values between training data and test data. Orange represents the control group, and blue indicates the treatment group. The outcome variable is death or dependency at 6 months. **a**, T-learner Logistic Regression. **b**, T-learner Random Forest. **c**, T-learner Support Vector Machine. **d**, T-learner XGBoost. **e**, T-learner Penalized Logistic Regression. **f**, S-learner Logistic Regression. **g**, S-learner Penalized Logistic Regression. **h**, S-learner XGBoost. **i**, S-learner BART. Abbreviations: IST: the International Stroke Trial, ITE: individualized treatment effect.

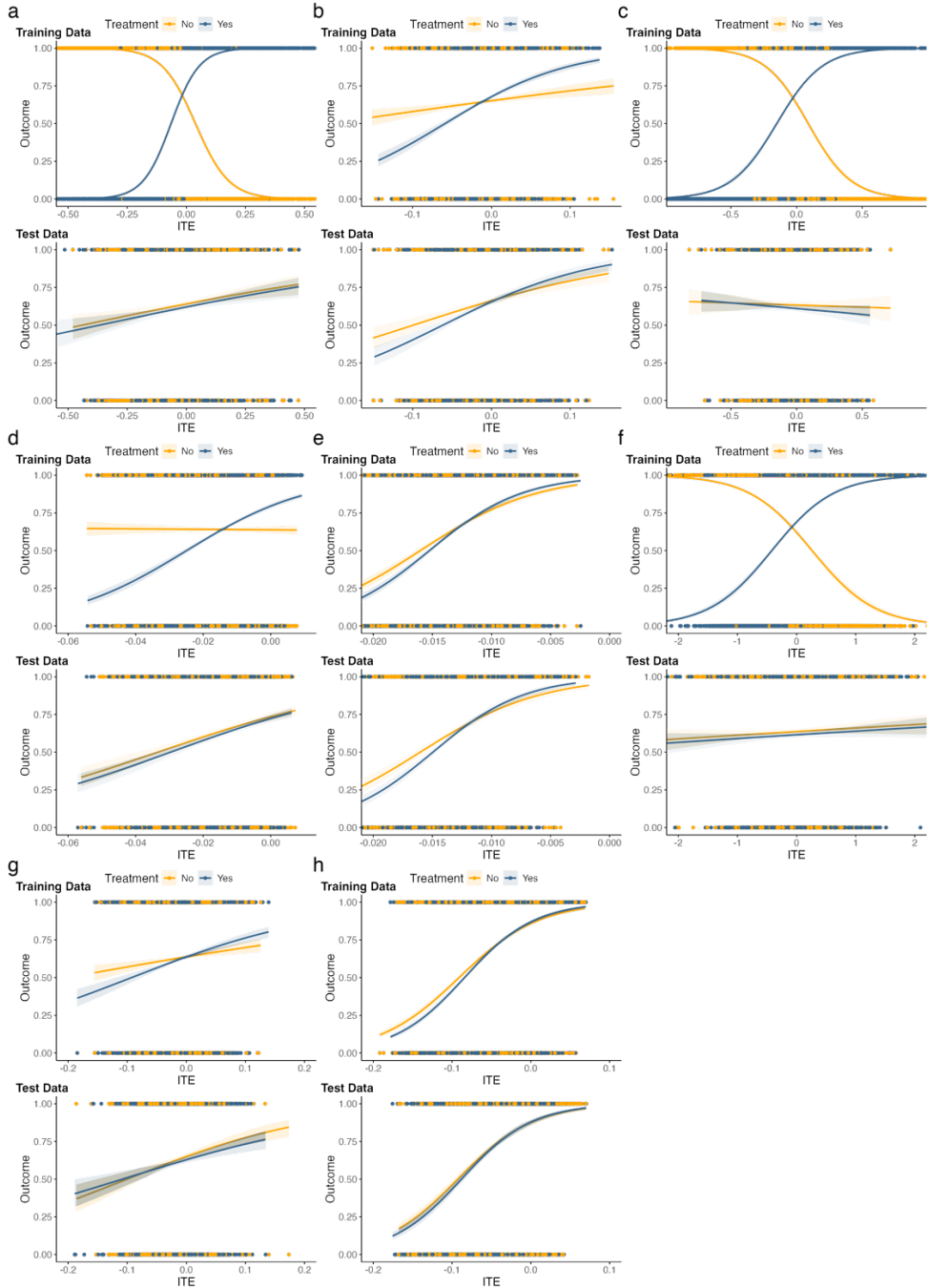

Figure S2.4.2. Internal validation on the IST dataset: outcome-ITE comparative analysis. Dot plots and line plots depict the true and fitted patient outcomes against estimated ITE values between training data and test data. Orange represents the control group, and blue indicates the treatment group. The outcome variable is death or dependency at 6 months. **a**, X-learner Random Forest. **b**, X-learner BART. **c**, DR-learner Random Forest. **d**, Causal Forest. **e**, Bayesian Causal Forest. **f**, Model-based Recursive Partitioning. **g**, CVAE. **h**, GANITE. Abbreviations: IST: the International Stroke Trial, ITE: individualized treatment effect.

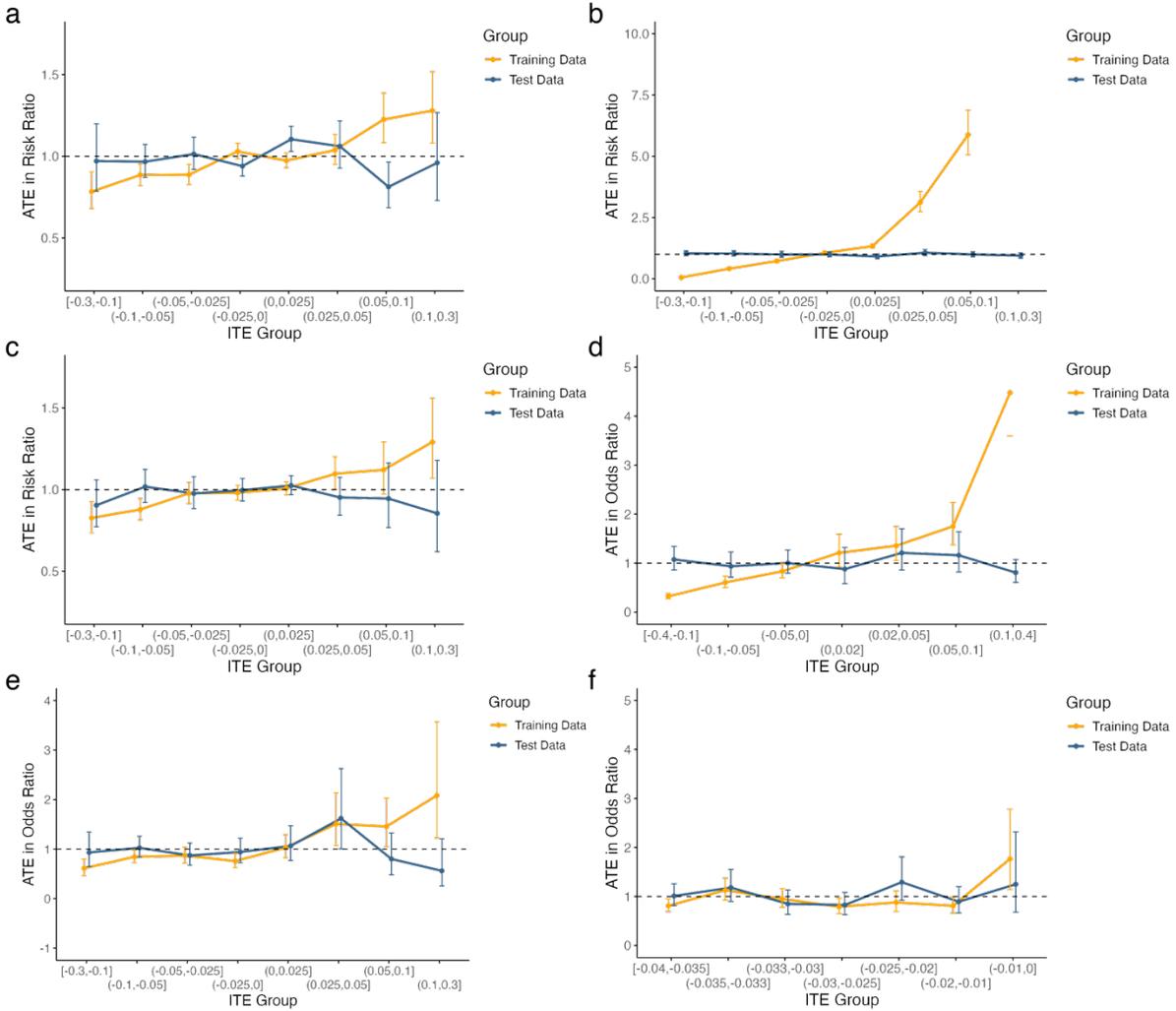

Figure S2.5.1. Internal validation on the IST dataset: ATE-ITE comparative analysis. Line plots depict ATE in risk ratio within different ITE subgroups and provide the confidence intervals at 95% level. Orange represents the training data, and blue indicates the test data. The horizontal dashed line at 1.0 means no treatment effects. The outcome variable is death or dependency at 6 months. **a**, T-learner Logistic Regression. **b**, T-learner Random Forest. **c**, T-learner Support Vector Machine. **d**, T-learner XGBoost. **e**, T-learner Penalized Logistic Regression. **f**, S-learner Logistic Regression. Abbreviations: IST: the International Stroke Trial, ITE: individualized treatment effect, ATE: average treatment effect.

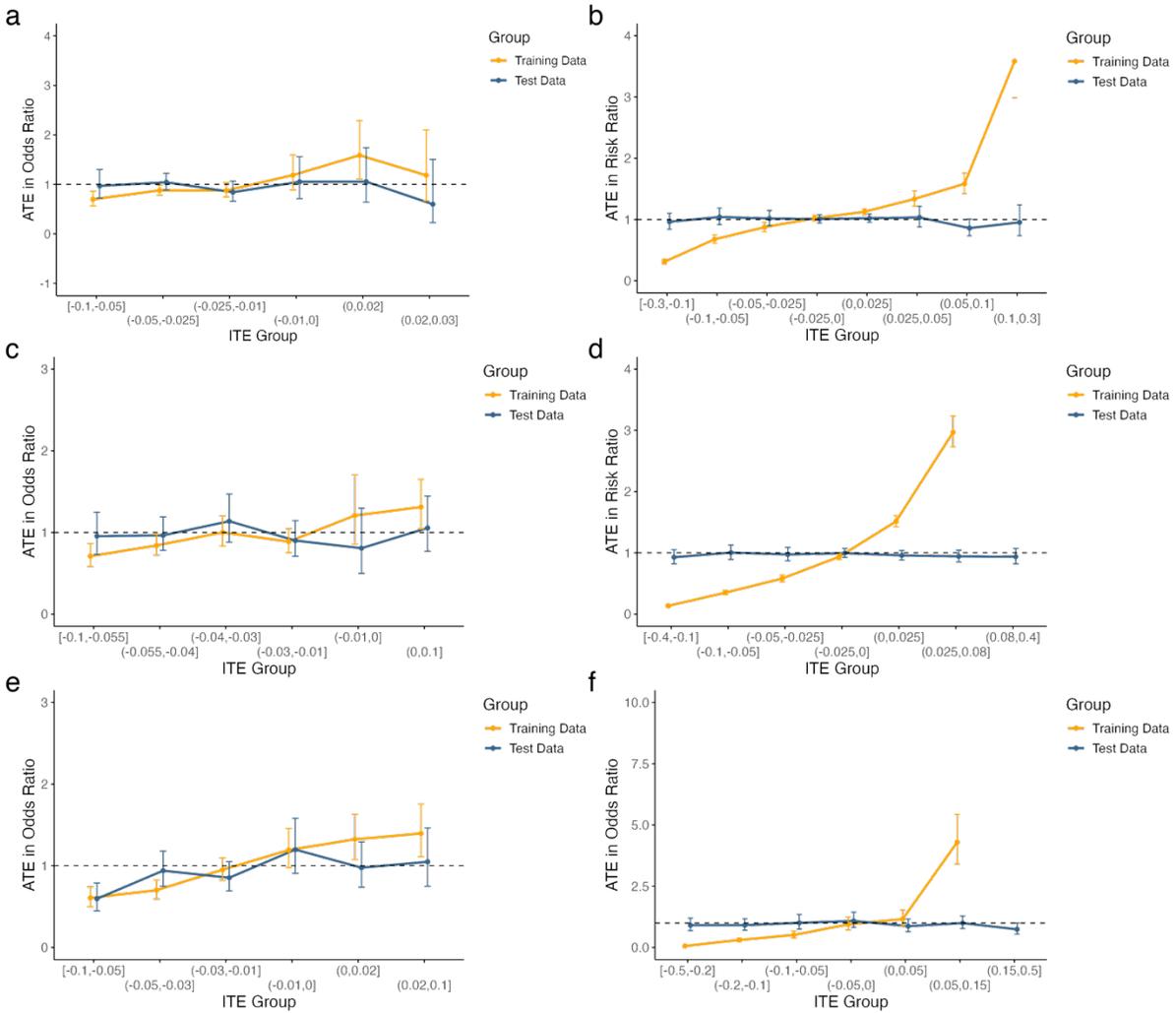

Figure S2.5.2. Internal validation on the IST dataset: ATE-ITE comparative analysis. Line plots depict ATE in risk ratio within different ITE subgroups and provide the confidence intervals at 95% level. Orange represents the training data, and blue indicates the test data. The horizontal dashed line at 1.0 means no treatment effects. The outcome variable is death or dependency at 6 months. **a**, S-learner Penalized Logistic Regression. **b**, S-learner XGBoost. **c**, S-learner BART. **d**, X-learner Random Forest. **e**, X-learner BART. **f**, DR-learner Random Forest. Abbreviations: IST: the International Stroke Trial, ITE: individualized treatment effect, ATE: average treatment effect.

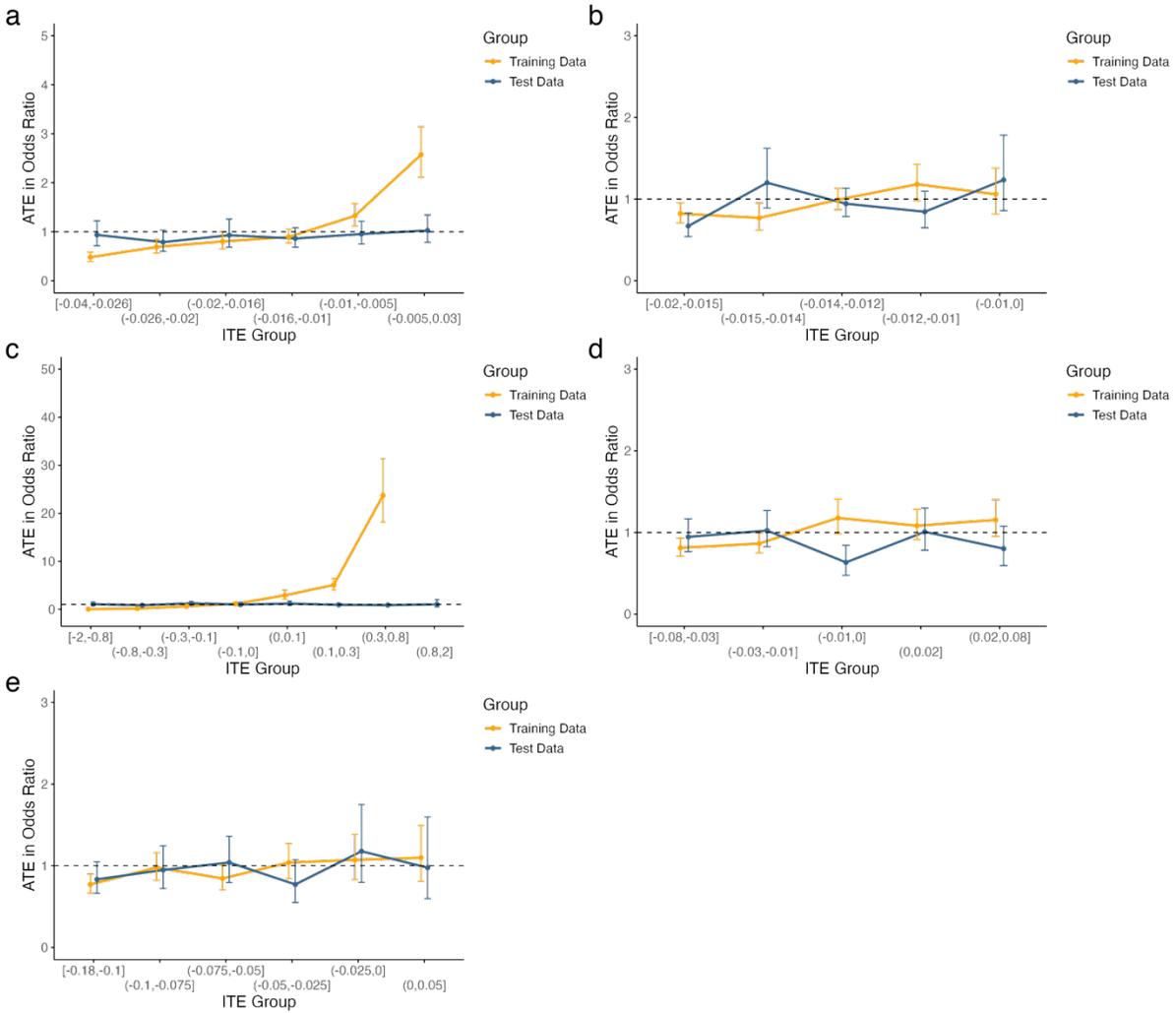

Figure S2.5.2. Internal validation on the IST dataset: ATE-ITE comparative analysis. Line plots depict ATE in risk ratio within different ITE subgroups and provide the confidence intervals at 95% level. Orange represents the training data, and blue indicates the test data. The horizontal dashed line at 1.0 means no treatment effects. The outcome variable is death or dependency at 6 months. **a**, Causal Forest. **b**, Bayesian Causal Forest. **c**, Model-based Recursive Partitioning. **d**, CVAE. **e**, GANITE. Abbreviations: IST: the International Stroke Trial, ITE: individualized treatment effect, ATE: average treatment effect.

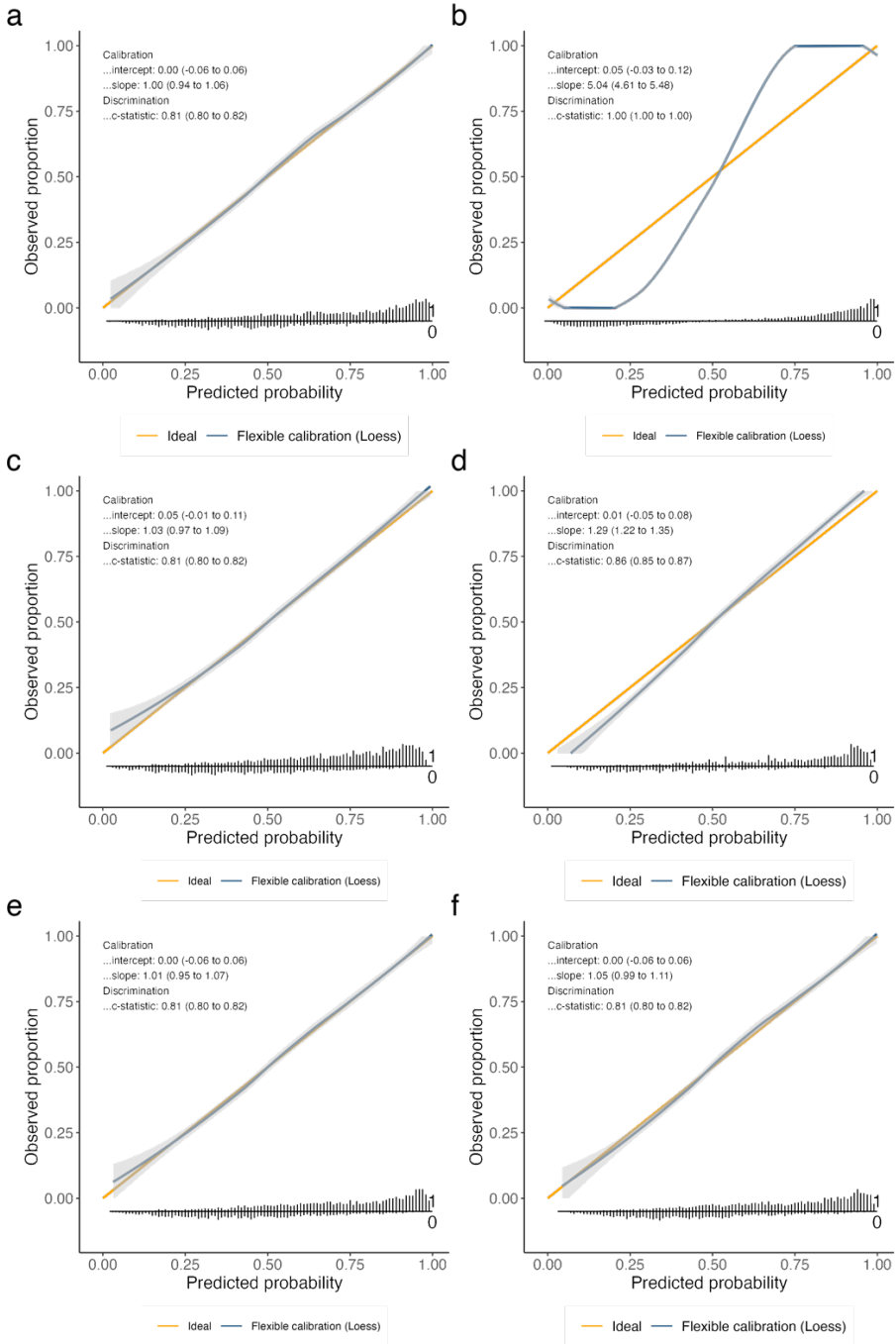

Figure S2.6.1. Internal validation on the IST training dataset: calibration plot of predicted outcome in the treatment group. The orange line indicates ideal calibration. The outcome variable is death or dependency at 6 months. **a**, T-learner Logistic Regression. **b**, T-learner Random Forest. **c**, T-learner Support Vector Machine. **d**, T-learner XGBoost. **e**, T-learner Penalized Logistic Regression. **f**, S-learner Logistic Regression. Abbreviations: IST: the International Stroke Trial.

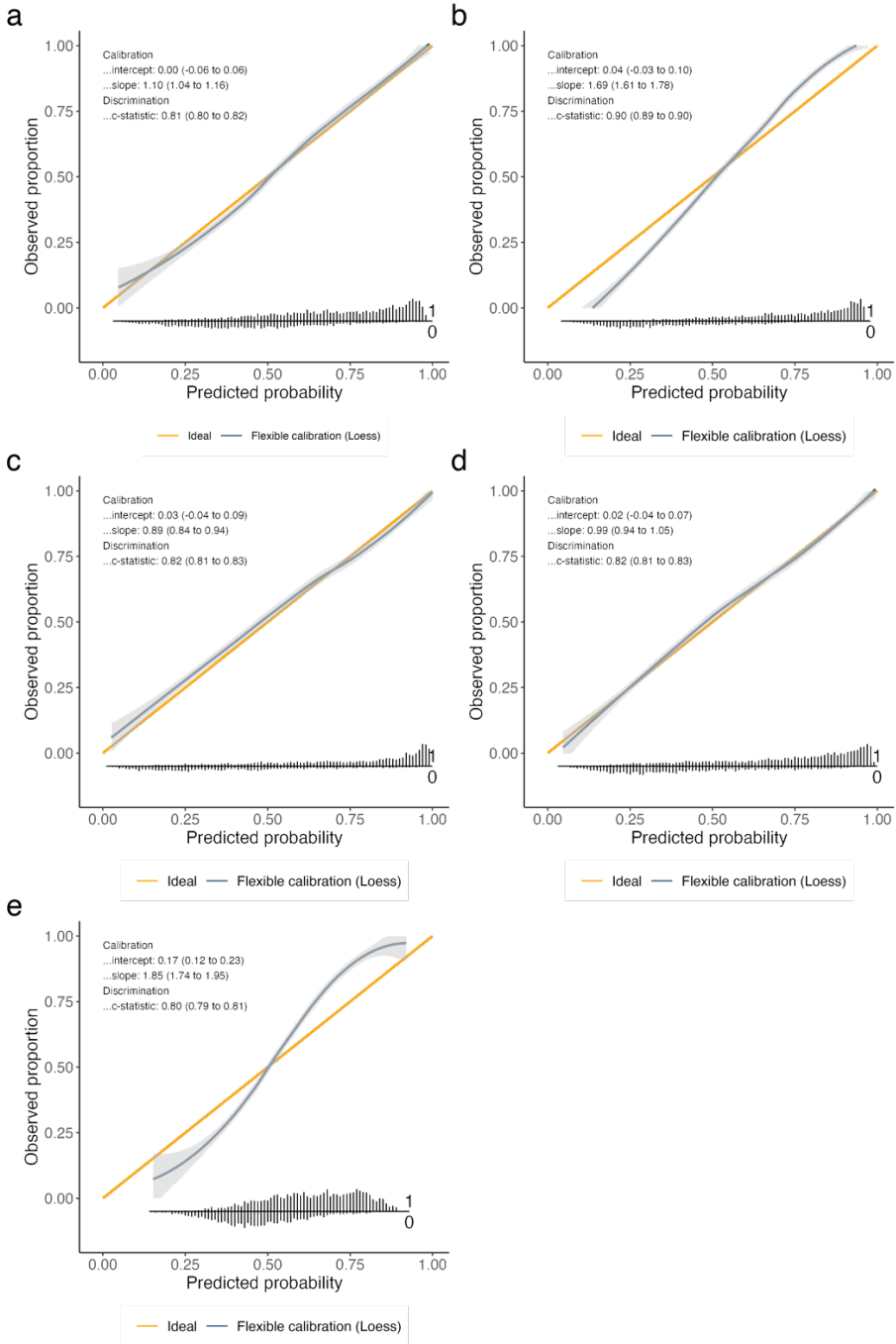

Figure S2.6.2. Internal validation on the IST training dataset: calibration plot of predicted outcome in the treatment group. The orange line indicates ideal calibration. The outcome variable is death or dependency at 6 months. **a**, S-learner Penalized Logistic Regression. **b**, S-learner XGBoost **c**, S-learner BART. **d**, CVAE. **e**, GANITE. Abbreviations: IST: the International Stroke Trial.

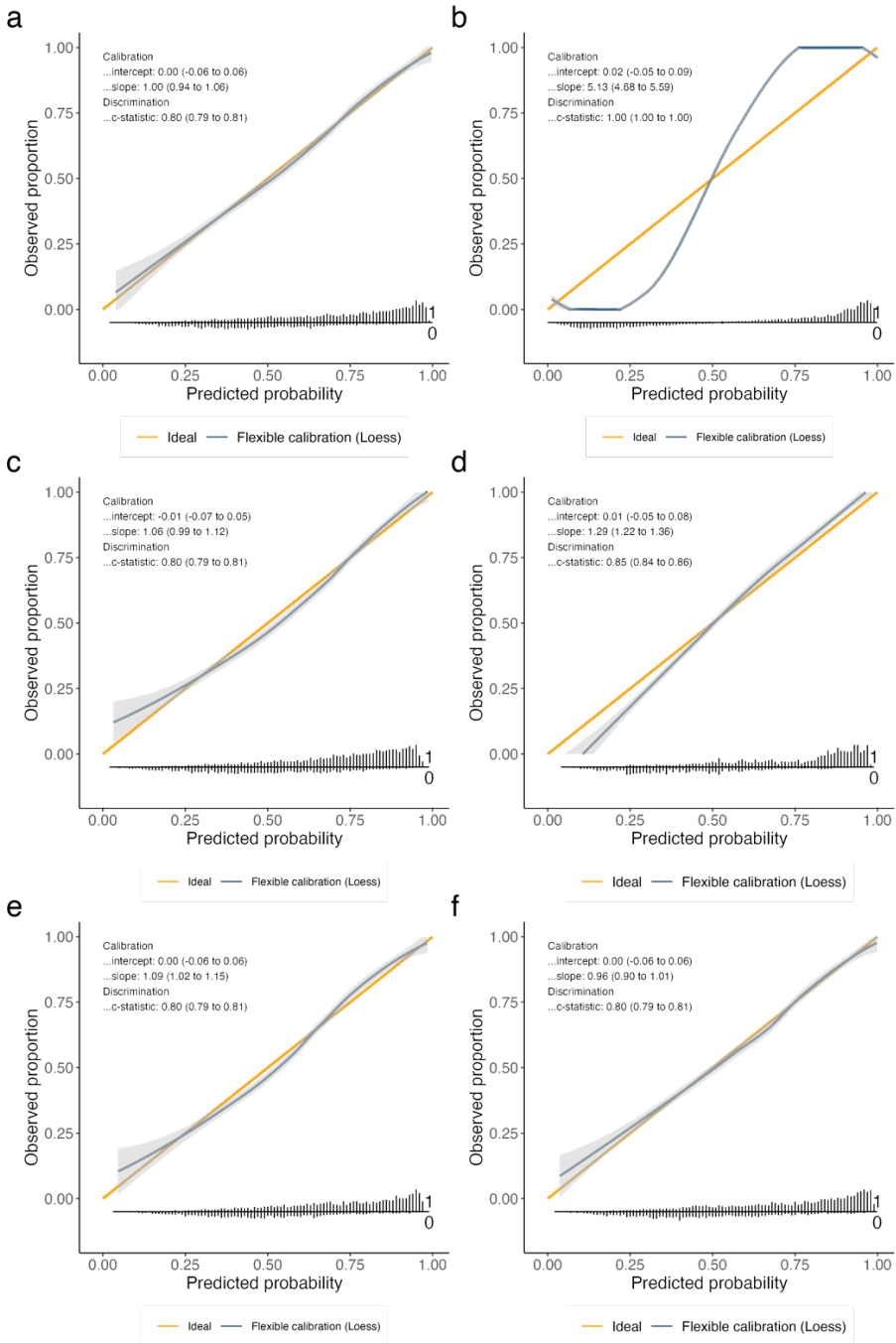

Figure S2.7.1. Internal validation on the IST training dataset: calibration plot of predicted outcome in the control group. The orange line indicates ideal calibration. The outcome variable is death or dependency at 6 months. **a**, T-learner Logistic Regression. **b**, T-learner Random Forest. **c**, T-learner Support Vector Machine. **d**, T-learner XGBoost. **e**, T-learner Penalized Logistic Regression. **f**, S-learner Logistic Regression. Abbreviations: IST: the International Stroke Trial.

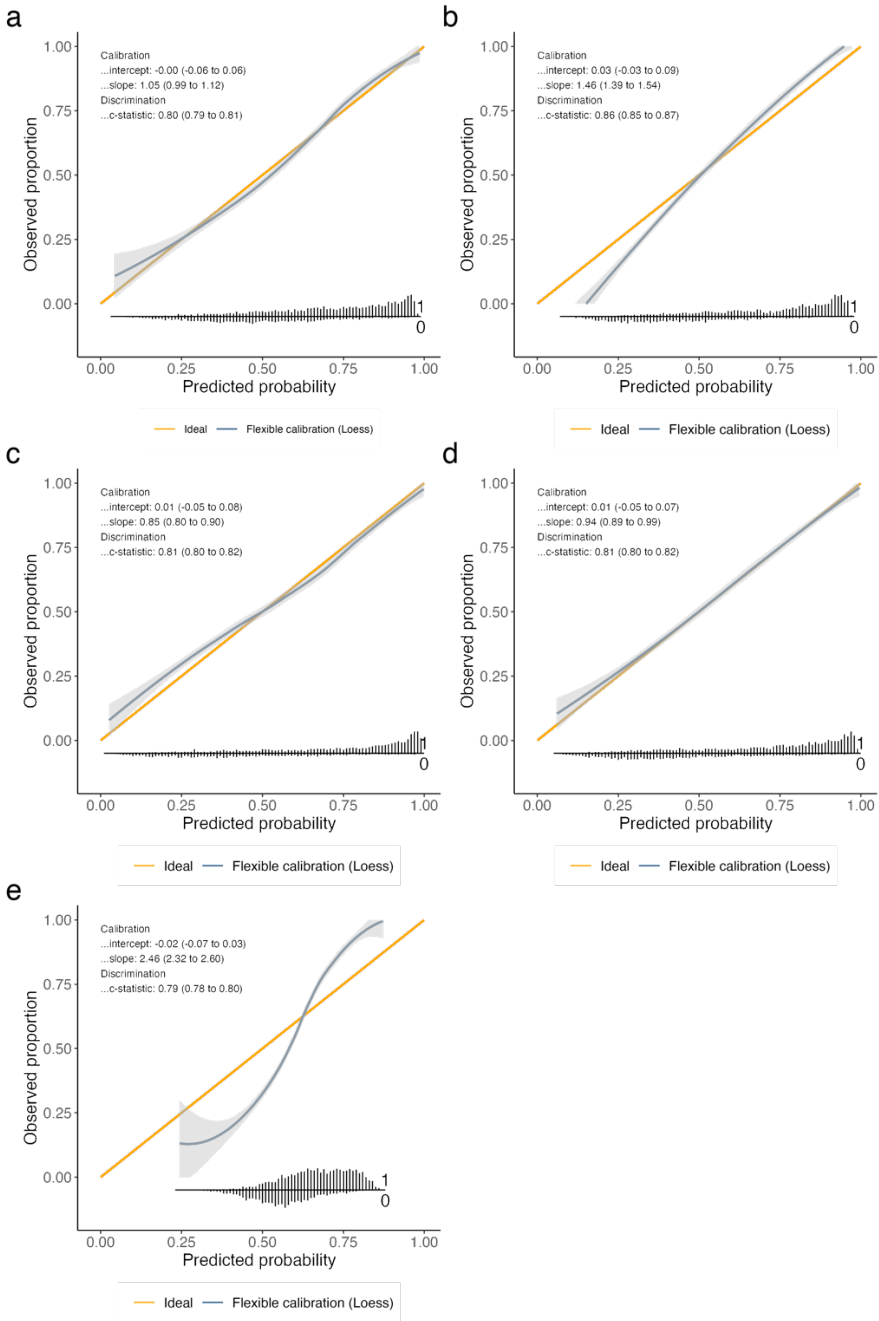

Figure S2.7.2. Internal validation on the IST training dataset: calibration plot of predicted outcome in the control group. The orange line indicates ideal calibration. The outcome variable is death or dependency at 6 months. **a**, S-learner Penalized Logistic Regression. **b**, S-learner XGBoost **c**, S-learner BART. **d**, CVAE. **e**, GANITE. Abbreviations: IST: the International Stroke Trial.

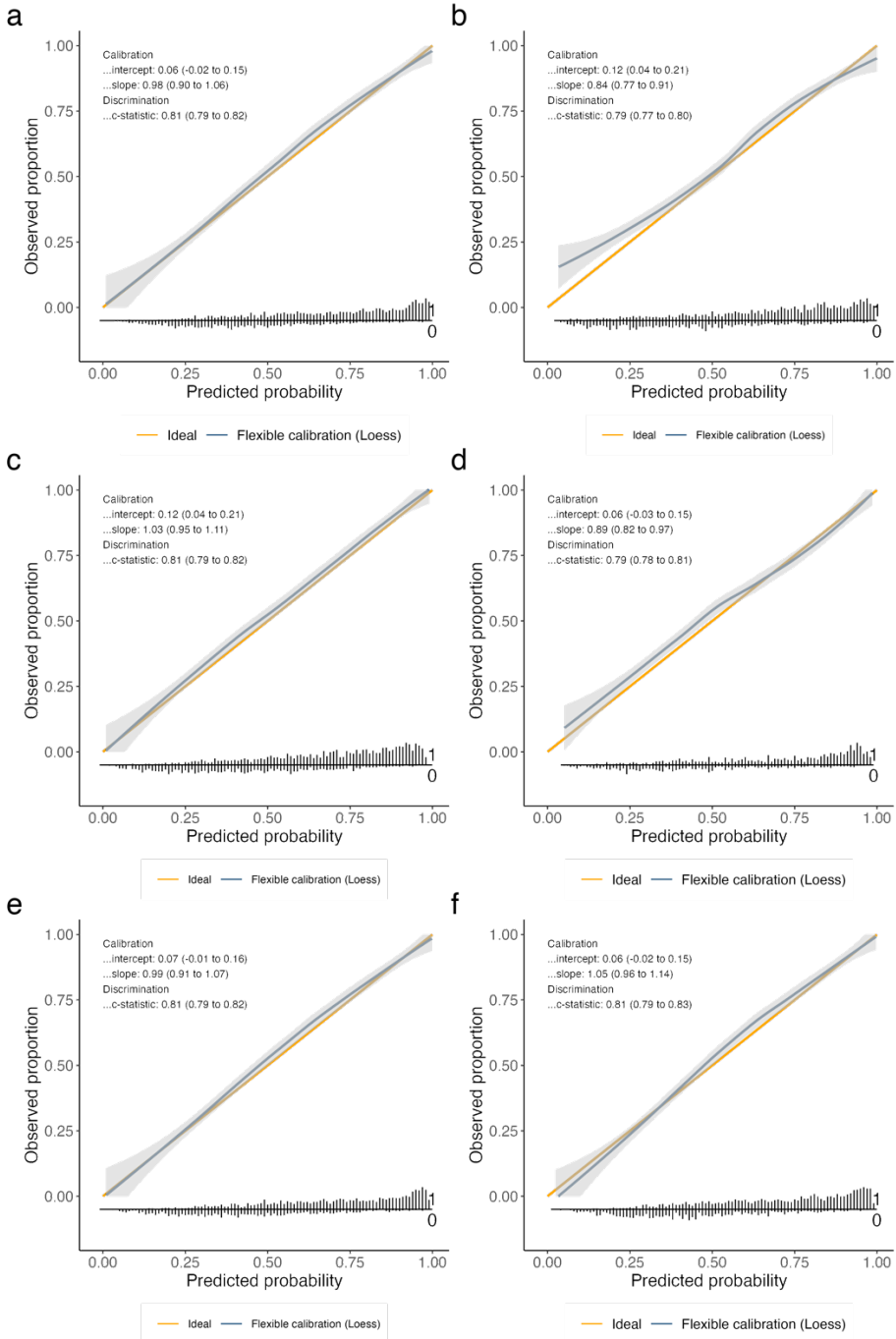

Figure S2.8.1. Internal validation on the IST test dataset: calibration plot of predicted outcome in the treatment group. The orange line indicates ideal calibration. The outcome variable is death or dependency at 6 months. **a**, T-learner Logistic Regression. **b**, T-learner Random Forest. **c**, T-learner Support Vector Machine. **d**, T-learner XGBoost. **e**, T-learner Penalized Logistic Regression. **f**, S-learner Logistic Regression. Abbreviations: IST: the International Stroke Trial.

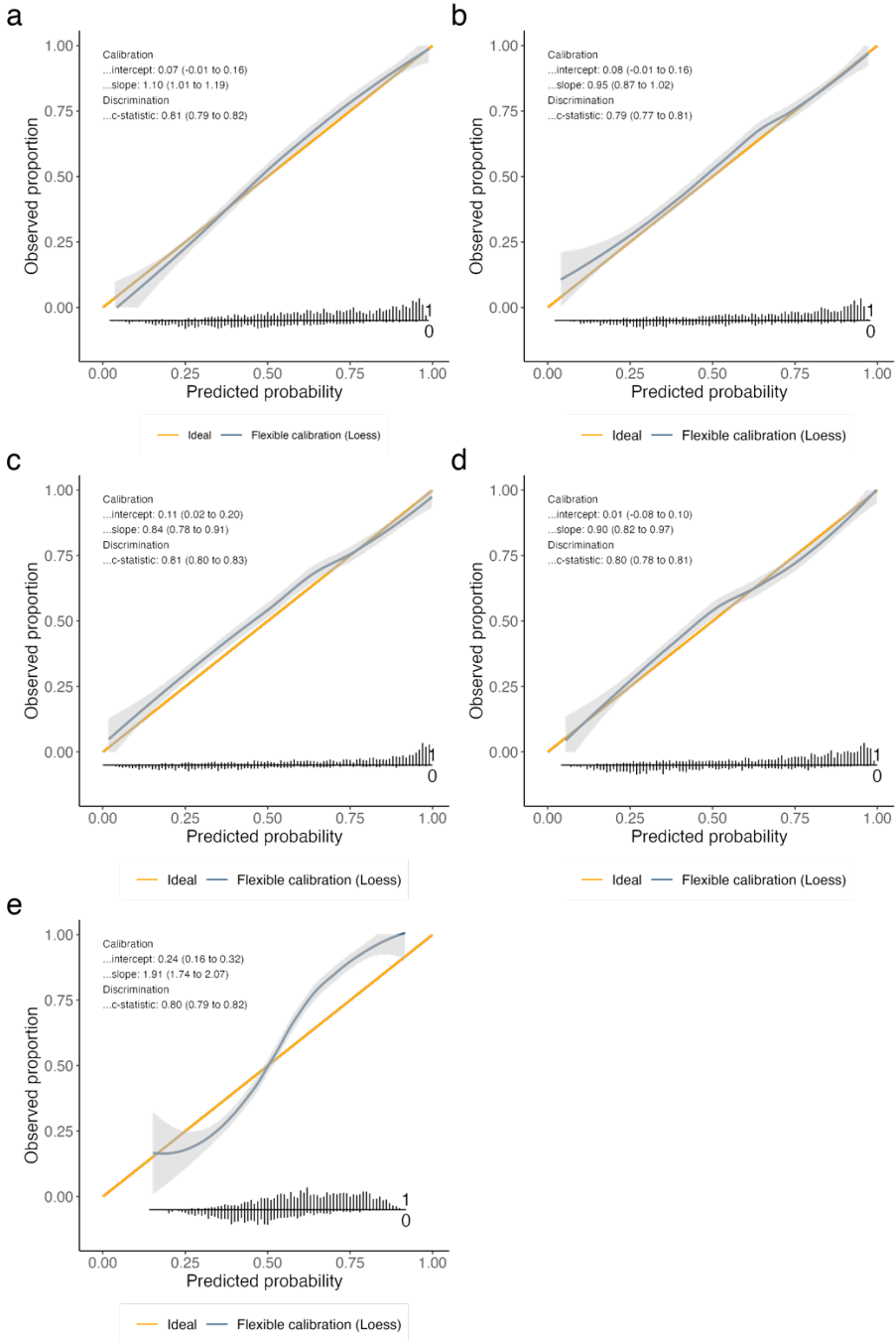

Figure S2.8.2. Internal validation on the IST test dataset: calibration plot of predicted outcome in the treatment group. The orange line indicates ideal calibration. The outcome variable is death or dependency at 6 months. **a**, S-learner Penalized Logistic Regression. **b**, S-learner XGBoost **c**, S-learner BART. **d**, CVAE. **e**, GANITE. Abbreviations: IST: the International Stroke Trial.

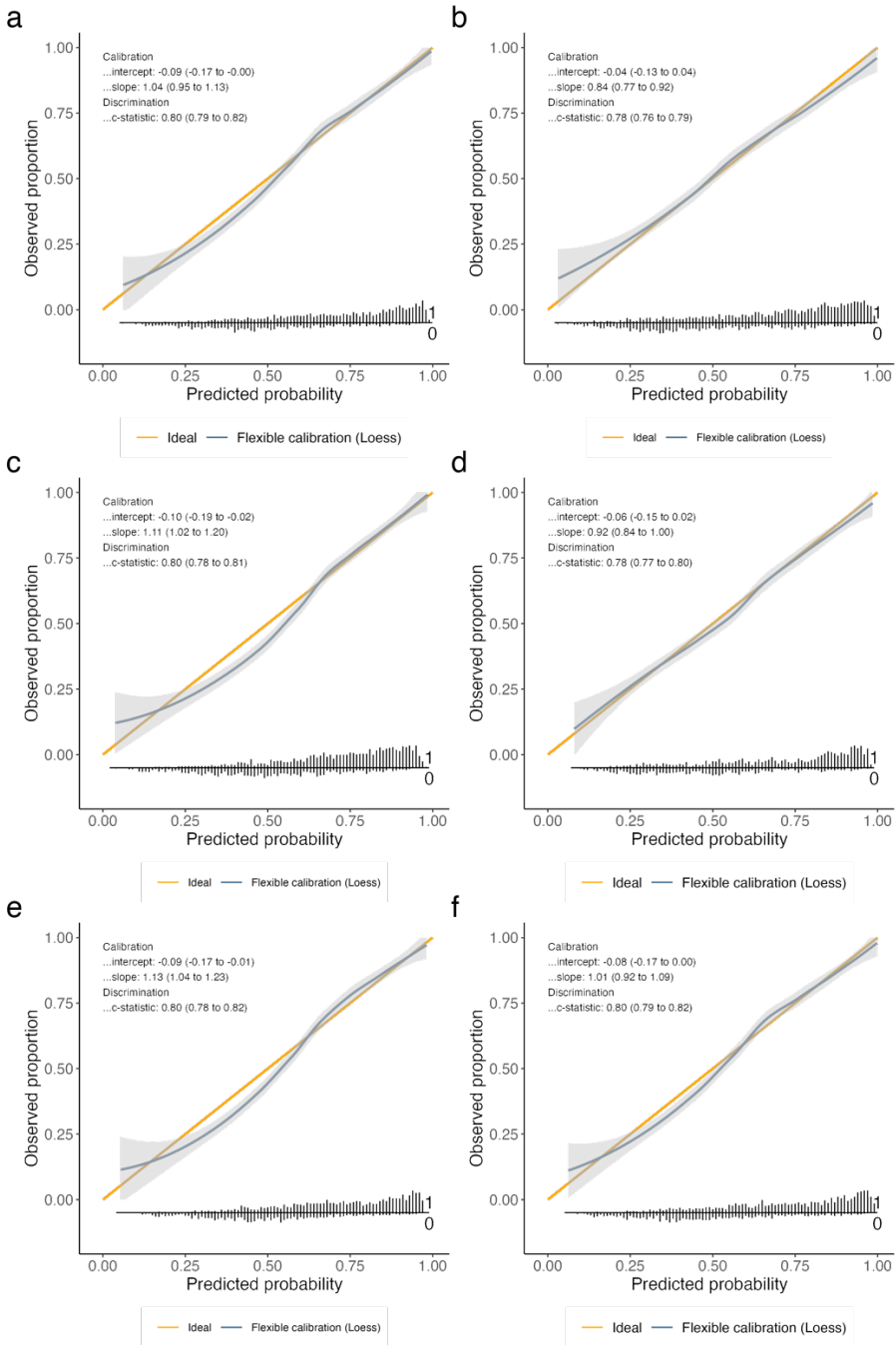

Figure S2.9.1. Internal validation on the IST test dataset: calibration plot of predicted outcome in the control group. The orange line indicates ideal calibration. The outcome variable is death or dependency at 6 months. **a**, T-learner Logistic Regression. **b**, T-learner Random Forest. **c**, T-learner Support Vector Machine. **d**, T-learner XGBoost. **e**, T-learner Penalized Logistic Regression. **f**, S-learner Logistic Regression. Abbreviations: IST: the International Stroke Trial.

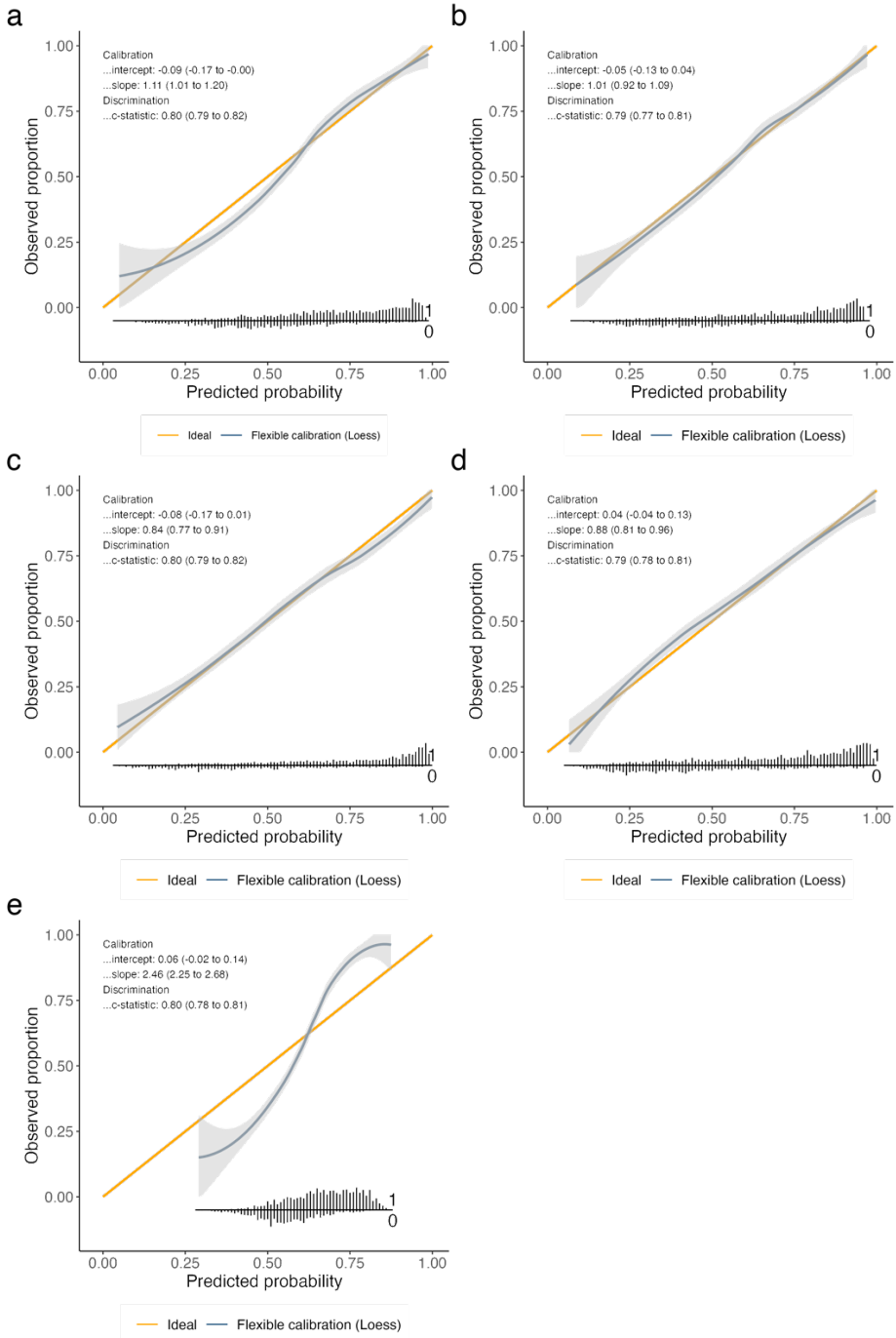

Figure S2.9.2. Internal validation on the IST test dataset: calibration plot of predicted outcome in the control group. The orange line indicates ideal calibration. The outcome variable is death or dependency at 6 months. **a**, S-learner Penalized Logistic Regression. **b**, S-learner XGBoost **c**, S-learner BART. **d**, CVAE. **e**, GANITE. Abbreviations: IST: the International Stroke Trial.

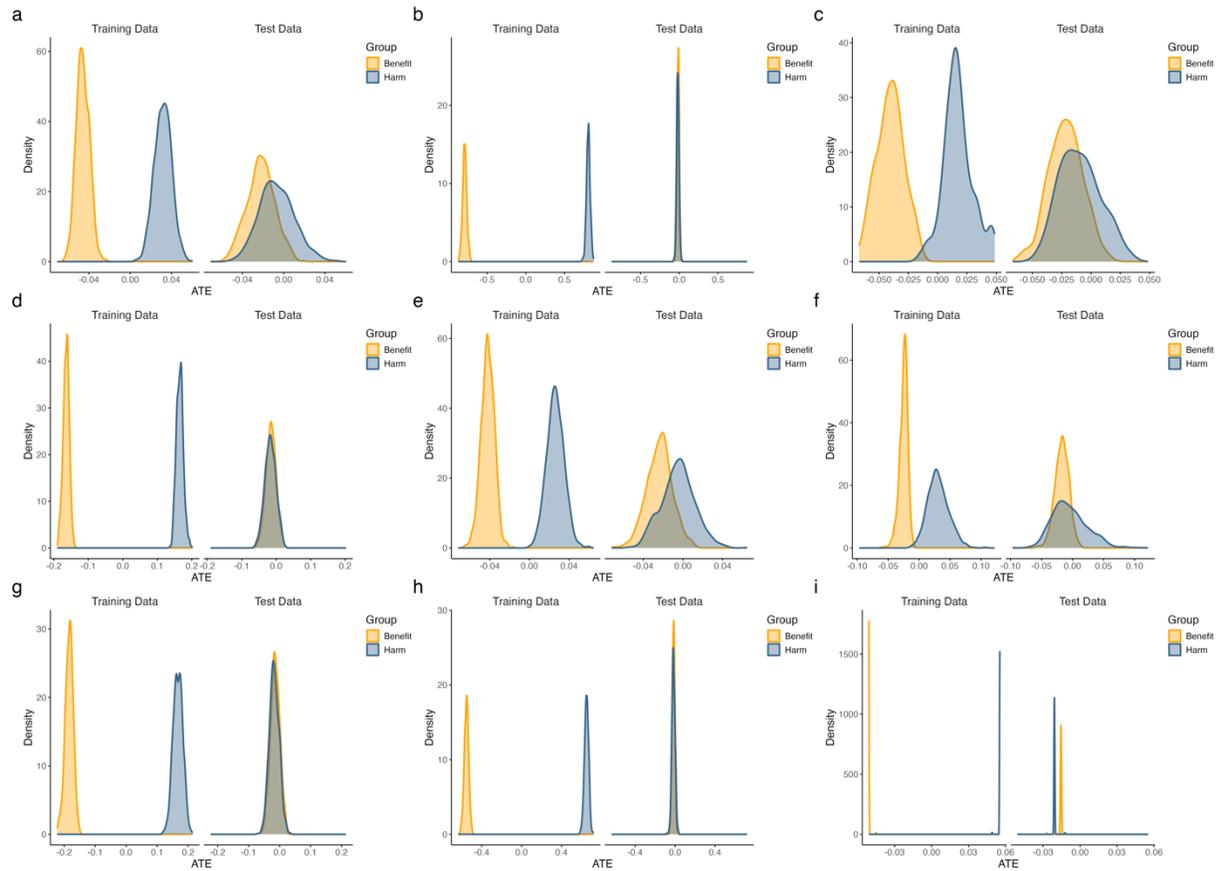

Figure S2.10.1. Internal validation on the IST dataset: density comparative analysis of causal machine learning-based individualized treatment effects. Density plots depict the distributions of ATE in training data and test data stratified by benefit (orange) and harm (blue) groups. The distribution of ATE comes from 1000 (**a, b, d, e, f**) or 100 (**c, i**) random train-test splits experiments. Benefit means negative ITE values and harm means positive ITE values. The outcome variable is death or dependency at 6 months. **a**, T-learner Logistic Regression. **b**, T-learner Random Forest. **c**, T-learner Support Vector Machine. **d**, T-learner XGBoost. **e**, T-learner Penalized Logistic Regression. **f**, S-learner Penalized Logistic Regression. **g**, S-learner XGBoost. **h**, X-learner Random Forest. **i**, X-learner BART. Abbreviations: IST: the International Stroke Trial, ITE: individualized treatment effect, ATE: average treatment effect.

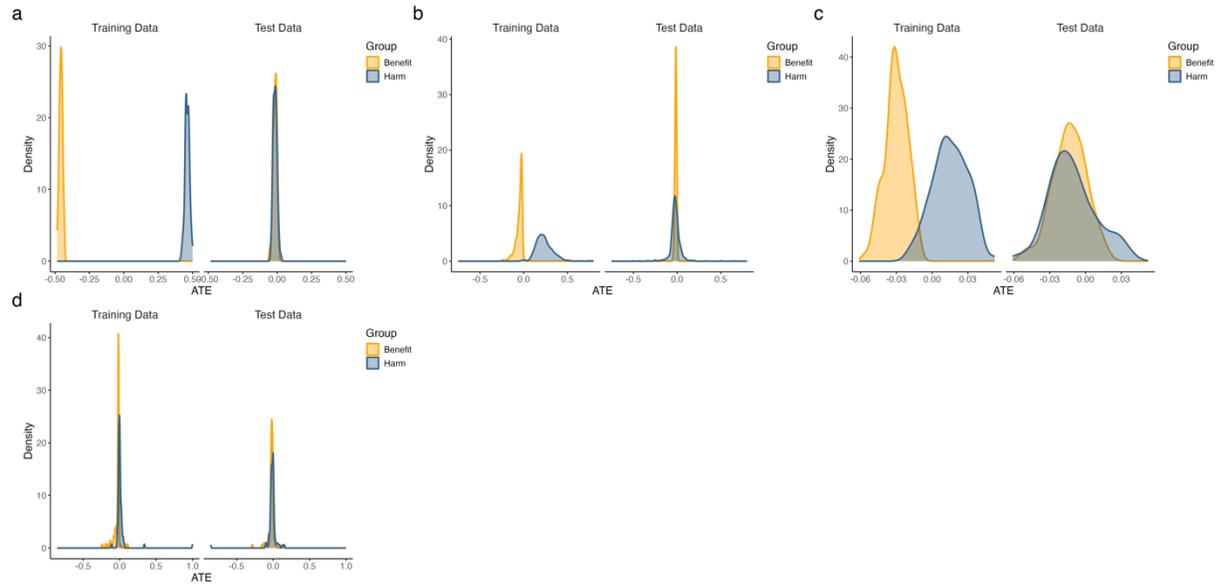

Figure S2.10.2. Internal validation on the IST dataset: density comparative analysis of causal machine learning-based individualized treatment effects. Density plots depict the distributions of ATE in training data and test data stratified by benefit (orange) and harm (blue) groups. The distribution of ATE comes from 1000 (**b**) or 100 (**a, c, d**) random train-test splits experiments. Benefit means negative ITE values and harm means positive ITE values. The outcome variable is death or dependency at 6 months. **a**, DR-learner Random Forest. **b**, Causal Forest. **c**, CVAE. **d**, GANITE. Abbreviations: IST: the International Stroke Trial, ITE: individualized treatment effect, ATE: average treatment effect.

**Results of internal validation on CAST dataset**

Table S1. Summary of quantitative validation metrics on CAST training and test dataset of 17 causal machine learning methods with 4 weeks' death or dependency outcome.

|  |  | c-for-benefit (training) | c-for-benefit (test) | mbcb (training) | mbcb (test) | Calibration-based pseudo R-squared (training) | Calibration-based pseudo R-squared (test) |
|---|---|---|---|---|---|---|---|
| **T-learner** | Logistic Regression | 0.539 | 0.505 | 0.542 | 0.542 | 0.925 | -1.102 |
|  | Random Forest | 0.963 | 0.494 | 0.714 | 0.633 | 0.255 | -3.034 |
|  | Support Vector Machine | 0.512 | 0.495 | 0.578 | 0.577 | -3.525 | -5.675 |
|  | XGBoost | 0.649 | 0.500 | 0.600 | 0.600 | 0.590 | -4.154 |
|  | Penalized Logistic Regression | 0.529 | 0.514 | 0.533 | 0.533 | 0.780 | 0.083 |
| **S-learner** | Logistic Regression | 0.502 | 0.494 | 0.505 | 0.505 | 0.125 | -0.088 |
|  | Penalized Logistic Regression | 0.520 | 0.504 | 0.517 | 0.517 | 0.611 | 0.077 |
|  | XGBoost | 0.660 | 0.499 | 0.558 | 0.556 | 0.322 | -4.402 |
|  | BART | 0.538 | 0.506 | 0.521 | 0.521 | 0.367 | -0.758 |
| **X-learner** | Random Forest | 0.901 | 0.502 | - | - | 0.216 | -18.731 |
|  | BART | 0.549 | 0.506 | - | - | 0.592 | -2.660 |
| **DR-learner** | Random Forest | 0.837 | 0.492 | - | - | 0.638 | -8.926 |
| **Tree-based methods** | Causal Forest | 0.569 | 0.504 | - | - | 0.112 | -0.065 |
|  | Bayesian Causal Forest | 0.526 | 0.509 | - | - | 0.092 | 0.014 |
|  | model-based recursive partitioning | 0.757 | 0.510 | - | - | - | - |
| **Deep Learning** | CVAE | 0.530 | 0.496 | 0.533 | 0.532 | 0.540 | -1.103 |
|  | GANITE | 0.519 | 0.492 | 0.525 | 0.525 | 0.290 | -1.243 |

Notes: Consistent values across training and test data that are closer to 1 indicate a better model fit. The dash symbol indicates instances where:

- The validation metric could not be assessed due to the model's mechanism.
- The metric was not applicable to the model.

Abbreviations: CAST: the Chinese Acute Stroke Trial.

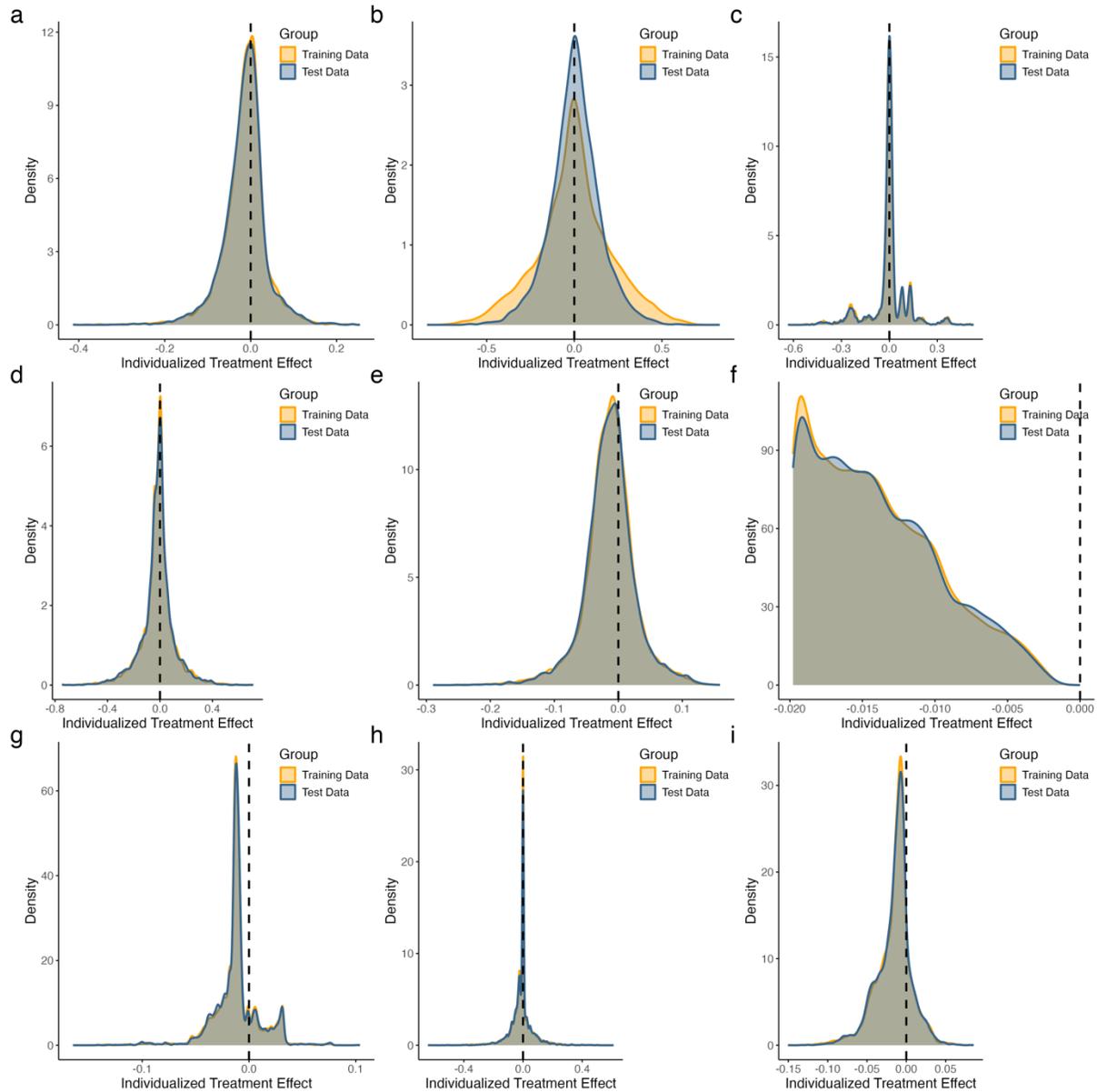

Figure S3.1.1. Internal validation on the CAST dataset: density plots of causal machine learning-based individualized treatment effects. Orange represents the training data, and blue indicates the test data. The outcome variable is death or dependency at 4 weeks. **a**, T-learner Logistic Regression. **b**, T-learner Random Forest. **c**, T-learner Support Vector Machine. **d**, T-learner XGBoost. **e**, T-learner Penalized Logistic Regression. **f**, S-learner Logistic Regression. **g**, S-learner Penalized Logistic Regression. **h**, S-learner XGBoost. **i**, S-learner BART. Abbreviations: CAST: the Chinese Acute Stroke Trial, ITE: individualized treatment effect.

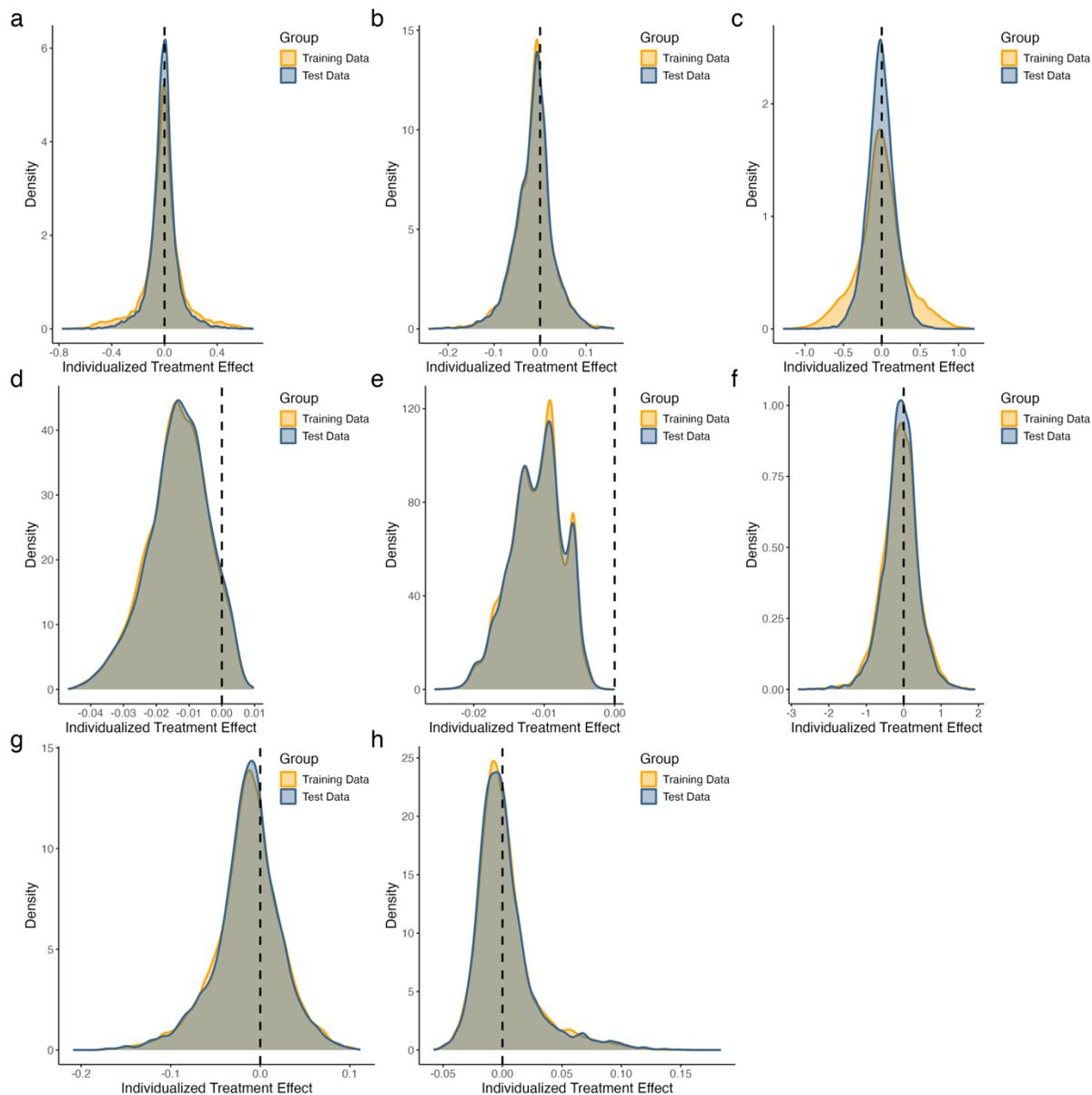

Figure S3.1.2. Internal validation on the CAST dataset: density plots of causal machine learning-based individualized treatment effects. Orange represents the training data, and blue indicates the test data. The outcome variable is death or dependency at 4 weeks. **a**, X-learner Random Forest. **b**, X-learner BART. **c**, DR-learner Random Forest. **d**, Causal Forest. **e**, Bayesian Causal Forest. **f**, Model-based Recursive Partitioning. **g**, CVAE. **h**, GANITE. Abbreviations: CAST: the Chinese Acute Stroke Trial, ITE: individualized treatment effect.

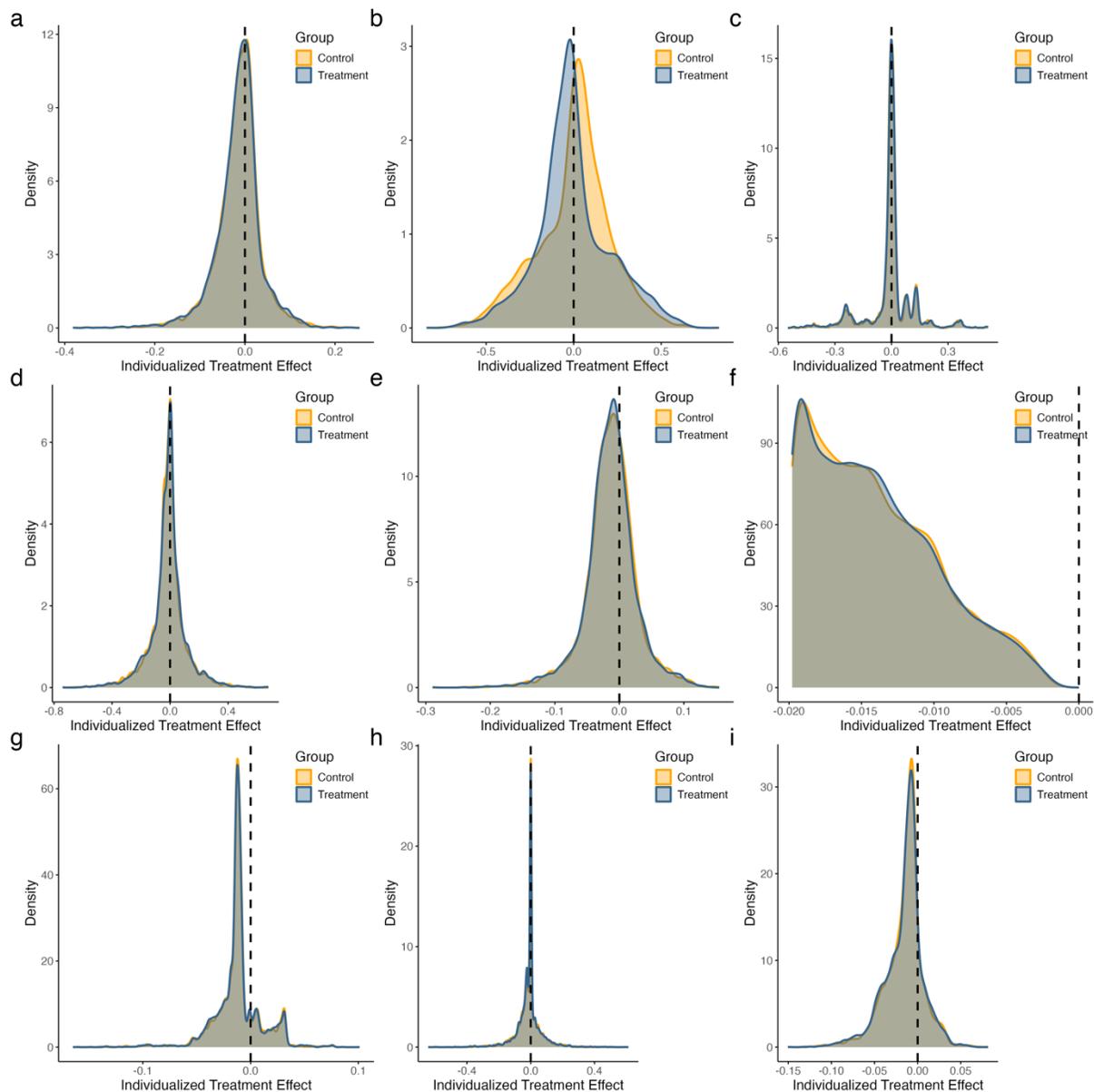

Figure S3.2.1. Internal validation on the CAST training dataset: density plots of causal machine learning-based individualized treatment effects. Orange represents the control group, and blue indicates the treatment group. The outcome variable is death or dependency at 4 weeks. **a**, T-learner Logistic Regression. **b**, T-learner Random Forest. **c**, T-learner Support Vector Machine. **d**, T-learner XGBoost. **e**, T-learner Penalized Logistic Regression. **f**, S-learner Logistic Regression. **g**, S-learner Penalized Logistic Regression. **h**, S-learner XGBoost. **i**, S-learner BART. Abbreviations: CAST: the Chinese Acute Stroke Trial, ITE: individualized treatment effect.

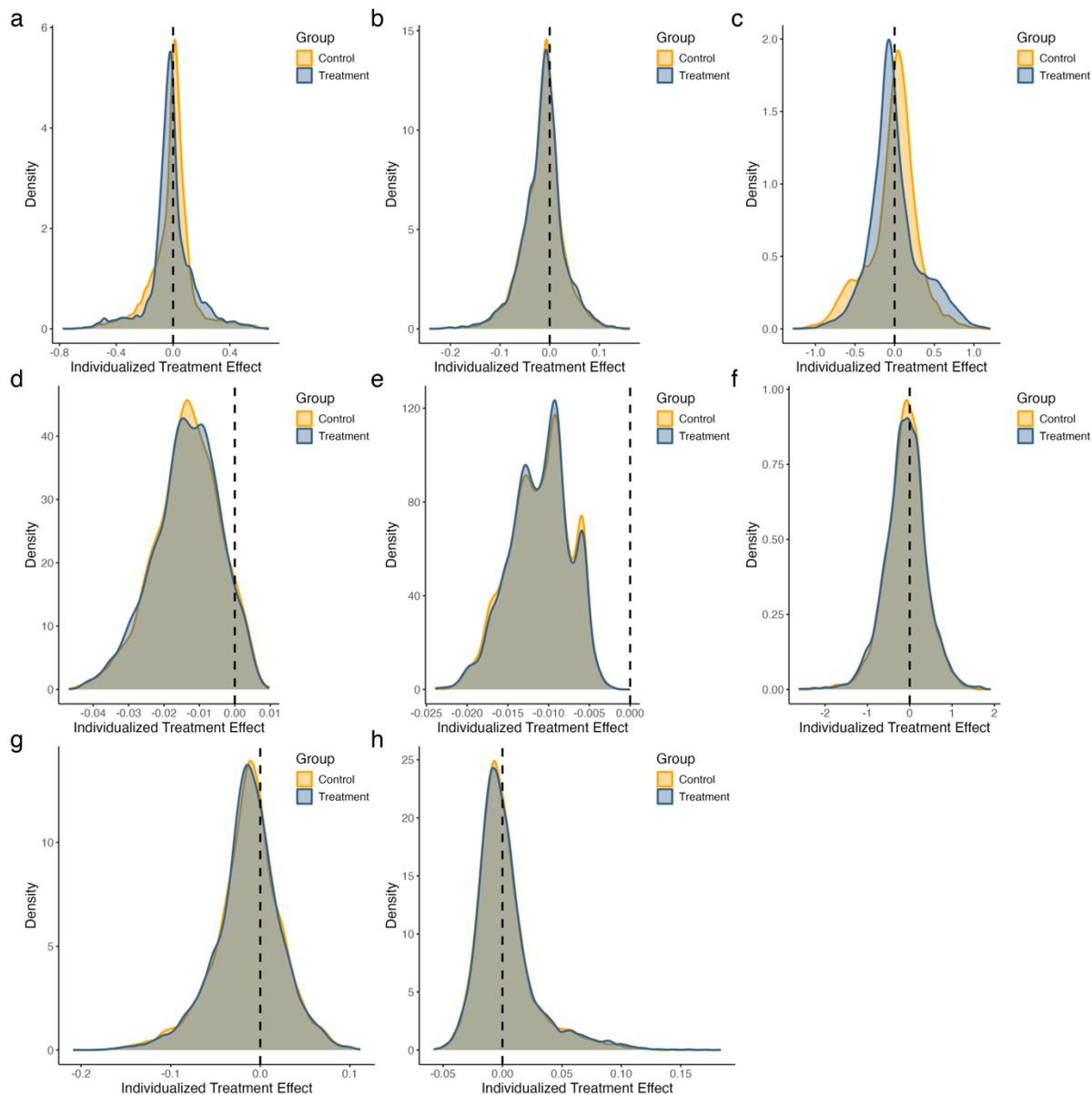

Figure S3.2.2. Internal validation on the CAST training dataset: density plots of causal machine learning-based individualized treatment effects. Orange represents the control group, and blue indicates the treatment group. The outcome variable is death or dependency at 4 weeks. **a**, X-learner Random Forest. **b**, X-learner BART. **c**, DR-learner Random Forest. **d**, Causal Forest. **e**, Bayesian Causal Forest. **f**, Model-based Recursive Partitioning. **g**, CVAE. **h**, GANITE. Abbreviations: CAST: the Chinese Acute Stroke Trial, ITE: individualized treatment effect.

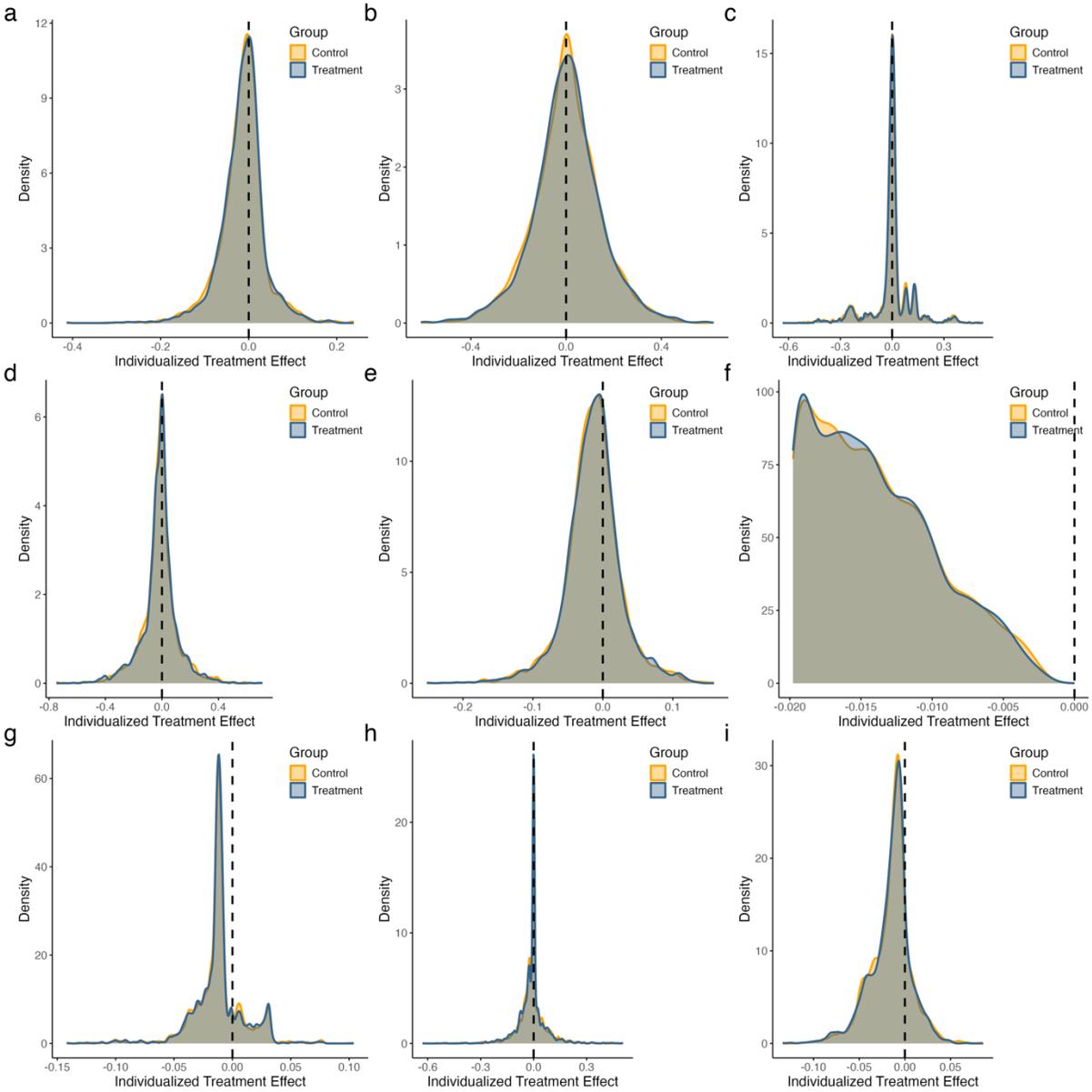

Figure S3.3.1. Internal validation on the CAST test dataset: density plots of causal machine learning-based individualized treatment effects. Orange represents the control group, and blue indicates the treatment group. The outcome variable is death or dependency at 4 weeks. **a**, T-learner Logistic Regression. **b**, T-learner Random Forest. **c**, T-learner Support Vector Machine. **d**, T-learner XGBoost. **e**, T-learner Penalized Logistic Regression. **f**, S-learner Logistic Regression. **g**, S-learner Penalized Logistic Regression. **h**, S-learner XGBoost. **i**, S-learner BART. Abbreviations: CAST: the Chinese Acute Stroke Trial, ITE: individualized treatment effect.

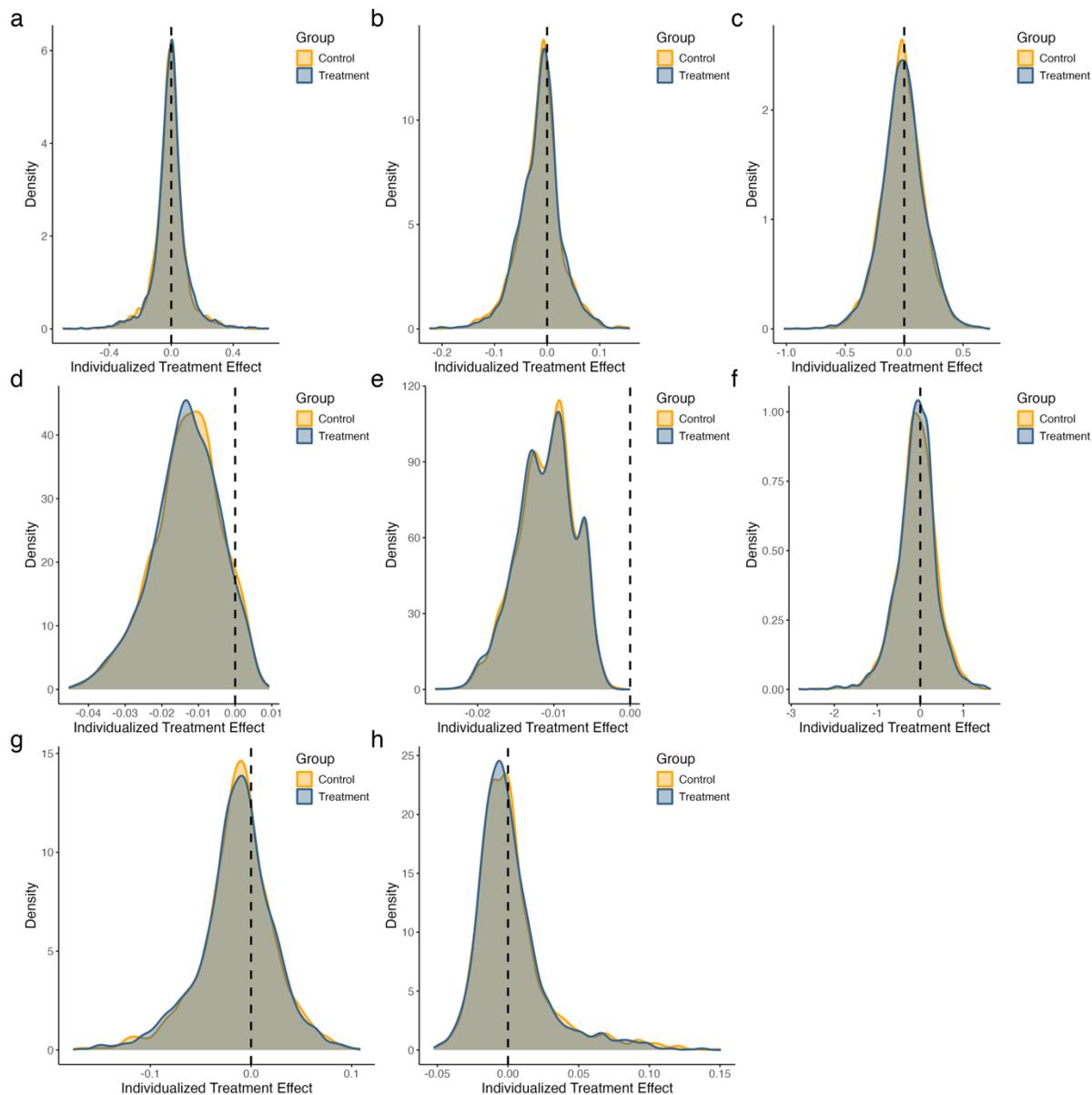

Figure S3.3.2. Internal validation on the CAST test dataset: density plots of causal machine learning-based individualized treatment effects. Orange represents the control group, and blue indicates the treatment group. The outcome variable is death or dependency at 4 weeks. **a**, X-learner Random Forest. **b**, X-learner BART. **c**, DR-learner Random Forest. **d**, Causal Forest. **e**, Bayesian Causal Forest. **f**, Model-based Recursive Partitioning. **g**, CVAE. **h**, GANITE. Abbreviations: CAST: the Chinese Acute Stroke Trial, ITE: individualized treatment effect.

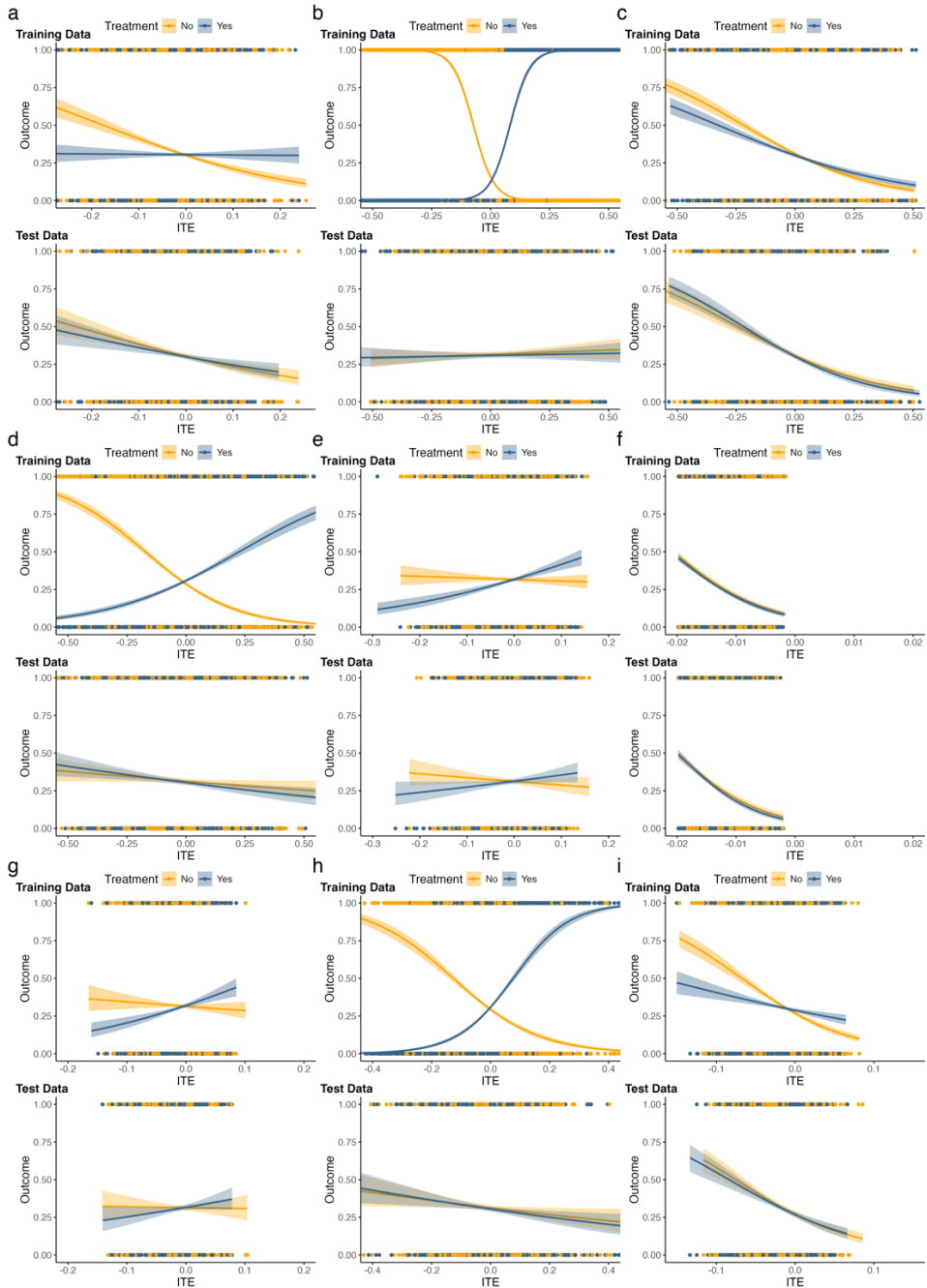

Figure S3.4.1. Internal validation on the CAST dataset: outcome-ITE comparative analysis. Dot plots and line plots depict the true and fitted patient outcomes against estimated ITE values between training data and test data. Orange represents the control group, and blue indicates the treatment group. The outcome variable is death or dependency at 4 weeks. **a**, T-learner Logistic Regression. **b**, T-learner Random Forest. **c**, T-learner Support Vector Machine. **d**, T-learner XGBoost. **e**, T-learner Penalized Logistic Regression. **f**, S-learner Logistic Regression. **g**, S-learner Penalized Logistic Regression. **h**, S-learner XGBoost. **i**, S-learner BART. Abbreviations: CAST: the Chinese Acute Stroke Trial, ITE: individualized treatment effect.

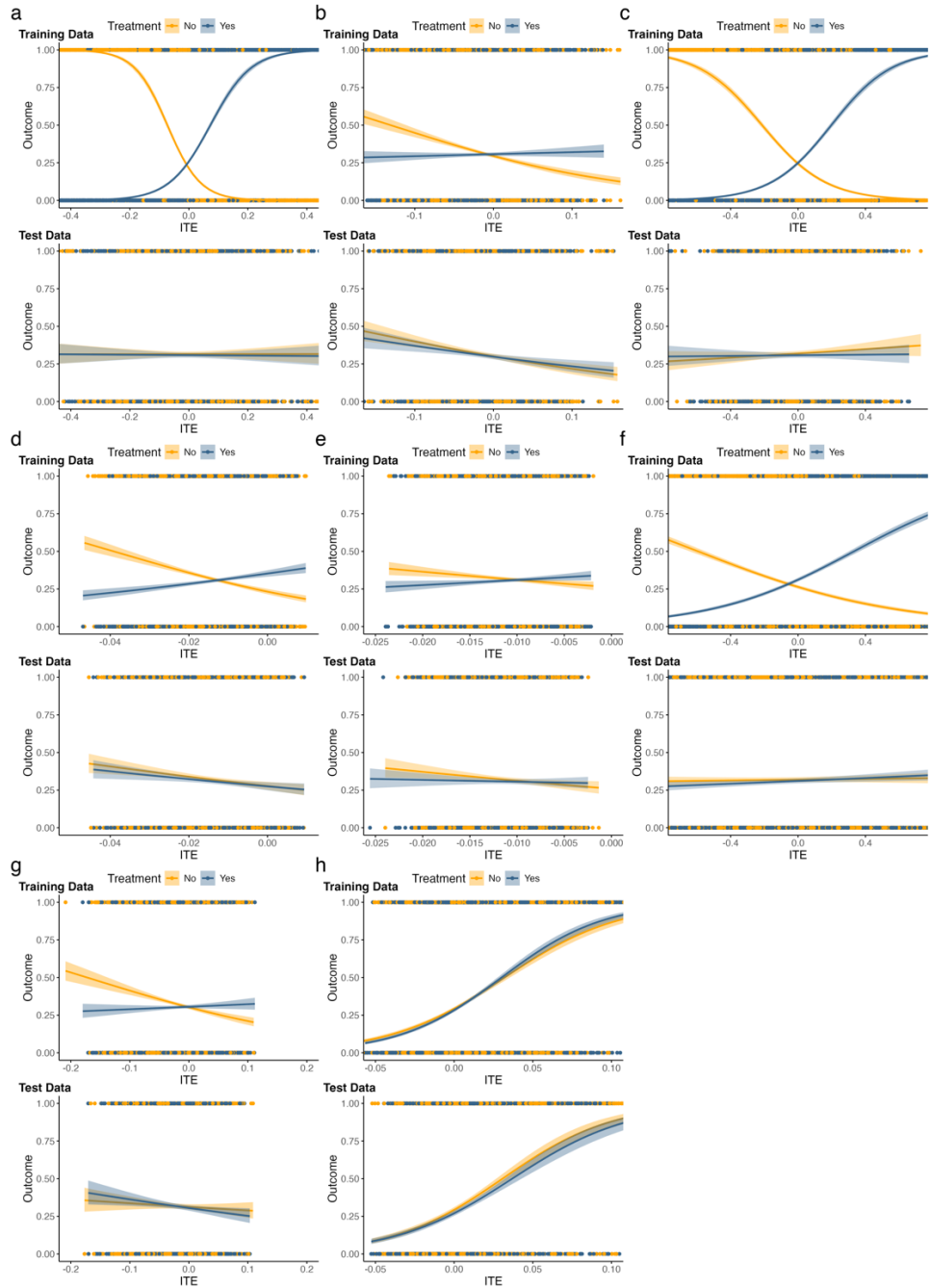

Figure S3.4.2. Internal validation on the CAST dataset: outcome-ITE comparative analysis. Dot plots and line plots depict the true and fitted patient outcomes against estimated ITE values between training data and test data. Orange represents the control group, and blue indicates the treatment group. The outcome variable is death or dependency at 4 weeks. **a**, X-learner Random Forest. **b**, X-learner BART. **c**, DR-learner Random Forest. **d**, Causal Forest. **e**, Bayesian Causal Forest. **f**, Model-based Recursive Partitioning. **g**, CVAE. **h**, GANITE. Abbreviations: CAST: the Chinese Acute Stroke Trial, ITE: individualized treatment effect.

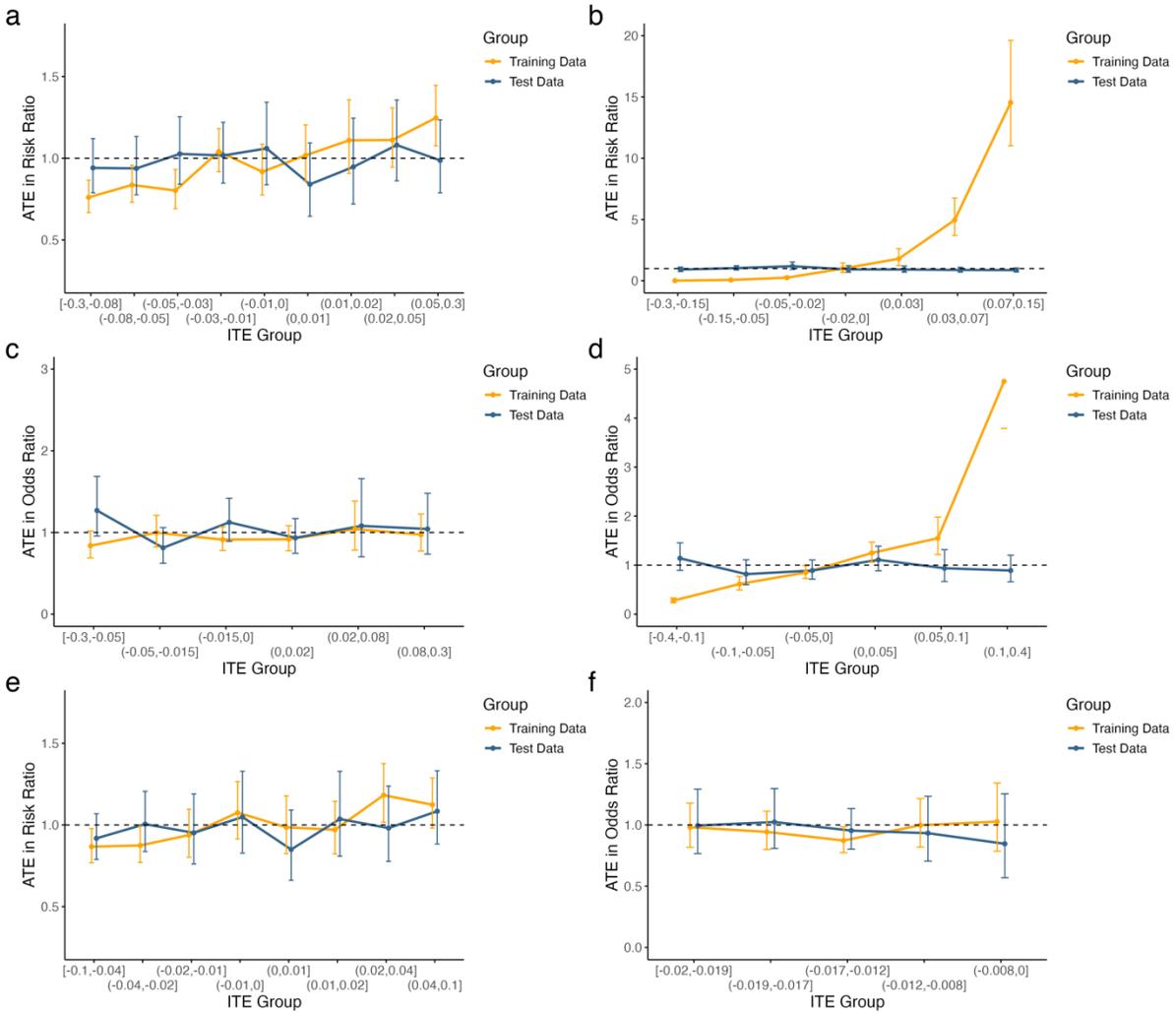

Figure S3.5.1. Internal validation on the CAST dataset: ATE-ITE comparative analysis. Line plots depict ATE in risk ratio within different ITE subgroups and provide the confidence intervals at 95% level. Orange represents the training data, and blue indicates the test data. The horizontal dashed line at 1.0 means no treatment effects. The outcome variable is death or dependency at 4 weeks. **a**, T-learner Logistic Regression. **b**, T-learner Random Forest. **c**, T-learner Support Vector Machine. **d**, T-learner XGBoost. **e**, T-learner Penalized Logistic Regression. **f**, S-learner Logistic Regression. Abbreviations: CAST: the Chinese Acute Stroke Trial, ITE: individualized treatment effect, ATE: average treatment effect.

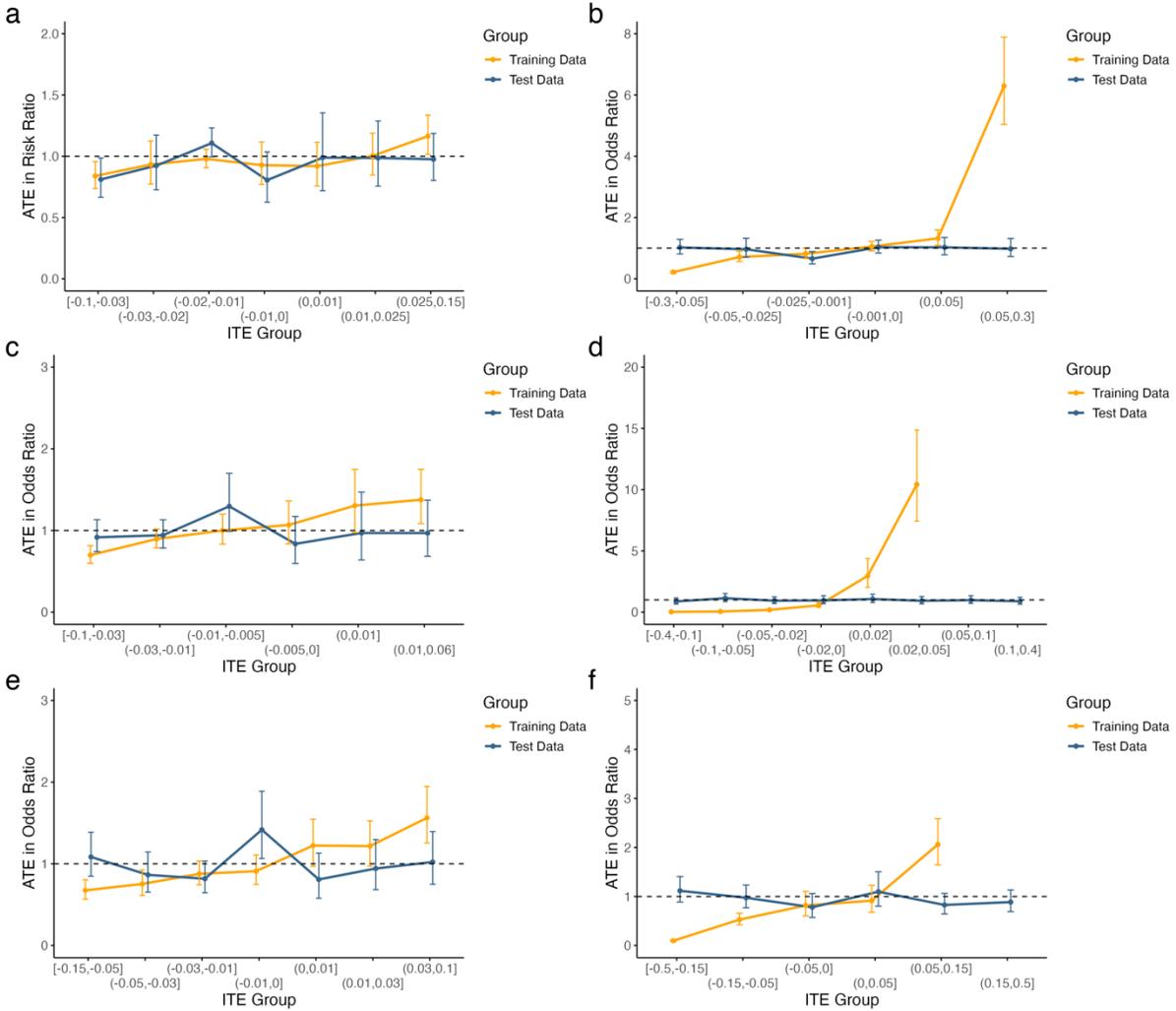

Figure S3.5.2. Internal validation on the CAST dataset: ATE-ITE comparative analysis. Line plots depict ATE in risk ratio within different ITE subgroups and provide the confidence intervals at 95% level. Orange represents the training data, and blue indicates the test data. The horizontal dashed line at 1.0 means no treatment effects. The outcome variable is death or dependency at 4 weeks. **a**, S-learner Penalized Logistic Regression. **b**, S-learner XGBoost. **c**, S-learner BART. **d**, X-learner Random Forest. **e**, X-learner BART. **f**, DR-learner Random Forest. Abbreviations: CAST: the Chinese Acute Stroke Trial, ITE: individualized treatment effect, ATE: average treatment effect.

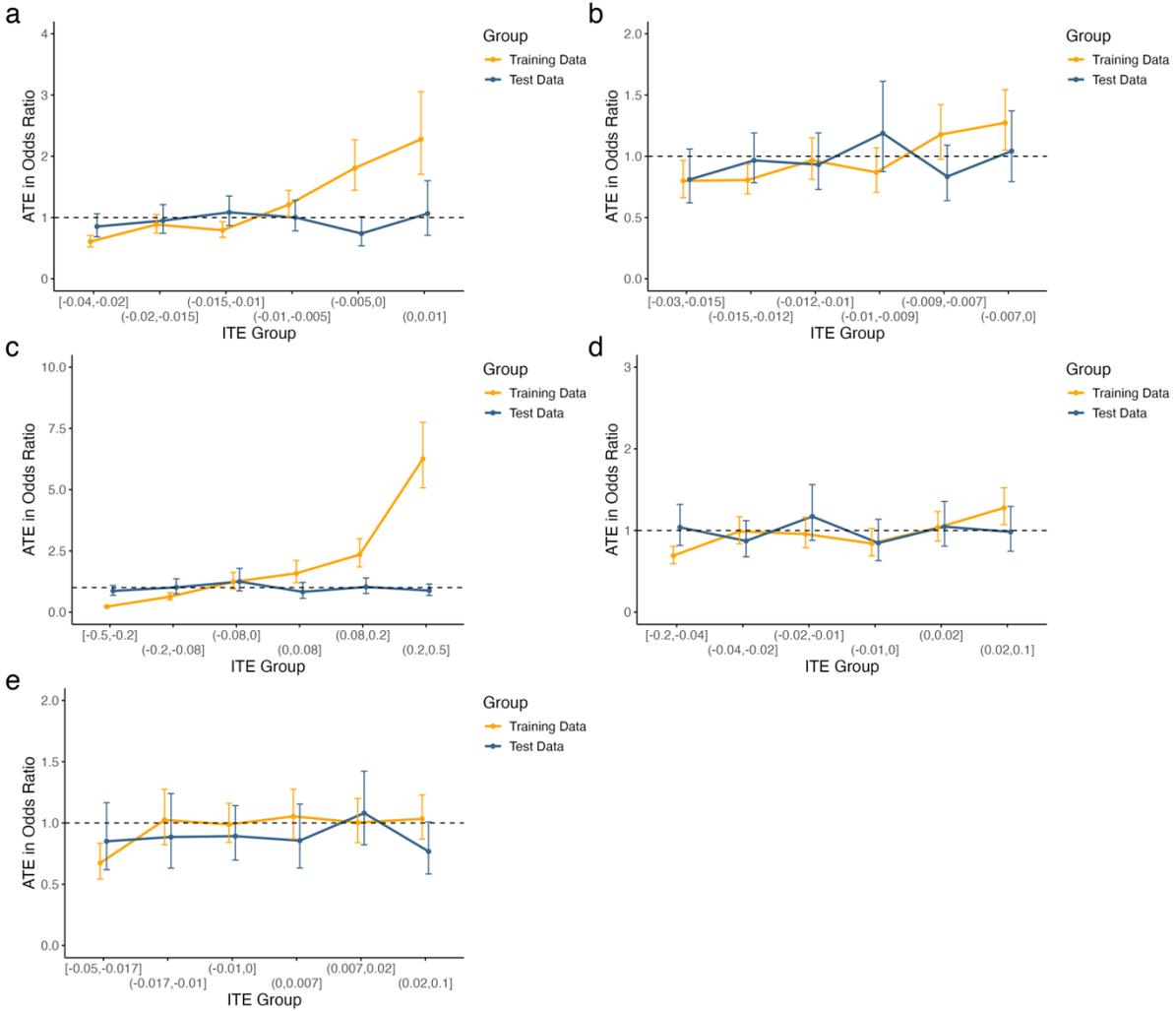

Figure S3.5.2. Internal validation on the CAST dataset: ATE-ITE comparative analysis. Line plots depict ATE in risk ratio within different ITE subgroups and provide the confidence intervals at 95% level. Orange represents the training data, and blue indicates the test data. The horizontal dashed line at 1.0 means no treatment effects. The outcome variable is death or dependency at 4 weeks. **a**, Causal Forest. **b**, Bayesian Causal Forest. **c**, Model-based Recursive Partitioning. **d**, CVAE. **e**, GANITE. Abbreviations: CAST: the Chinese Acute Stroke Trial, ITE: individualized treatment effect, ATE: average treatment effect.

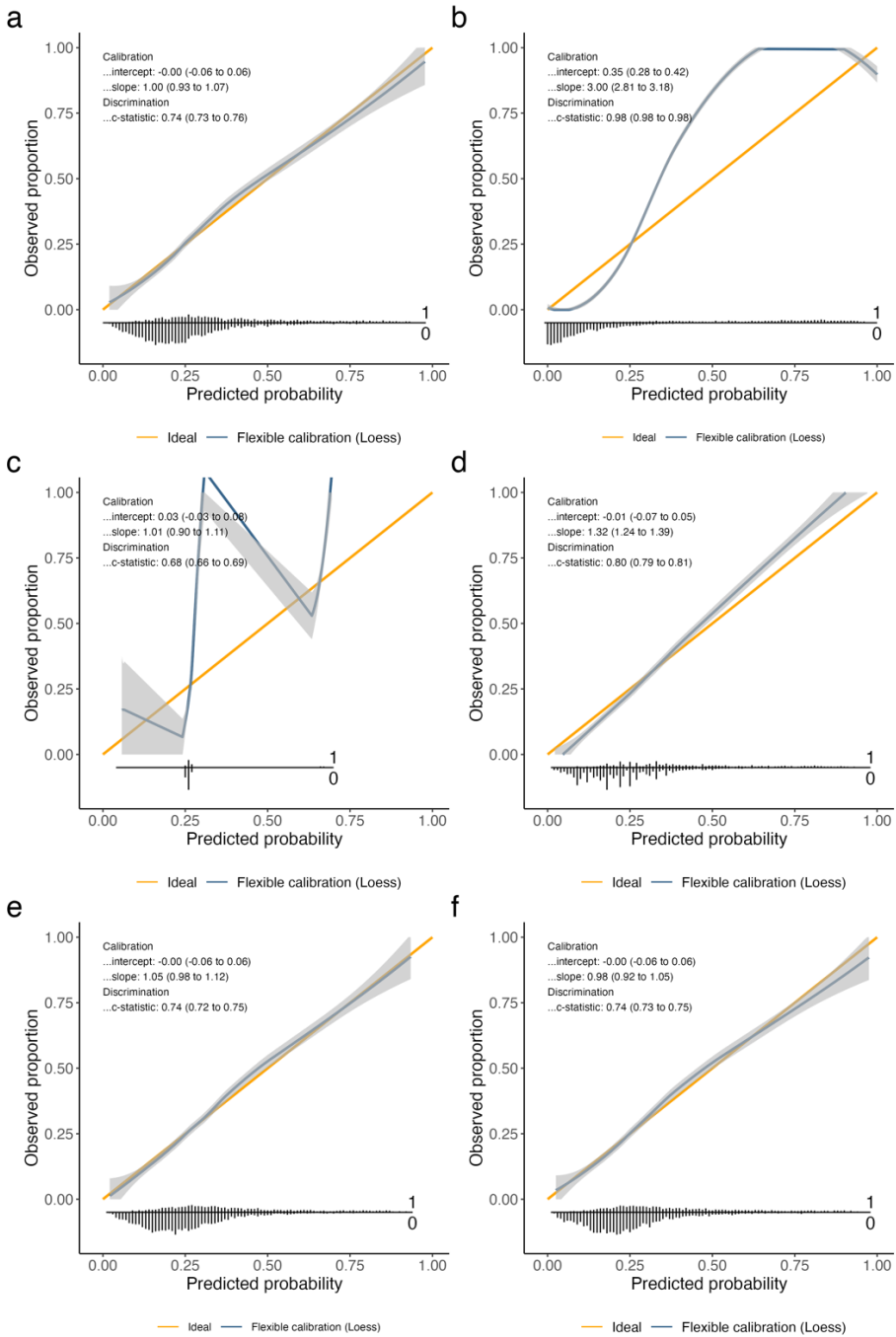

Figure S3.6.1. Internal validation on the CAST training dataset: calibration plot of predicted outcome in the treatment group. The orange line indicates ideal calibration. The outcome variable is death or dependency at 4 weeks. **a**, T-learner Logistic Regression. **b**, T-learner Random Forest. **c**, T-learner Support Vector Machine. **d**, T-learner XGBoost. **e**, T-learner Penalized Logistic Regression. **f**, S-learner Logistic Regression. Abbreviations: CAST: the Chinese Acute Stroke Trials.

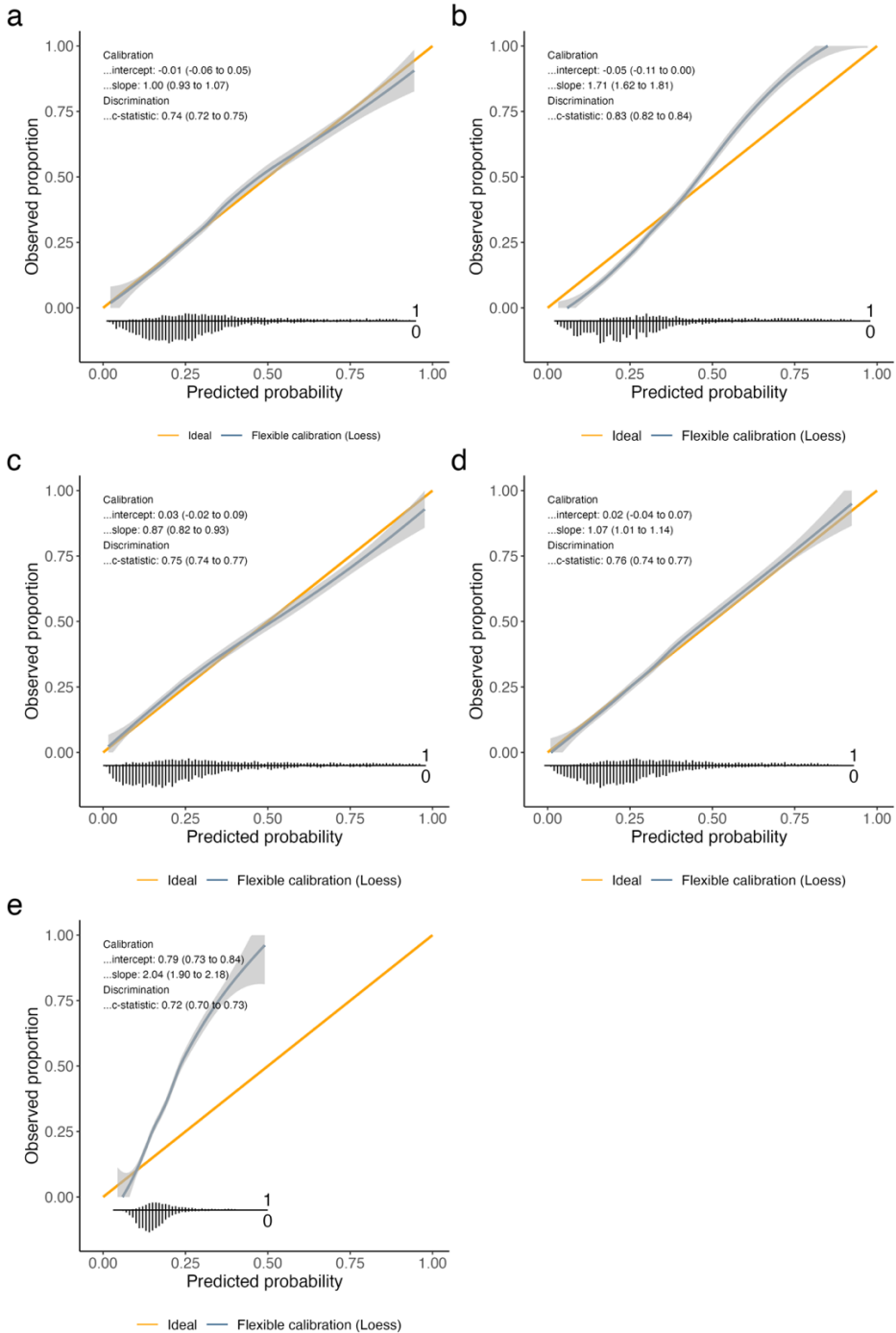

Figure S3.6.2. Internal validation on the CAST training dataset: calibration plot of predicted outcome in the treatment group. The orange line indicates ideal calibration. The outcome variable is death or dependency at 4 weeks. **a**, S-learner Penalized Logistic Regression. **b**, S-learner XGBoost **c**, S-learner BART. **d**, CVAE. **e**, GANITE. Abbreviations: CAST: the Chinese Acute Stroke Trial.

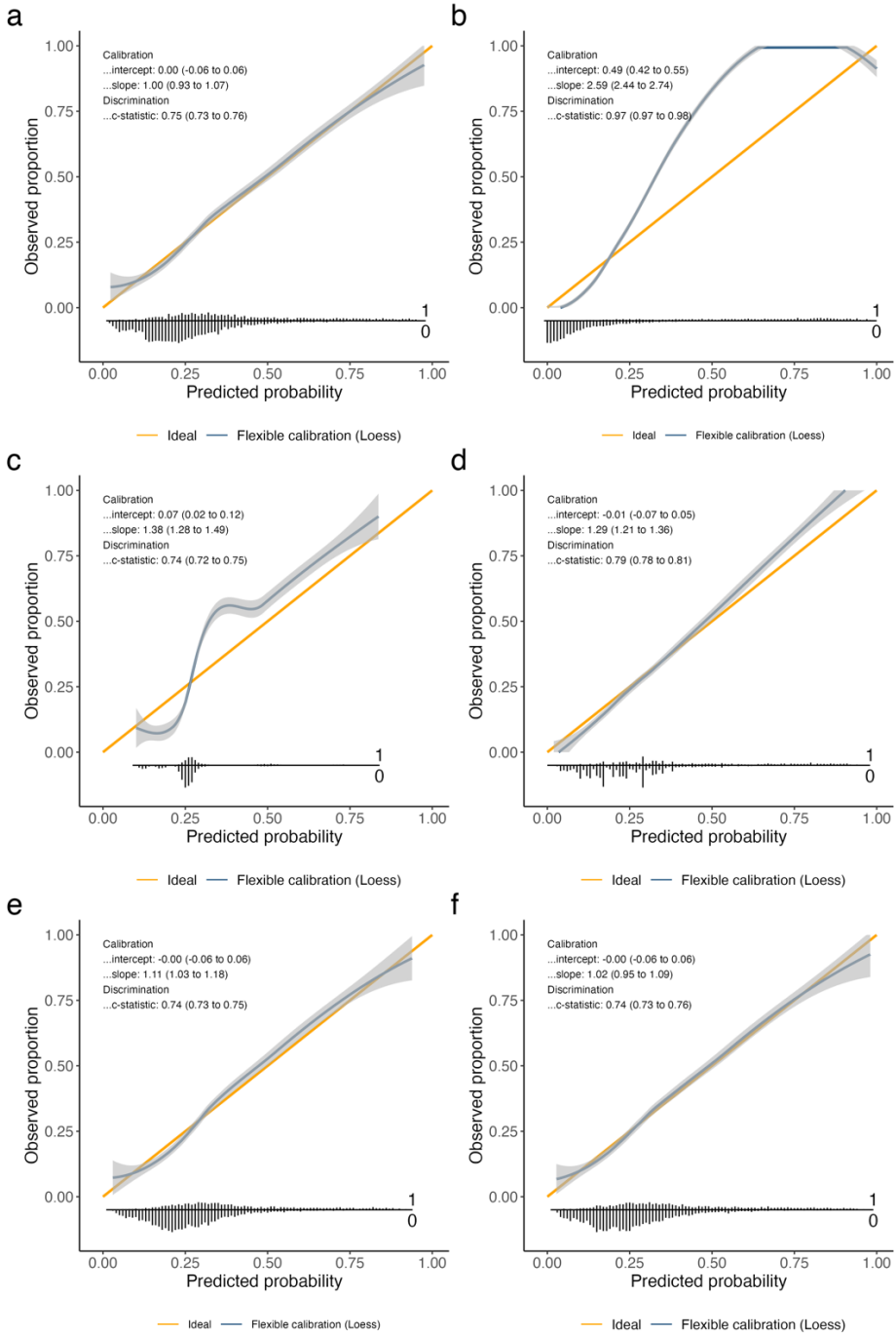

Figure S3.7.1. Internal validation on the CAST training dataset: calibration plot of predicted outcome in the control group. The orange line indicates ideal calibration. The outcome variable is death or dependency at 4 weeks. **a**, T-learner Logistic Regression. **b**, T-learner Random Forest. **c**, T-learner Support Vector Machine. **d**, T-learner XGBoost. **e**, T-learner Penalized Logistic Regression. **f**, S-learner Logistic Regression. Abbreviations: CAST: the Chinese Acute Stroke Trial.

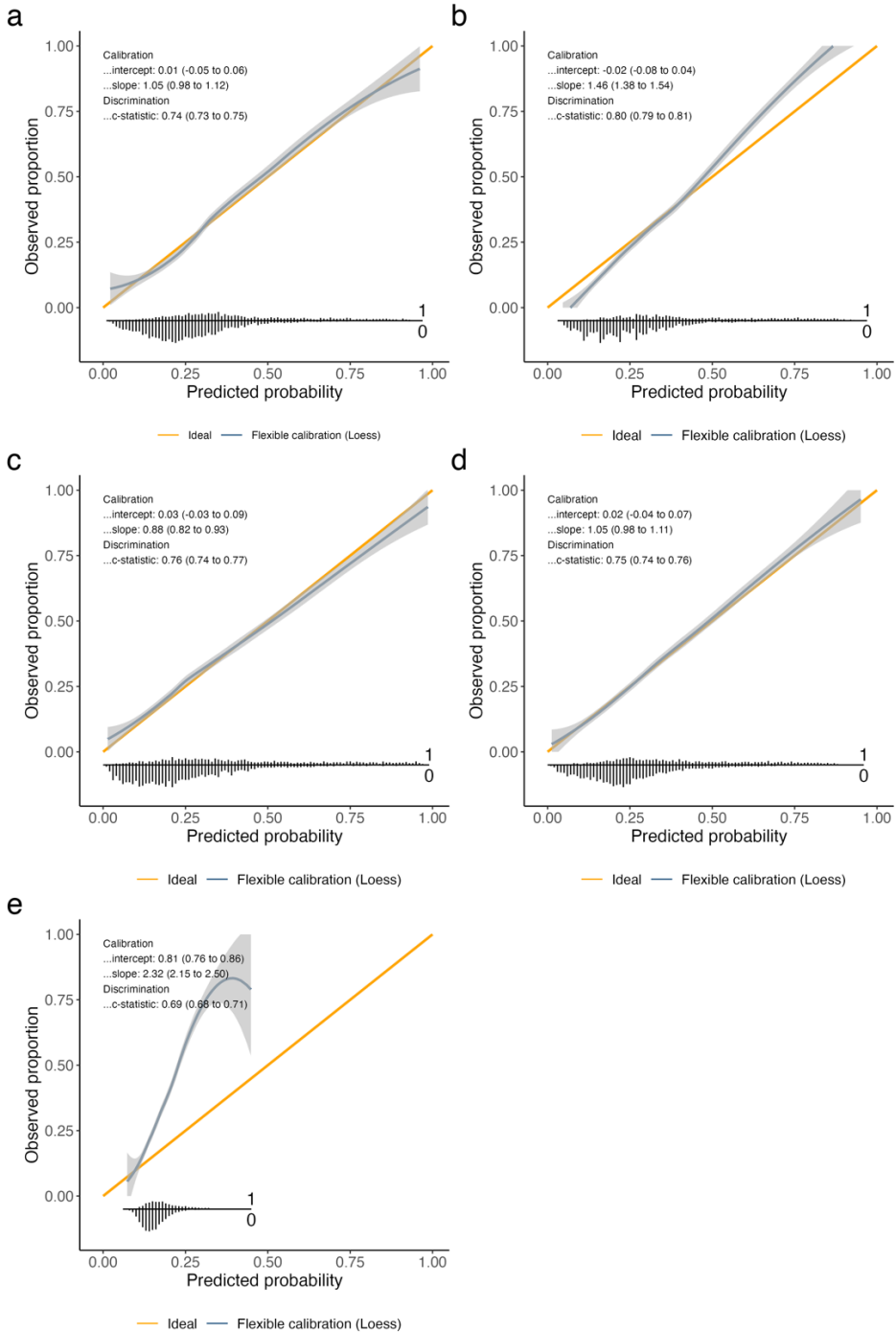

Figure S3.7.2. Internal validation on the CAST training dataset: calibration plot of predicted outcome in the control group. The orange line indicates ideal calibration. The outcome variable is death or dependency at 4 weeks. **a**, S-learner Penalized Logistic Regression. **b**, S-learner XGBoost **c**, S-learner BART. **d**, CVAE. **e**, GANITE. Abbreviations: CAST: the Chinese Acute Stroke Trial.

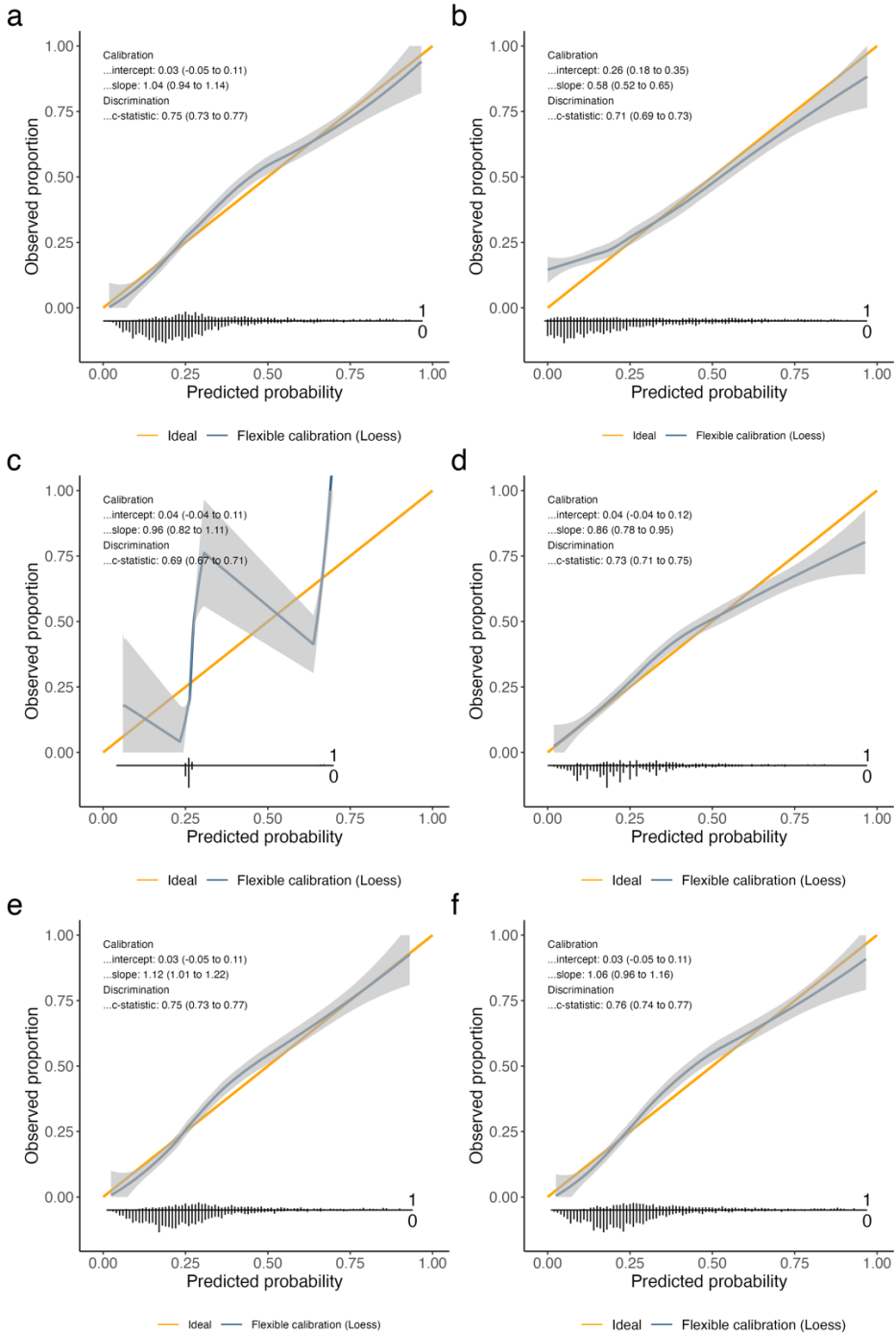

Figure S3.8.1. Internal validation on the CAST test dataset: calibration plot of predicted outcome in the treatment group. The orange line indicates ideal calibration. The outcome variable is death or dependency at 4 weeks. **a**, T-learner Logistic Regression. **b**, T-learner Random Forest. **c**, T-learner Support Vector Machine. **d**, T-learner XGBoost. **e**, T-learner Penalized Logistic Regression. **f**, S-learner Logistic Regression. Abbreviations: CAST: the Chinese Acute Stroke Trial.

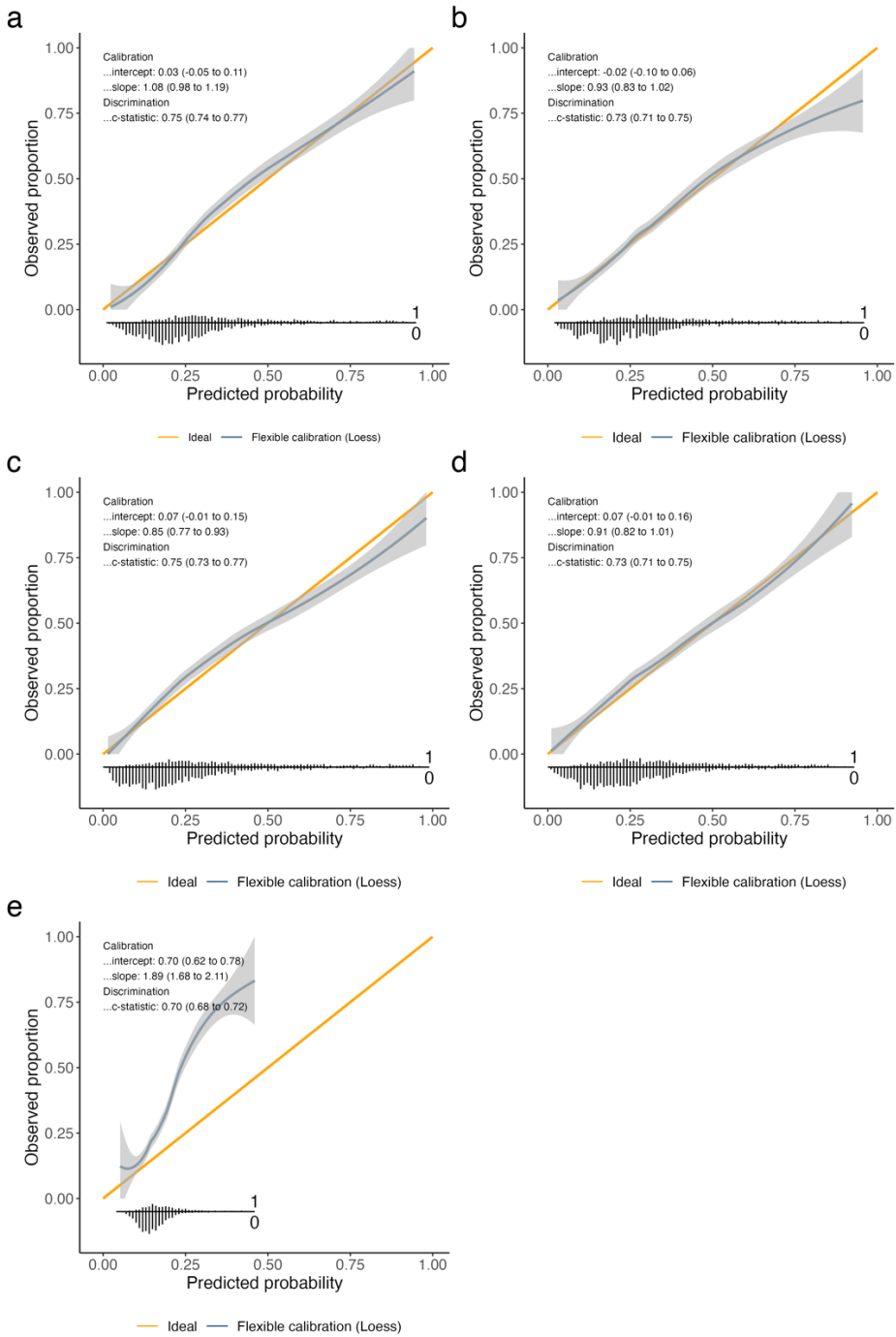

Figure S3.8.2. Internal validation on the CAST test dataset: calibration plot of predicted outcome in the treatment group. The orange line indicates ideal calibration. The outcome variable is death or dependency at 4 weeks. **a**, S-learner Penalized Logistic Regression. **b**, S-learner XGBoost **c**, S-learner BART. **d**, CVAE. **e**, GANITE. Abbreviations: CAST: the Chinese Acute Stroke Trial.

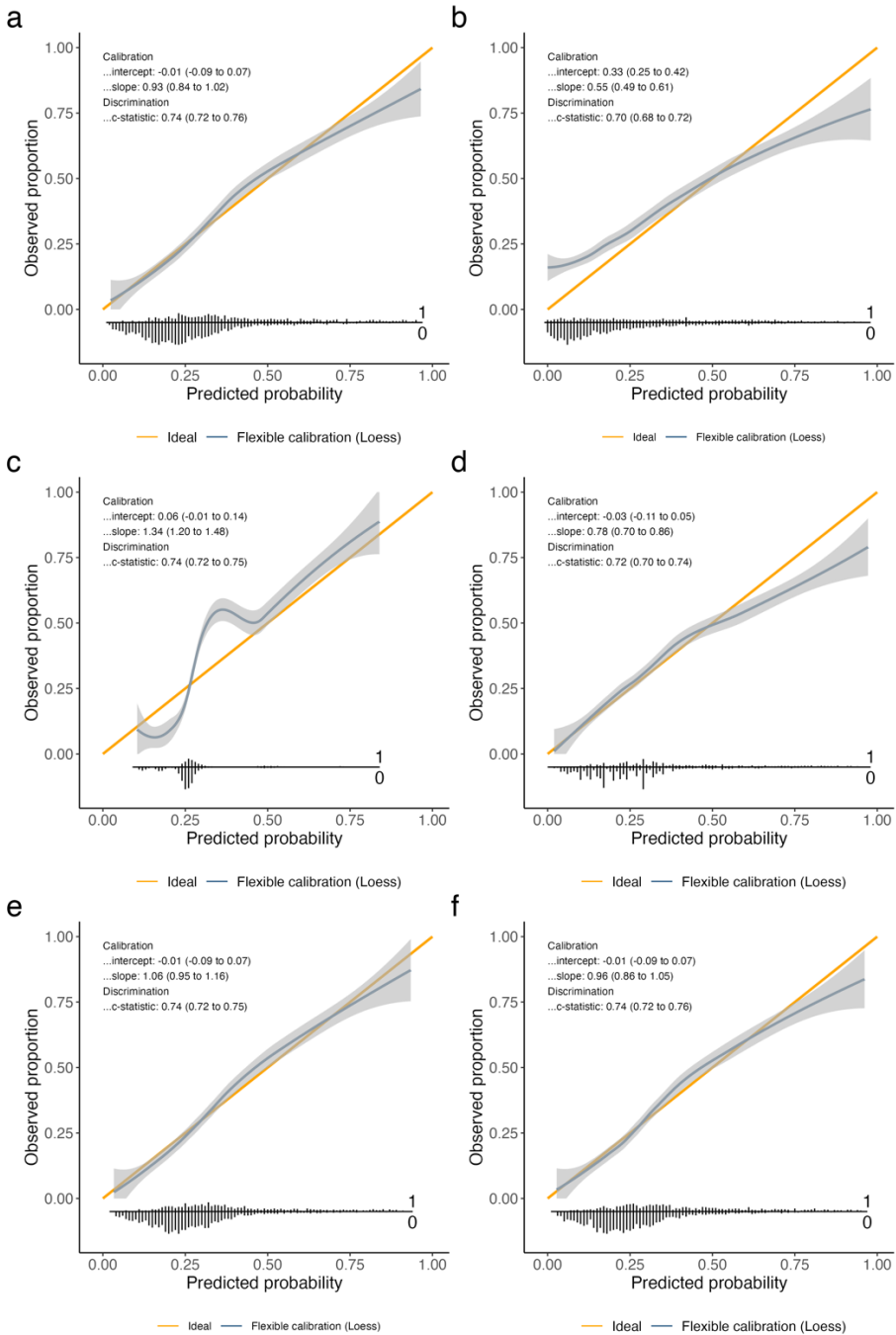

Figure S3.9.1. Internal validation on the CAST test dataset: calibration plot of predicted outcome in the control group. The orange line indicates ideal calibration. The outcome variable is death or dependency at 4 weeks. **a**, T-learner Logistic Regression. **b**, T-learner Random Forest. **c**, T-learner Support Vector Machine. **d**, T-learner XGBoost. **e**, T-learner Penalized Logistic Regression. **f**, S-learner Logistic Regression. Abbreviations: CAST: the Chinese Acute Stroke Trial.

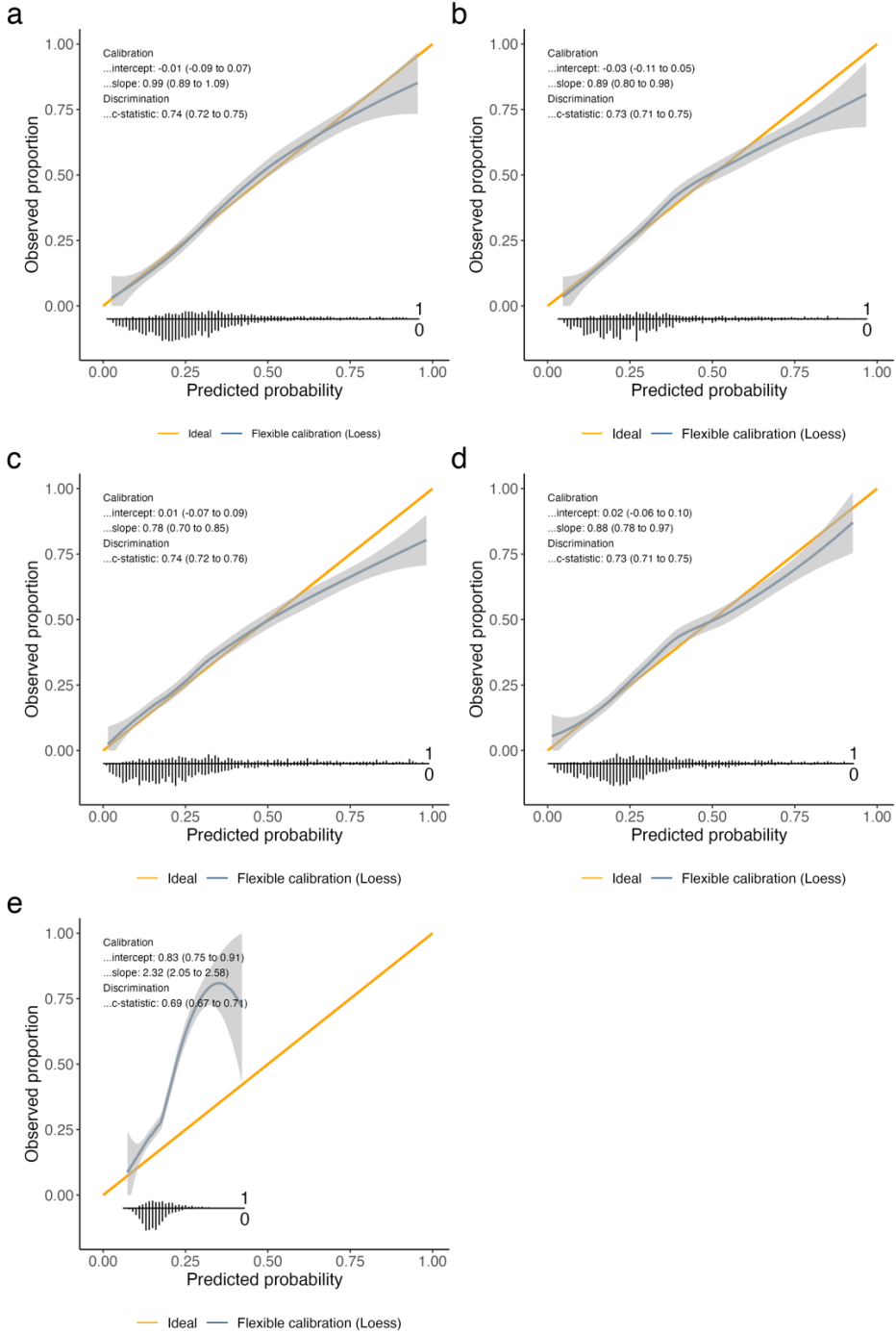

Figure S3.9.2. Internal validation on the CAST test dataset: calibration plot of predicted outcome in the control group. The orange line indicates ideal calibration. The outcome variable is death or dependency at 4 weeks. **a**, S-learner Penalized Logistic Regression. **b**, S-learner XGBoost **c**, S-learner BART. **d**, CVAE. **e**, GANITE. Abbreviations: CAST: the Chinese Acute Stroke Trial.

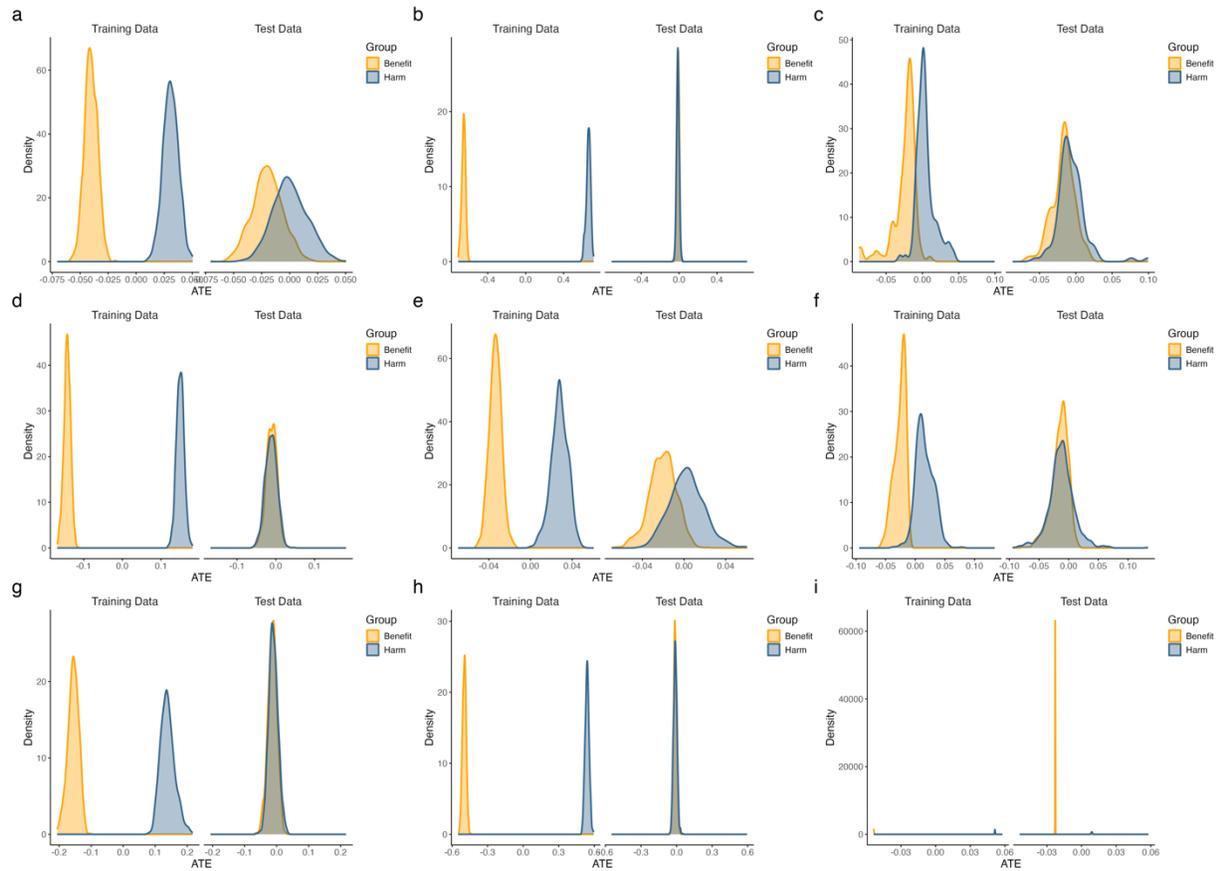

Figure S3.10.1. Internal validation on the CAST dataset: density comparative analysis of causal machine learning-based individualized treatment effects. Density plots depict the distributions of ATE in training data and test data stratified by benefit (orange) and harm (blue) groups. The distribution of ATE comes from 1000 (**a, b, d, e, f**) or 100 (**c, i**) random train-test splits experiments. Benefit means negative ITE values and harm means positive ITE values. The outcome variable is death or dependency at 4 weeks. **a**, T-learner Logistic Regression. **b**, T-learner Random Forest. **c**, T-learner Support Vector Machine. **d**, T-learner XGBoost. **e**, T-learner Penalized Logistic Regression. **f**, S-learner Penalized Logistic Regression. **g**, S-learner XGBoost. **h**, X-learner Random Forest. **i**, X-learner BART. Abbreviations: CAST: the Chinese Acute Stroke Trial, ITE: individualized treatment effect, ATE: average treatment effect.

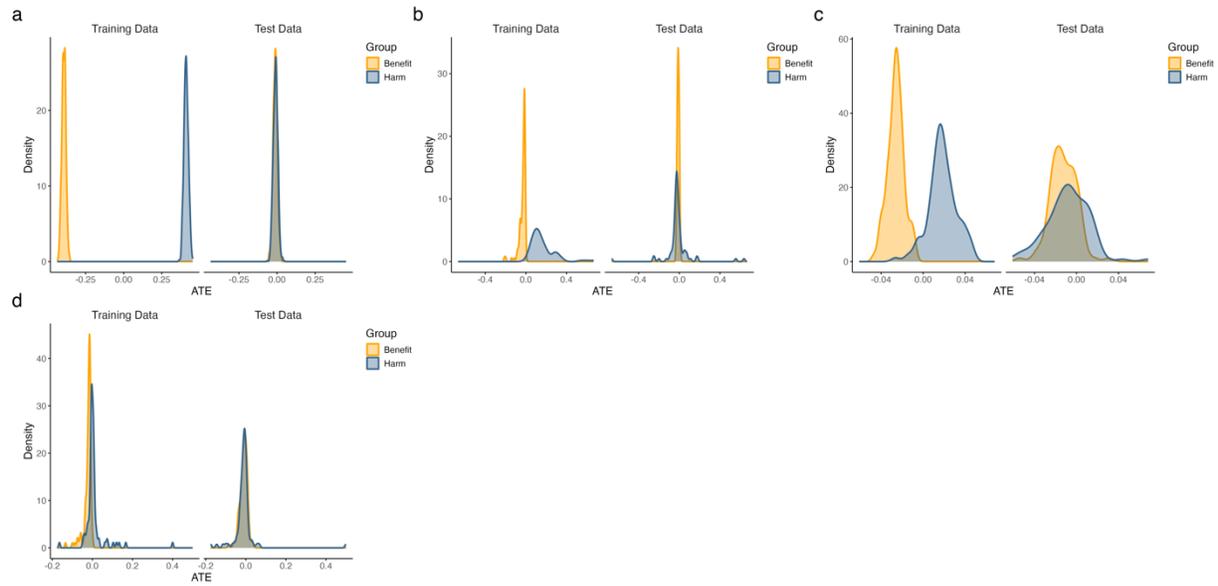

Figure S3.10.2. Internal validation on the CAST dataset: density comparative analysis of causal machine learning-based individualized treatment effects. Density plots depict the distributions of ATE in training data and test data stratified by benefit (orange) and harm (blue) groups. The distribution of ATE comes from 1000 (**b**) or 100 (**a, c, d**) random train-test splits experiments. Benefit means negative ITE values and harm means positive ITE values. The outcome variable is death or dependency at 4 weeks. **a**, DR-learner Random Forest. **b**, Causal Forest. **c**, CVAE. **d**, GANITE. Abbreviations: CAST: the Chinese Acute Stroke Trial, ITE: individualized treatment effect, ATE: average treatment effect.

**Results of internal validation on the combined IST and CAST dataset**

Table S2. Summary of quantitative validation metrics on the combined IST and CAST dataset of 17 causal machine learning methods. The outcome variable is death at 14 days for IST and death at 4 weeks for CAST.

|  |  | c-for-benefit (training) | c-for-benefit (test) | mbcb (training) | mbcb (test) | Calibration-based pseudo R-squared (training) | Calibration-based pseudo R-squared (test) |
|---|---|---|---|---|---|---|---|
| **T-learner** | Logistic Regression | 0.544 | 0.515 | 0.540 | 0.539 | 0.668 | -1.565 |
|  | Random Forest | 0.998 | 0.480 | 0.785 | 0.657 | 0.216 | -3.684 |
|  | Support Vector Machine | 0.540 | 0.496 | 0.554 | 0.554 | 0.578 | -1.579 |
|  | XGBoost | 0.734 | 0.498 | 0.620 | 0.617 | 0.528 | -2.591 |
|  | Penalized Logistic Regression | 0.500 | 0.501 | - | - | - | - |
| **S-learner** | Logistic Regression | 0.517 | 0.501 | 0.514 | 0.514 | 0.625 | -0.413 |
|  | Penalized Logistic Regression | 0.500 | 0.500 | - | - | - | - |
|  | XGBoost | 0.742 | 0.507 | 0.564 | 0.561 | 0.316 | -2.114 |
|  | BART | 0.553 | 0.511 | 0.529 | 0.528 | 0.404 | -0.545 |
| **X-learner** | Random Forest | 0.986 | 0.479 | - | - | 0.139 | -2.297 |
|  | BART | 0.612 | 0.505 | - | - | 0.395 | -0.669 |
| **DR-learner** | Random Forest | 0.793 | 0.506 | - | - | 0.715 | -5.365 |
| **Tree-based methods** | Causal Forest | 0.551 | 0.492 | - | - | 0.112 | -0.177 |
|  | Bayesian Causal Forest | 0.531 | 0.496 | - | - | 0.131 | -0.134 |
|  | model-based recursive partitioning | - | - | - | - | - | - |
| **Deep Learning** | CVAE | 0.516 | 0.484 | 0.535 | 0.536 | -0.263 | -1.488 |
|  | GANITE | 0.508 | 0.484 | 0.531 | 0.531 | -40.604 | -21.871 |

Notes: Consistent values across training and test data that are closer to 1 indicate a better model fit. The dash symbol indicates instances where:

- The validation metric could not be assessed due to the model's mechanism.
- The model failed to estimate valid individualized treatment effects.
- The metric was not applicable to the model.

Abbreviations: IST: the International Stroke Trial, CAST: the Chinese Acute Stroke Trial.

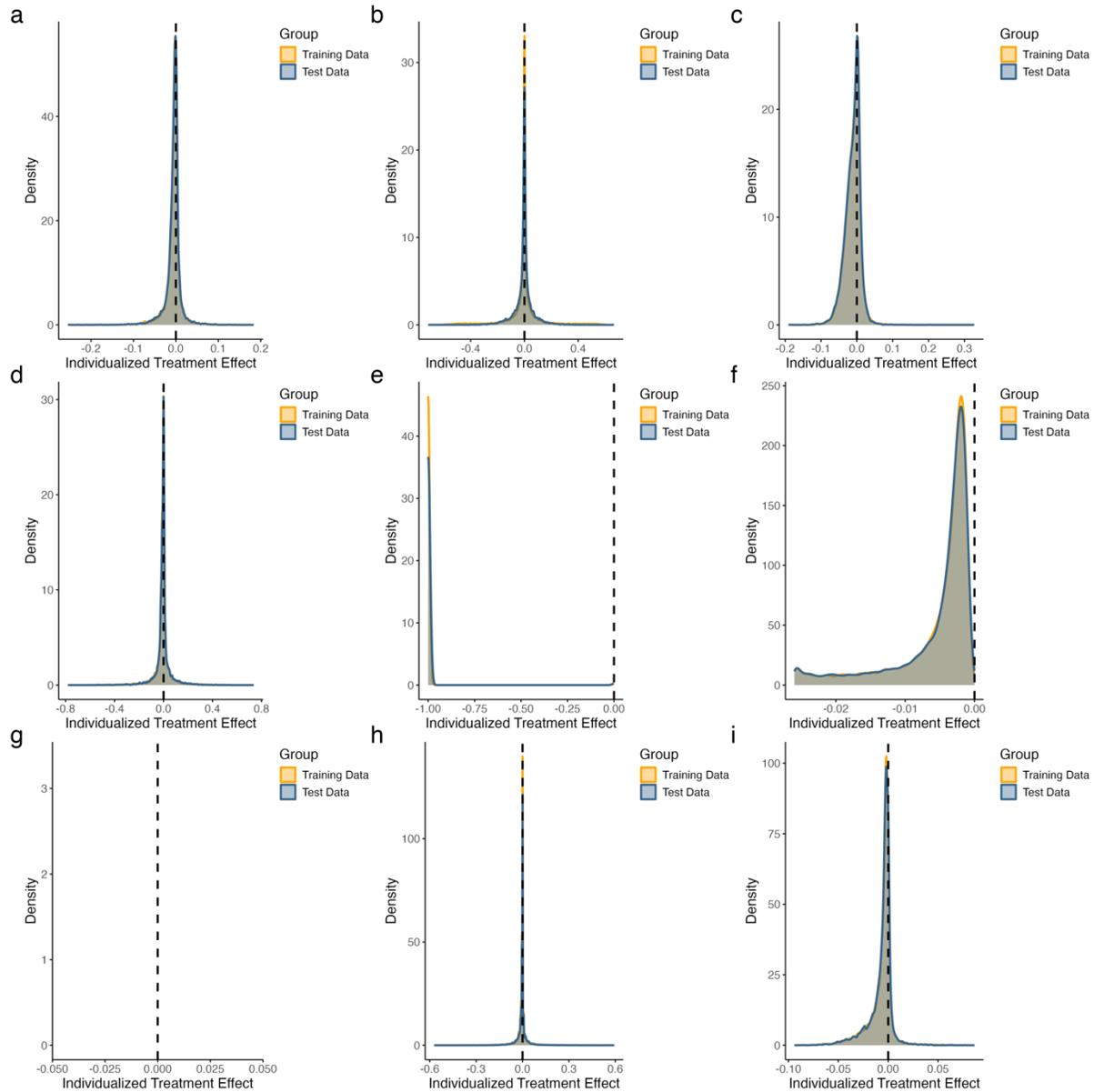

Figure S4.1.1. Internal validation on the combined IST and CAST dataset: density plots of causal machine learning-based individualized treatment effects. Orange represents the training data, and blue indicates the test data. The outcome variable is death at 14 days for IST and death at 4 weeks for CAST. **a**, T-learner Logistic Regression. **b**, T-learner Random Forest. **c**, T-learner Support Vector Machine. **d**, T-learner XGBoost. **e**, T-learner Penalized Logistic Regression. **f**, S-learner Logistic Regression. **g**, S-learner Penalized Logistic Regression. **h**, S-learner XGBoost. **i**, S-learner BART. Abbreviations: IST: the International Stroke Trial, CAST: the Chinese Acute Stroke Trial, ITE: individualized treatment effect.

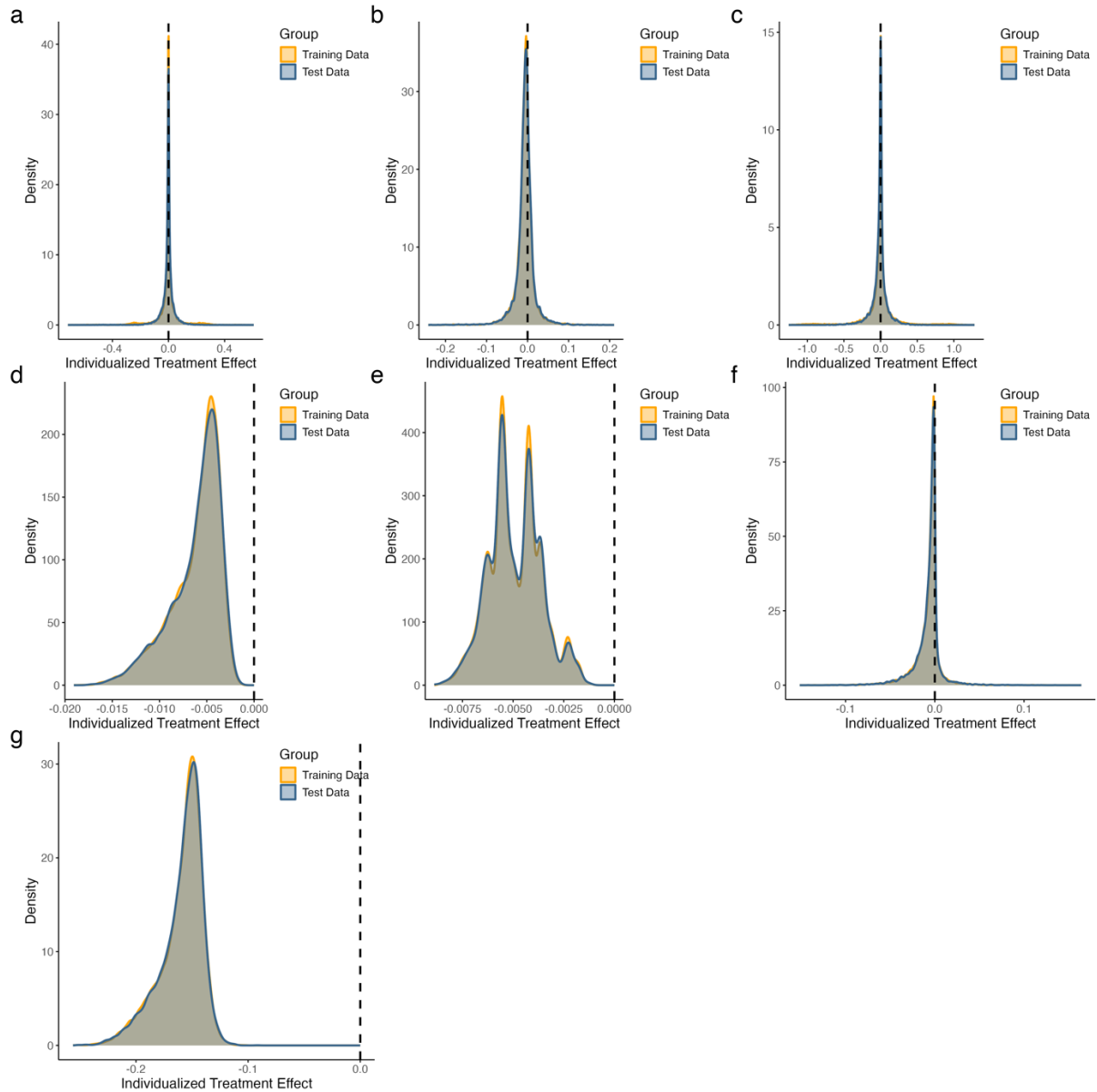

Figure S4.1.2. Internal validation on the combined IST and CAST dataset: density plots of causal machine learning-based individualized treatment effects. Orange represents the training data, and blue indicates the test data. The outcome variable is death at 14 days for IST and death at 4 weeks for CAST. **a**, X-learner Random Forest. **b**, X-learner BART. **c**, DR-learner Random Forest. **d**, Causal Forest. **e**, Bayesian Causal Forest. **f**, CVAE. **g**, GANITE. Abbreviations: IST: the International Stroke Trial, CAST: the Chinese Acute Stroke Trial, ITE: individualized treatment effect.

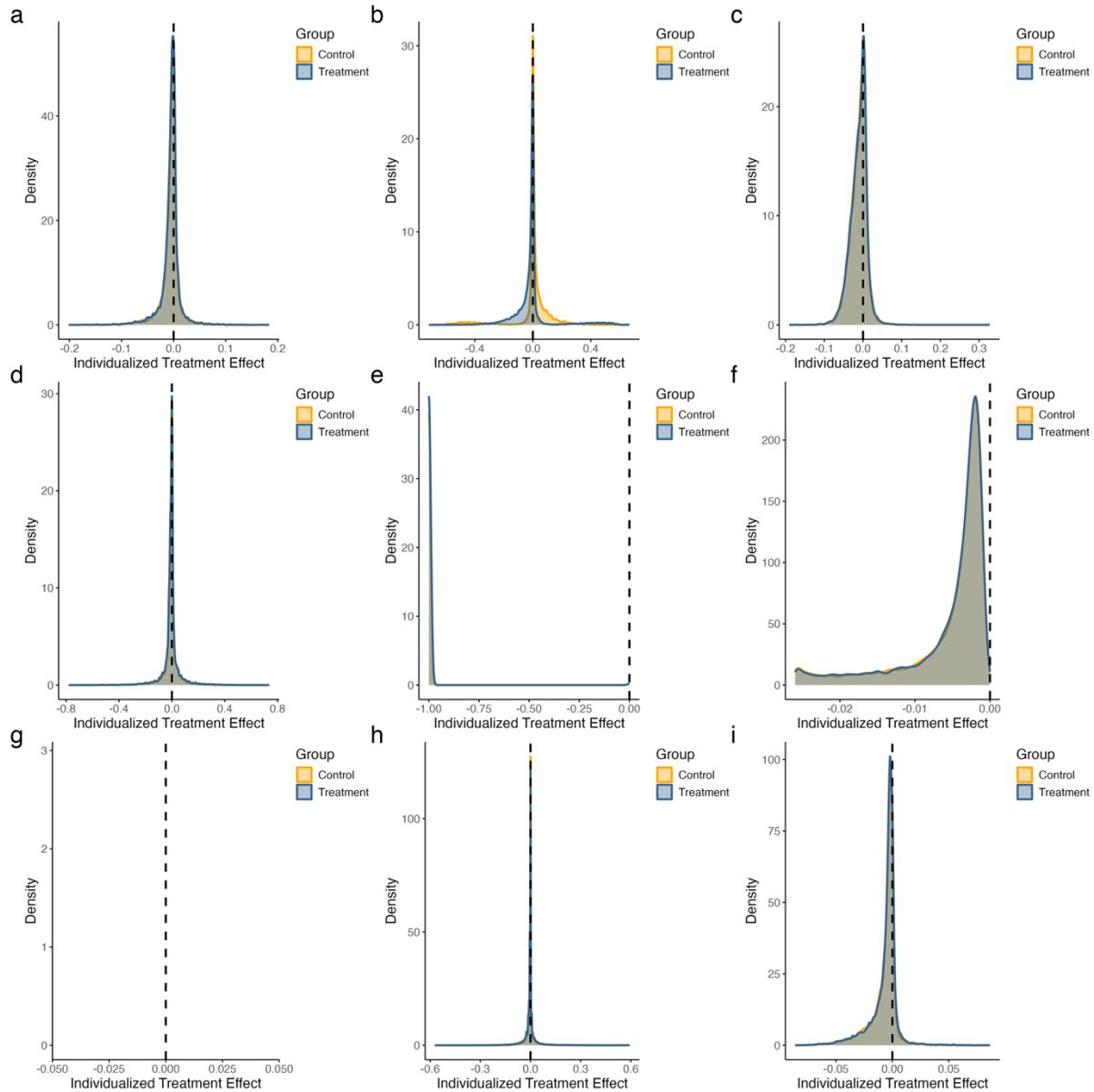

Figure S4.2.1. Internal validation on the combined IST and CAST training dataset: density plots of causal machine learning-based individualized treatment effects. Orange represents the control group, and blue indicates the treatment group. The outcome variable is death at 14 days for IST and death at 4 weeks for CAST. **a**, T-learner Logistic Regression. **b**, T-learner Random Forest. **c**, T-learner Support Vector Machine. **d**, T-learner XGBoost. **e**, T-learner Penalized Logistic Regression. **f**, S-learner Logistic Regression. **g**, S-learner Penalized Logistic Regression. **h**, S-learner XGBoost. **i**, S-learner BART. Abbreviations: IST: the International Stroke Trial, CAST: the Chinese Acute Stroke Trial, ITE: individualized treatment effect.

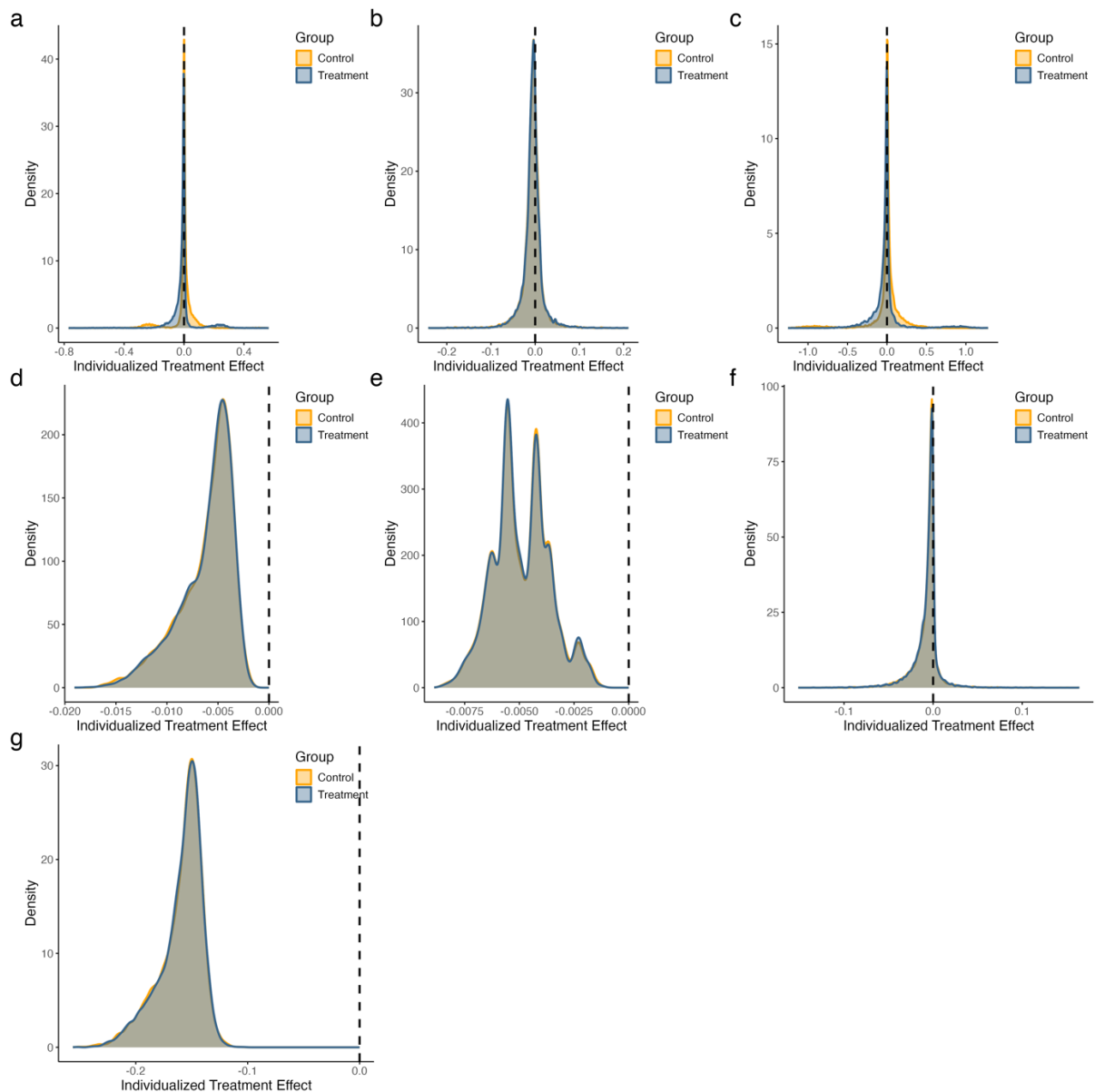

Figure S4.2.2. Internal validation on the combined IST and CAST training dataset: density plots of causal machine learning-based individualized treatment effects. Orange represents the control group, and blue indicates the treatment group. The outcome variable is death at 14 days for IST and death at 4 weeks for CAST. **a**, X-learner Random Forest. **b**, X-learner BART. **c**, DR-learner Random Forest. **d**, Causal Forest. **e**, Bayesian Causal Forest. **f**, CVAE. **g**, GANITE. Abbreviations: IST: the International Stroke Trial, CAST: the Chinese Acute Stroke Trial, ITE: individualized treatment effect.

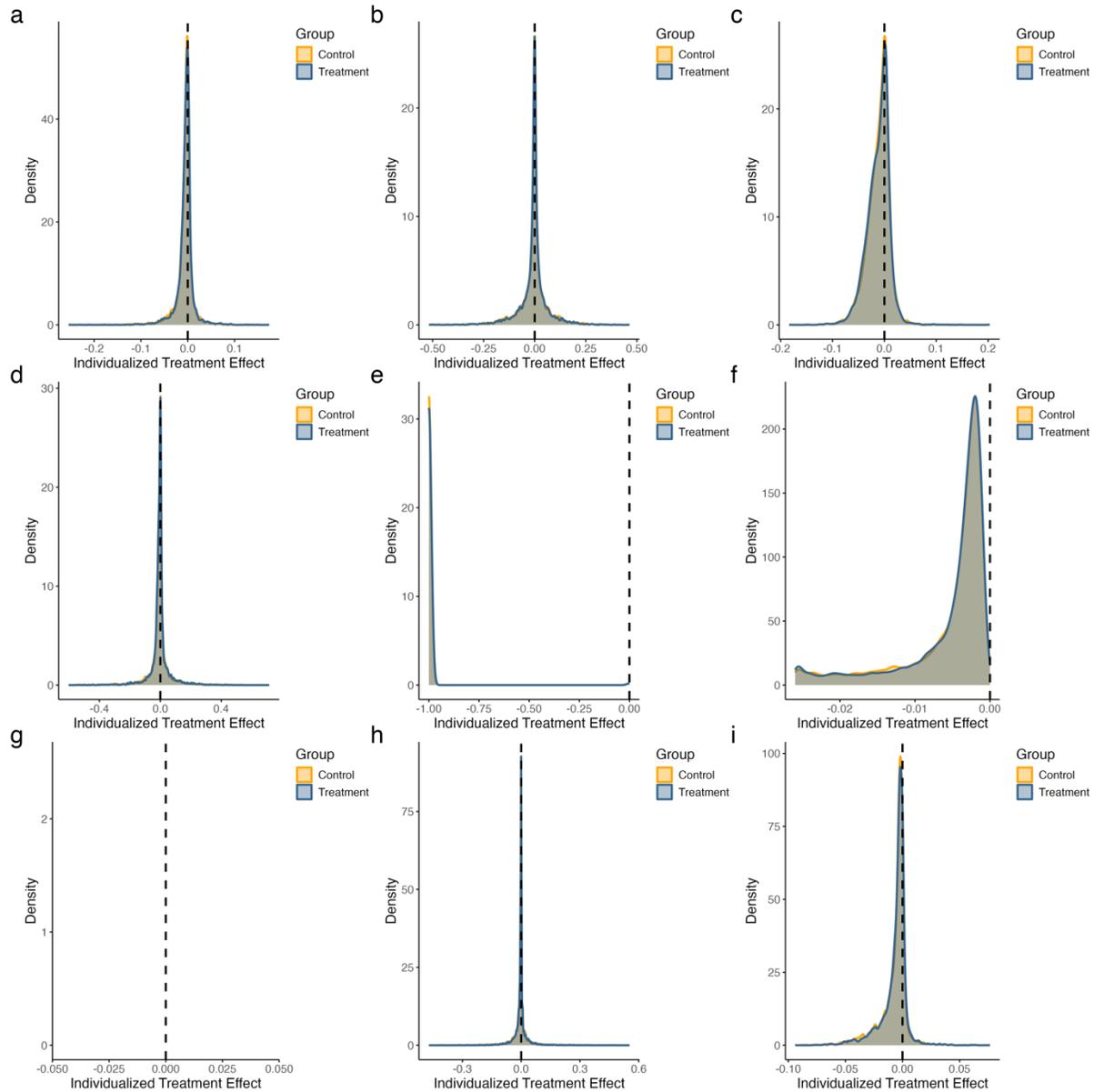

Figure S4.3.1. Internal validation on the combined IST and CAST test dataset: density plots of causal machine learning-based individualized treatment effects. Orange represents the control group, and blue indicates the treatment group. The outcome variable is death at 14 days for IST and death at 4 weeks for CAST. **a**, T-learner Logistic Regression. **b**, T-learner Random Forest. **c**, T-learner Support Vector Machine. **d**, T-learner XGBoost. **e**, T-learner Penalized Logistic Regression. **f**, S-learner Logistic Regression. **g**, S-learner Penalized Logistic Regression. **h**, S-learner XGBoost. **i**, S-learner BART. Abbreviations: IST: the International Stroke Trial, CAST: the Chinese Acute Stroke Trial, ITE: individualized treatment effect.

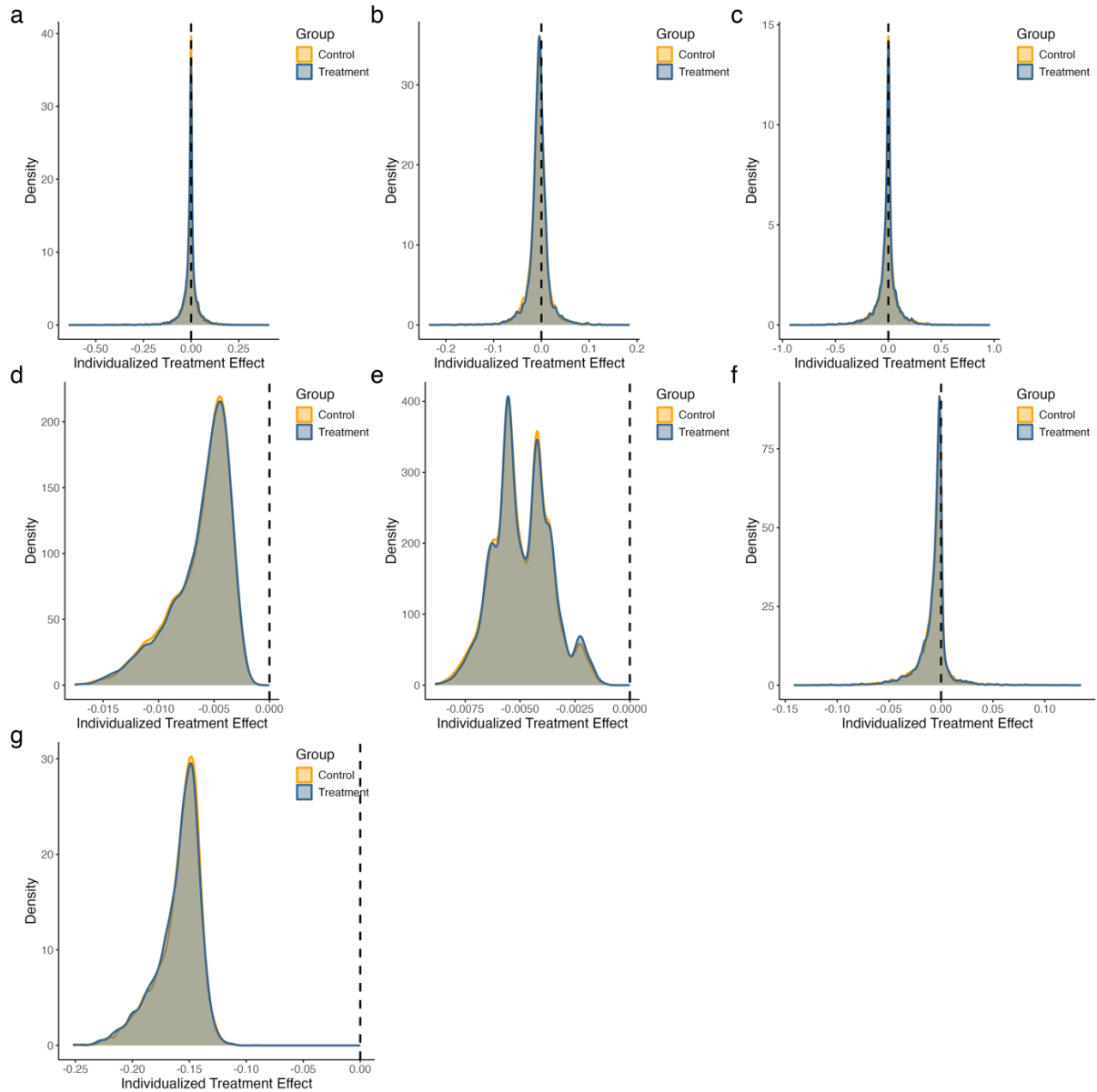

Figure S4.3.2. Internal validation on the combined IST and CAST test dataset: density plots of causal machine learning-based individualized treatment effects. Orange represents the control group, and blue indicates the treatment group. The outcome variable is death at 14 days for IST and death at 4 weeks for CAST. **a**, X-learner Random Forest. **b**, X-learner BART. **c**, DR-learner Random Forest. **d**, Causal Forest. **e**, Bayesian Causal Forest. **f**, CVAE. **g**, GANITE. Abbreviations: IST: the International Stroke Trial, CAST: the Chinese Acute Stroke Trial, ITE: individualized treatment effect.

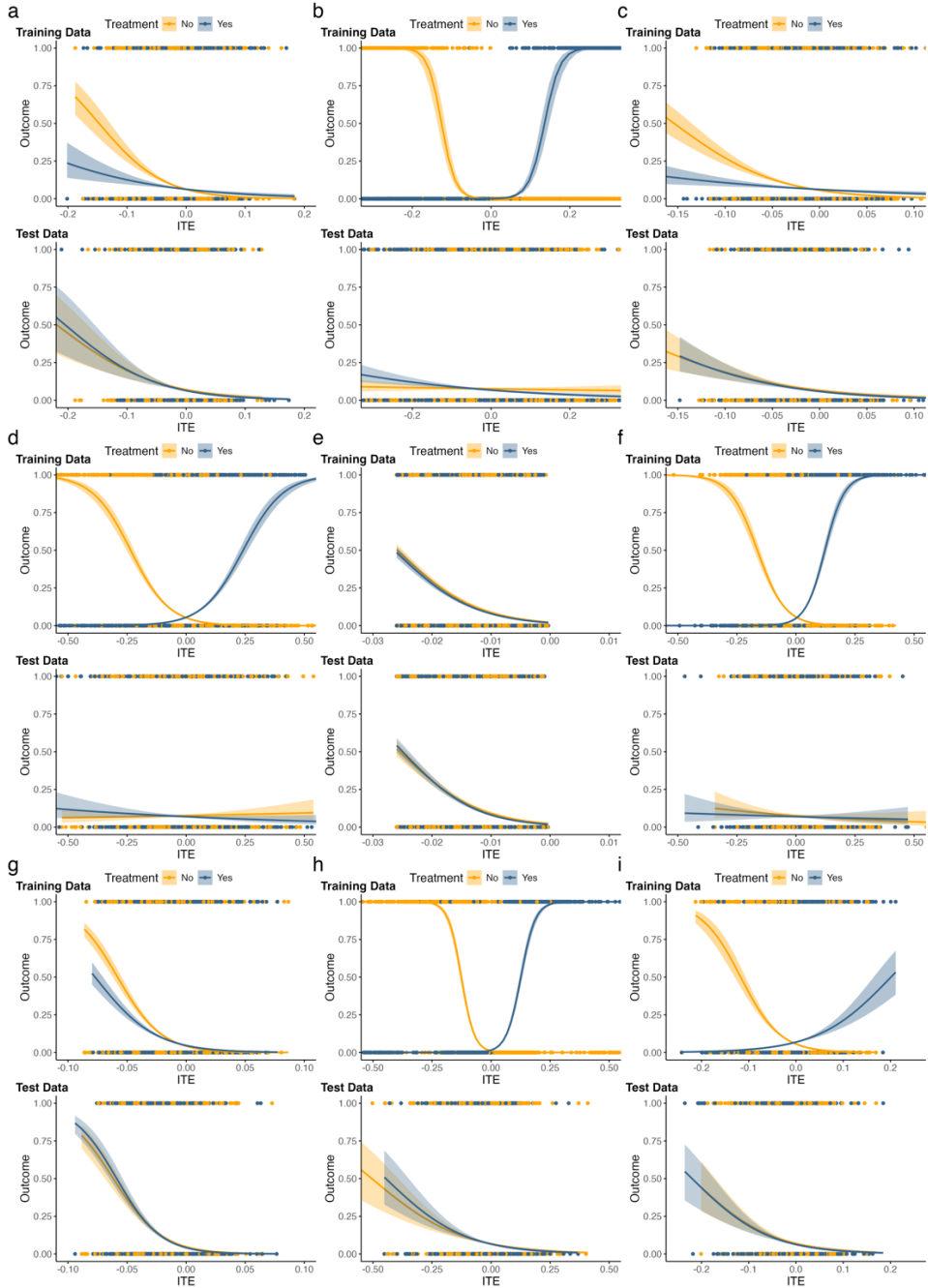

Figure S4.4.1. Internal validation on the combined IST and CAST dataset: outcome-ITE comparative analysis. Dot plots and line plots depict the true and fitted patient outcomes against estimated ITE values between training data and test data. Orange represents the control group, and blue indicates the treatment group. The outcome variable is death at 14 days for IST and death at 4 weeks for CAST. **a**, T-learner Logistic Regression. **b**, T-learner Random Forest. **c**, T-learner Support Vector Machine. **d**, T-learner XGBoost. **e**, S-learner Logistic Regression. **f**, S-learner XGBoost. **g**, S-learner BART. **h**, X-learner Random Forest. **i**, X-learner BART. Abbreviations: IST: the International Stroke Trial, CAST: the Chinese Acute Stroke Trial, ITE: individualized treatment effect.

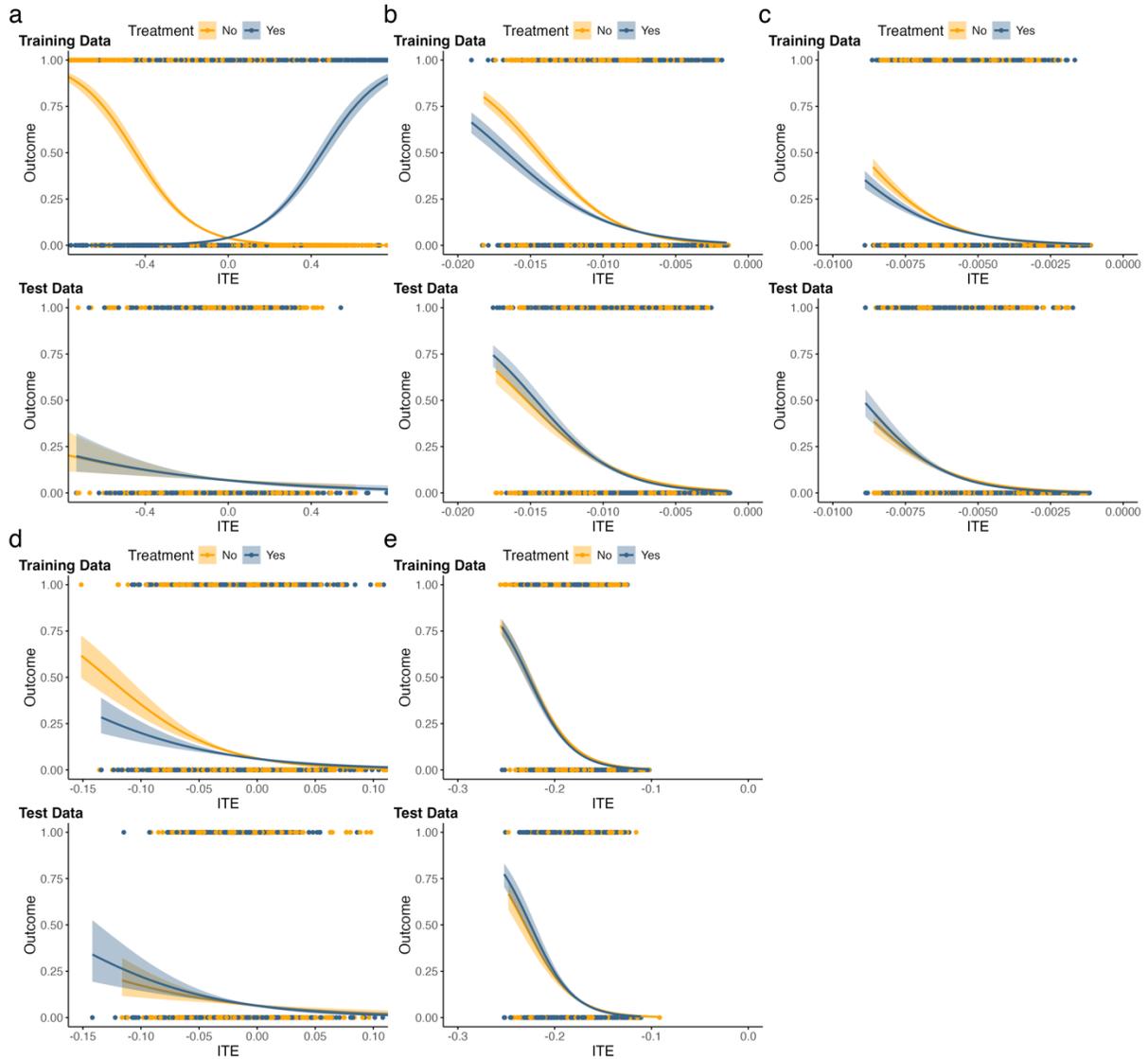

Figure S4.4.2. Internal validation on the combined IST and CAST dataset: outcome-ITE comparative analysis. Dot plots and line plots depict the true and fitted patient outcomes against estimated ITE values between training data and test data. Orange represents the control group, and blue indicates the treatment group. The outcome variable is death at 14 days for IST and death at 4 weeks for CAST. **a**, DR-learner Random Forest. **b**, Causal Forest. **c**, Bayesian Causal Forest. **d**, CVAE. **e**, GANITE. Abbreviations: IST: the International Stroke Trial, CAST: the Chinese Acute Stroke Trial, ITE: individualized treatment effect.

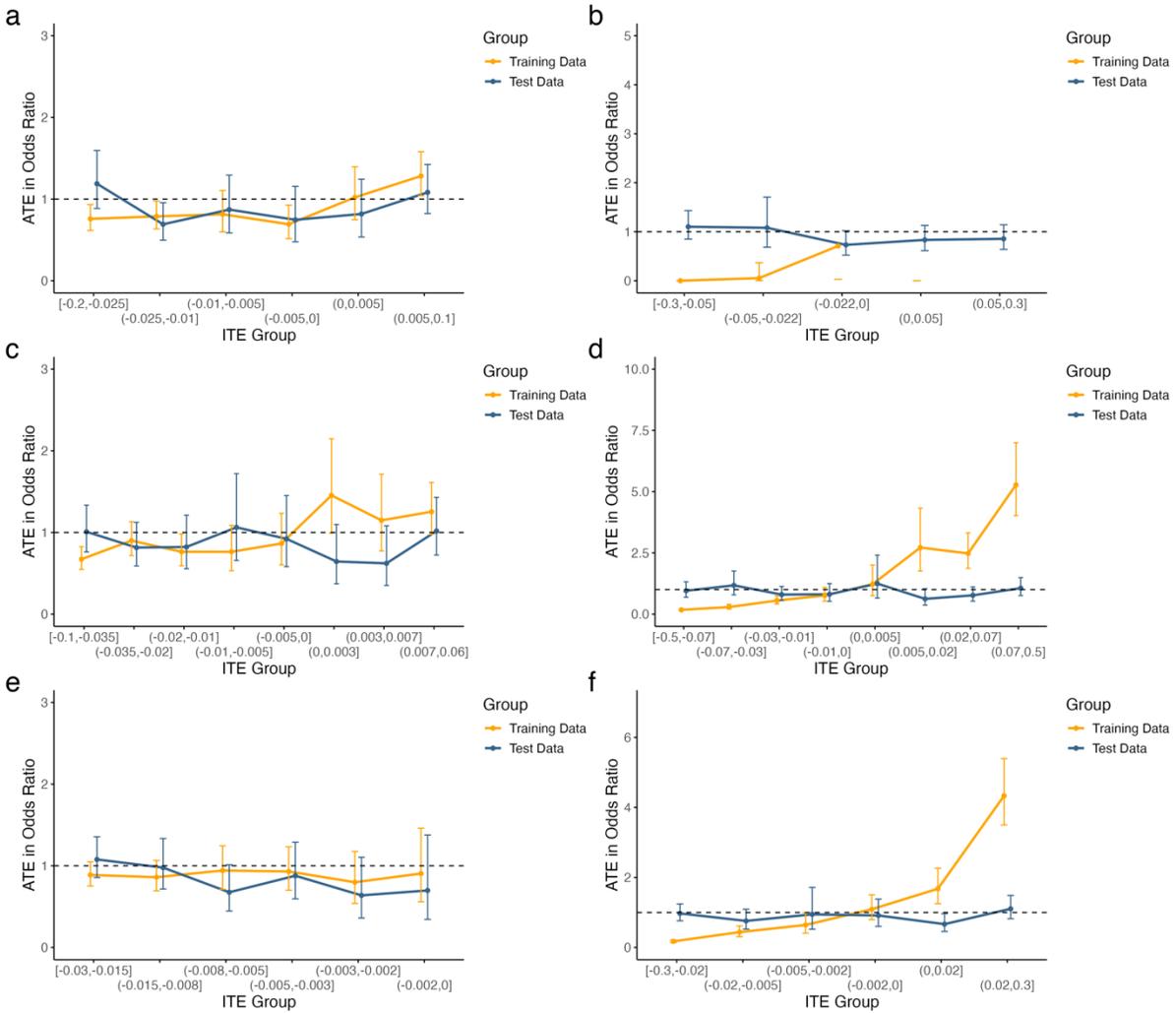

Figure S4.5.1. Internal validation on the combined IST and CAST dataset: ATE-ITE comparative analysis. Line plots depict ATE in risk ratio within different ITE subgroups and provide the confidence intervals at 95% level. Orange represents the training data, and blue indicates the test data. The horizontal dashed line at 1.0 means no treatment effects. The outcome variable is death at 14 days for IST and death at 4 weeks for CAST. **a**, T-learner Logistic Regression. **b**, T-learner Random Forest. **c**, T-learner Support Vector Machine. **d**, T-learner XGBoost. **e**, S-learner Logistic Regression. **f**, S-learner XGBoost. Abbreviations: IST: the International Stroke Trial, CAST: the Chinese Acute Stroke Trial, ITE: individualized treatment effect, ATE: average treatment effect.

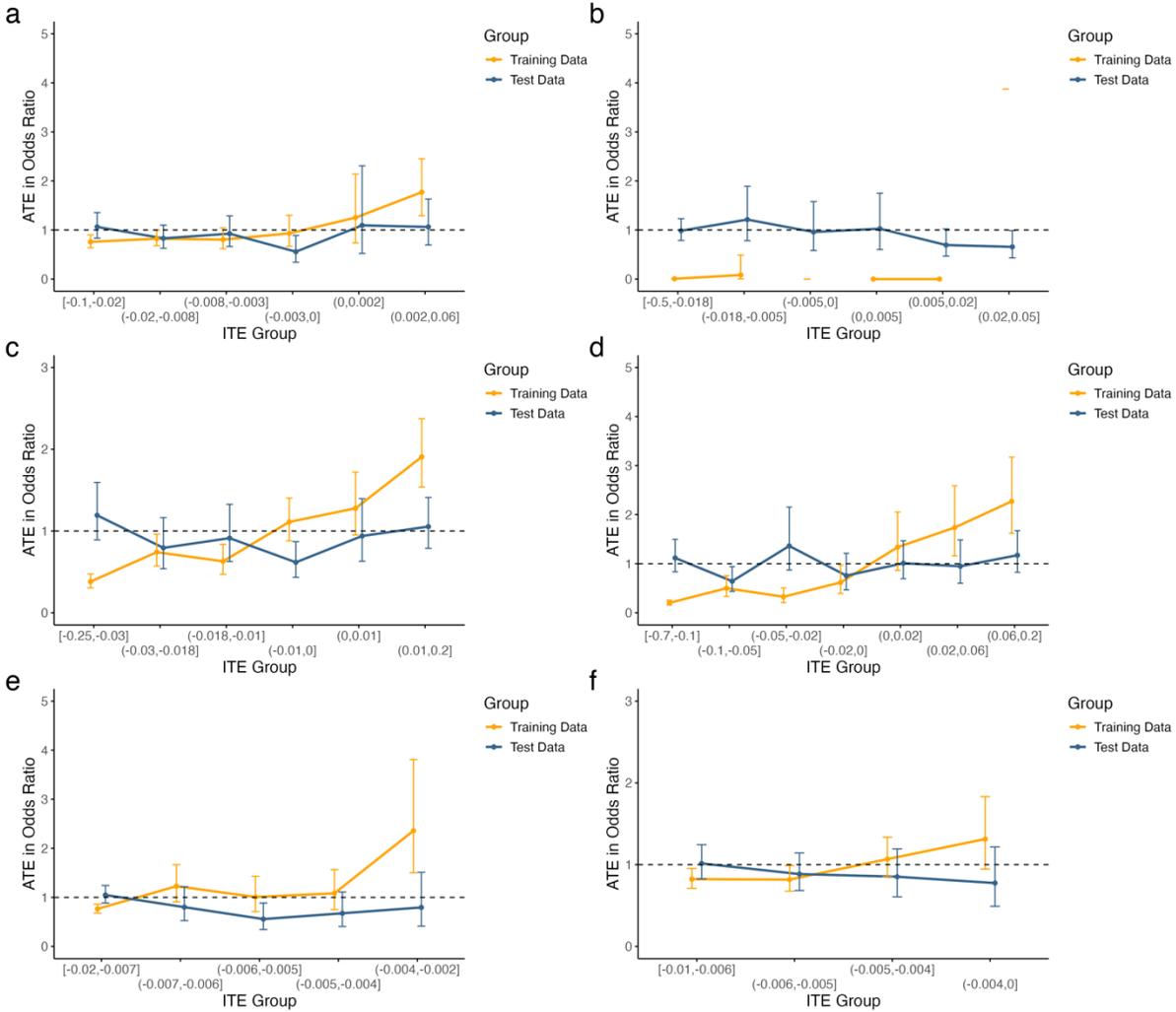

Figure S4.5.2. Internal validation on the combined IST and CAST dataset: ATE-ITE comparative analysis. Line plots depict ATE in risk ratio within different ITE subgroups and provide the confidence intervals at 95% level. Orange represents the training data, and blue indicates the test data. The horizontal dashed line at 1.0 means no treatment effects. The outcome variable is death at 14 days for IST and death at 4 weeks for CAST. **a**, S-learner BART. **b**, X-learner Random Forest. **c**, X-learner BART. **d**, DR-learner Random Forest. **e**, Causal Forest. **f**, Bayesian Causal Forest. Abbreviations: IST: the International Stroke Trial, CAST: the Chinese Acute Stroke Trial, ITE: individualized treatment effect, ATE: average treatment effect.

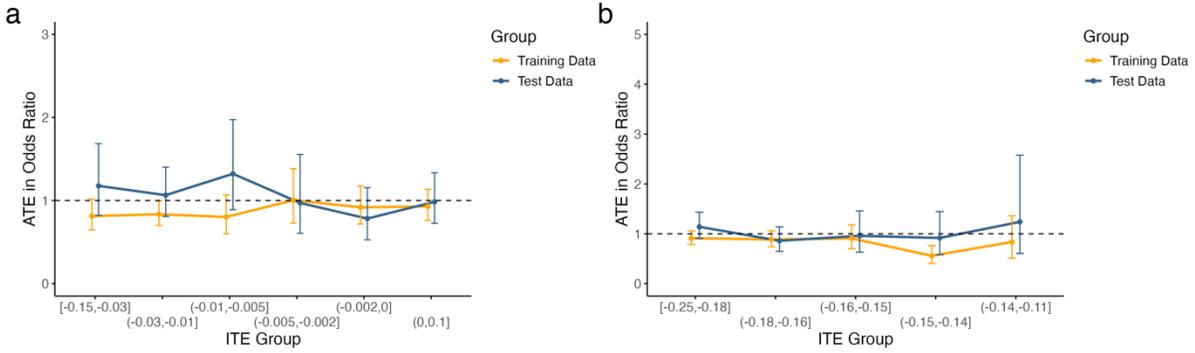

Figure S4.5.2. Internal validation on the combined IST and CAST dataset: ATE-ITE comparative analysis. Line plots depict ATE in risk ratio within different ITE subgroups and provide the confidence intervals at 95% level. Orange represents the training data, and blue indicates the test data. The horizontal dashed line at 1.0 means no treatment effects. The outcome variable is death at 14 days for IST and death at 4 weeks for CAST. **a**, CVAE. **b**, GANITE. Abbreviations: IST: the International Stroke Trial, CAST: the Chinese Acute Stroke Trial, ITE: individualized treatment effect, ATE: average treatment effect.

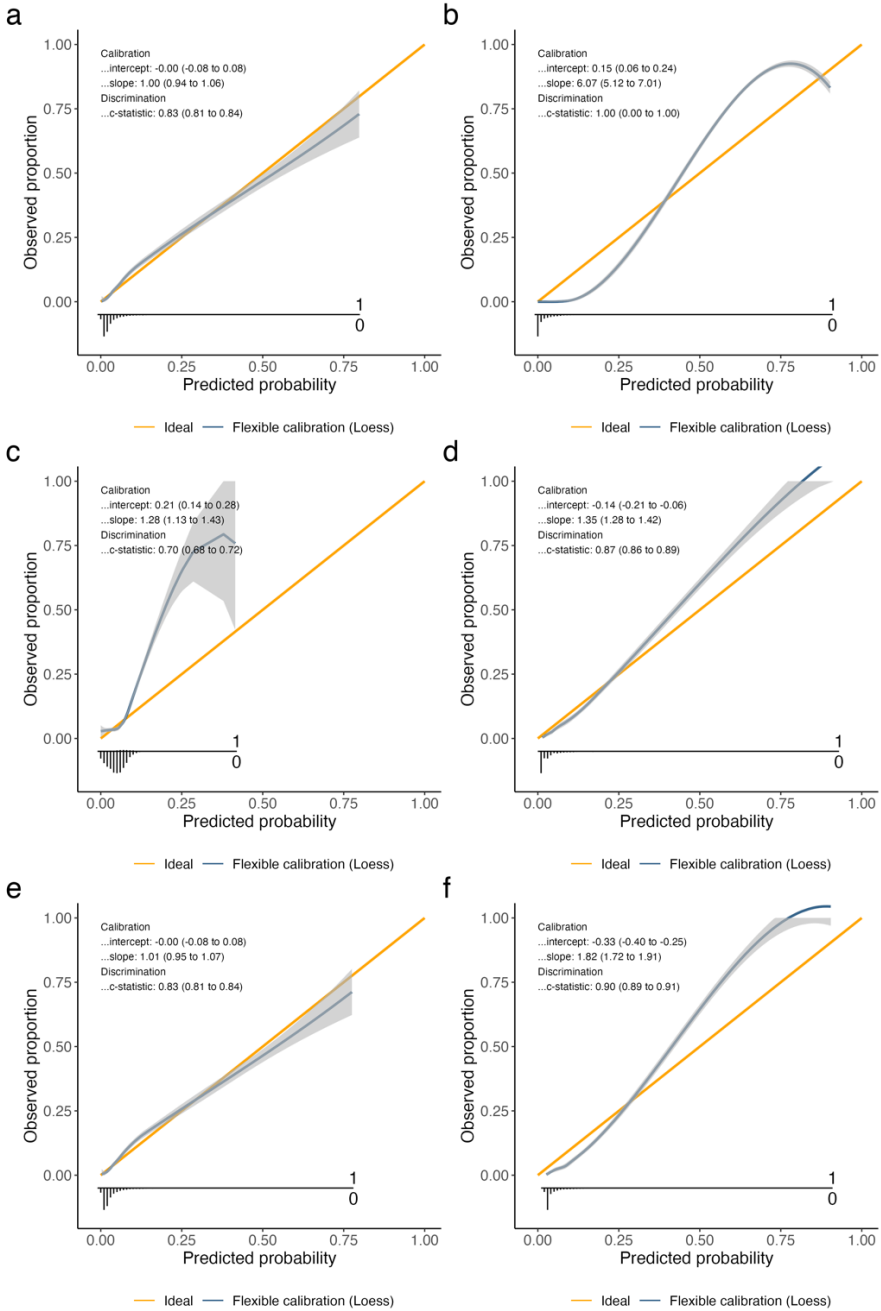

Figure S4.6.1. Internal validation on the combined IST and CAST training dataset: calibration plot of predicted outcome in the treatment group. The orange line indicates ideal calibration. The outcome variable is death at 14 days for IST and death at 4 weeks for CAST. **a**, T-learner Logistic Regression. **b**, T-learner Random Forest. **c**, T-learner Support Vector Machine. **d**, T-learner XGBoost. **e**, S-learner Logistic Regression. **f**, S-learner XGBoost. Abbreviations: IST: the International Stroke Trial, CAST: the Chinese Acute Stroke Trials.

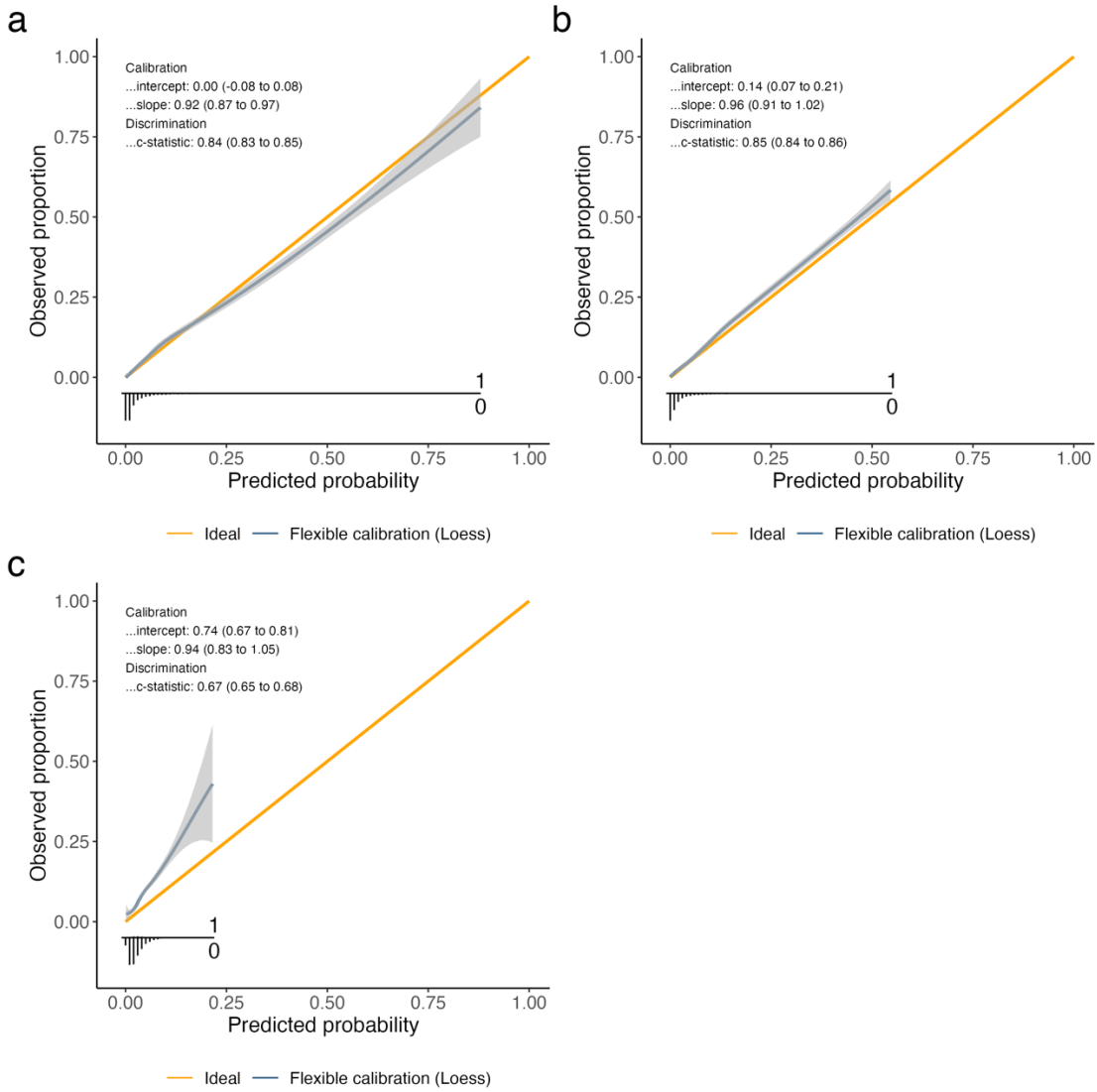

Figure S4.6.2. Internal validation on the combined IST and CAST training dataset: calibration plot of predicted outcome in the treatment group. The orange line indicates ideal calibration. The outcome variable is death at 14 days for IST and death at 4 weeks for CAST. **a**, S-learner BART. **b**, CVAE. **c**, GANITE. Abbreviations: IST: the International Stroke Trial, CAST: the Chinese Acute Stroke Trial.

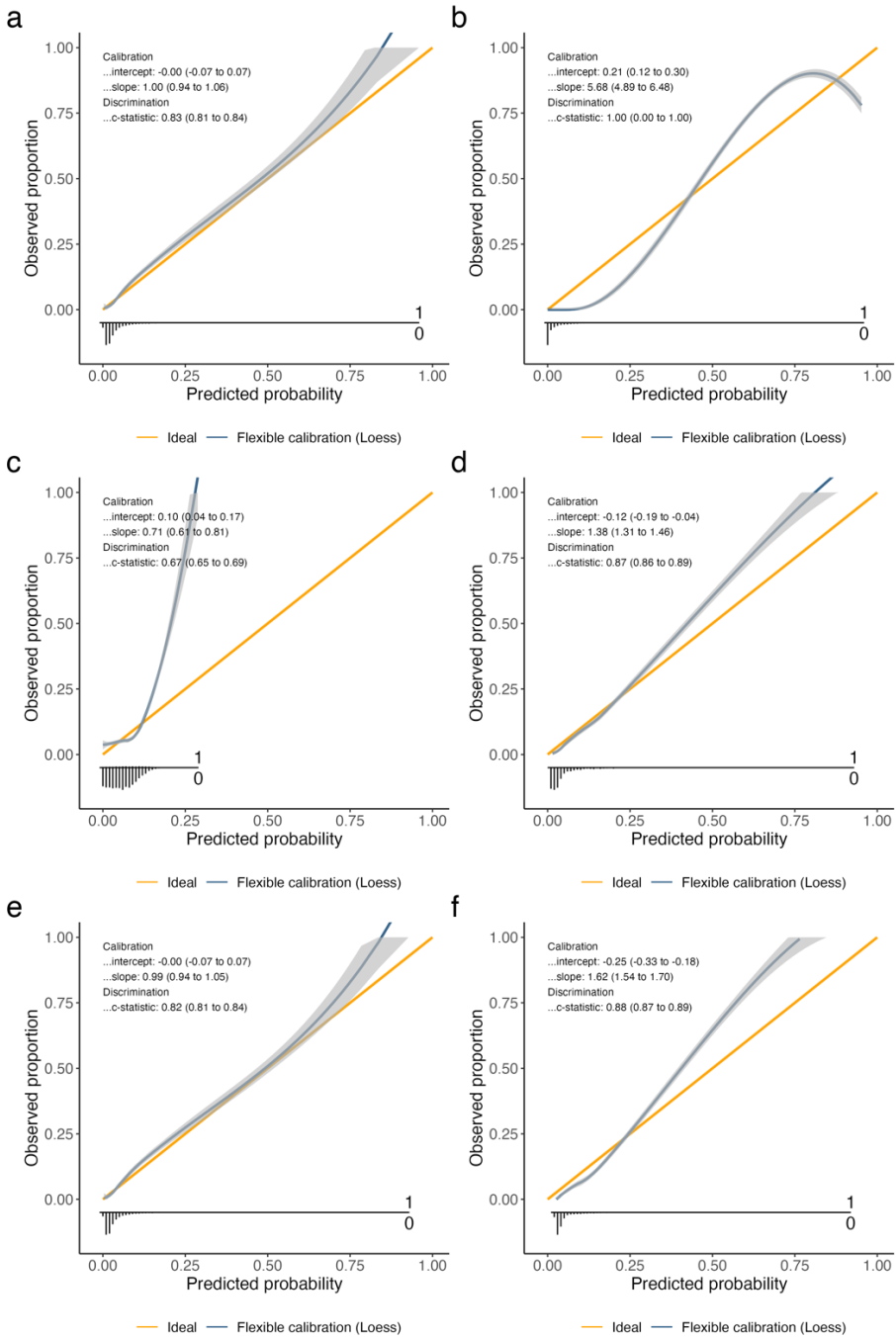

Figure S4.7.1. Internal validation on the combined IST and CAST training dataset: calibration plot of predicted outcome in the control group. The orange line indicates ideal calibration. The outcome variable is death at 14 days for IST and death at 4 weeks for CAST. **a**, T-learner Logistic Regression. **b**, T-learner Random Forest. **c**, T-learner Support Vector Machine. **d**, T-learner XGBoost. **e**, S-learner Logistic Regression. **f**, S-learner XGBoost. Abbreviations: IST: the International Stroke Trial, CAST: the Chinese Acute Stroke Trial.

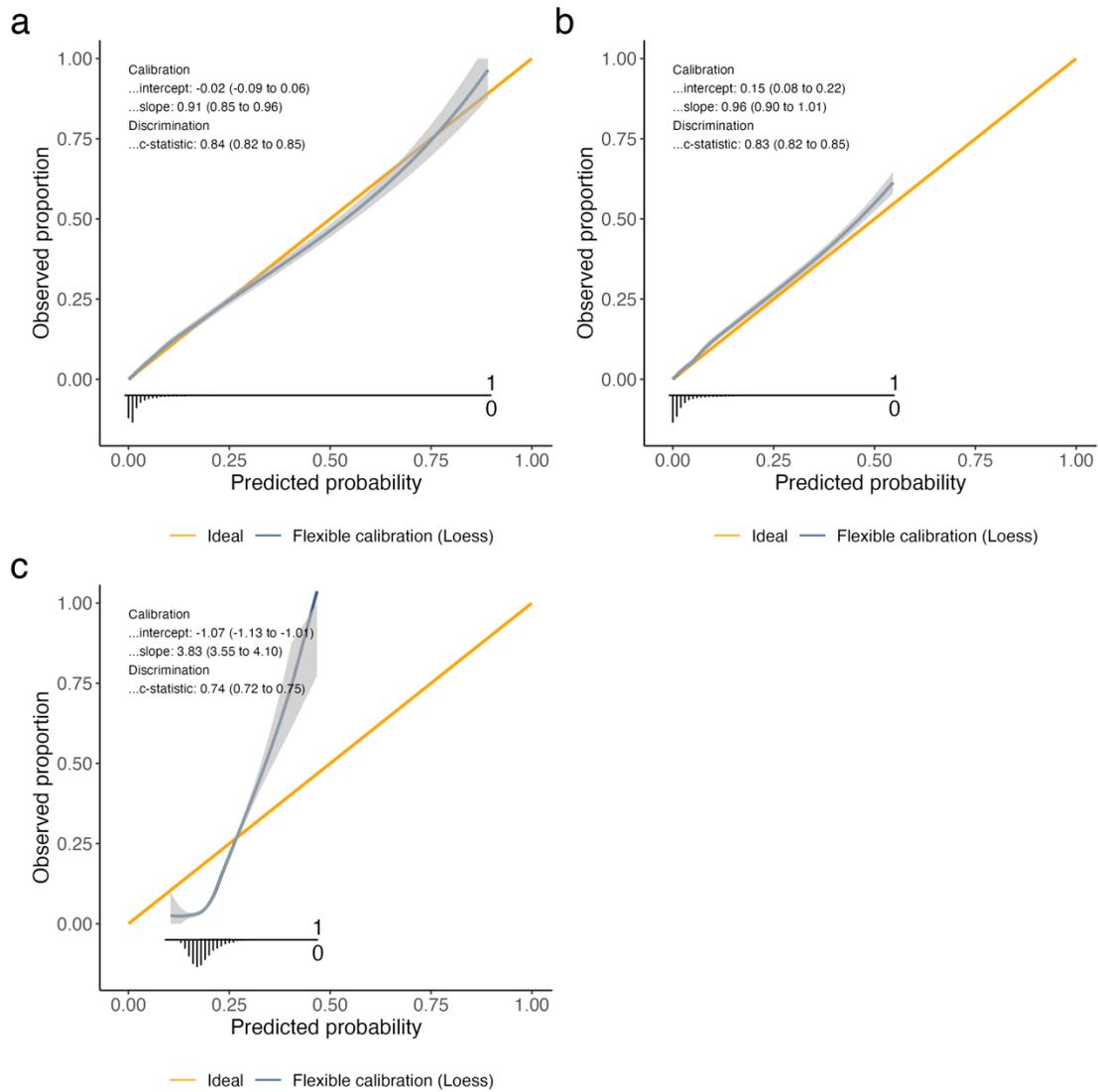

Figure S4.7.2. Internal validation on the combined IST and CAST training dataset: calibration plot of predicted outcome in the control group. The orange line indicates ideal calibration. The outcome variable is death at 14 days for IST and death at 4 weeks for CAST. **a**, S-learner BART. **b**, CVAE. **c**, GANITE. Abbreviations: IST: the International Stroke Trial, CAST: the Chinese Acute Stroke Trial.

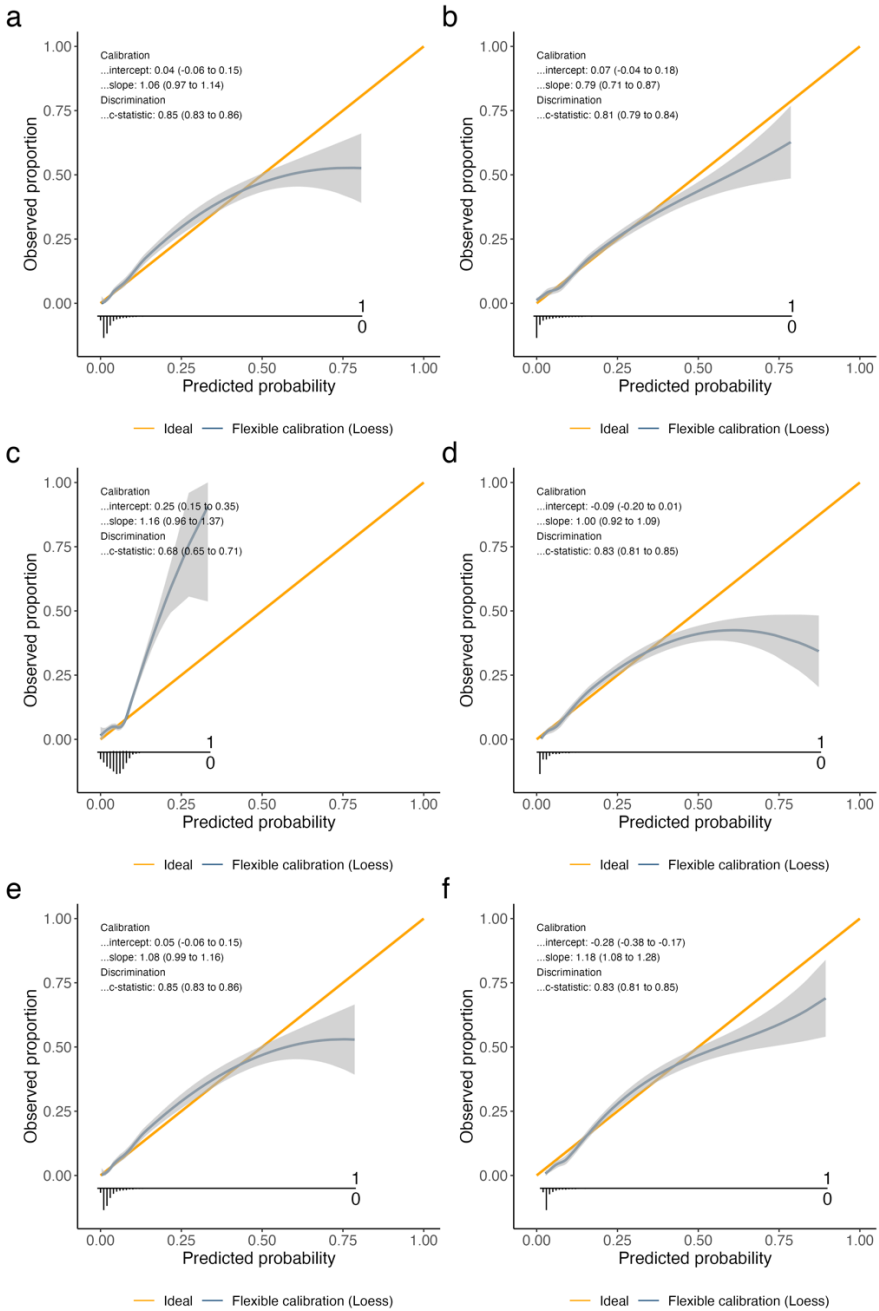

Figure S4.8.1. Internal validation on the combined IST and CAST test dataset: calibration plot of predicted outcome in the treatment group. The orange line indicates ideal calibration. The outcome variable is death at 14 days for IST and death at 4 weeks for CAST. **a**, T-learner Logistic Regression. **b**, T-learner Random Forest. **c**, T-learner Support Vector Machine. **d**, T-learner XGBoost. **e**, S-learner Logistic Regression. **f**, S-learner XGBoost. Abbreviations: IST: the International Stroke Trial, CAST: the Chinese Acute Stroke Trial.

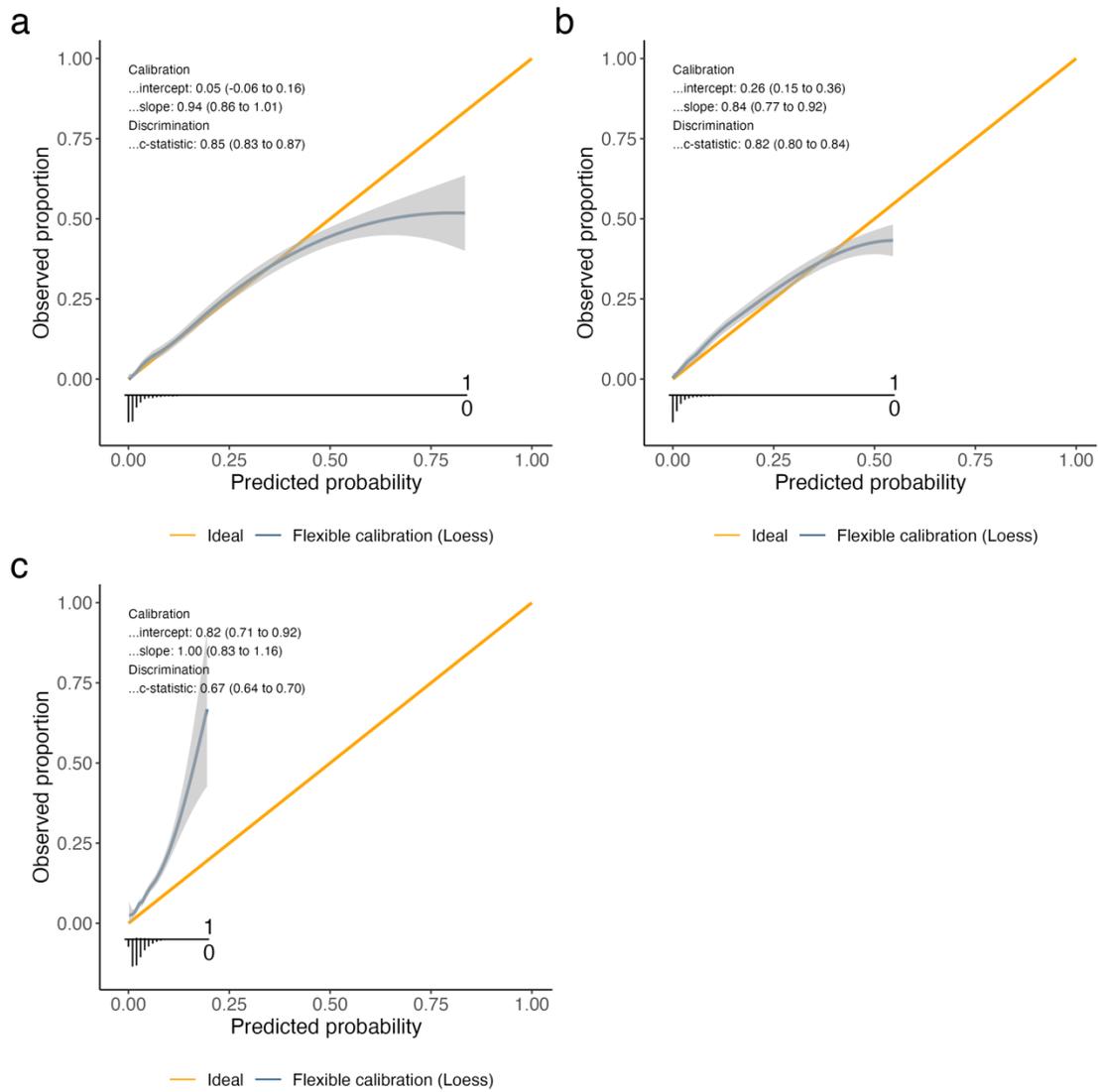

Figure S4.8.2. Internal validation on the combined IST and CAST test dataset: calibration plot of predicted outcome in the treatment group. The orange line indicates ideal calibration. The outcome variable is death at 14 days for IST and death at 4 weeks for CAST. **a**, S-learner BART. **b**, CVAE. **c**, GANITE. Abbreviations: IST: the International Stroke Trial, CAST: the Chinese Acute Stroke Trial.

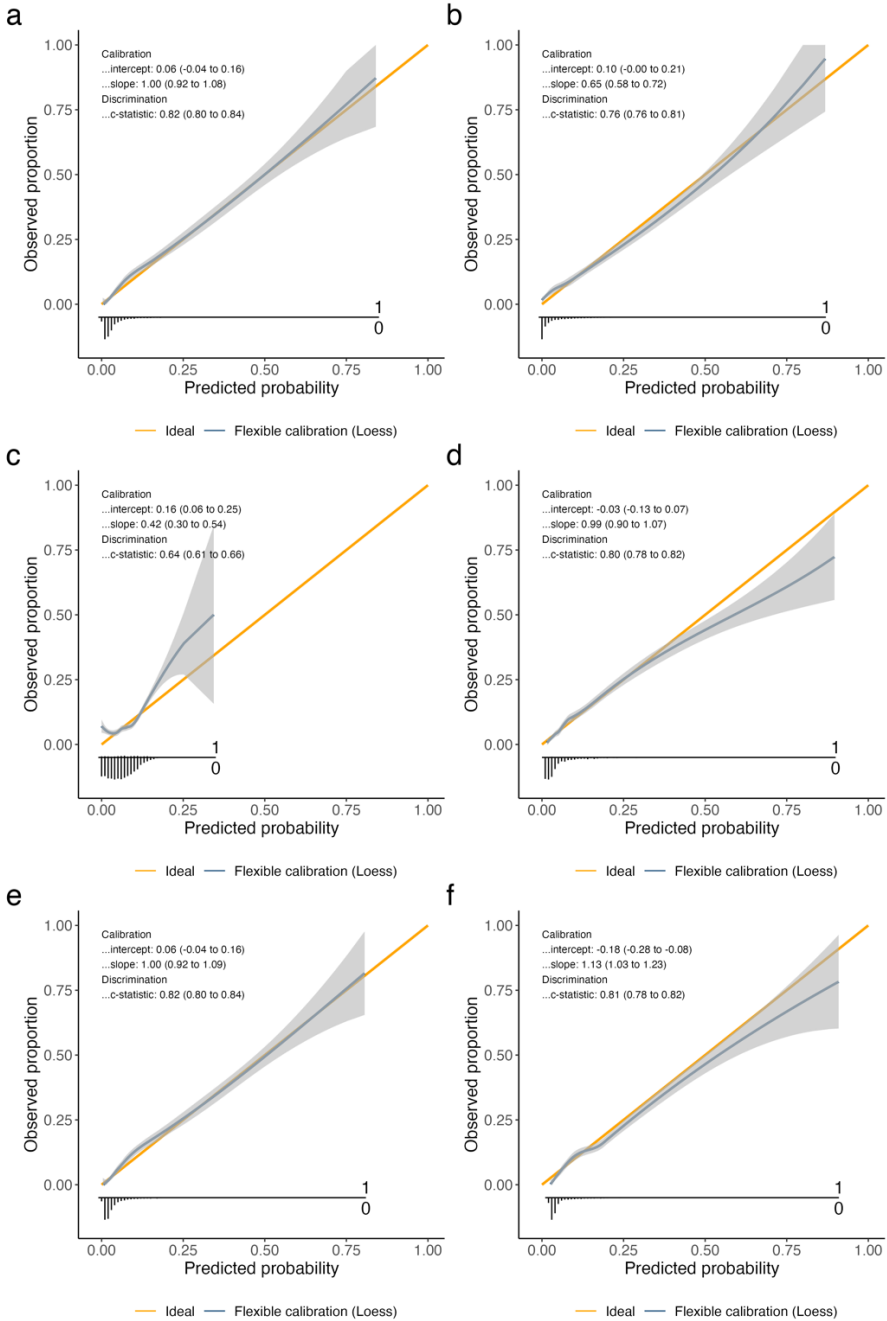

Figure S4.9.1. Internal validation on the combined IST and CAST test dataset: calibration plot of predicted outcome in the control group. The orange line indicates ideal calibration. The outcome variable is death at 14 days for IST and death at 4 weeks for CAST. **a**, T-learner Logistic Regression. **b**, T-learner Random Forest. **c**, T-learner Support Vector Machine. **d**, T-learner XGBoost. **e**, S-learner Logistic Regression. **f**, S-learner XGBoost. Abbreviations: IST: the International Stroke Trial, CAST: the Chinese Acute Stroke Trial.

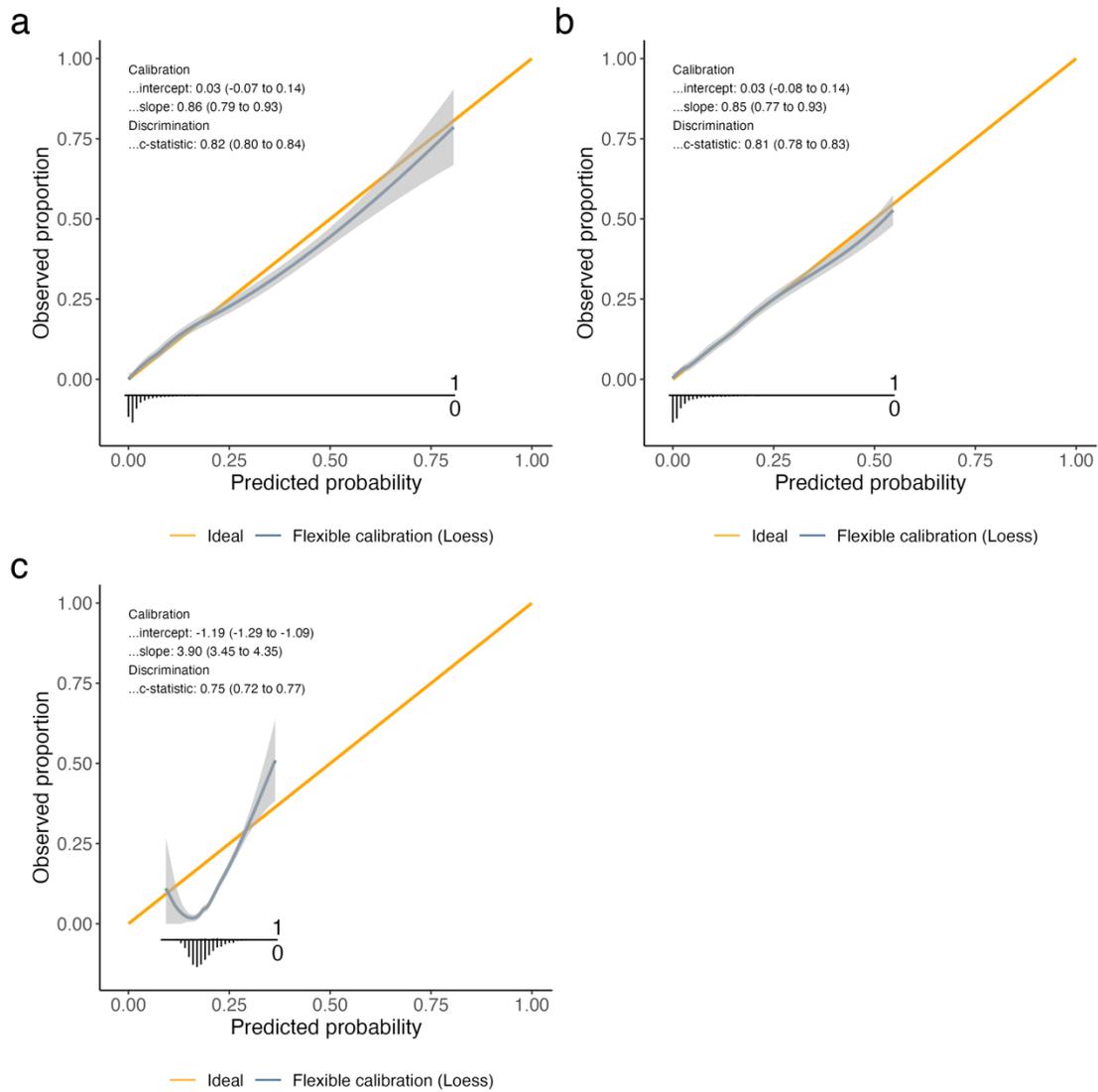

Figure S4.9.2. Internal validation on the combined IST and CAST test dataset: calibration plot of predicted outcome in the control group. The orange line indicates ideal calibration. The outcome variable is death at 14 days for IST and death at 4 weeks for CAST. **a**, S-learner BART. **b**, CVAE. **c**, GANITE. Abbreviations: IST: the International Stroke Trial, CAST: the Chinese Acute Stroke Trial.

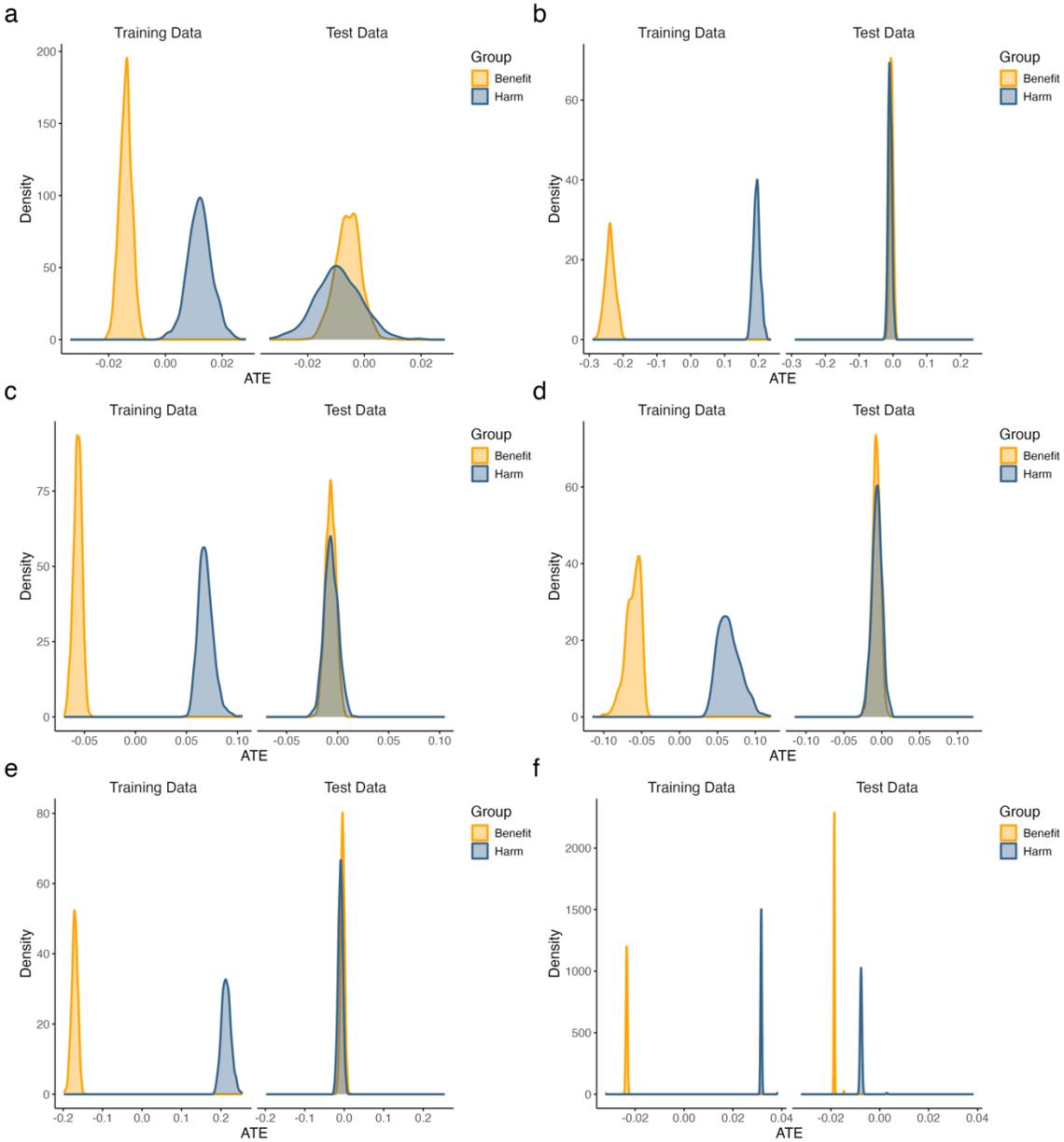

Figure S4.10.1. Internal validation on the combined IST and CAST dataset: density comparative analysis of causal machine learning-based individualized treatment effects. Density plots depict the distributions of ATE in training data and test data stratified by benefit (orange) and harm (blue) groups. The distribution of ATE comes from 1000 (**a, b, d, e, f**) or 100 (**c, i**) random train-test splits experiments. Benefit means negative ITE values and harm means positive ITE values. The outcome variable is death at 14 days for IST and death at 4 weeks for CAST. **a**, T-learner Logistic Regression. **b**, T-learner Random Forest. **c**, T-learner XGBoost. **d**, S-learner XGBoost. **e**, X-learner Random Forest. **f**, X-learner BART. Abbreviations: IST: the International Stroke Trial, CAST: the Chinese Acute Stroke Trial, ITE: individualized treatment effect, ATE: average treatment effect.

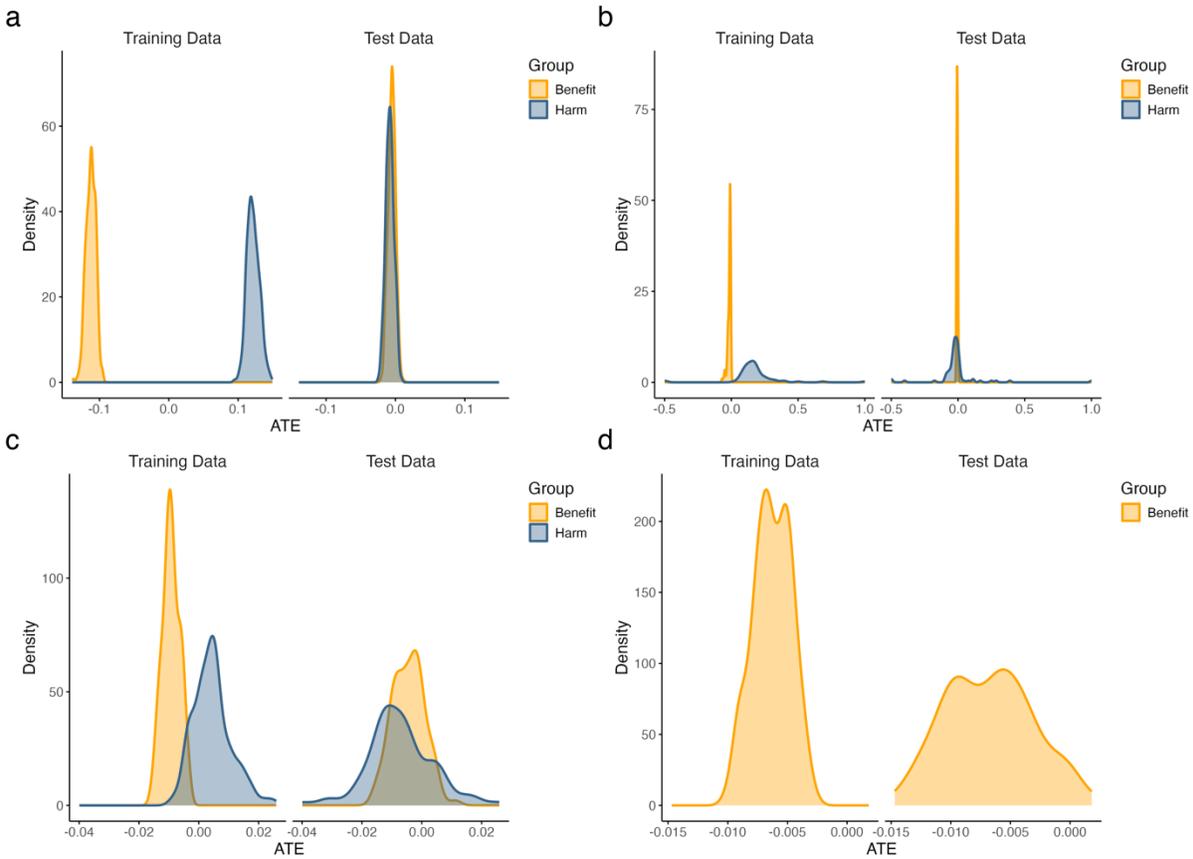

Figure S4.10.2. Internal validation on the combined IST and CAST dataset: density comparative analysis of causal machine learning-based individualized treatment effects. Density plots depict the distributions of ATE in training data and test data stratified by benefit (orange) and harm (blue) groups. The distribution of ATE comes from 1000 (**b**) or 100 (**a, c, d**) random train-test splits experiments. Benefit means negative ITE values and harm means positive ITE values. The outcome variable is death at 14 days for IST and death at 4 weeks for CAST. **a**, DR-learner Random Forest. **b**, Causal Forest. **c**, CVAE. **d**, GANITE. Abbreviations: IST: the International Stroke Trial, CAST: the Chinese Acute Stroke Trial, ITE: individualized treatment effect, ATE: average treatment effect.

## Results of external validation – IST as training data and CAST as test data

Table S3. Summary of quantitative validation metrics on external datasets (IST as training data and CAST as test data) of 17 causal machine learning methods. The outcome variable is death at 14 days for IST and death at 4 weeks for CAST.

| | | c-for-benefit (training) | c-for-benefit (test) | mbcb (training) | mbcb (test) | Calibration-based pseudo R-squared (training) | Calibration-based pseudo R-squared (test) |
|---|---|---|---|---|---|---|---|
| **T-learner** | Logistic Regression | 0.538 | 0.504 | 0.546 | 0.547 | 0.684 | -2.740 |
| | Random Forest | 0.998 | 0.486 | 0.773 | 0.672 | 0.223 | -11.254 |
| | Support Vector Machine | 0.534 | 0.514 | 0.607 | 0.615 | -0.976 | -8.961 |
| | XGBoost | 0.717 | 0.480 | 0.622 | 0.627 | 0.546 | -4.505 |
| | Penalized Logistic Regression | 0.500 | 0.500 | - | - | - | - |
| **S-learner** | Logistic Regression | 0.503 | 0.529 | 0.509 | 0.508 | -0.401 | 0.326 |
| | Penalized Logistic Regression | 0.500 | 0.500 | - | - | - | - |
| | XGBoost | 0.724 | 0.502 | 0.564 | 0.566 | 0.272 | -4.479 |
| | BART | 0.551 | 0.488 | 0.523 | 0.527 | 0.299 | -0.496 |
| **X-learner** | Random Forest | 0.976 | 0.484 | - | - | 0.142 | -3.295 |
| | BART | 0.602 | 0.492 | - | - | 0.394 | -1.678 |
| **DR-learner** | Random Forest | 0.813 | 0.485 | - | - | 0.657 | -7.144 |
| **Tree-based methods** | Causal Forest | 0.665 | 0.484 | - | - | 0.094 | -1.374 |
| | Bayesian Causal Forest | 0.515 | 0.485 | - | - | 0.072 | -0.118 |
| | model-based recursive partitioning | 0.802 | 0.481 | - | - | - | - |
| **Deep Learning** | CVAE | 0.527 | 0.523 | 0.525 | 0.525 | 0.467 | 0.492 |
| | GANITE | 0.495 | 0.515 | 0.532 | 0.528 | -17.126 | -21.885 |

Notes: Consistent values across training and test data that are closer to 1 indicate a better model fit. The dash symbol indicates instances where:

- The validation metric could not be assessed due to the model's mechanism.
- The model failed to estimate valid individualized treatment effects.
- The metric was not applicable to the model.

Abbreviations: IST: the International Stroke Trial, CAST: the Chinese Acute Stroke Trial.

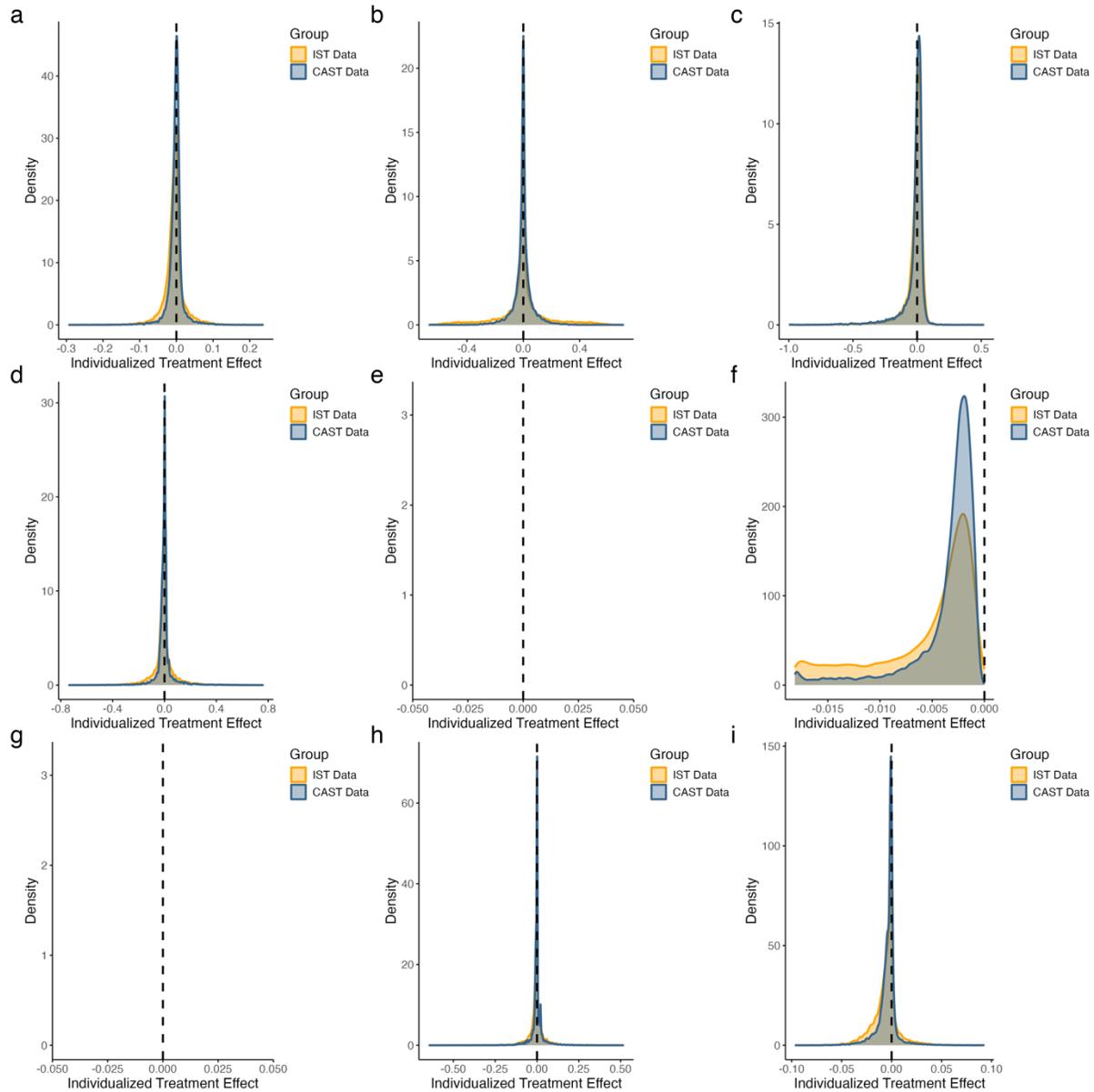

Figure S5.1.1. External validation with IST as training data and CAST as test data: density plots of causal machine learning-based individualized treatment effects. Orange represents the training data, and blue indicates the test data. The outcome variable is death at 14 days for IST and death at 4 weeks for CAST. **a**, T-learner Logistic Regression. **b**, T-learner Random Forest. **c**, T-learner Support Vector Machine. **d**, T-learner XGBoost. **e**, T-learner Penalized Logistic Regression. **f**, S-learner Logistic Regression. **g**, S-learner Penalized Logistic Regression. **h**, S-learner XGBoost. **i**, S-learner BART. Abbreviations: IST: the International Stroke Trial, CAST: the Chinese Acute Stroke Trial, ITE: individualized treatment effect.

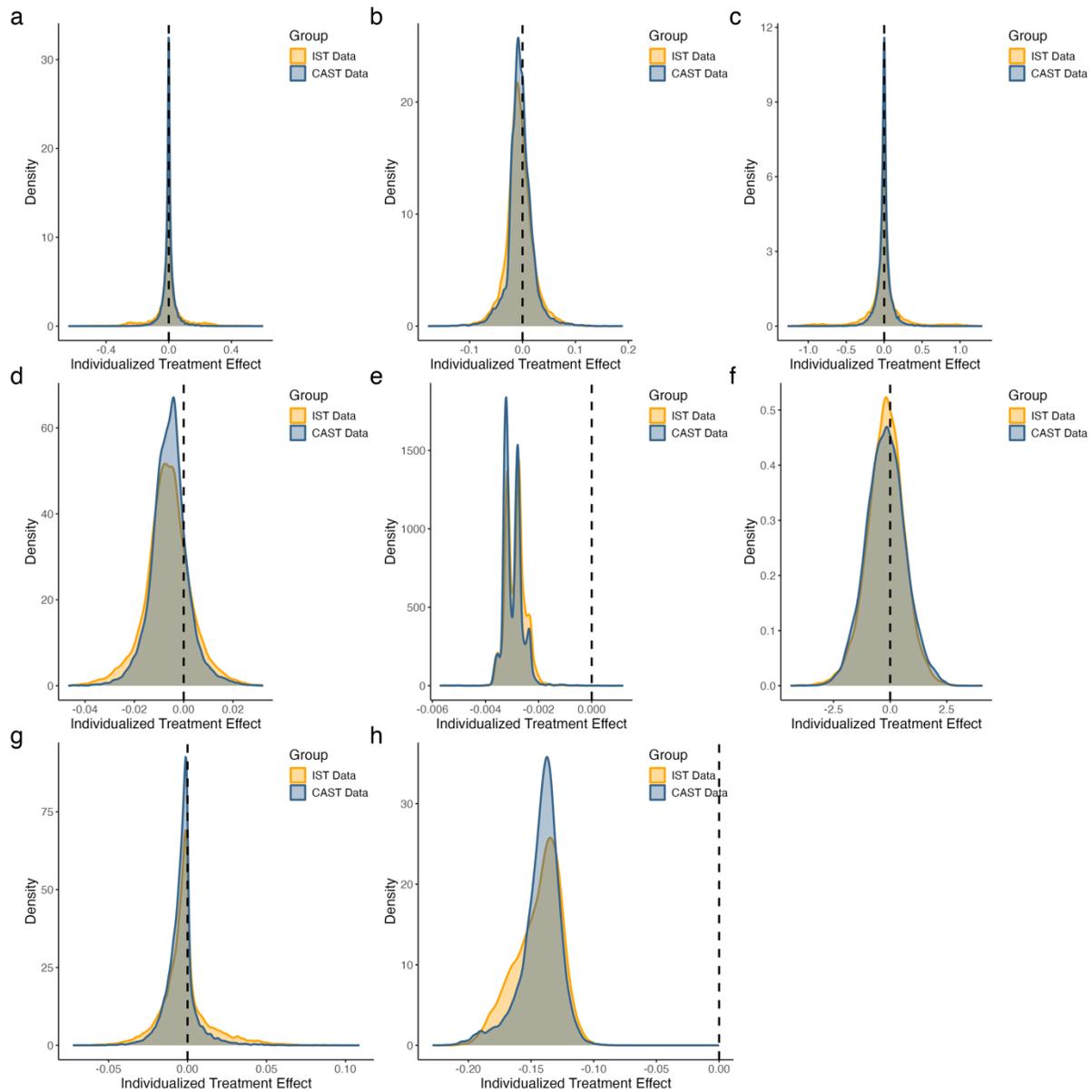

Figure S5.1.2. External validation with IST as training data and CAST as test data: density plots of causal machine learning-based individualized treatment effects. Orange represents the training data, and blue indicates the test data. The outcome variable is death at 14 days for IST and death at 4 weeks for CAST. **a**, X-learner Random Forest. **b**, X-learner BART. **c**, DR-learner Random Forest. **d**, Causal Forest. **e**, Bayesian Causal Forest. **f**, Model-based Recursive Partitioning. **g**, CVAE. **h**, GANITE. Abbreviations: IST: the International Stroke Trial, CAST: the Chinese Acute Stroke Trial, ITE: individualized treatment effect.

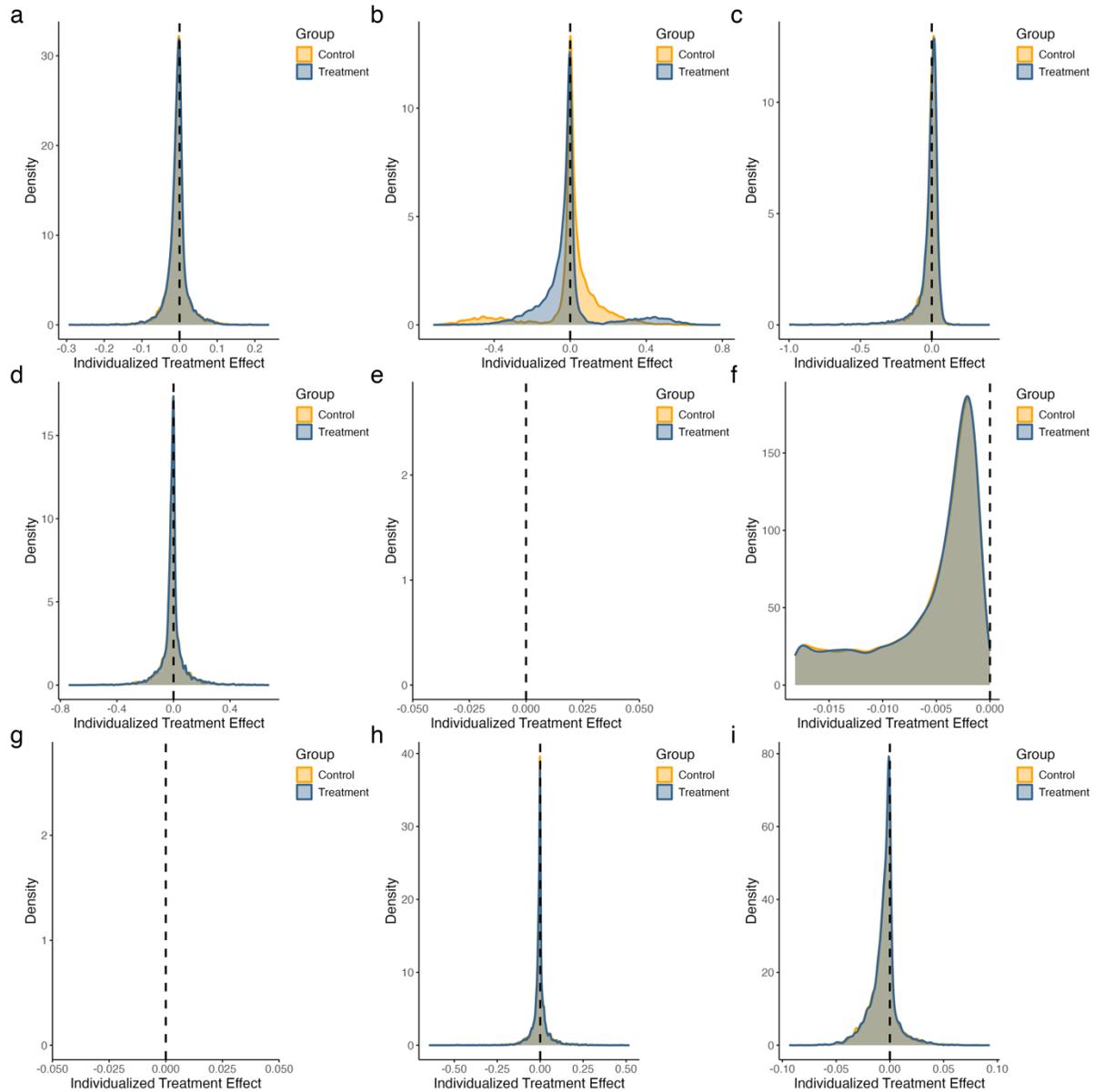

Figure S5.2.1. IST as training dataset: density plots of causal machine learning-based individualized treatment effects. Orange represents the control group, and blue indicates the treatment group. The outcome variable is death at 14 days for IST and death at 4 weeks for CAST. **a**, T-learner Logistic Regression. **b**, T-learner Random Forest. **c**, T-learner Support Vector Machine. **d**, T-learner XGBoost. **e**, T-learner Penalized Logistic Regression. **f**, S-learner Logistic Regression. **g**, S-learner Penalized Logistic Regression. **h**, S-learner XGBoost. **i**, S-learner BART. Abbreviations: IST: the International Stroke Trial, CAST: the Chinese Acute Stroke Trial, ITE: individualized treatment effect.

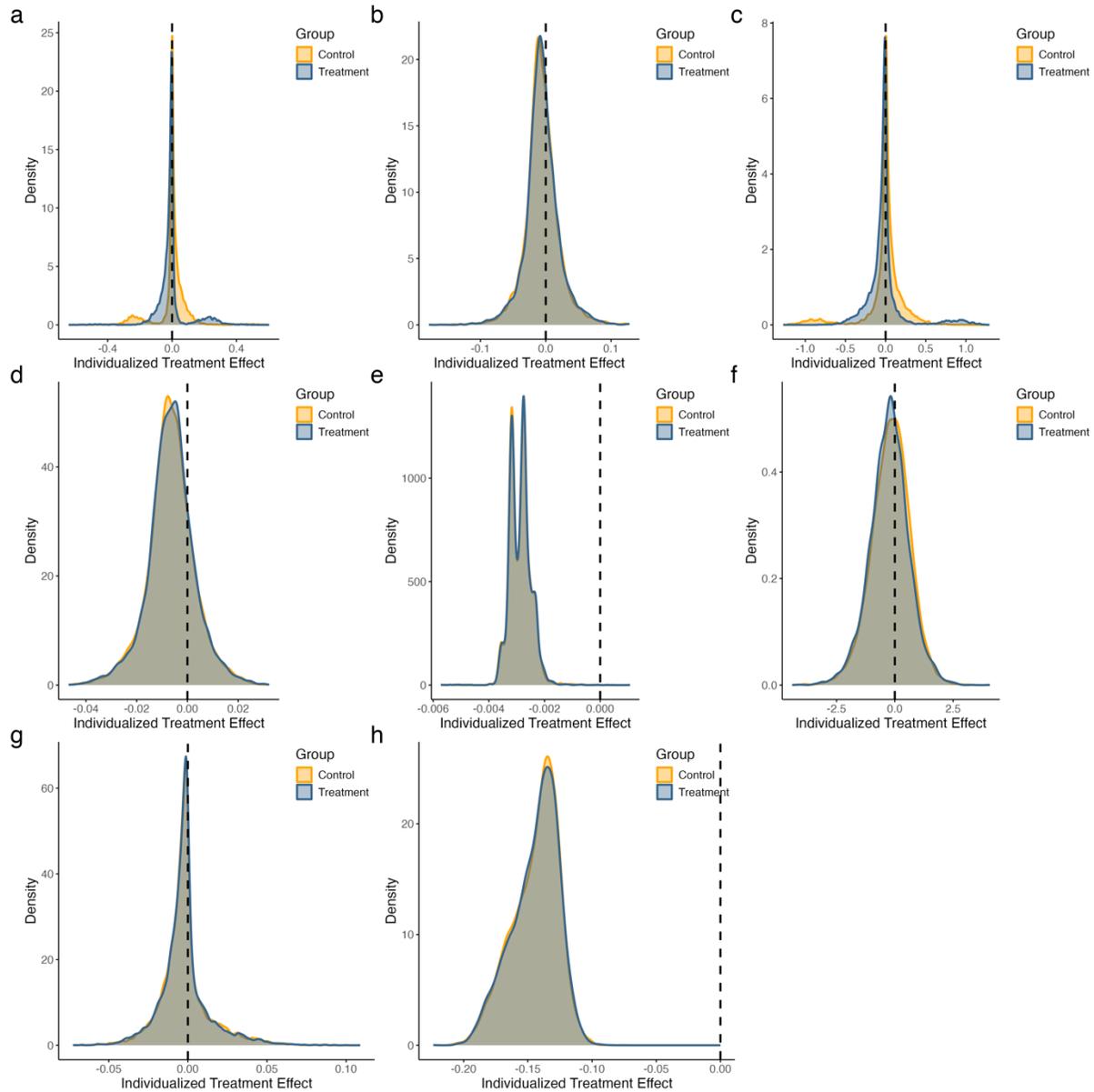

Figure S5.2.2. IST as training dataset: density plots of causal machine learning-based individualized treatment effects. Orange represents the control group, and blue indicates the treatment group. The outcome variable is death at 14 days for IST and death at 4 weeks for CAST. **a**, X-learner Random Forest. **b**, X-learner BART. **c**, DR-learner Random Forest. **d**, Causal Forest. **e**, Bayesian Causal Forest. **f**, Model-based Recursive Partitioning. **g**, CVAE. **h**, GANITE. Abbreviations: IST: the International Stroke Trial, CAST: the Chinese Acute Stroke Trial, ITE: individualized treatment effect.

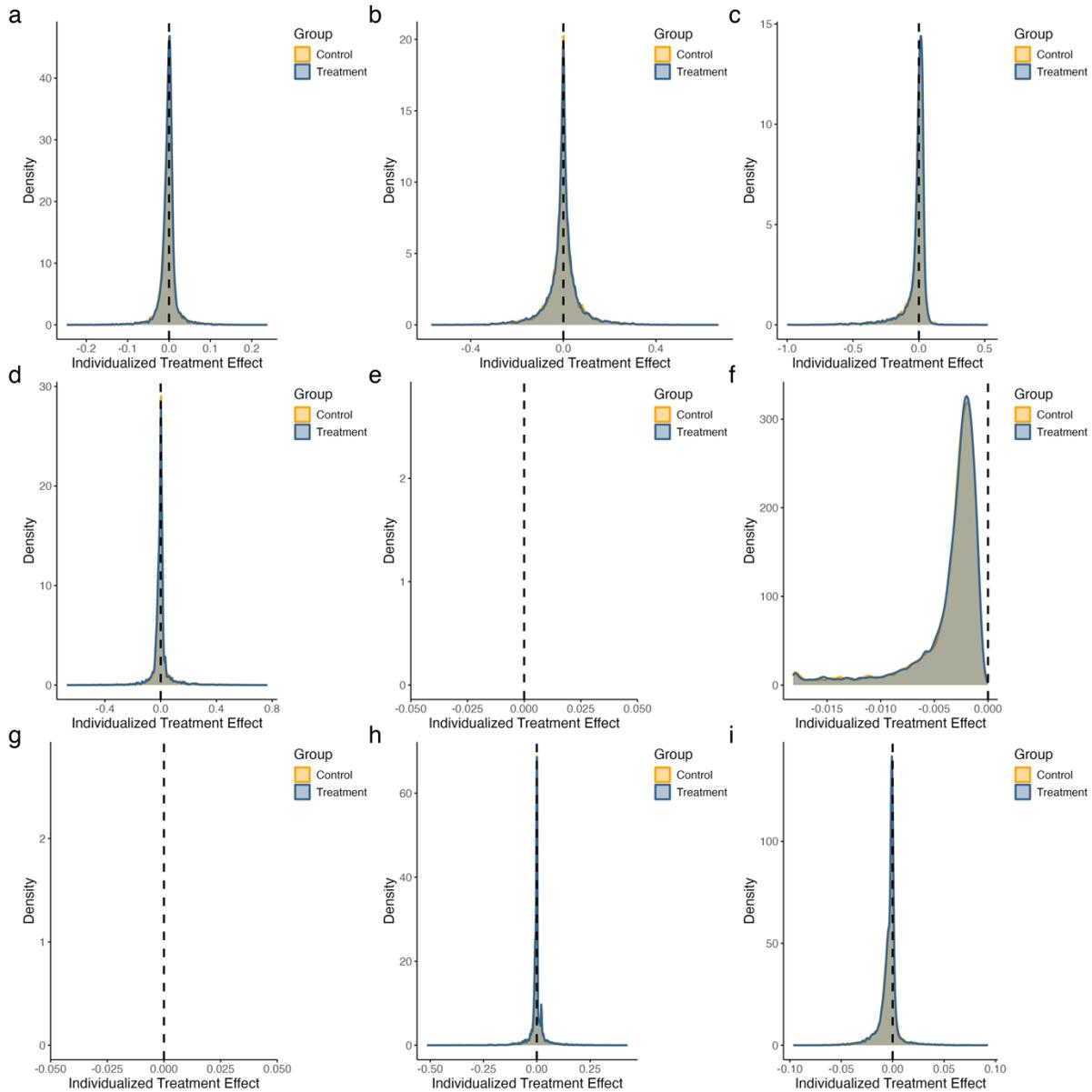

Figure S5.3.1. CAST as test dataset: density plots of causal machine learning-based individualized treatment effects. Orange represents the control group, and blue indicates the treatment group. The outcome variable is death at 14 days for IST and death at 4 weeks for CAST. **a**, T-learner Logistic Regression. **b**, T-learner Random Forest. **c**, T-learner Support Vector Machine. **d**, T-learner XGBoost. **e**, T-learner Penalized Logistic Regression. **f**, S-learner Logistic Regression. **g**, S-learner Penalized Logistic Regression. **h**, S-learner XGBoost. **i**, S-learner BART. Abbreviations: IST: the International Stroke Trial, CAST: the Chinese Acute Stroke Trial, ITE: individualized treatment effect.

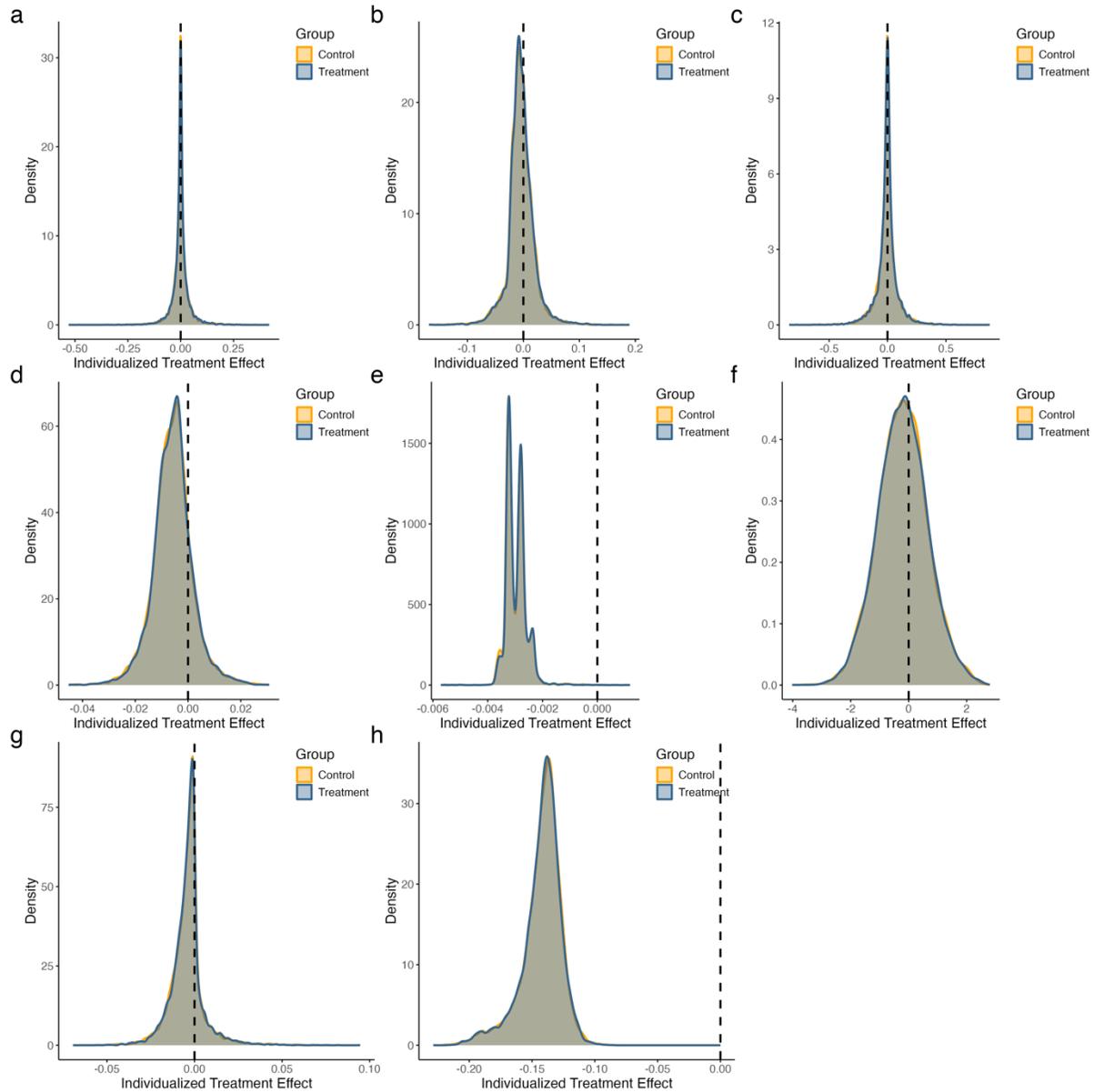

Figure S5.3.2. CAST as test dataset: density plots of causal machine learning-based individualized treatment effects. Orange represents the control group, and blue indicates the treatment group. The outcome variable is death at 14 days for IST and death at 4 weeks for CAST. **a**, X-learner Random Forest. **b**, X-learner BART. **c**, DR-learner Random Forest. **d**, Causal Forest. **e**, Bayesian Causal Forest. **f**, Model-based Recursive Partitioning. **g**, CVAE. **h**, GANITE. Abbreviations: IST: the International Stroke Trial, CAST: the Chinese Acute Stroke Trial, ITE: individualized treatment effect.

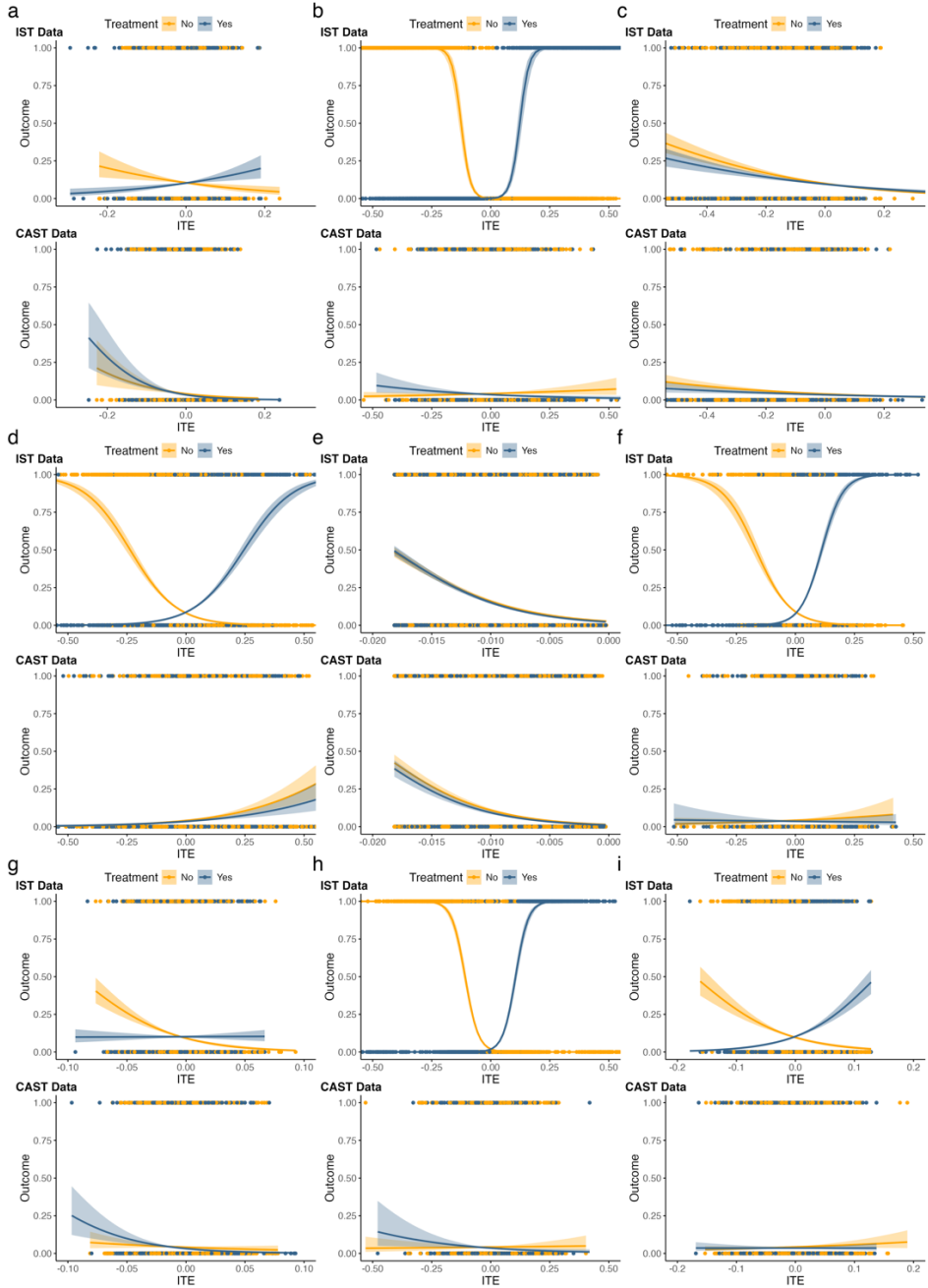

Figure S5.4.1. External validation with IST as training data and CAST as test data: outcome-ITE comparative analysis. Dot plots and line plots depict the true and fitted patient outcomes against estimated ITE values between training data and test data. Orange represents the control group, and blue indicates the treatment group. The outcome variable is death at 14 days for IST and death at 4 weeks for CAST. **a**, T-learner Logistic Regression. **b**, T-learner Random Forest. **c**, T-learner Support Vector Machine. **d**, T-learner XGBoost. **e**, S-learner Logistic Regression. **f**, S-learner XGBoost. **g**, S-learner BART. **h**, X-learner Random Forest. **i**, X-learner BART. Abbreviations: IST: the International Stroke Trial, CAST: the Chinese Acute Stroke Trial, ITE: individualized treatment effect.

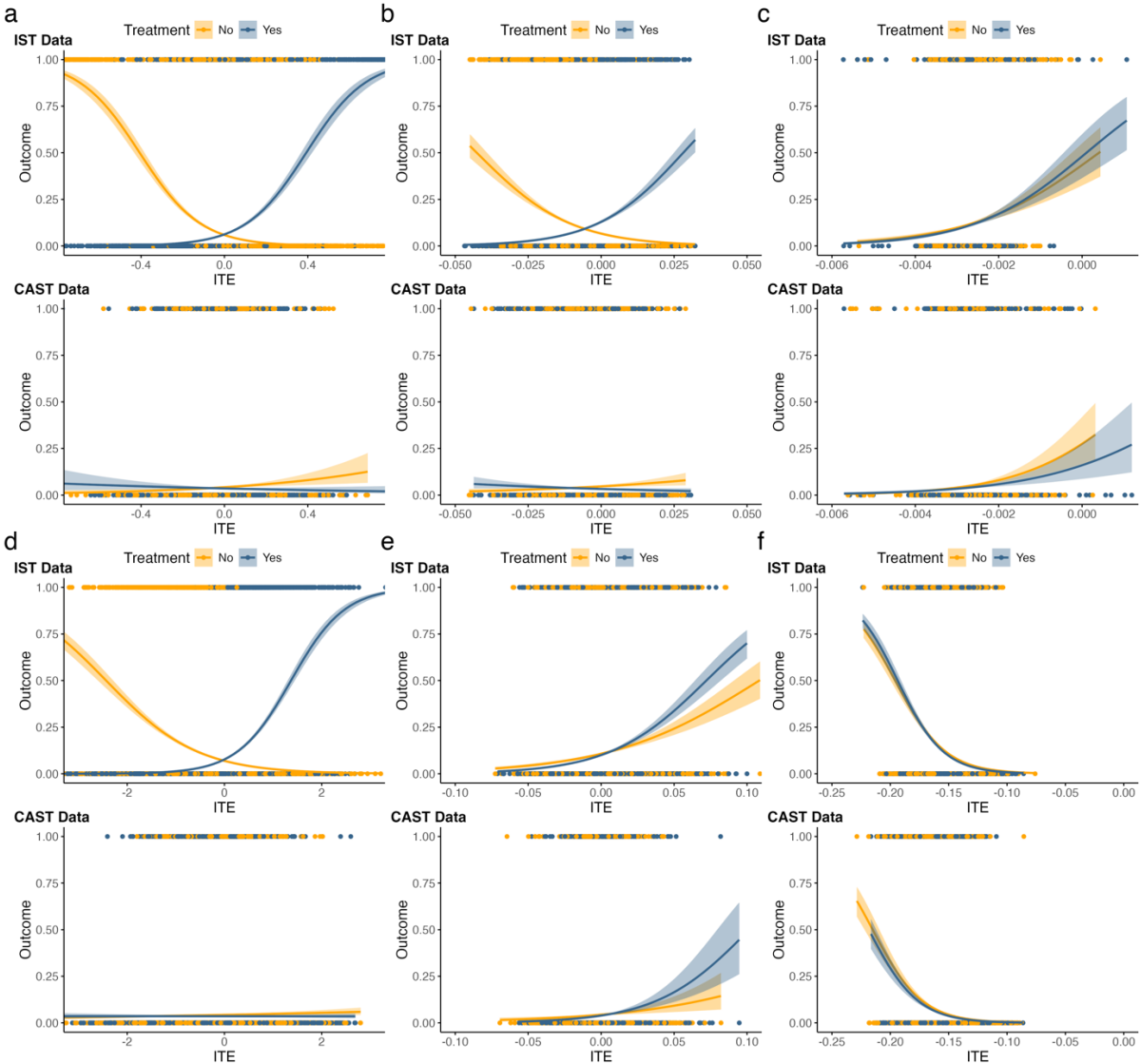

Figure S5.4.2. External validation with IST as training data and CAST as test data: outcome-ITE comparative analysis. Dot plots and line plots depict the true and fitted patient outcomes against estimated ITE values between training data and test data. Orange represents the control group, and blue indicates the treatment group. The outcome variable is death at 14 days for IST and death at 4 weeks for CAST. **a**, DR-learner Random Forest. **b**, Causal Forest. **c**, Bayesian Causal Forest. **d**, Model-based Recursive Partitioning. **e**, CVAE. **f**, GANITE. Abbreviations: IST: the International Stroke Trial, CAST: the Chinese Acute Stroke Trial, ITE: individualized treatment effect.

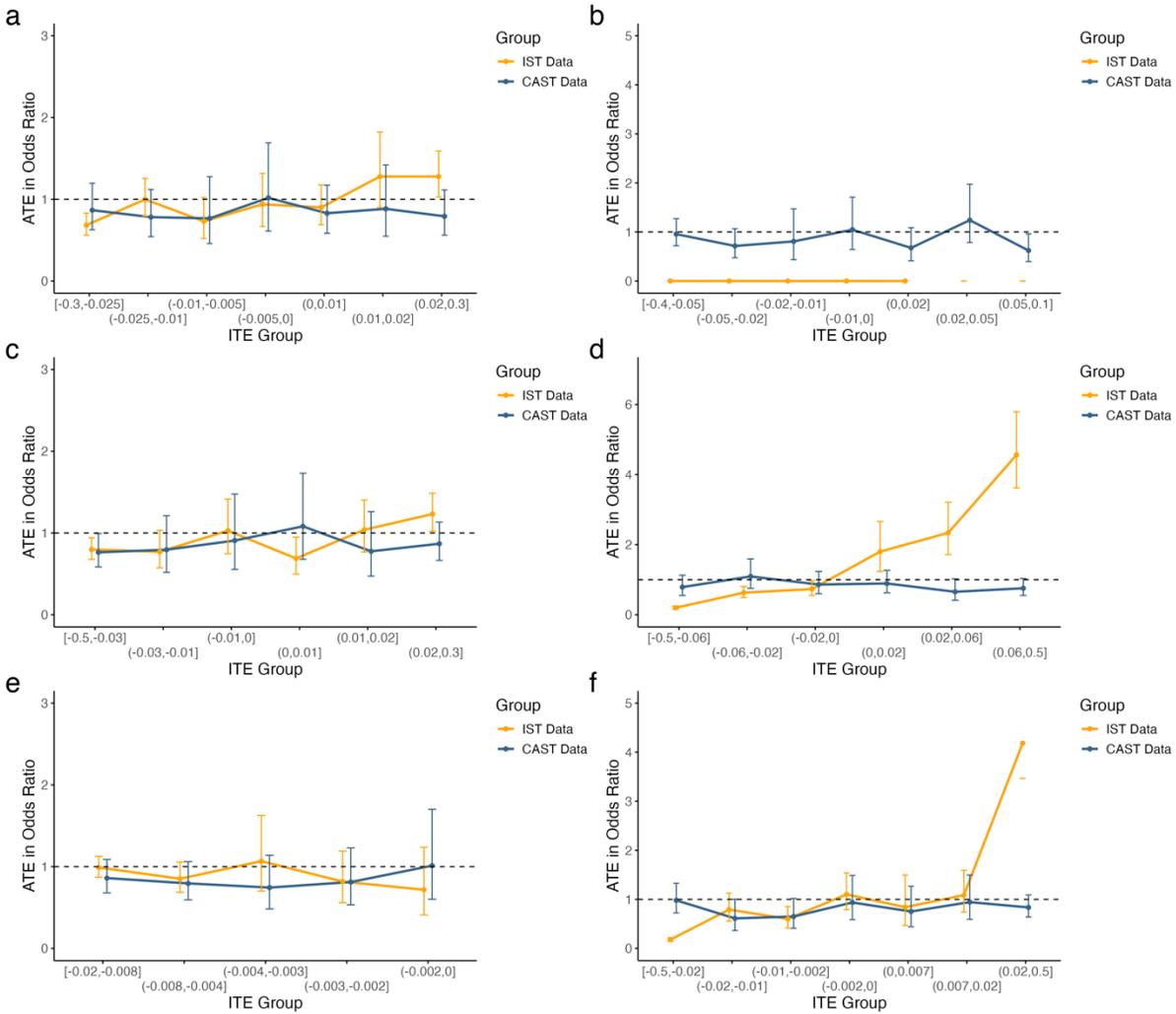

Figure S5.5.1. External validation with IST as training data and CAST as test data: ATE-ITE comparative analysis. Line plots depict ATE in risk ratio within different ITE subgroups and provide the confidence intervals at 95% level. Orange represents the training data, and blue indicates the test data. The horizontal dashed line at 1.0 means no treatment effects. The outcome variable is death at 14 days for IST and death at 4 weeks for CAST. **a**, T-learner Logistic Regression. **b**, T-learner Random Forest. **c**, T-learner Support Vector Machine. **d**, T-learner XGBoost. **e**, S-learner Logistic Regression. **f**, S-learner XGBoost. Abbreviations: IST: the International Stroke Trial, CAST: the Chinese Acute Stroke Trial, ITE: individualized treatment effect, ATE: average treatment effect.

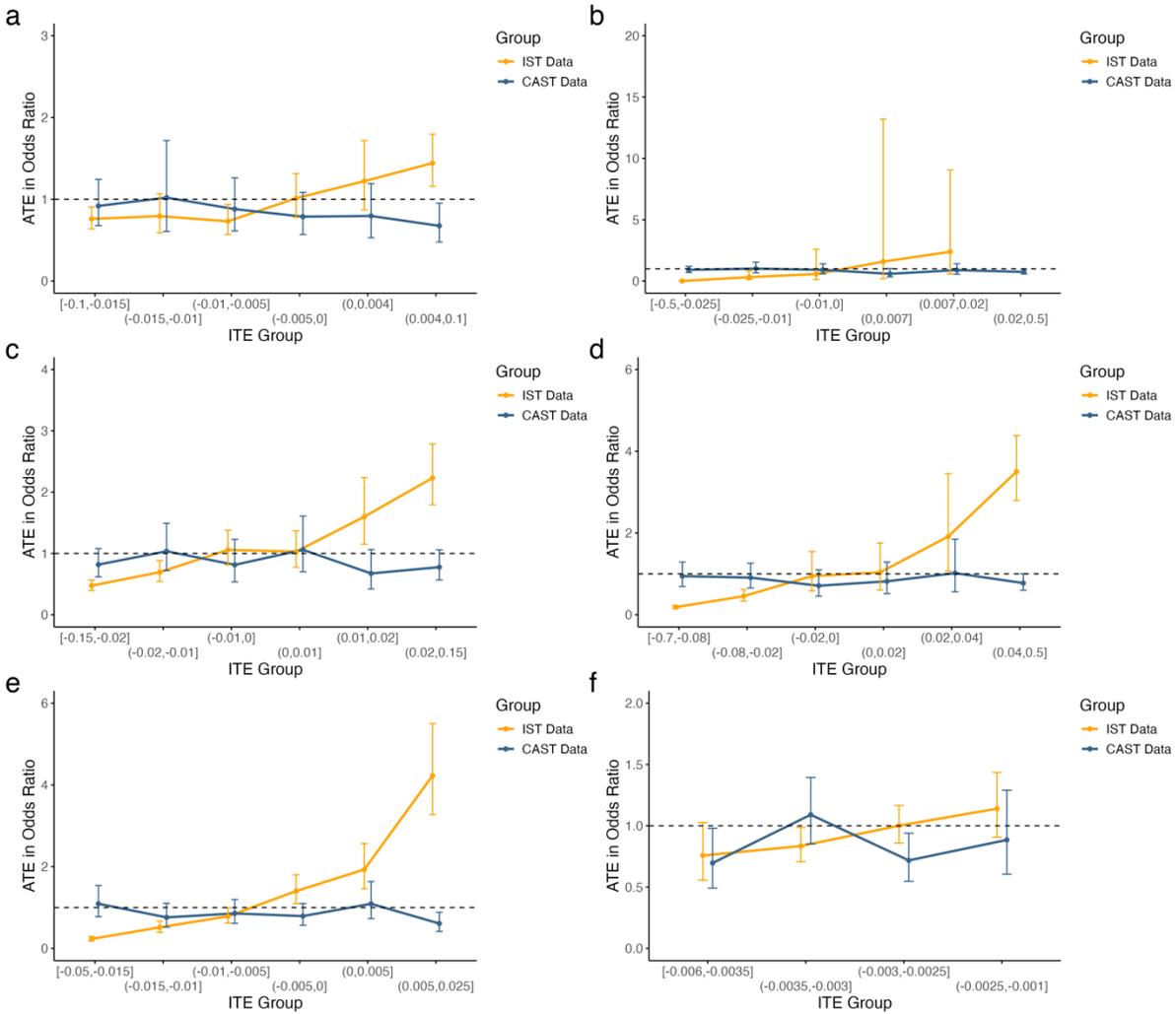

Figure S5.5.2. External validation with IST as training data and CAST as test data: ATE-ITE comparative analysis. Line plots depict ATE in risk ratio within different ITE subgroups and provide the confidence intervals at 95% level. Orange represents the training data, and blue indicates the test data. The horizontal dashed line at 1.0 means no treatment effects. The outcome variable is death at 14 days for IST and death at 4 weeks for CAST. **a**, S-learner BART. **b**, X-learner Random Forest. **c**, X-learner BART. **d**, DR-learner Random Forest. **e**, Causal Forest. **f**, Bayesian Causal Forest. Abbreviations: IST: the International Stroke Trial, CAST: the Chinese Acute Stroke Trial, ITE: individualized treatment effect, ATE: average treatment effect.

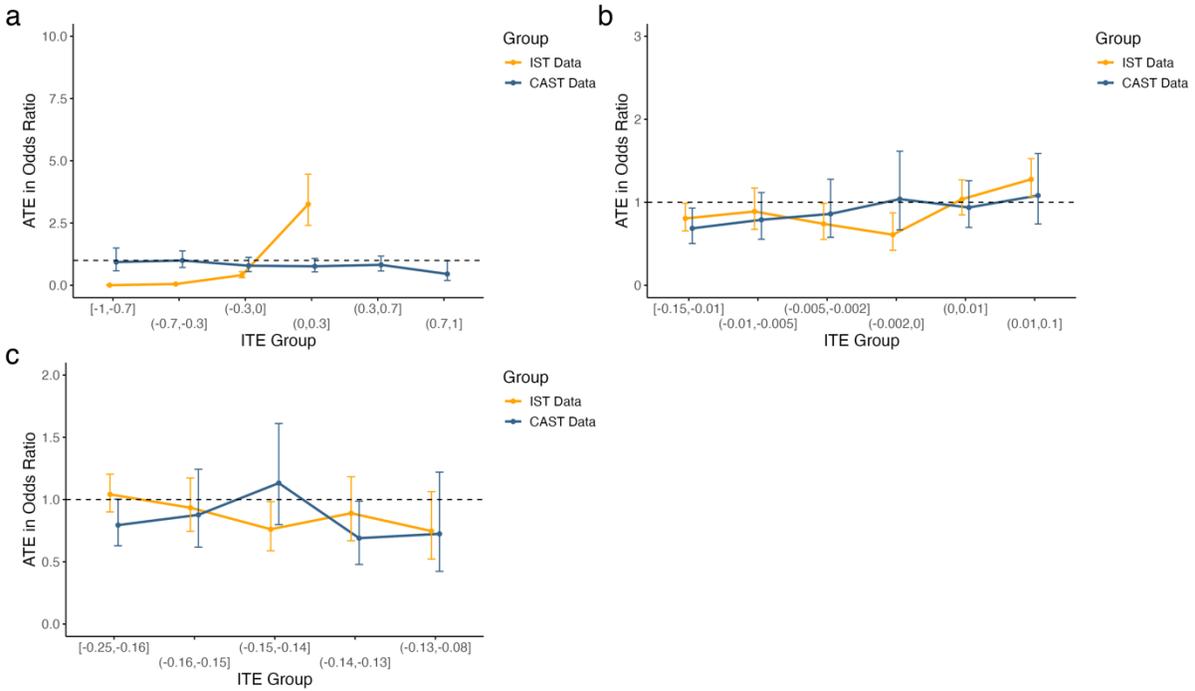

Figure S5.5.2. External validation with IST as training data and CAST as test data: ATE-ITE comparative analysis. Line plots depict ATE in risk ratio within different ITE subgroups and provide the confidence intervals at 95% level. Orange represents the training data, and blue indicates the test data. The horizontal dashed line at 1.0 means no treatment effects. The outcome variable is death at 14 days for IST and death at 4 weeks for CAST. **a**, Model-based Recursive Partitioning. **b**, CVAE. **c**, GANITE. Abbreviations: IST: the International Stroke Trial, CAST: the Chinese Acute Stroke Trial, ITE: individualized treatment effect, ATE: average treatment effect.

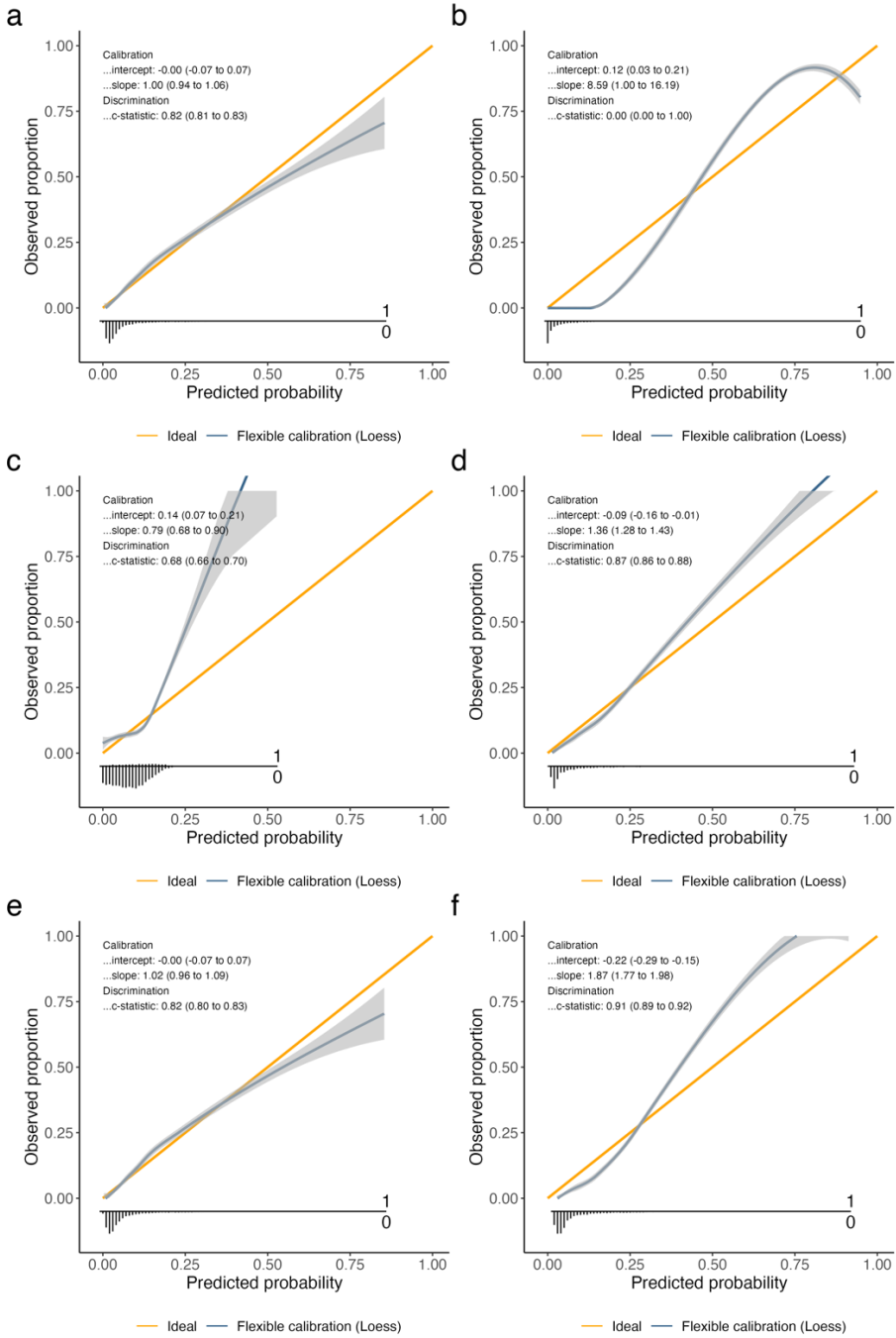

Figure S5.6.1. IST as training dataset: calibration plot of predicted outcome in the treatment group. The orange line indicates ideal calibration. The outcome variable is death at 14 days for IST and death at 4 weeks for CAST. **a**, T-learner Logistic Regression. **b**, T-learner Random Forest. **c**, T-learner Support Vector Machine. **d**, T-learner XGBoost. **e**, S-learner Logistic Regression. **f**, S-learner XGBoost. Abbreviations: IST: the International Stroke Trial.

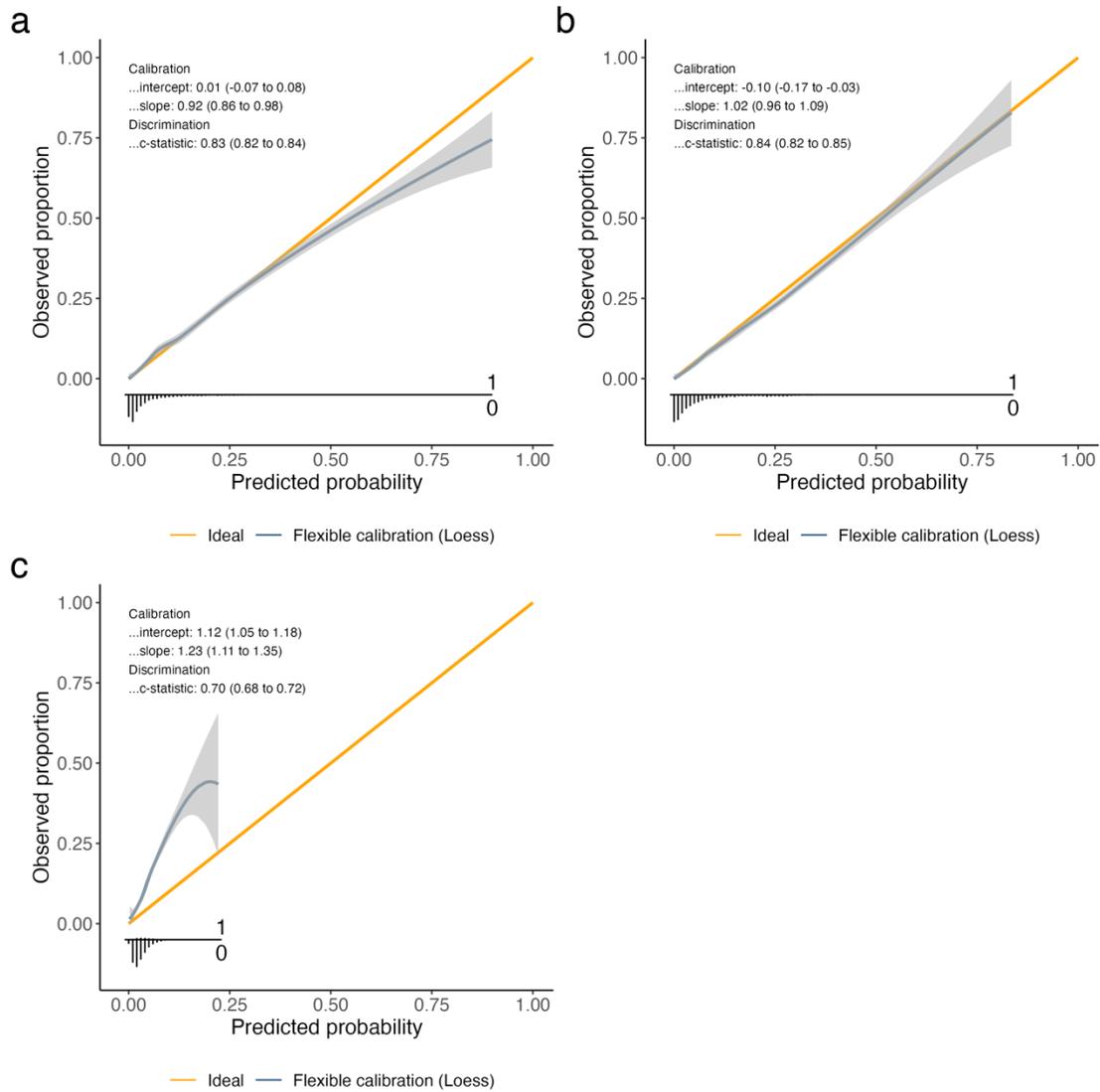

Figure S5.6.2. IST as training dataset: calibration plot of predicted outcome in the treatment group. The orange line indicates ideal calibration. The outcome variable is death at 14 days for IST and death at 4 weeks for CAST. **a**, S-learner BART. **b**, CVAE. **c**, GANITE. Abbreviations: IST: the International Stroke Trial.

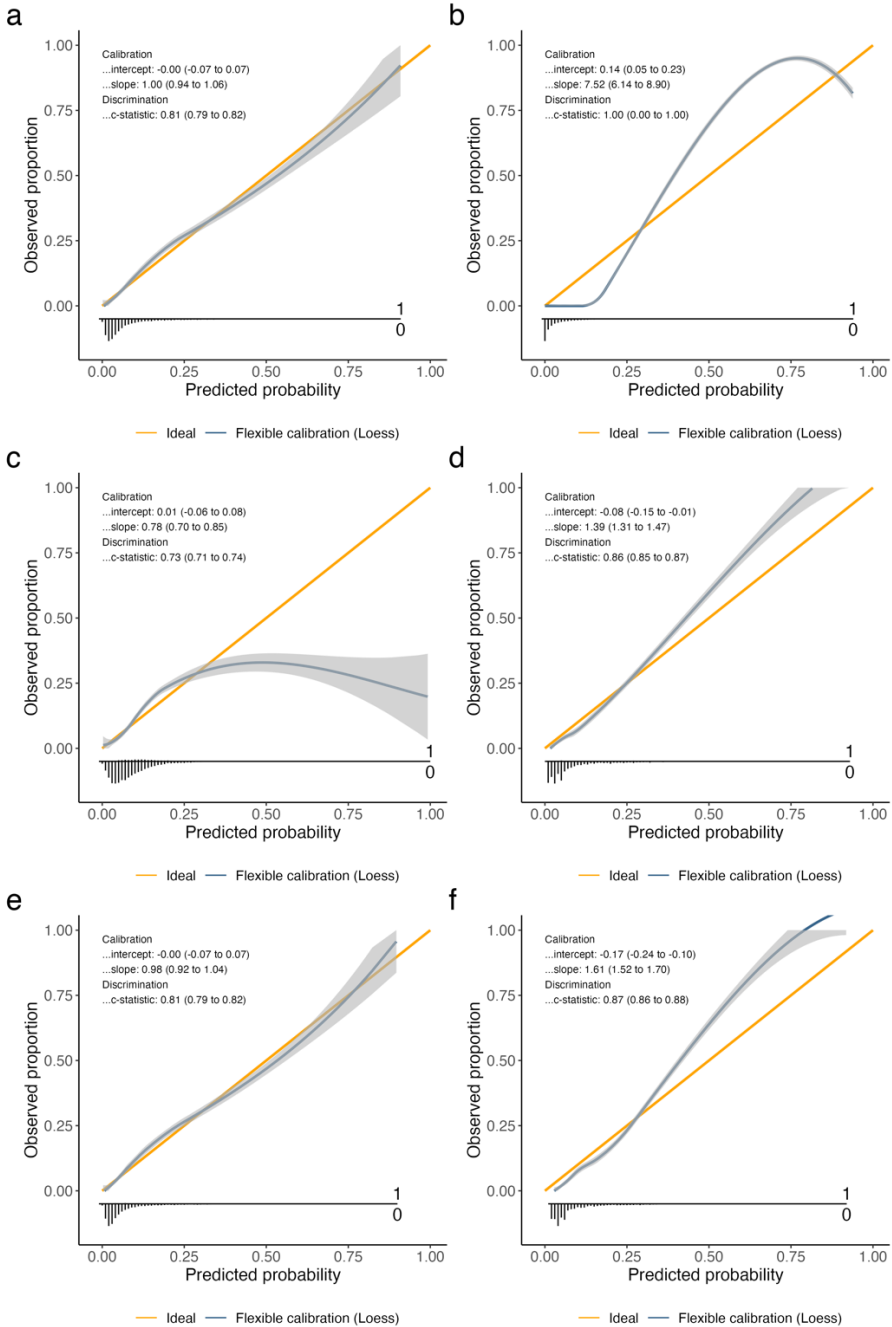

Figure S5.7.1. IST as training dataset: calibration plot of predicted outcome in the control group. The orange line indicates ideal calibration. The outcome variable is death at 14 days for IST and death at 4 weeks for CAST. **a**, T-learner Logistic Regression. **b**, T-learner Random Forest. **c**, T-learner Support Vector Machine. **d**, T-learner XGBoost. **e**, S-learner Logistic Regression. **f**, S-learner XGBoost. Abbreviations: IST: the International Stroke Trial.

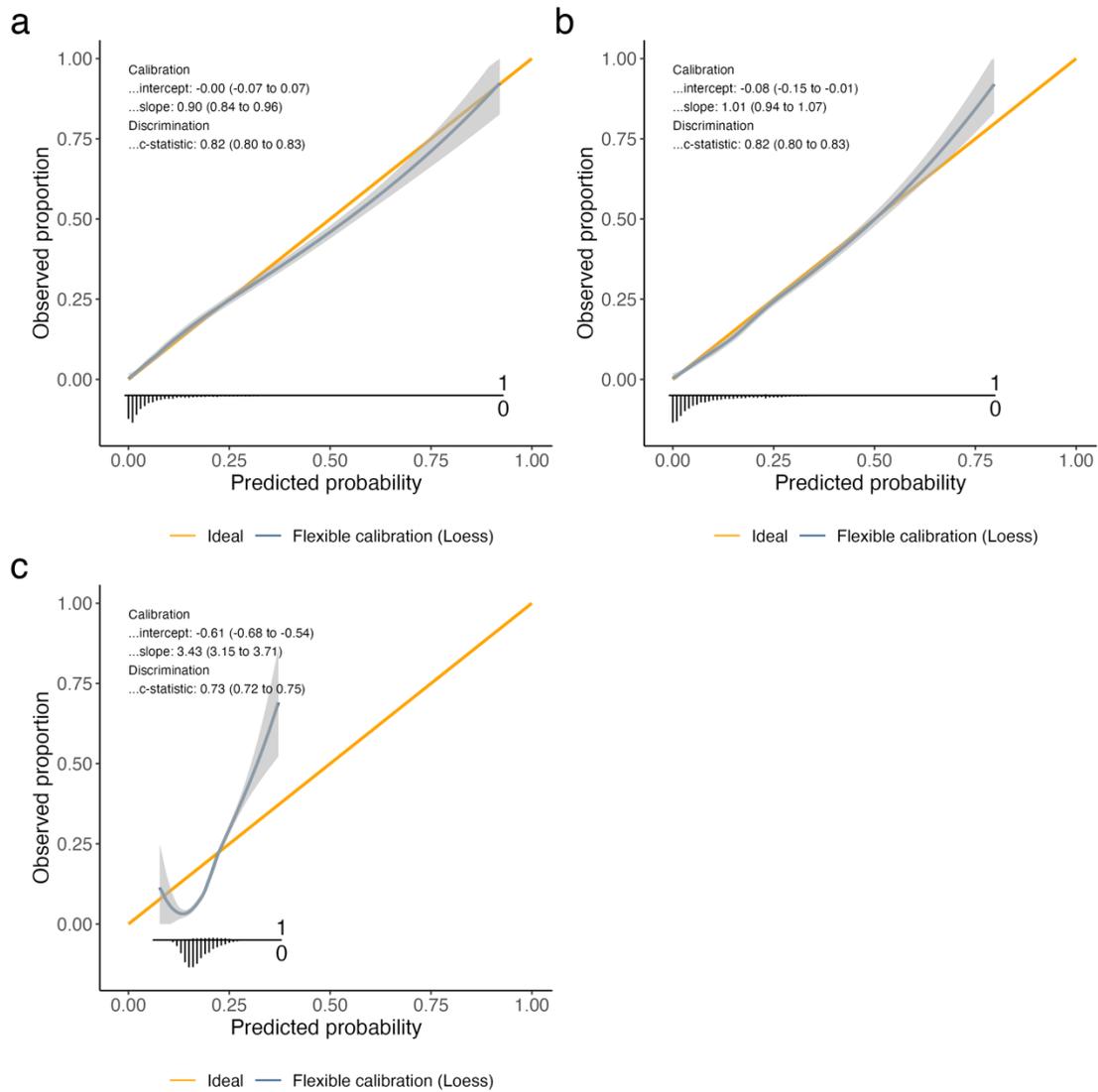

Figure S5.7.2. IST as training dataset: calibration plot of predicted outcome in the control group. The orange line indicates ideal calibration. The outcome variable is death at 14 days for IST and death at 4 weeks for CAST. **a**, S-learner BART. **b**, CVAE. **c**, GANITE. Abbreviations: IST: the International Stroke Trial.

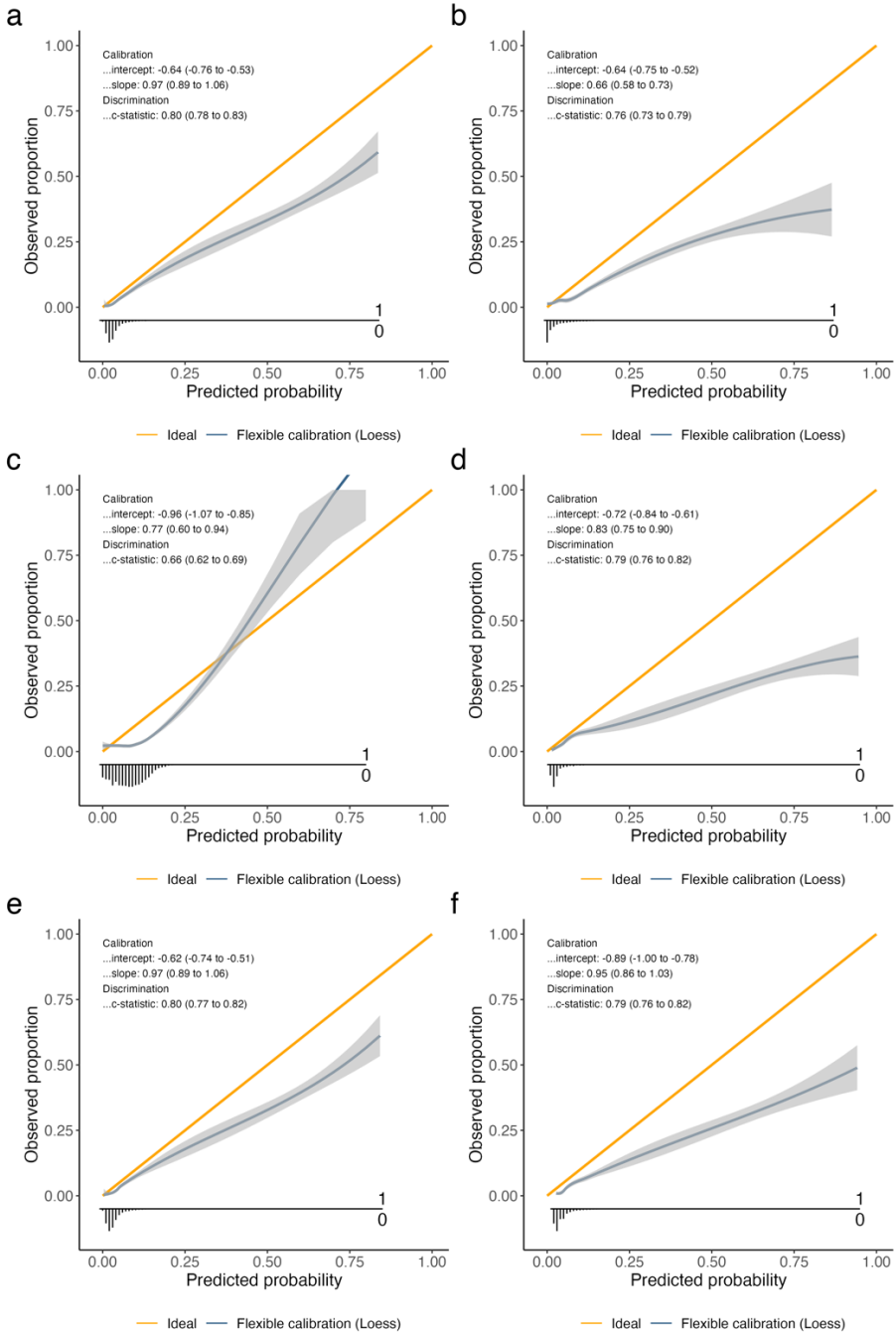

Figure S5.8.1. CAST as test dataset: calibration plot of predicted outcome in the treatment group. The orange line indicates ideal calibration. The outcome variable is death at 14 days for IST and death at 4 weeks for CAST. **a**, T-learner Logistic Regression. **b**, T-learner Random Forest. **c**, T-learner Support Vector Machine. **d**, T-learner XGBoost. **e**, S-learner Logistic Regression. **f**, S-learner XGBoost. Abbreviations: CAST: the Chinese Acute Stroke Trial.

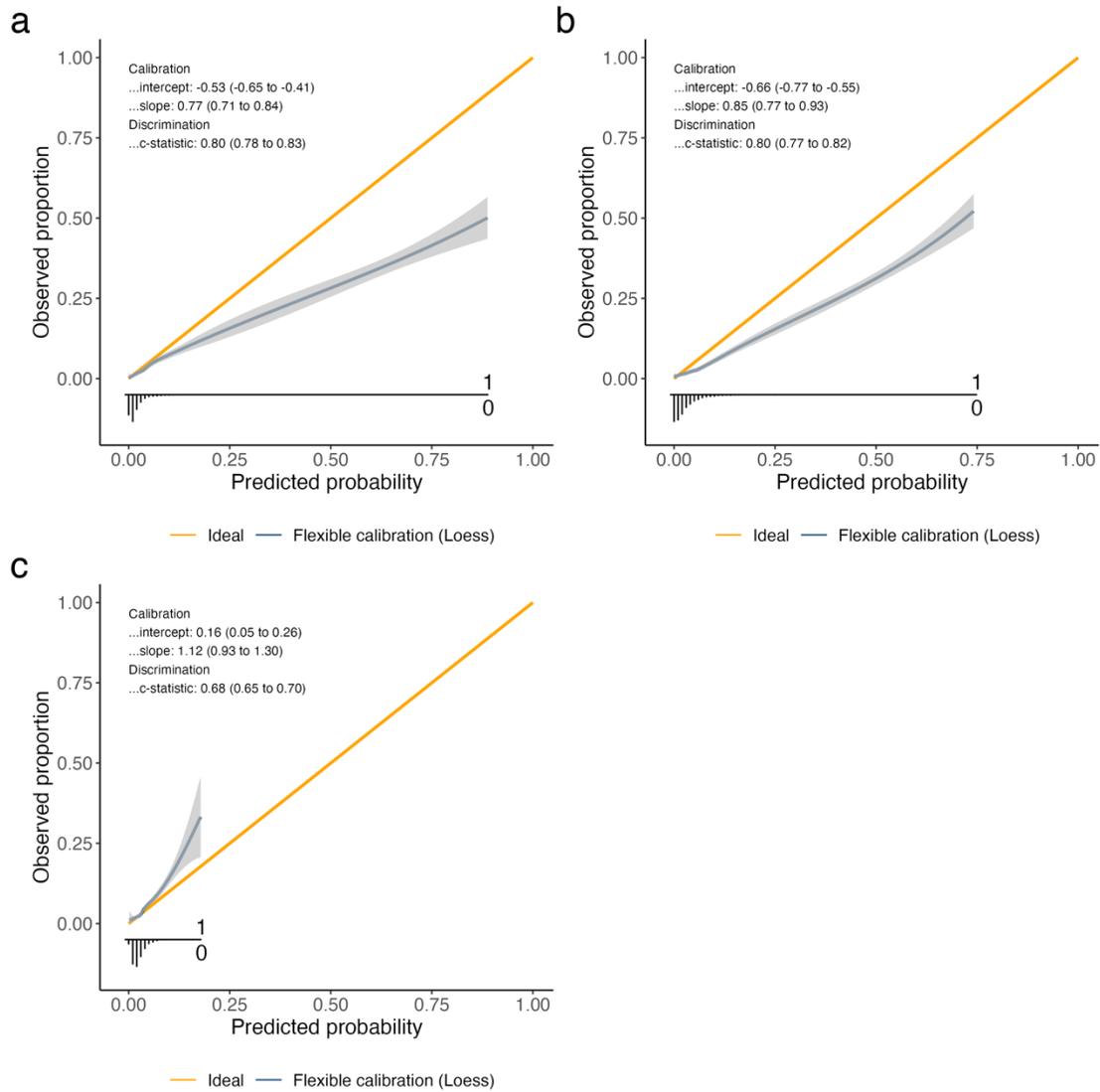

Figure S5.8.2. CAST as test dataset: calibration plot of predicted outcome in the treatment group. The orange line indicates ideal calibration. The outcome variable is death at 14 days for IST and death at 4 weeks for CAST. **a**, S-learner BART. **b**, CVAE. **c**, GANITE. Abbreviations: CAST: the Chinese Acute Stroke Trial.

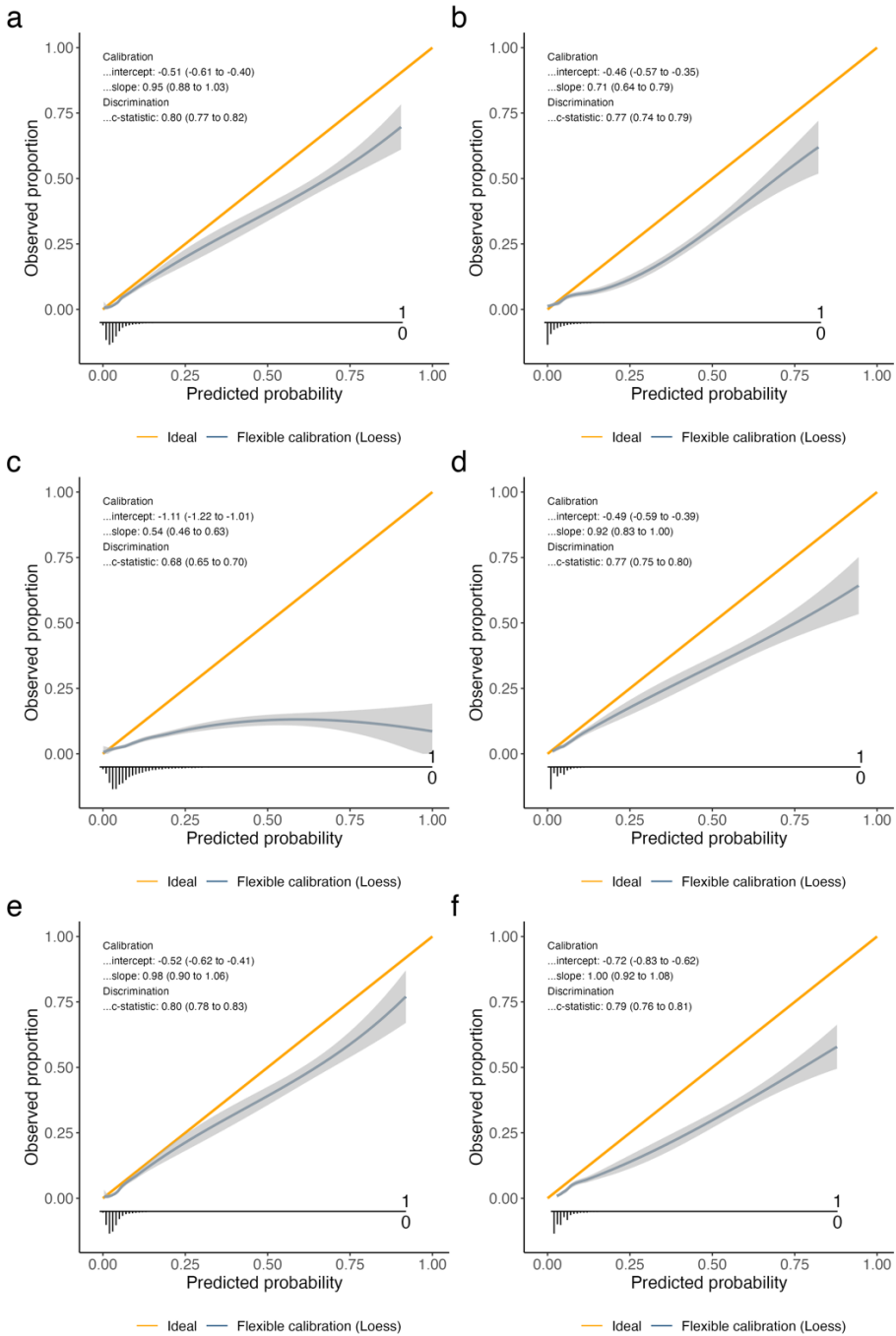

Figure S5.9.1. CAST as test dataset: calibration plot of predicted outcome in the control group. The orange line indicates ideal calibration. The outcome variable is death at 14 days for IST and death at 4 weeks for CAST. **a**, T-learner Logistic Regression. **b**, T-learner Random Forest. **c**, T-learner Support Vector Machine. **d**, T-learner XGBoost. **e**, S-learner Logistic Regression. **f**, S-learner XGBoost. Abbreviations: CAST: the Chinese Acute Stroke Trial.

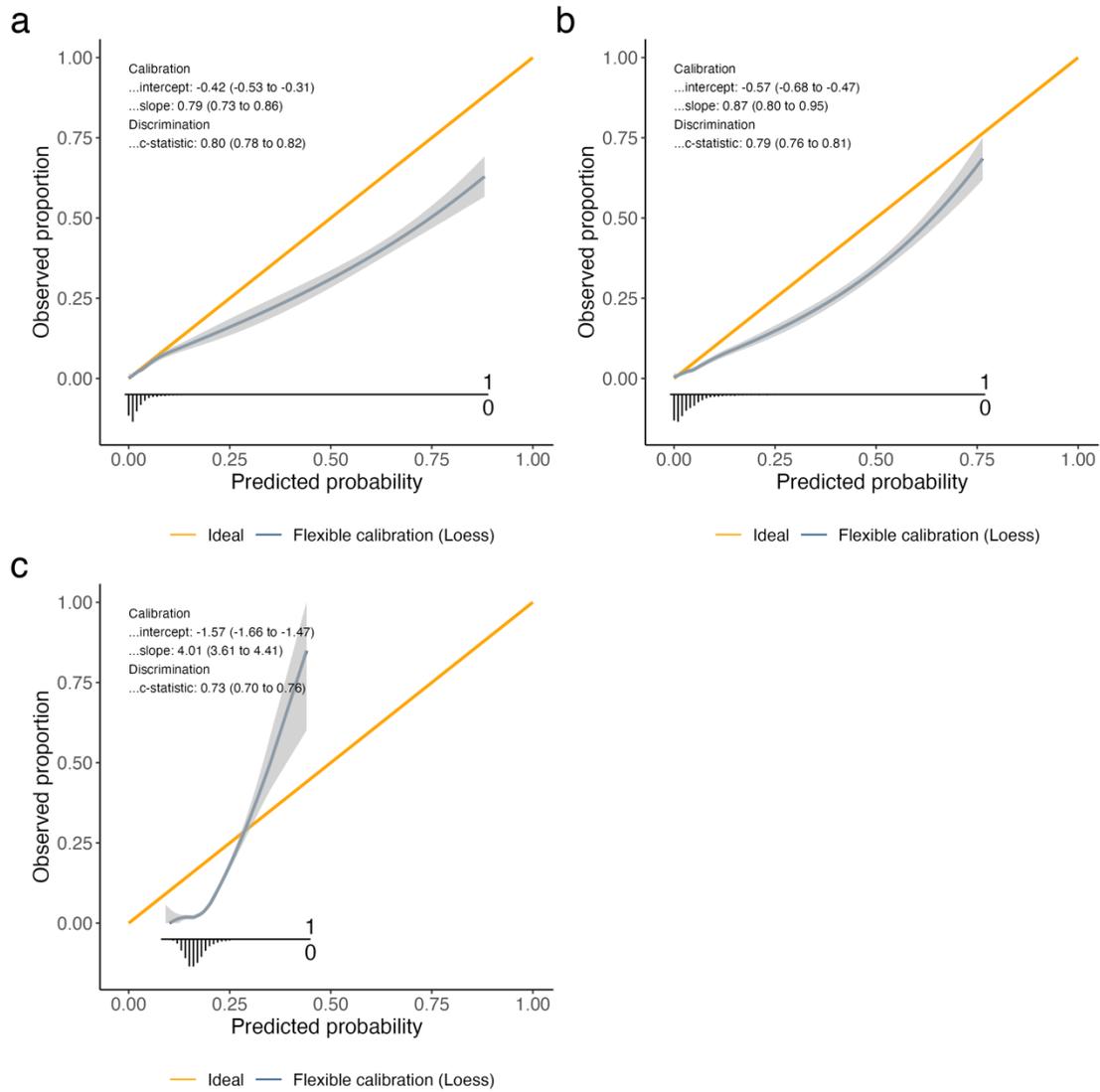

Figure S5.9.2. CAST as test dataset: calibration plot of predicted outcome in the control group. The orange line indicates ideal calibration. The outcome variable is death at 14 days for IST and death at 4 weeks for CAST. **a**, S-learner BART. **b**, CVAE. **c**, GANITE. Abbreviations: CAST: the Chinese Acute Stroke Trial.

## Results of external validation – CAST as training data and IST as test data

Table S4. Summary of quantitative validation metrics on external datasets (CAST as training data and IST as test data) of 17 causal machine learning methods. The outcome variable is death at 14 days for IST and death at 4 weeks for CAST.

|  |  | c-for-benefit (training) | c-for-benefit (test) | mbcb (training) | mbcb (test) | Calibration-based pseudo R-squared (training) | Calibration-based pseudo R-squared (test) |
|---|---|---|---|---|---|---|---|
| **T-learner** | Logistic Regression | 0.576 | 0.498 | 0.572 | 0.592 | 0.635 | -3.824 |
|  | Random Forest | 0.993 | 0.497 | 0.806 | 0.655 | 0.270 | -8.679 |
|  | Support Vector Machine | 0.523 | 0.497 | 0.578 | 0.580 | -0.349 | -3.363 |
|  | XGBoost | 0.803 | 0.504 | 0.628 | 0.679 | 0.482 | -9.664 |
|  | Penalized Logistic Regression | 0.500 | 0.500 | - | - | - | - |
| **S-learner** | Logistic Regression | 0.530 | 0.500 | 0.525 | 0.528 | 0.635 | -1.047 |
|  | Penalized Logistic Regression | 0.500 | 0.500 | - | - | - | - |
|  | XGBoost | 0.790 | 0.493 | 0.566 | 0.593 | 0.325 | -3.351 |
|  | BART | 0.564 | 0.500 | 0.548 | 0.561 | 0.683 | -4.418 |
| **X-learner** | Random Forest | 0.994 | 0.504 | - | - | 0.188 | -2.441 |
|  | BART | 0.644 | 0.492 | - | - | 0.534 | -3.154 |
| **DR-learner** | Random Forest | 0.776 | 0.518 | - | - | 0.778 | -4.909 |
| **Tree-based methods** | Causal Forest | 0.600 | 0.499 | - | - | 0.190 | -0.116 |
|  | Bayesian Causal Forest | 0.570 | 0.493 | - | - | 0.640 | -1.839 |
|  | model-based recursive partitioning | - | - | - | - | - | - |
| **Deep Learning** | CVAE | 0.523 | 0.492 | 0.562 | 0.551 | -0.747 | -3.190 |
|  | GANITE | 0.518 | 0.488 | 0.531 | 0.532 | -20.976 | -13.067 |

Notes: Consistent values across training and test data that are closer to 1 indicate a better model fit. The dash symbol indicates instances where:

- The validation metric could not be assessed due to the model's mechanism.
- The model failed to estimate valid individualized treatment effects.
- The metric was not applicable to the model.

Abbreviations: IST: the International Stroke Trial, CAST: the Chinese Acute Stroke Trial.

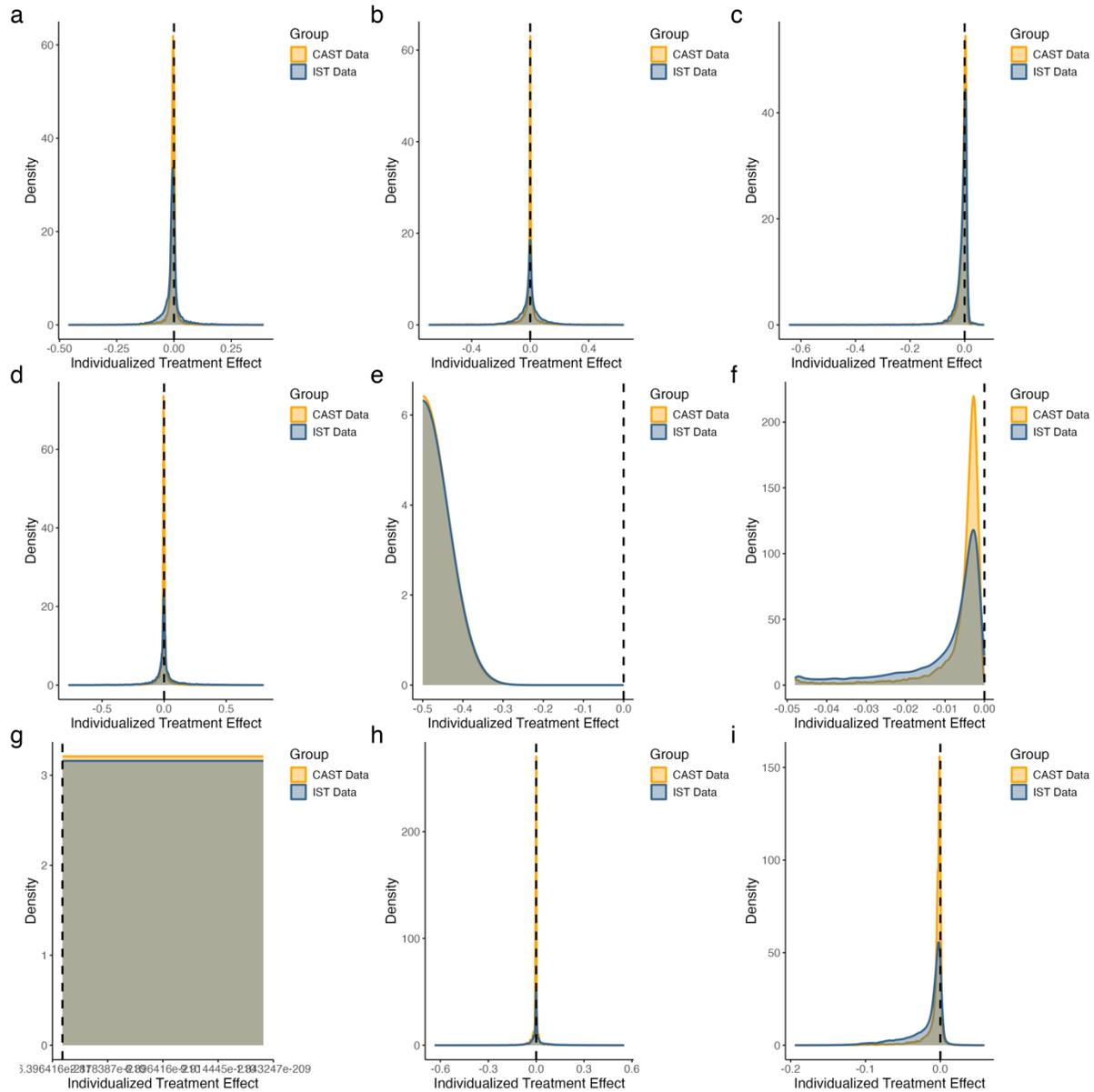

Figure S6.1.1. External validation with CAST as training data and IST as test data: density plots of causal machine learning-based individualized treatment effects. Orange represents the training data, and blue indicates the test data. The outcome variable is death at 14 days for IST and death at 4 weeks for CAST. **a**, T-learner Logistic Regression. **b**, T-learner Random Forest. **c**, T-learner Support Vector Machine. **d**, T-learner XGBoost. **e**, T-learner Penalized Logistic Regression. **f**, S-learner Logistic Regression. **g**, S-learner Penalized Logistic Regression. **h**, S-learner XGBoost. **i**, S-learner BART. Abbreviations: IST: the International Stroke Trial, CAST: the Chinese Acute Stroke Trial, ITE: individualized treatment effect.

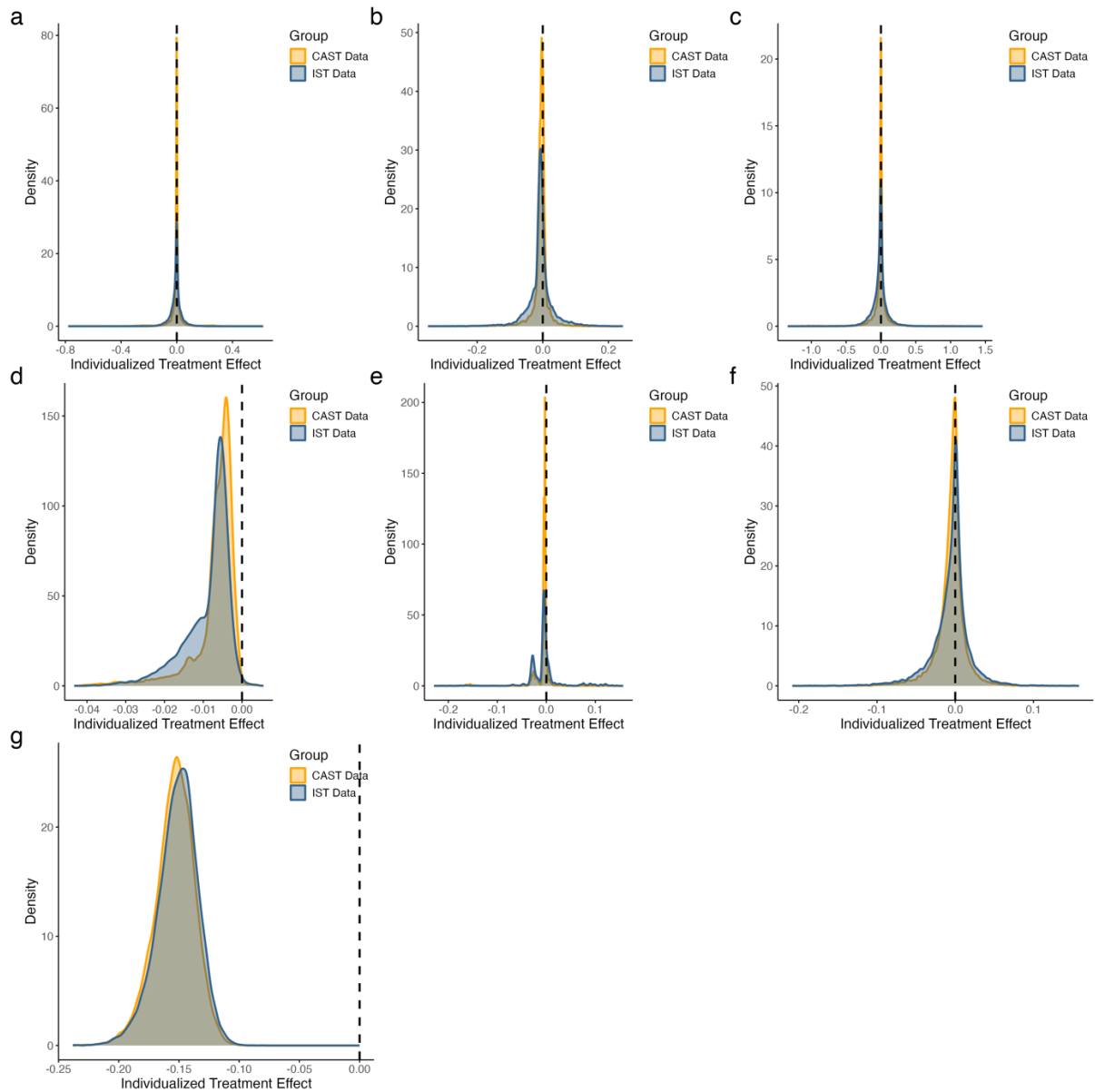

Figure S6.1.2. External validation with CAST as training data and IST as test data: density plots of causal machine learning-based individualized treatment effects. Orange represents the training data, and blue indicates the test data. The outcome variable is death at 14 days for IST and death at 4 weeks for CAST. **a**, X-learner Random Forest. **b**, X-learner BART. **c**, DR-learner Random Forest. **d**, Causal Forest. **e**, Bayesian Causal Forest. **f**, CVAE. **g**, GANITE. Abbreviations: IST: the International Stroke Trial, CAST: the Chinese Acute Stroke Trial, ITE: individualized treatment effect.

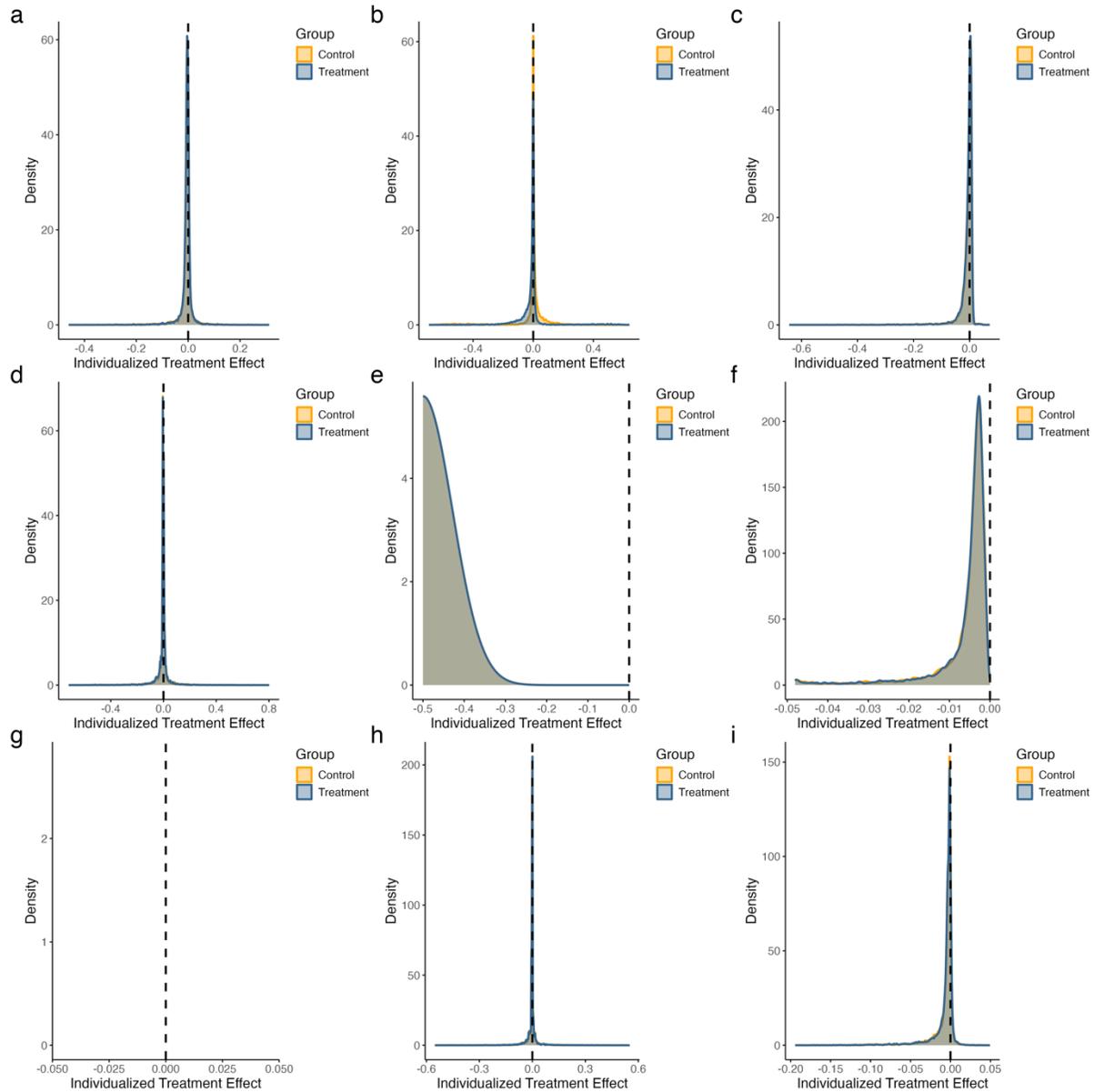

Figure S6.2.1. CAST as training dataset: density plots of causal machine learning-based individualized treatment effects. Orange represents the control group, and blue indicates the treatment group. The outcome variable is death at 14 days for IST and death at 4 weeks for CAST. **a**, T-learner Logistic Regression. **b**, T-learner Random Forest. **c**, T-learner Support Vector Machine. **d**, T-learner XGBoost. **e**, T-learner Penalized Logistic Regression. **f**, S-learner Logistic Regression. **g**, S-learner Penalized Logistic Regression. **h**, S-learner XGBoost. **i**, S-learner BART. Abbreviations: IST: the International Stroke Trial, CAST: the Chinese Acute Stroke Trial, ITE: individualized treatment effect.

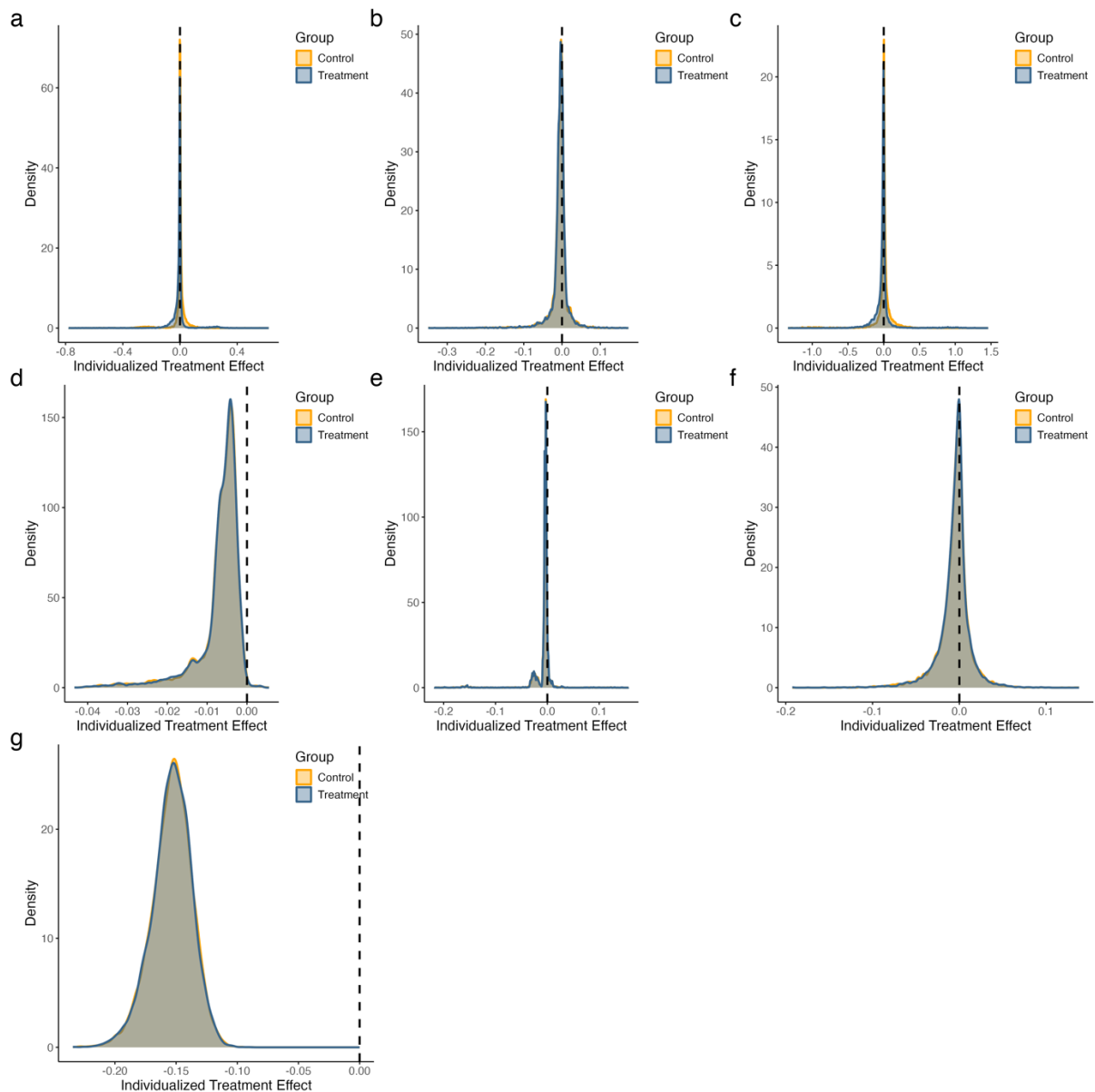

Figure S6.2.2. CAST as training dataset: density plots of causal machine learning-based individualized treatment effects. Orange represents the control group, and blue indicates the treatment group. The outcome variable is death at 14 days for IST and death at 4 weeks for CAST. **a**, X-learner Random Forest. **b**, X-learner BART. **c**, DR-learner Random Forest. **d**, Causal Forest. **e**, Bayesian Causal Forest. **f**, CVAE. **g**, GANITE. Abbreviations: IST: the International Stroke Trial, CAST: the Chinese Acute Stroke Trial, ITE: individualized treatment effect.

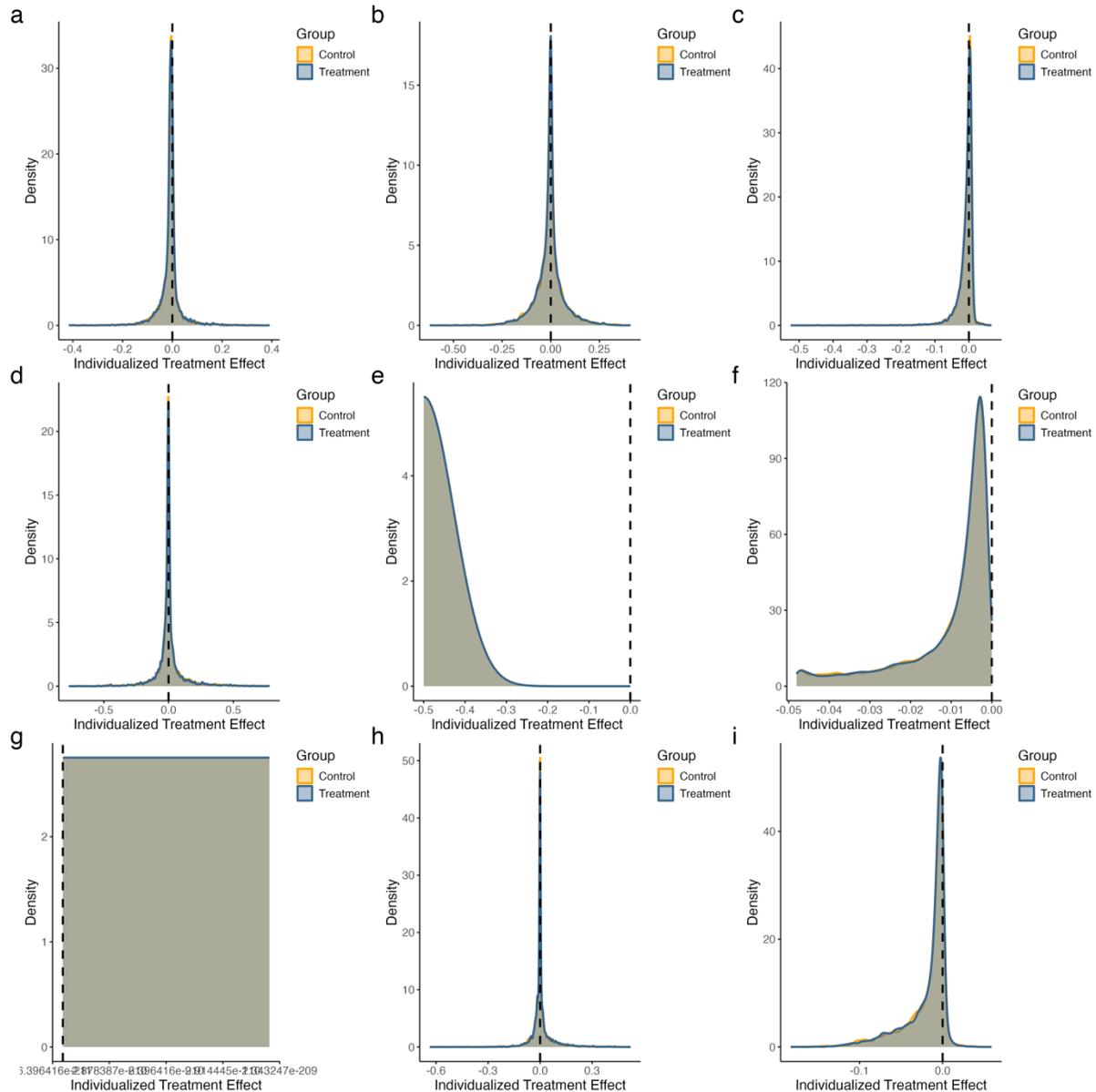

Figure S6.3.1. IST as test dataset: density plots of causal machine learning-based individualized treatment effects. Orange represents the control group, and blue indicates the treatment group. The outcome variable is death at 14 days for IST and death at 4 weeks for CAST. **a**, T-learner Logistic Regression. **b**, T-learner Random Forest. **c**, T-learner Support Vector Machine. **d**, T-learner XGBoost. **e**, T-learner Penalized Logistic Regression. **f**, S-learner Logistic Regression. **g**, S-learner Penalized Logistic Regression. **h**, S-learner XGBoost. **i**, S-learner BART. Abbreviations: IST: the International Stroke Trial, CAST: the Chinese Acute Stroke Trial, ITE: individualized treatment effect.

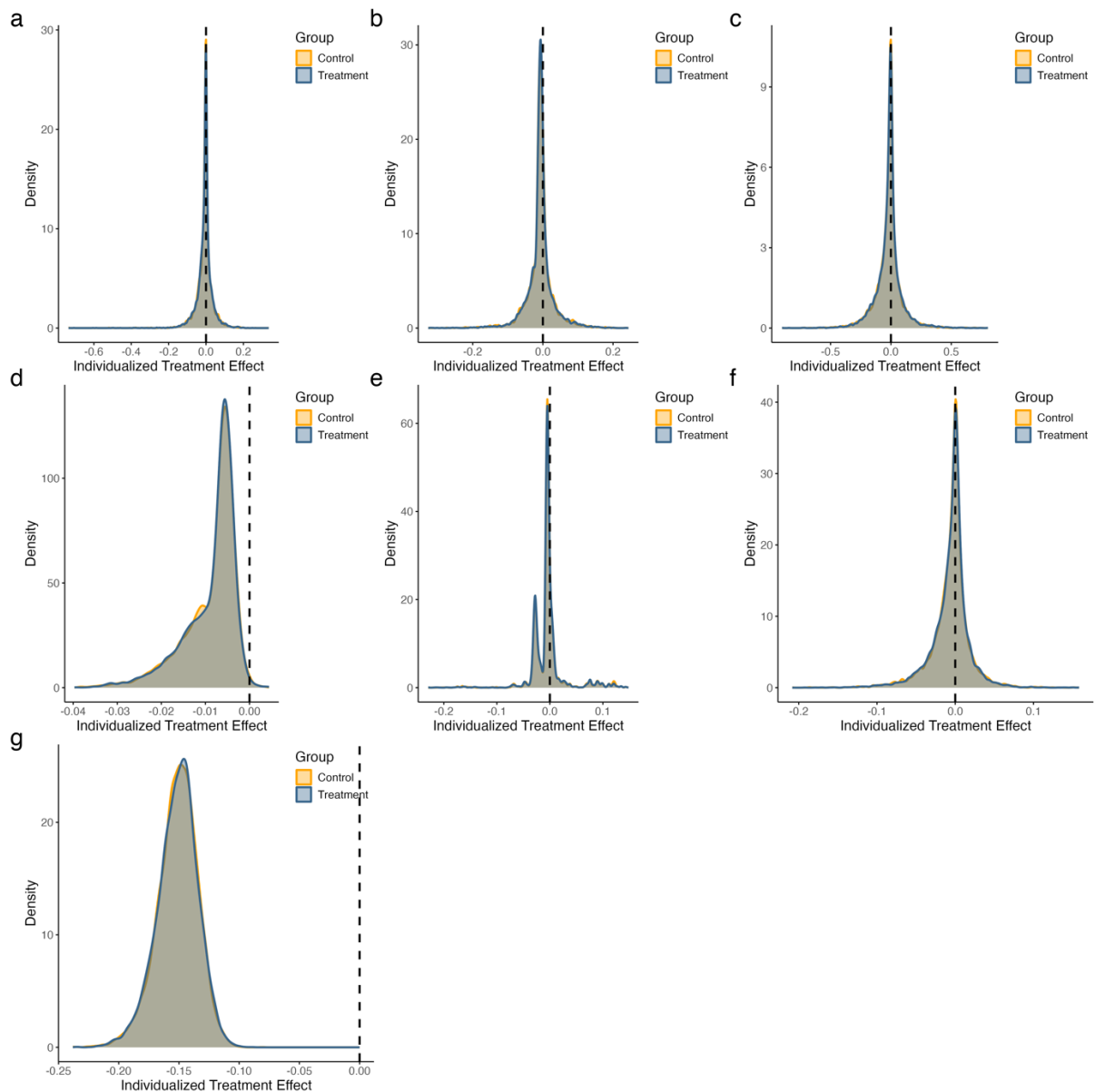

Figure S6.3.2. IST as test dataset: density plots of causal machine learning-based individualized treatment effects. Orange represents the control group, and blue indicates the treatment group. The outcome variable is death at 14 days for IST and death at 4 weeks for CAST. **a**, X-learner Random Forest. **b**, X-learner BART. **c**, DR-learner Random Forest. **d**, Causal Forest. **e**, Bayesian Causal Forest. **f**, CVAE. **g**, GANITE. Abbreviations: IST: the International Stroke Trial, CAST: the Chinese Acute Stroke Trial, ITE: individualized treatment effect.

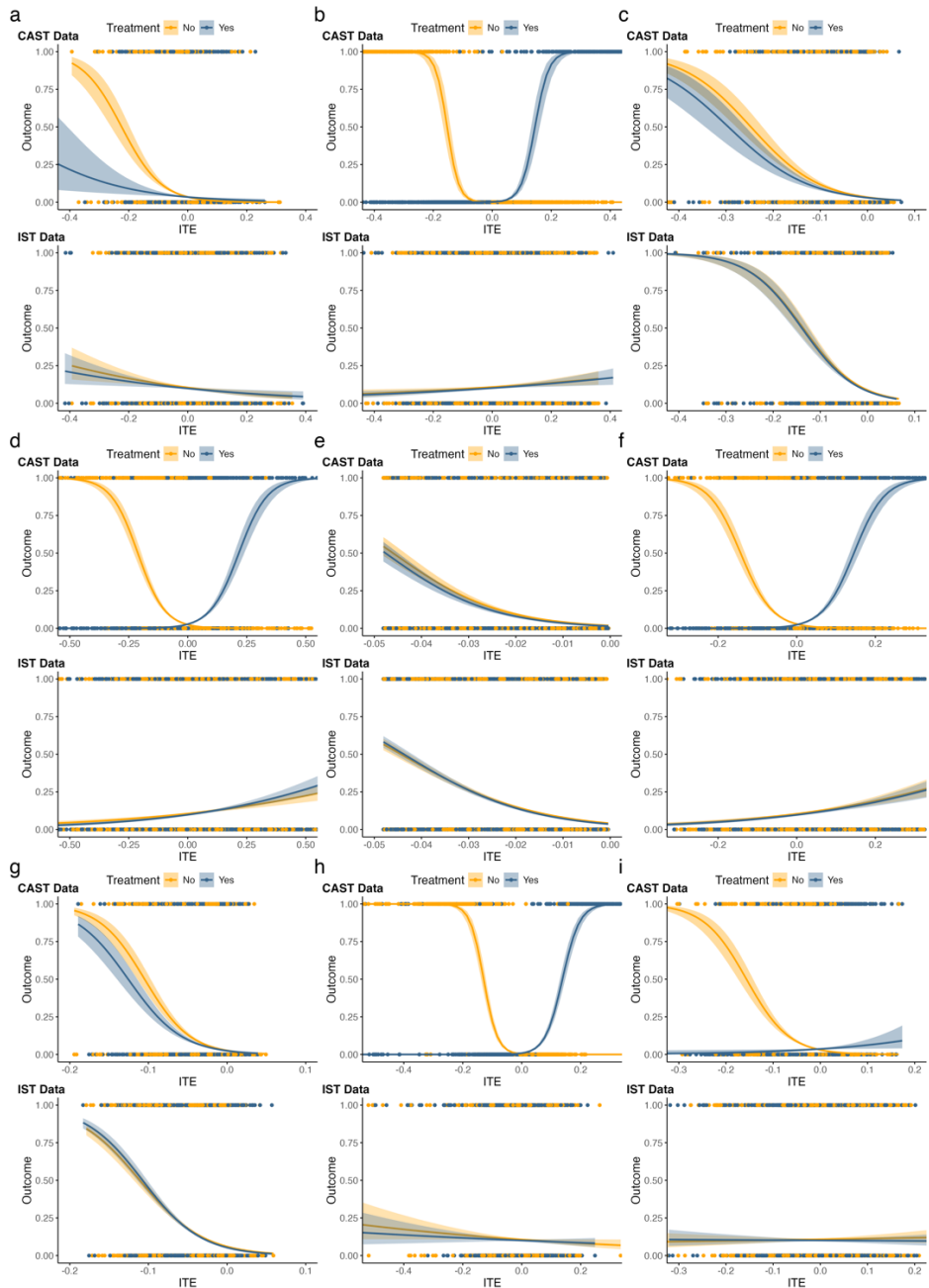

Figure S6.4.1. External validation with CAST as training data and IST as test data: outcome-ITE comparative analysis. Dot plots and line plots depict the true and fitted patient outcomes against estimated ITE values between training data and test data. Orange represents the control group, and blue indicates the treatment group. The outcome variable is death at 14 days for IST and death at 4 weeks for CAST. **a**, T-learner Logistic Regression. **b**, T-learner Random Forest. **c**, T-learner Support Vector Machine. **d**, T-learner XGBoost. **e**, S-learner Logistic Regression. **f**, S-learner XGBoost. **g**, S-learner BART. **h**, X-learner Random Forest. **i**, X-learner BART. Abbreviations: IST: the International Stroke Trial, CAST: the Chinese Acute Stroke Trial, ITE: individualized treatment effect.

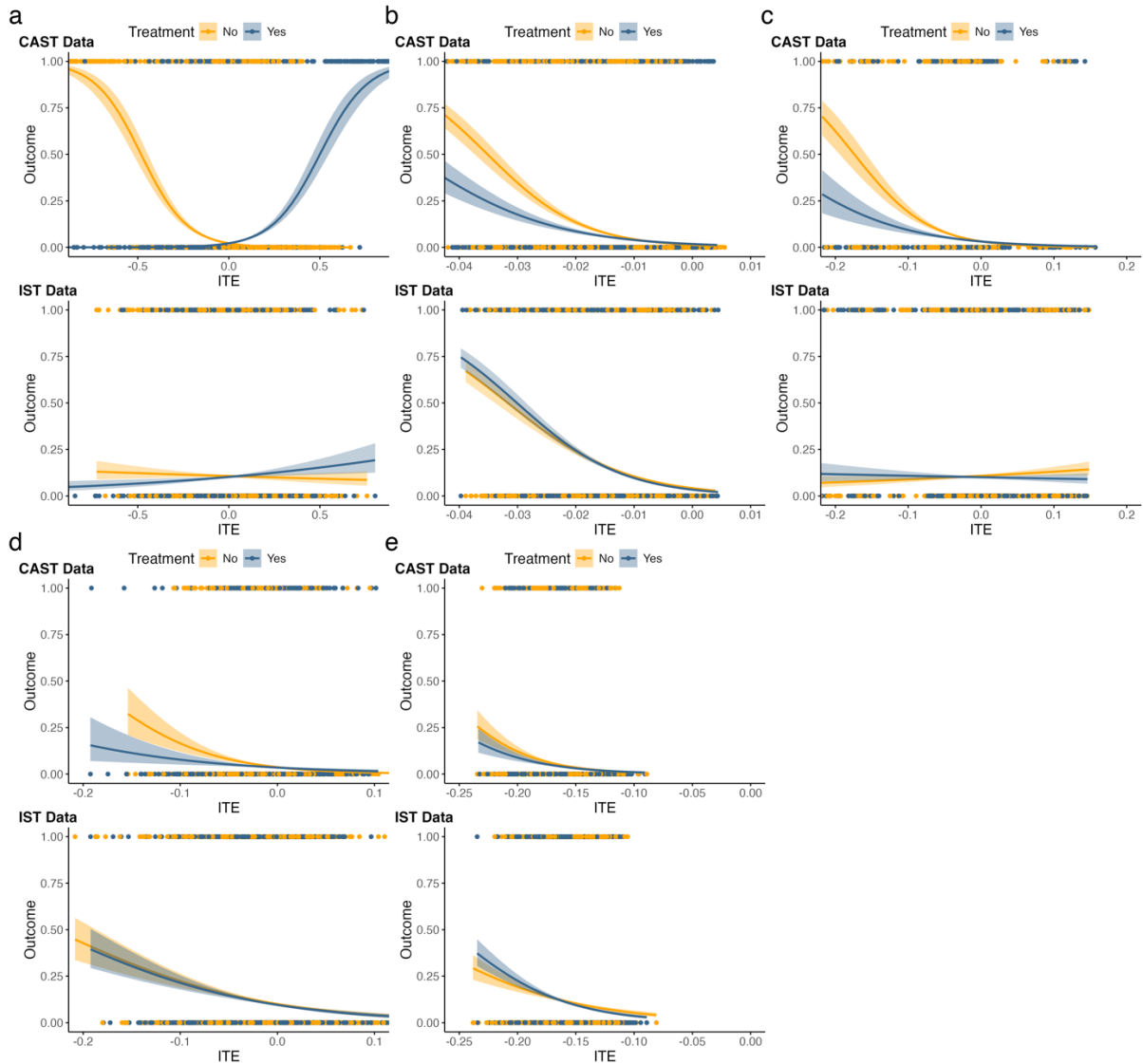

Figure S6.4.2. External validation with CAST as training data and IST as test data: outcome-ITE comparative analysis. Dot plots and line plots depict the true and fitted patient outcomes against estimated ITE values between training data and test data. Orange represents the control group, and blue indicates the treatment group. The outcome variable is death at 14 days for IST and death at 4 weeks for CAST. **a**, DR-learner Random Forest. **b**, Causal Forest. **c**, Bayesian Causal Forest. **d**, CVAE. **e**, GANITE. Abbreviations: IST: the International Stroke Trial, CAST: the Chinese Acute Stroke Trial, ITE: individualized treatment effect.

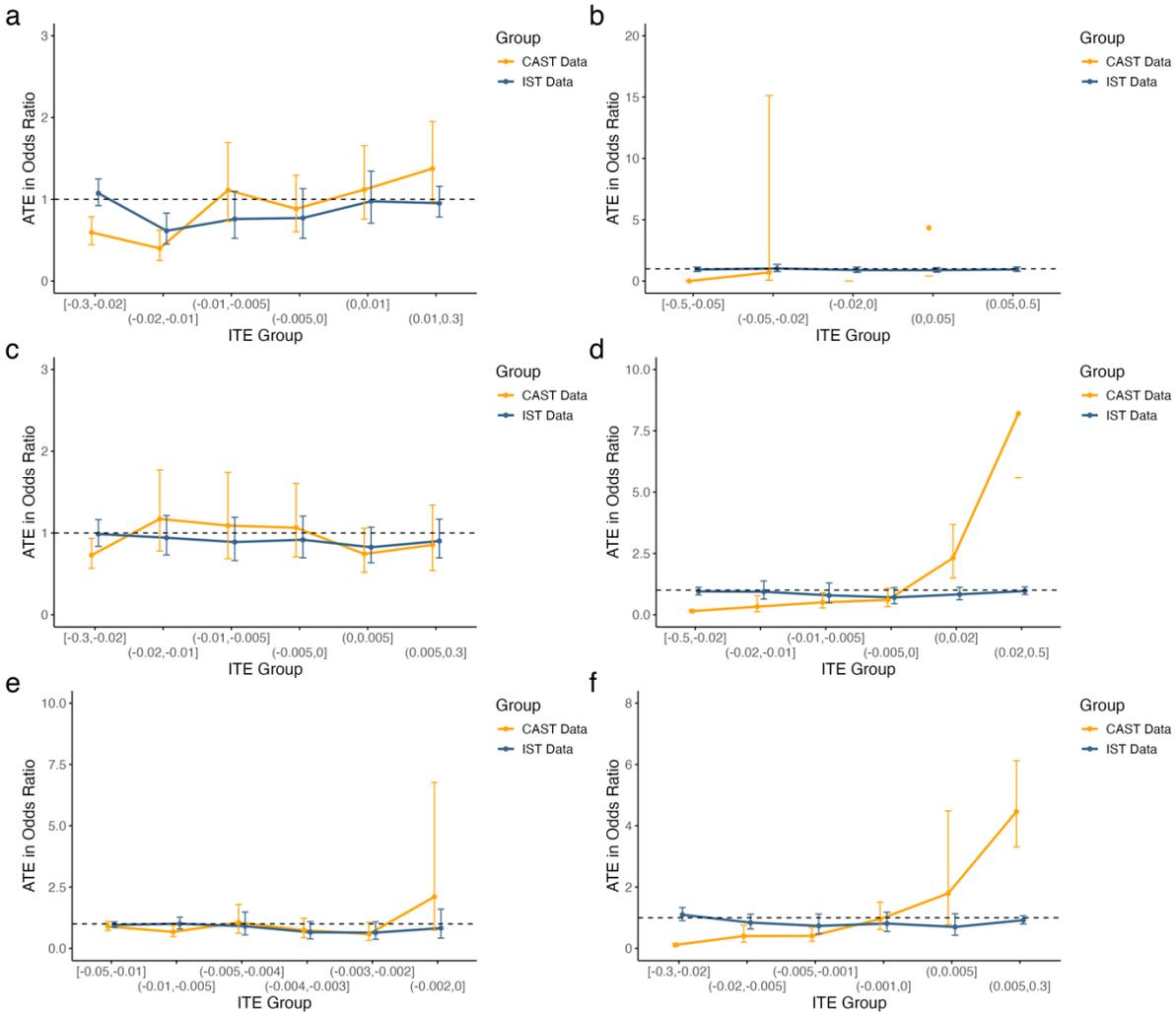

Figure S6.5.1. External validation with CAST as training data and IST as test data: ATE-ITE comparative analysis. Line plots depict ATE in risk ratio within different ITE subgroups and provide the confidence intervals at 95% level. Orange represents the training data, and blue indicates the test data. The horizontal dashed line at 1.0 means no treatment effects. The outcome variable is death at 14 days for IST and death at 4 weeks for CAST. **a**, T-learner Logistic Regression. **b**, T-learner Random Forest. **c**, T-learner Support Vector Machine. **d**, T-learner XGBoost. **e**, S-learner Logistic Regression. **f**, S-learner XGBoost. Abbreviations: IST: the International Stroke Trial, CAST: the Chinese Acute Stroke Trial, ITE: individualized treatment effect, ATE: average treatment effect.

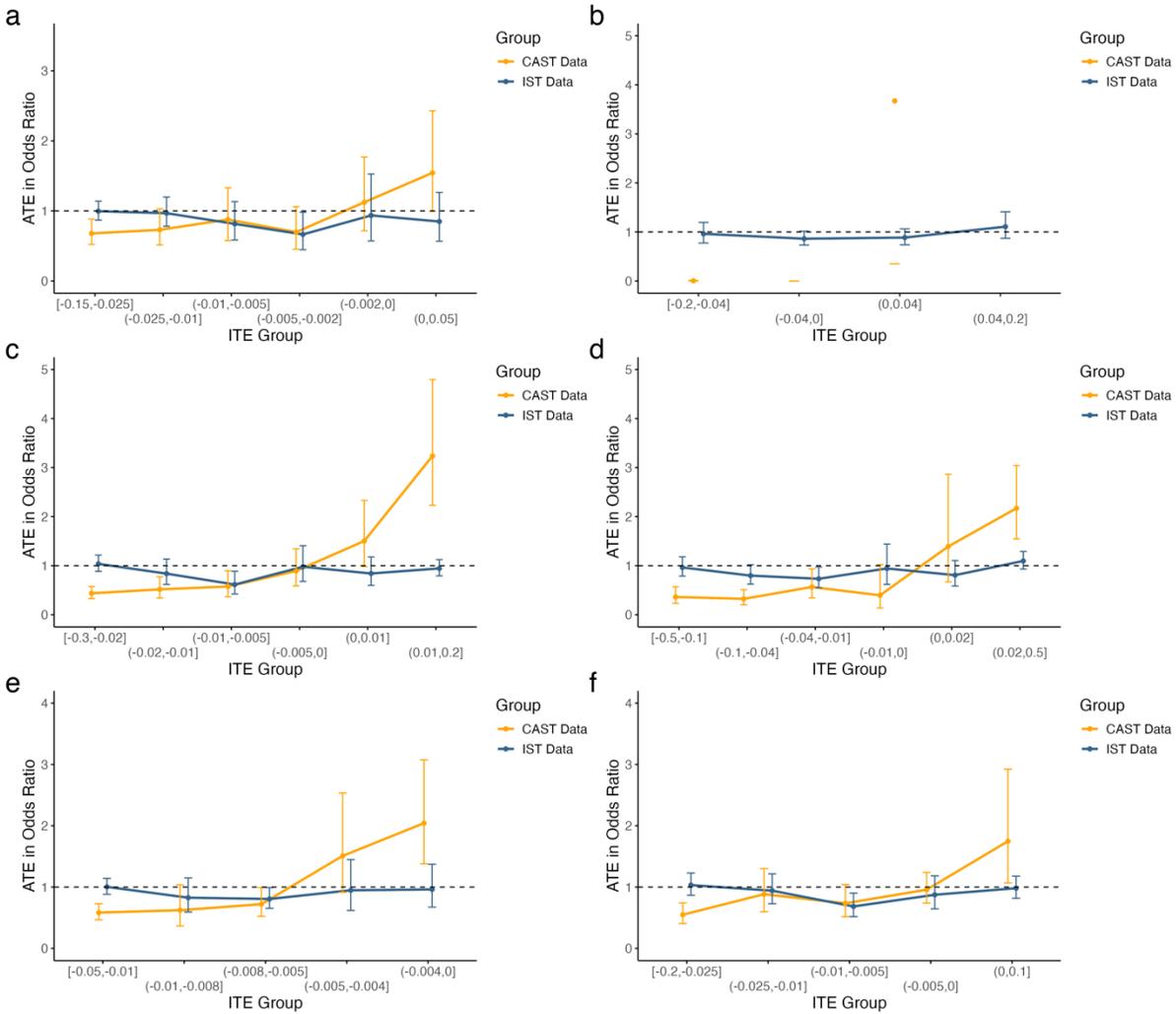

Figure S6.5.2. External validation with CAST as training data and IST as test data: ATE-ITE comparative analysis. Line plots depict ATE in risk ratio within different ITE subgroups and provide the confidence intervals at 95% level. Orange represents the training data, and blue indicates the test data. The horizontal dashed line at 1.0 means no treatment effects. The outcome variable is death at 14 days for IST and death at 4 weeks for CAST. **a**, S-learner BART. **b**, X-learner Random Forest. **c**, X-learner BART. **d**, DR-learner Random Forest. **e**, Causal Forest. **f**, Bayesian Causal Forest. Abbreviations: IST: the International Stroke Trial, CAST: the Chinese Acute Stroke Trial, ITE: individualized treatment effect, ATE: average treatment effect.

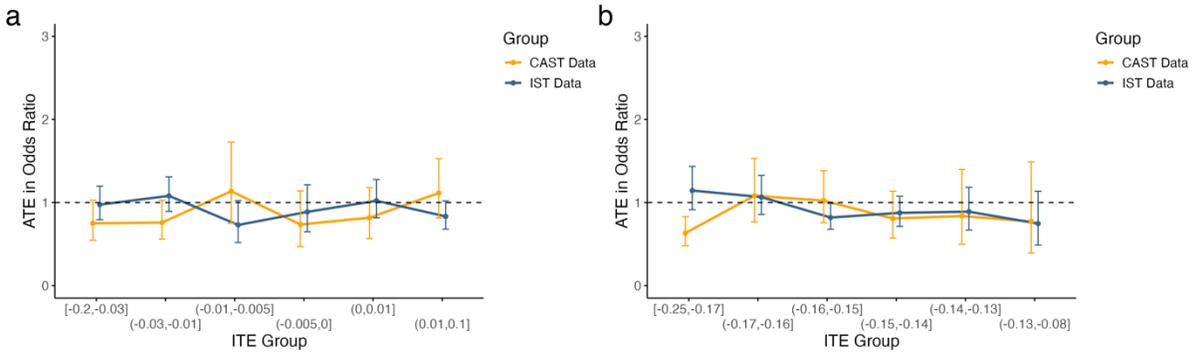

Figure S6.5.2. External validation with CAST as training data and IST as test data: ATE-ITE comparative analysis. Line plots depict ATE in risk ratio within different ITE subgroups and provide the confidence intervals at 95% level. Orange represents the training data, and blue indicates the test data. The horizontal dashed line at 1.0 means no treatment effects. The outcome variable is death at 14 days for IST and death at 4 weeks for CAST. **a**. CVAE. **b**, GANITE. Abbreviations: IST: the International Stroke Trial, CAST: the Chinese Acute Stroke Trial, ITE: individualized treatment effect, ATE: average treatment effect.

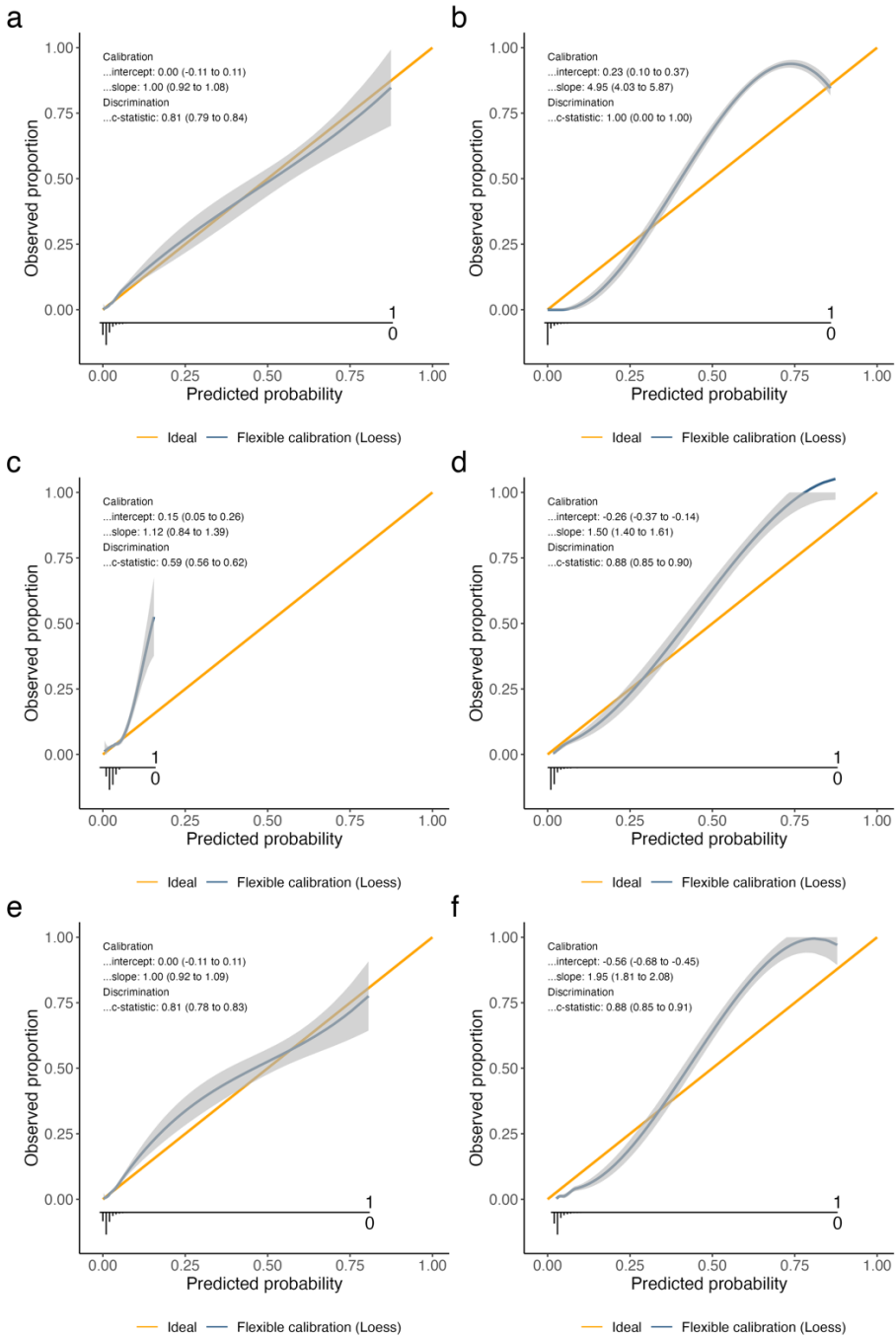

Figure S6.6.1. CAST as training dataset: calibration plot of predicted outcome in the treatment group. The orange line indicates ideal calibration. The outcome variable is death at 14 days for IST and death at 4 weeks for CAST. **a**, T-learner Logistic Regression. **b**, T-learner Random Forest. **c**, T-learner Support Vector Machine. **d**, T-learner XGBoost. **e**, S-learner Logistic Regression. **f**, S-learner XGBoost. Abbreviations: CAST: the Chinese Acute Stroke Trials.

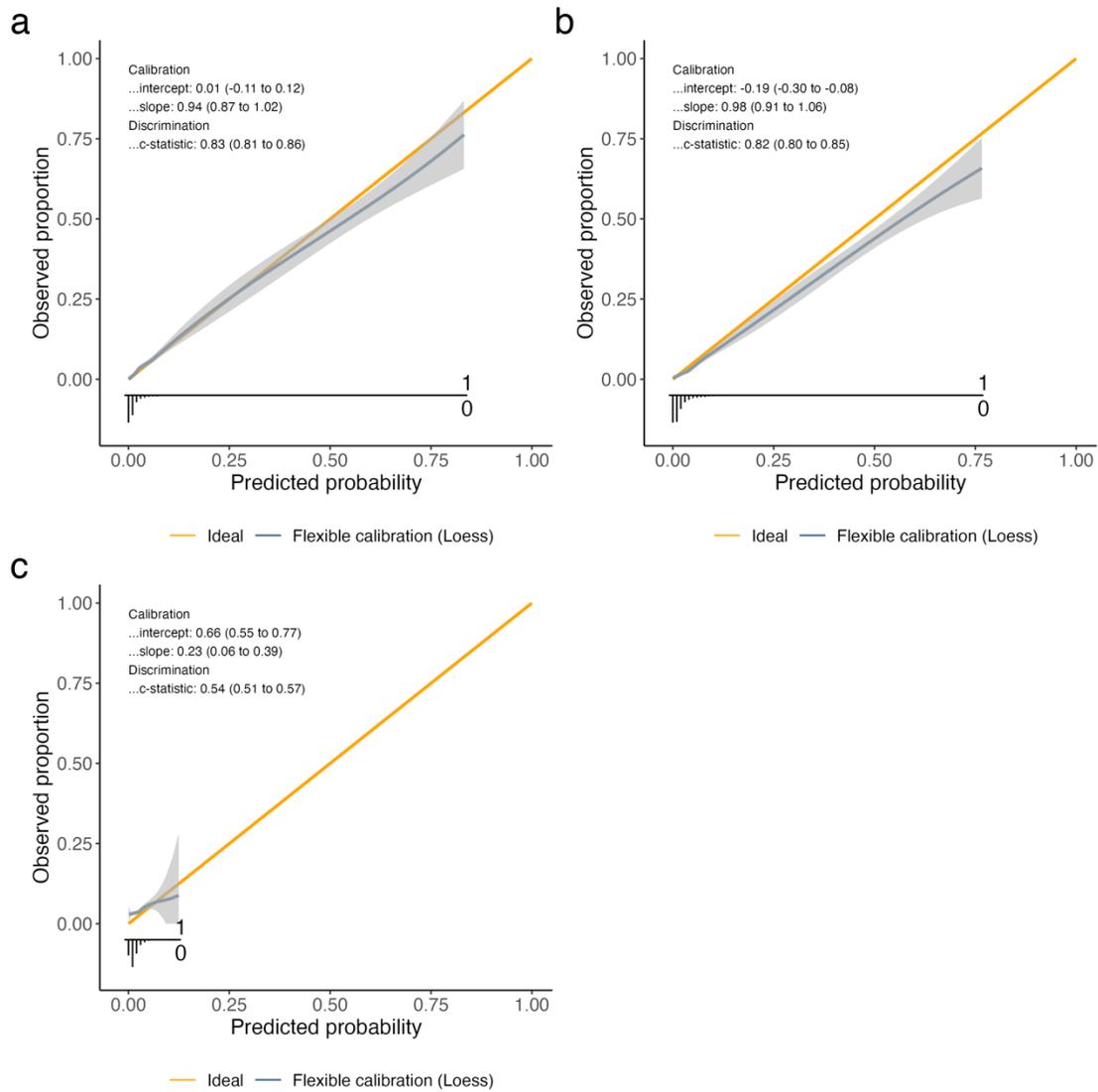

Figure S6.6.2. CAST as training dataset: calibration plot of predicted outcome in the treatment group. The orange line indicates ideal calibration. The outcome variable is death at 14 days for IST and death at 4 weeks for CAST. **a**, S-learner BART. **b**, CVAE. **c**, GANITE. Abbreviations: CAST: the Chinese Acute Stroke Trial.

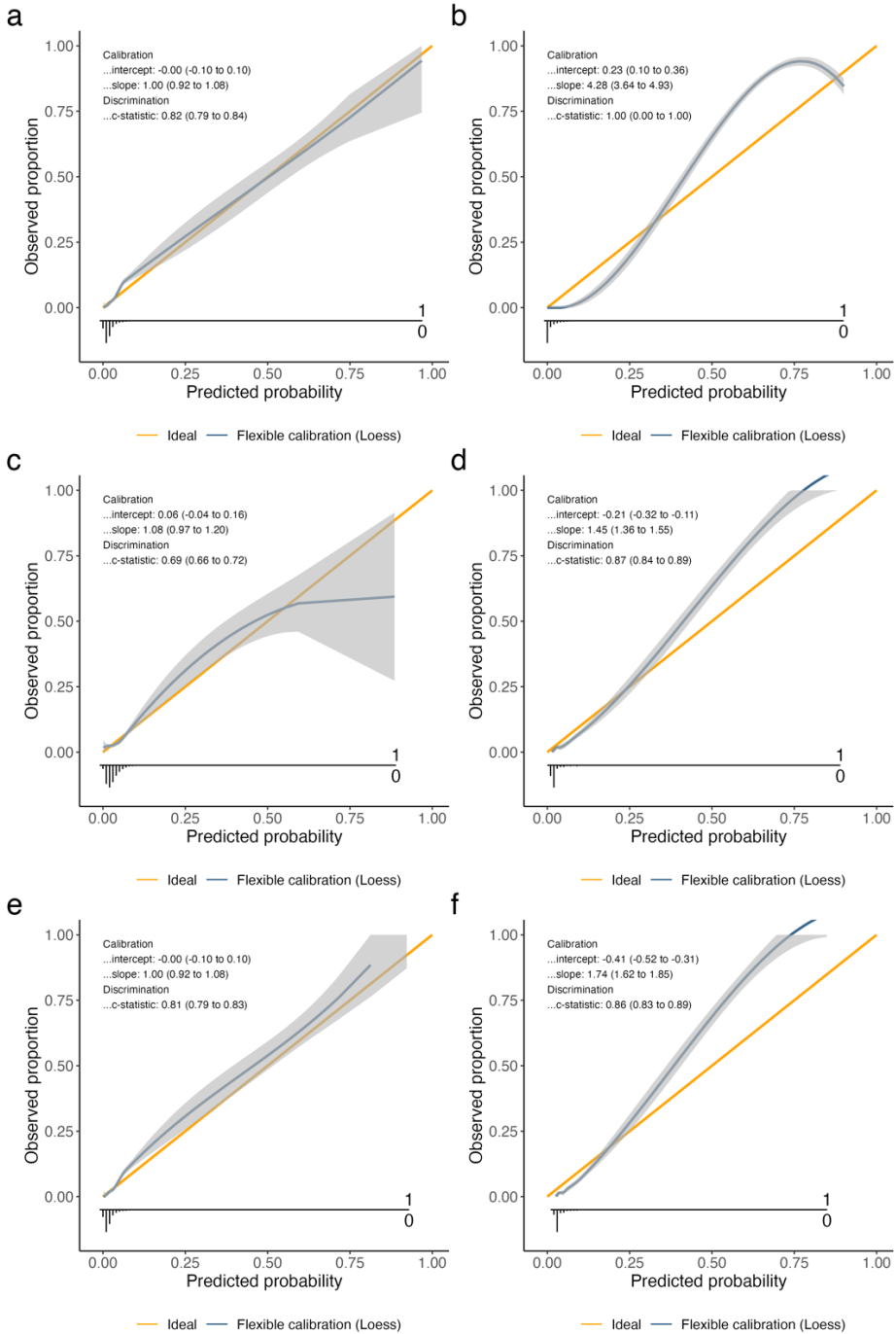

Figure S6.7.1. CAST as training dataset: calibration plot of predicted outcome in the control group. The orange line indicates ideal calibration. The outcome variable is death at 14 days for IST and death at 4 weeks for CAST. **a**, T-learner Logistic Regression. **b**, T-learner Random Forest. **c**, T-learner Support Vector Machine. **d**, T-learner XGBoost. **e**, S-learner Logistic Regression. **f**, S-learner XGBoost. Abbreviations: CAST: the Chinese Acute Stroke Trial.

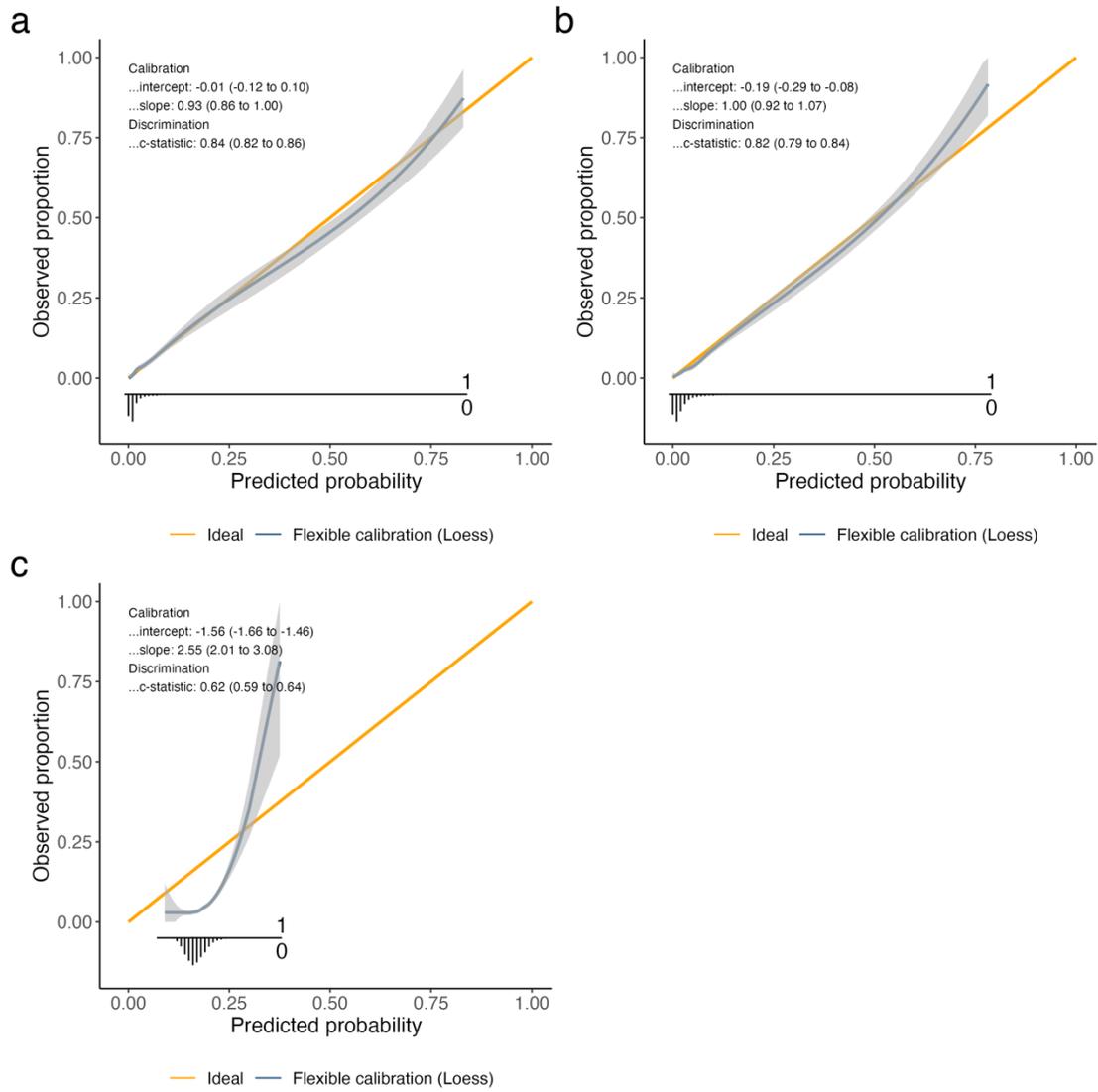

Figure S6.7.2. CAST as training dataset: calibration plot of predicted outcome in the control group. The orange line indicates ideal calibration. The outcome variable is death at 14 days for IST and death at 4 weeks for CAST. **a**, S-learner BART. **b**, CVAE. **c**, GANITE. Abbreviations: CAST: the Chinese Acute Stroke Trial.

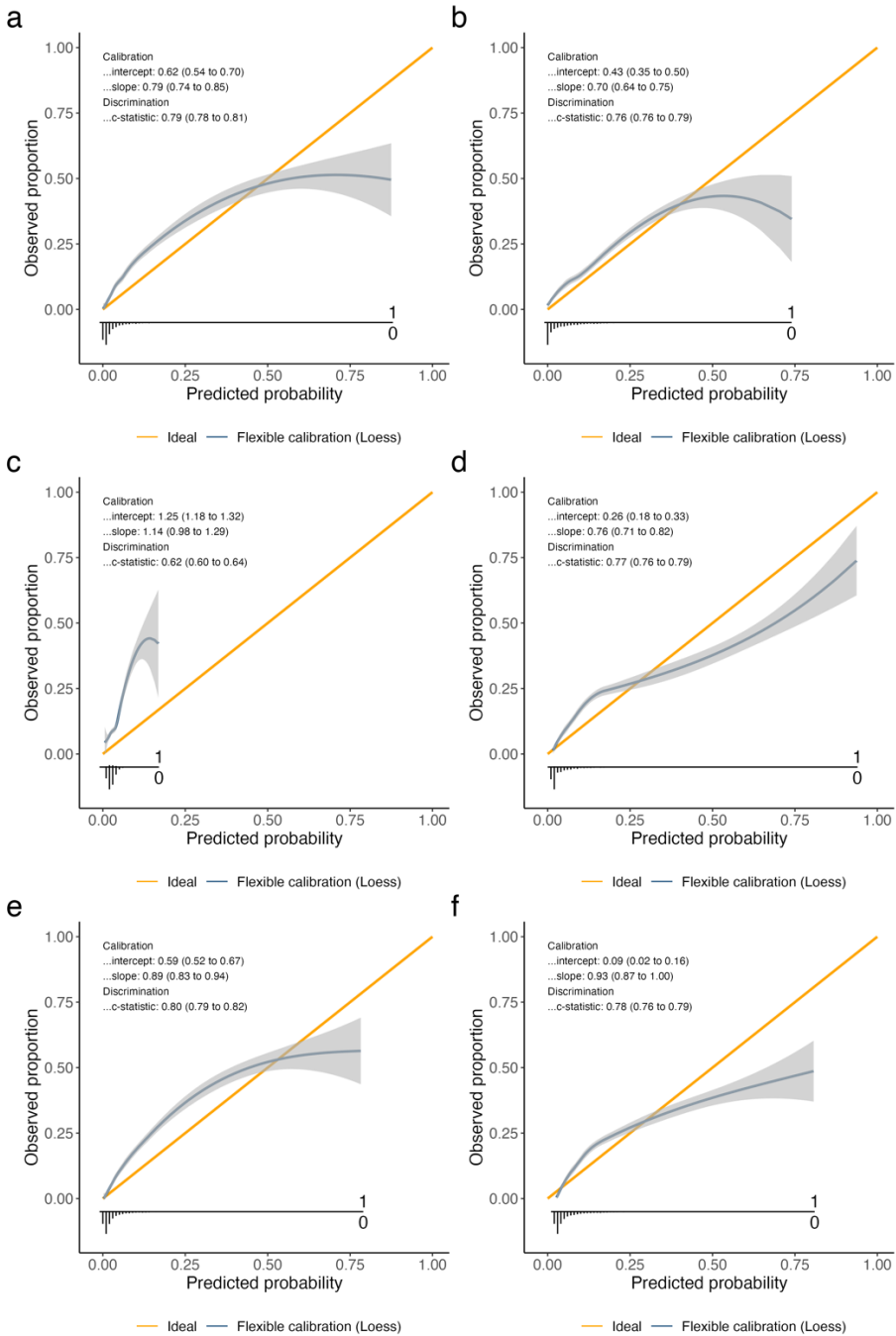

Figure S6.8.1. IST as test dataset: calibration plot of predicted outcome in the treatment group. The orange line indicates ideal calibration. The outcome variable is death at 14 days for IST and death at 4 weeks for CAST. **a**, T-learner Logistic Regression. **b**, T-learner Random Forest. **c**, T-learner Support Vector Machine. **d**, T-learner XGBoost. **e**, S-learner Logistic Regression. **f**, S-learner XGBoost. Abbreviations: IST: the International Stroke Trial.

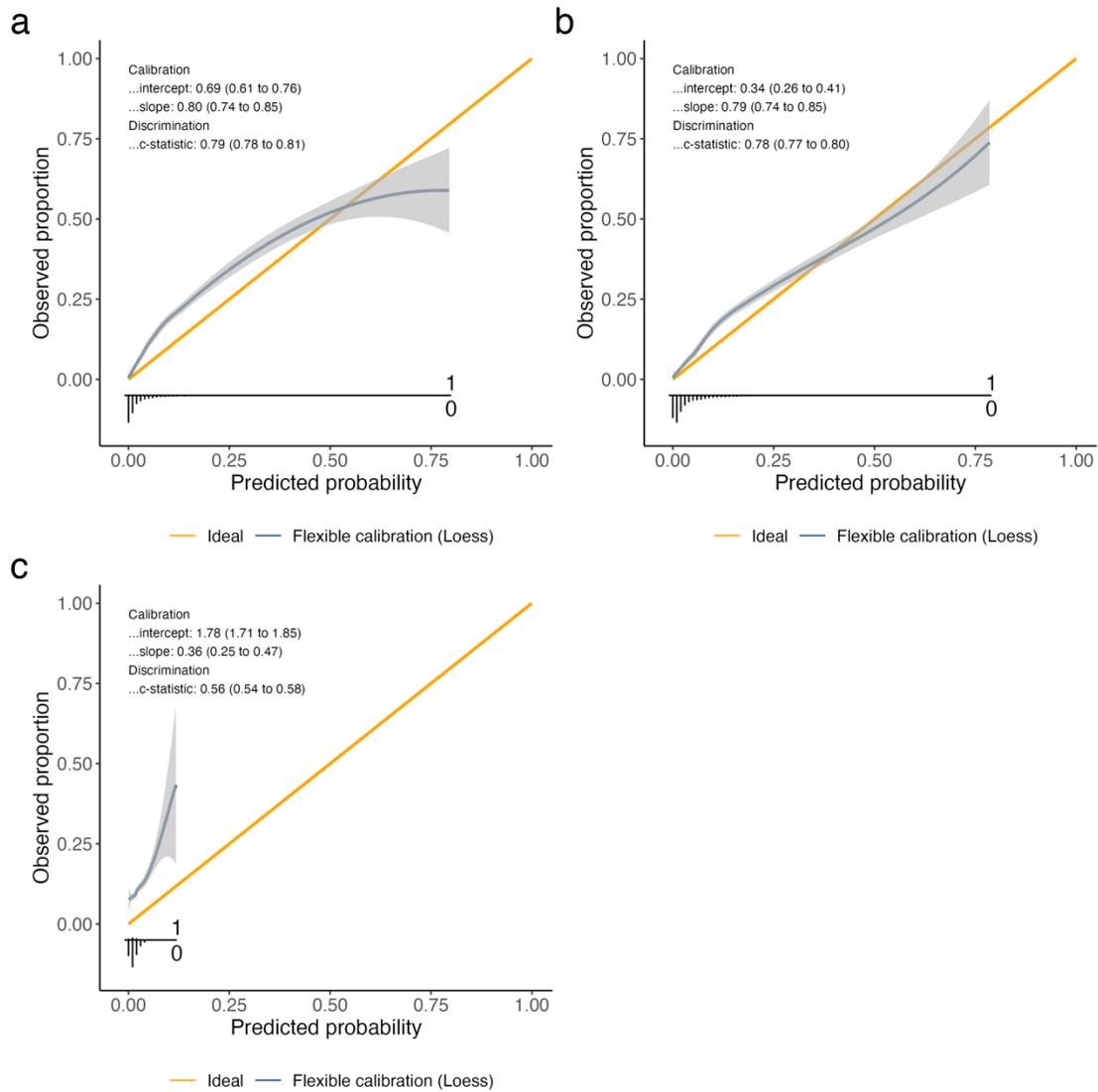

Figure S6.8.2. IST as test dataset: calibration plot of predicted outcome in the treatment group. The orange line indicates ideal calibration. The outcome variable is death at 14 days for IST and death at 4 weeks for CAST. **a**, S-learner BART. **b**, CVAE. **c**, GANITE. Abbreviations: IST: the International Stroke Trial.

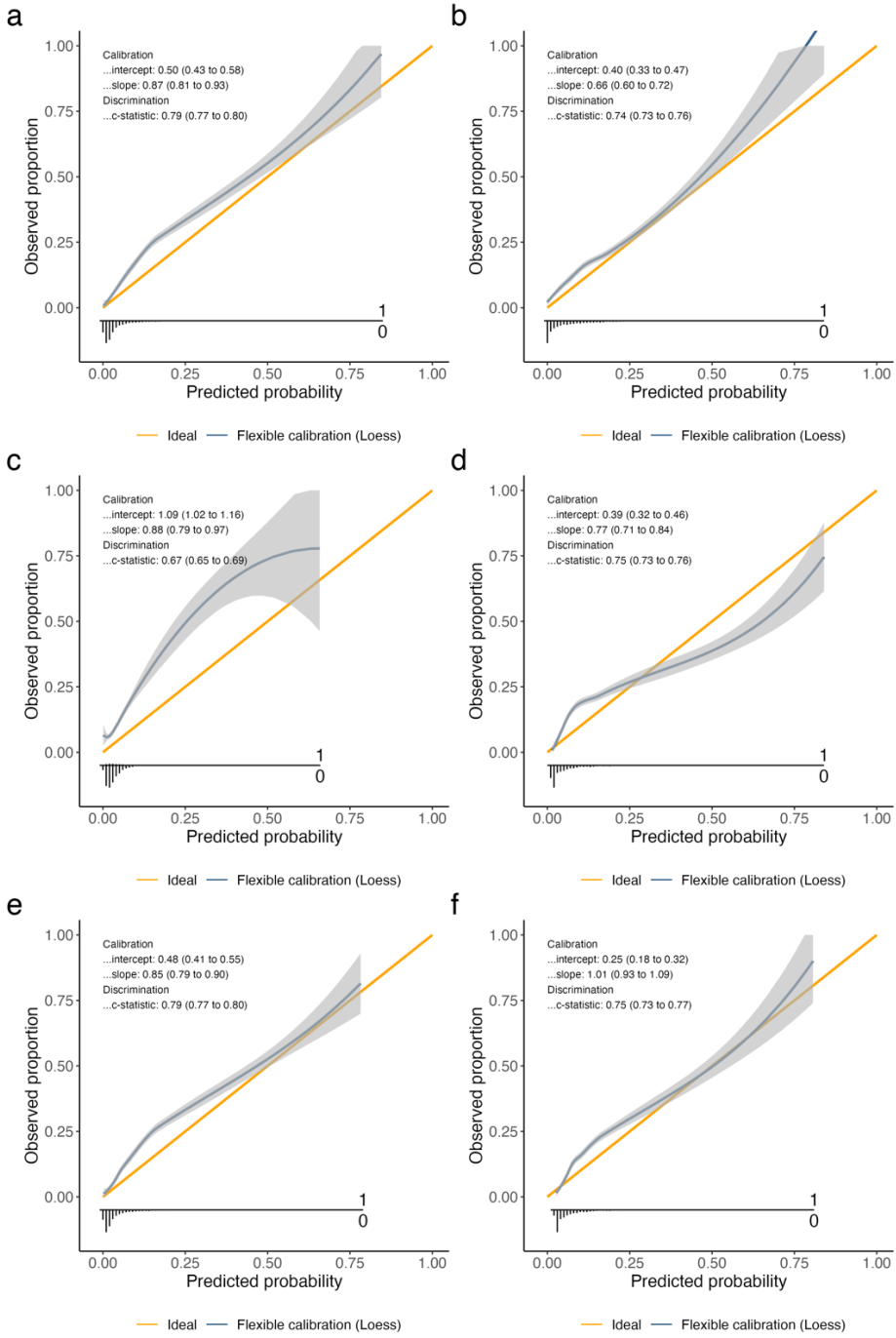

Figure S6.9.1. IST as test dataset: calibration plot of predicted outcome in the control group. The orange line indicates ideal calibration. The outcome variable is death at 14 days for IST and death at 4 weeks for CAST. **a**, T-learner Logistic Regression. **b**, T-learner Random Forest. **c**, T-learner Support Vector Machine. **d**, T-learner XGBoost. **e**, S-learner Logistic Regression. **f**, S-learner XGBoost. Abbreviations: IST: the International Stroke Trial.

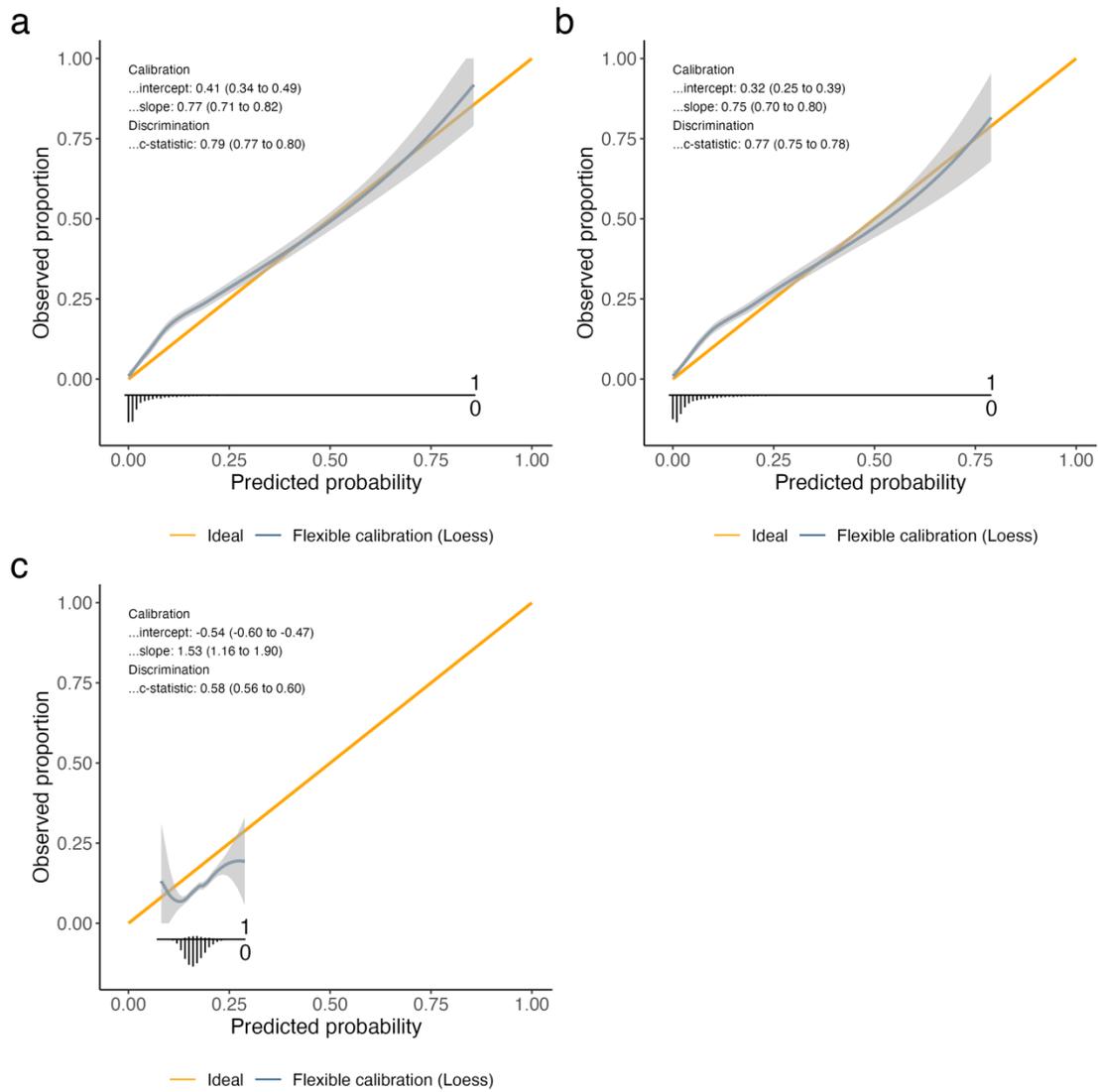

Figure S6.9.2. IST as test dataset: calibration plot of predicted outcome in the control group. The orange line indicates ideal calibration. The outcome variable is death at 14 days for IST and death at 4 weeks for CAST. **a**, S-learner BART. **b**, CVAE. **c**, GANITE. Abbreviations: IST: the International Stroke Trial.

## Simulation study

**Settings I**: 20 continuous covariates, binary treatment, binary outcome, utilizing all the covariates to model ITEs, sample size as 20000. The true ITEs are modelled by logistic function with a global treatment effect and covariates-interactive treatment effects.

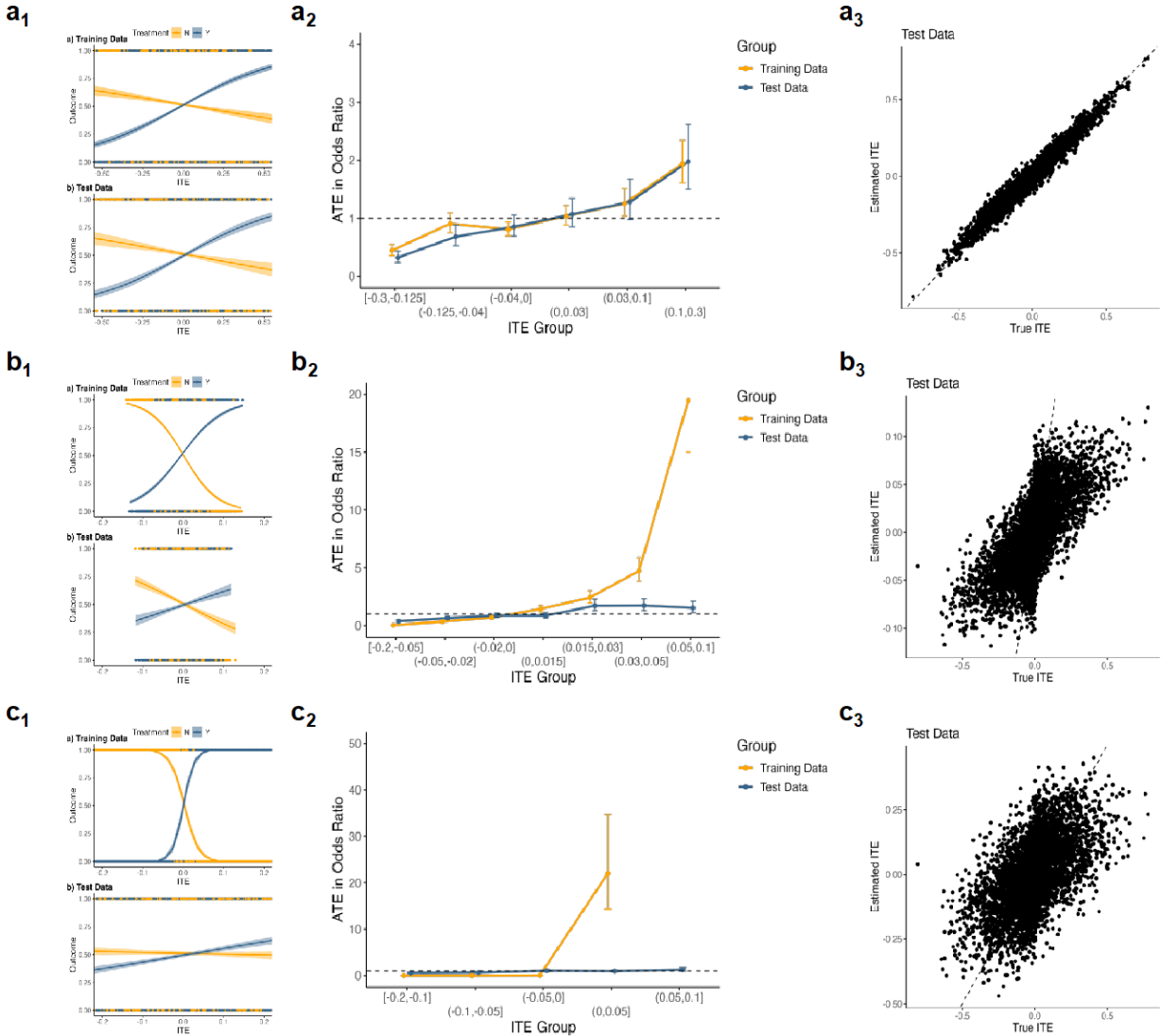

Figure S7.1.1. Simulation study: comparative analysis of causal machine learning-based individualized treatment effects. $a_1$-$a_3$, T learner Logistic Regression. $b_1$-$b_3$, Causal Forest. $c_1$-$c_3$, T learner Random Forest. $a_1,b_1,c_1$, Comparing patient outcomes against estimated ITE values between training data and test data. $a_2,b_2,c_2$, Line plots depict ATE in odds ratio within different ITE subgroups and provide the confidence intervals at 95% level. The horizontal dashed line at 1.0 means no treatment effects. $a_3,b_3,c_3$, Dot plots of estimated ITEs against the true ITEs in the test data. The black dashed line indicates ideal prediction when estimated ITEs equal to true ITEs. Abbreviations: ITE: individualized treatment effect, ATE: average treatment effect.

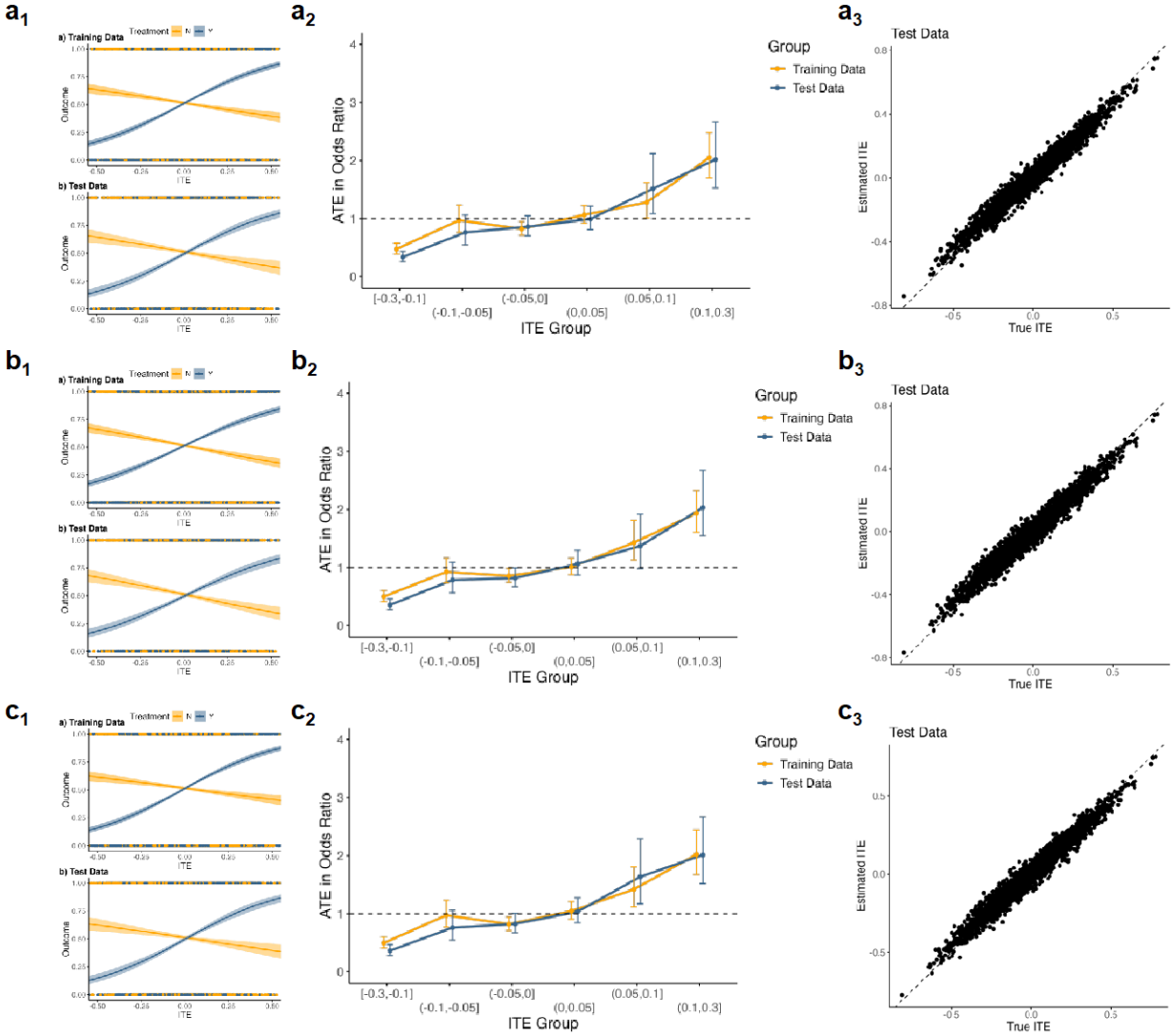

Figure S7.1.2. Simulation study: comparative analysis of causal machine learning-based individualized treatment effects. **$a_1$-$a_3$**, T learner Support Vector Machine. **$b_1$-$b_3$**, T learner Penalized Logistic Regression. **$c_1$-$c_3$**, S learner Penalized Logistic Regression. **$a_1$,$b_1$,$c_1$**, Comparing patient outcomes against estimated ITE values between training data and test data. **$a_2$,$b_2$,$c_2$**, Line plots depict ATE in odds ratio within different ITE subgroups and provide the confidence intervals at 95% level. A horizontal dashed line at 1.0 means null treatment effects. **$a_3$,$b_3$,$c_3$**, Dot plots of estimated ITEs against the true ITEs in the test data. The black dashed line indicates ideal prediction when estimated ITEs equal to true ITEs. Abbreviations: ITE: individualized treatment effect, ATE: average treatment effect.

**Settings II**: 20 continuous covariates, binary treatment, binary outcome, utilizing only 6 less significant covariates with minimal influence on ITEs to train the causal machine learning models, sample size as 20000. The true ITEs are modelled by logistic function with a global treatment effect and covariates-interactive treatment effects.

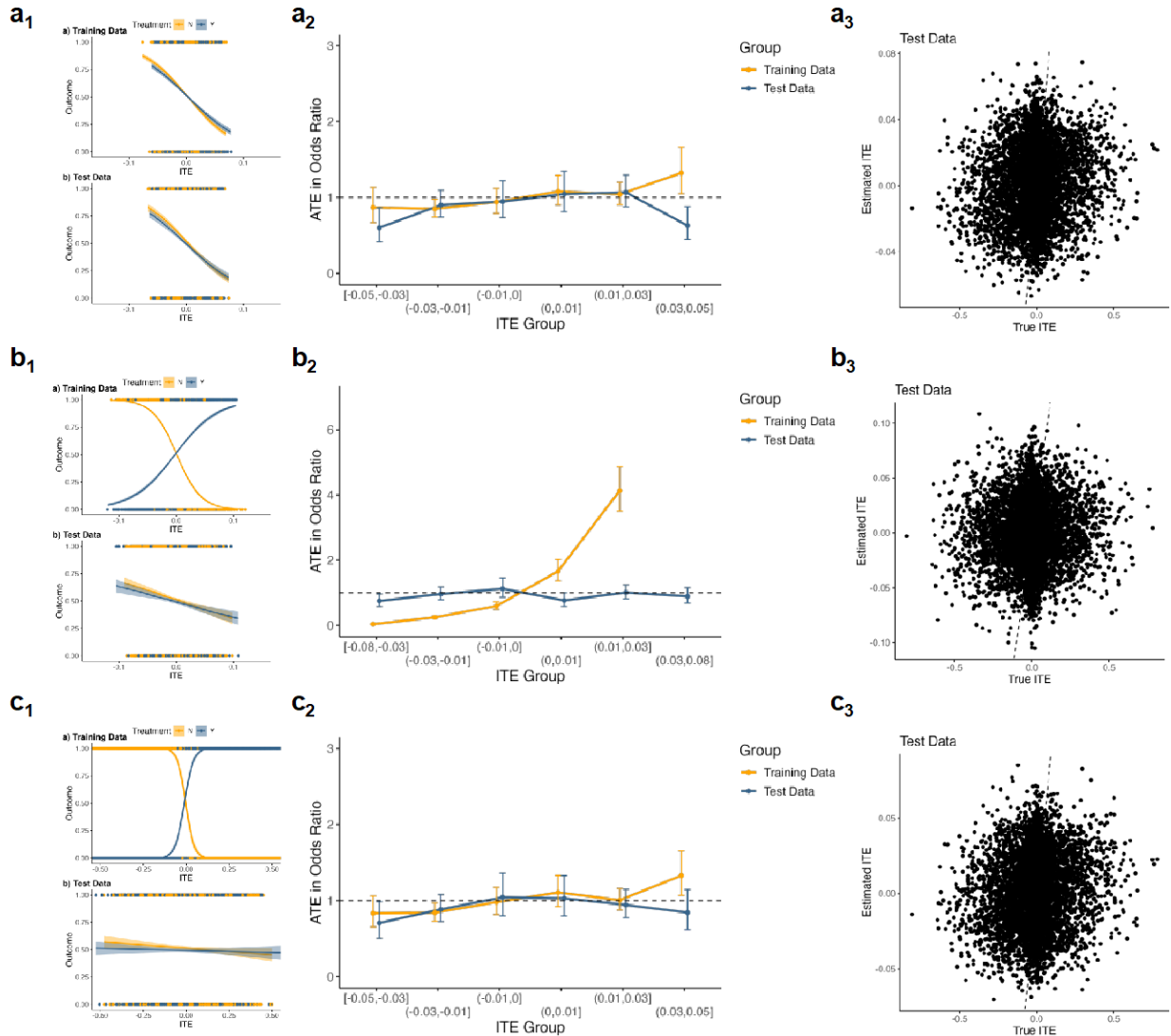

Figure S7.2.1. Simulation study: comparative analysis of causal machine learning-based individualized treatment effects. $a_1$-$a_3$, T learner Logistic Regression. $b_1$-$b_3$, Causal Forest. $c_1$-$c_3$, T learner Random Forest. $a_1,b_1,c_1$, Comparing patient outcomes against estimated ITE values between training data and test data. $a_2,b_2,c_2$, Line plots depict ATE in odds ratio within different ITE subgroups and provide the confidence intervals at 95% level. A horizontal dashed line at 1.0 means null treatment effects. $a_3,b_3,c_3$, Dot plots of estimated ITEs against the true ITEs in the test data. The black dashed line indicates ideal prediction when estimated ITEs equal to true ITEs. Abbreviations: ITE: individualized treatment effect, ATE: average treatment effect.

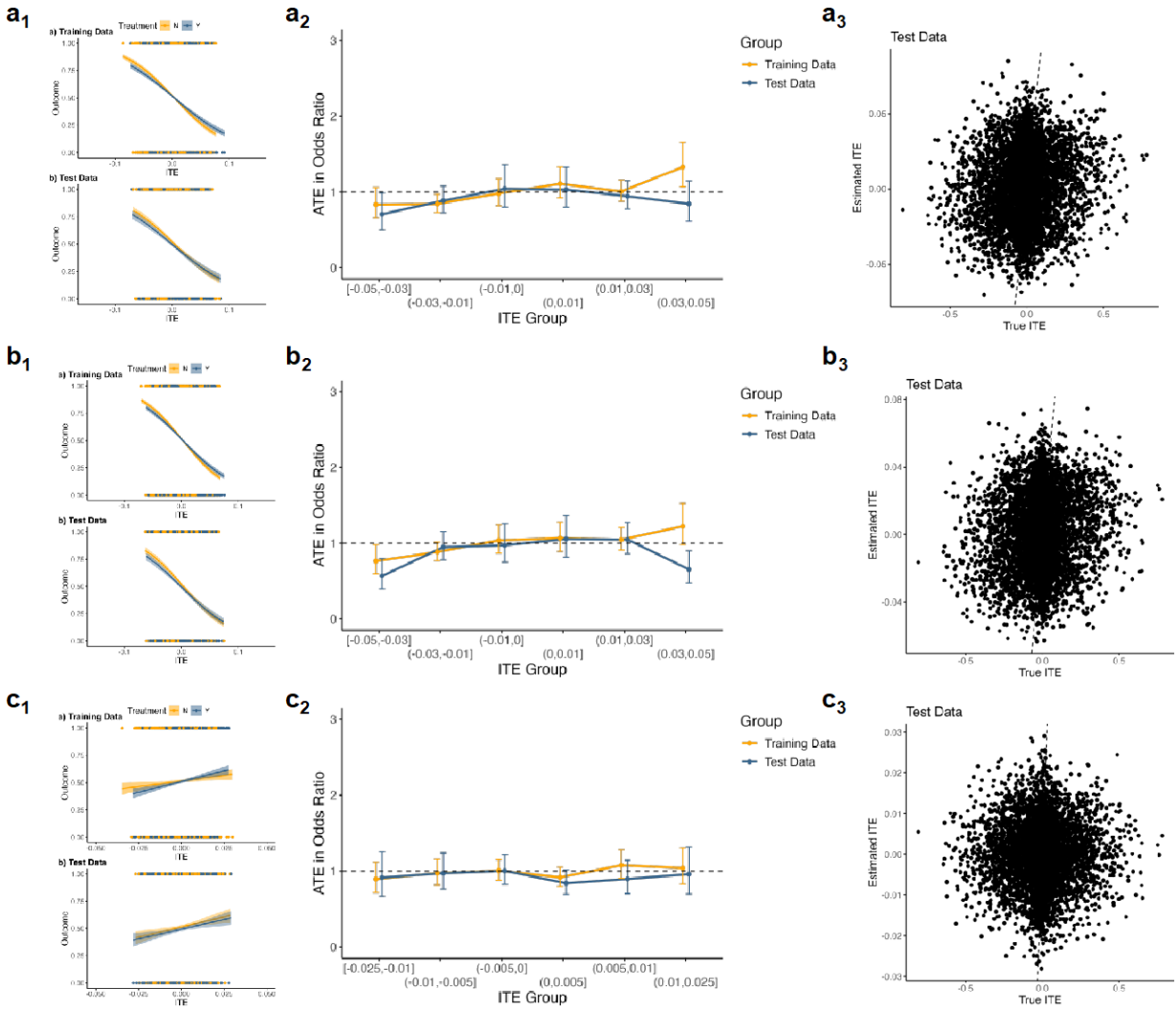

Figure S7.2.2. Simulation study: comparative analysis of causal machine learning-based individualized treatment effects. $a_1$-$a_3$, T learner Support Vector Machine. $b_1$-$b_3$, T learner Penalized Logistic Regression. $c_1$-$c_3$, S learner Penalized Logistic Regression. $a_1,b_1,c_1$, Comparing patient outcomes against estimated ITE values between training data and test data. $a_2,b_2,c_2$, Line plots depict ATE in odds ratio within different ITE subgroups and provide the confidence intervals at 95% level. A horizontal dashed line at 1.0 means null treatment effects. $a_3,b_3,c_3$, Dot plots of estimated ITEs against the true ITEs in the test data. The black dashed line indicates ideal prediction when estimated ITEs equal to true ITEs. Abbreviations: ITE: individualized treatment effect, ATE: average treatment effect.

**Settings III**: 10 continuous covariates and 10 discrete covariates, binary treatment, binary outcome, utilizing all the covariates to model ITEs, sample size as 20000. The true ITEs are modelled by logistic function with a global treatment effect and covariates-interactive treatment effects.

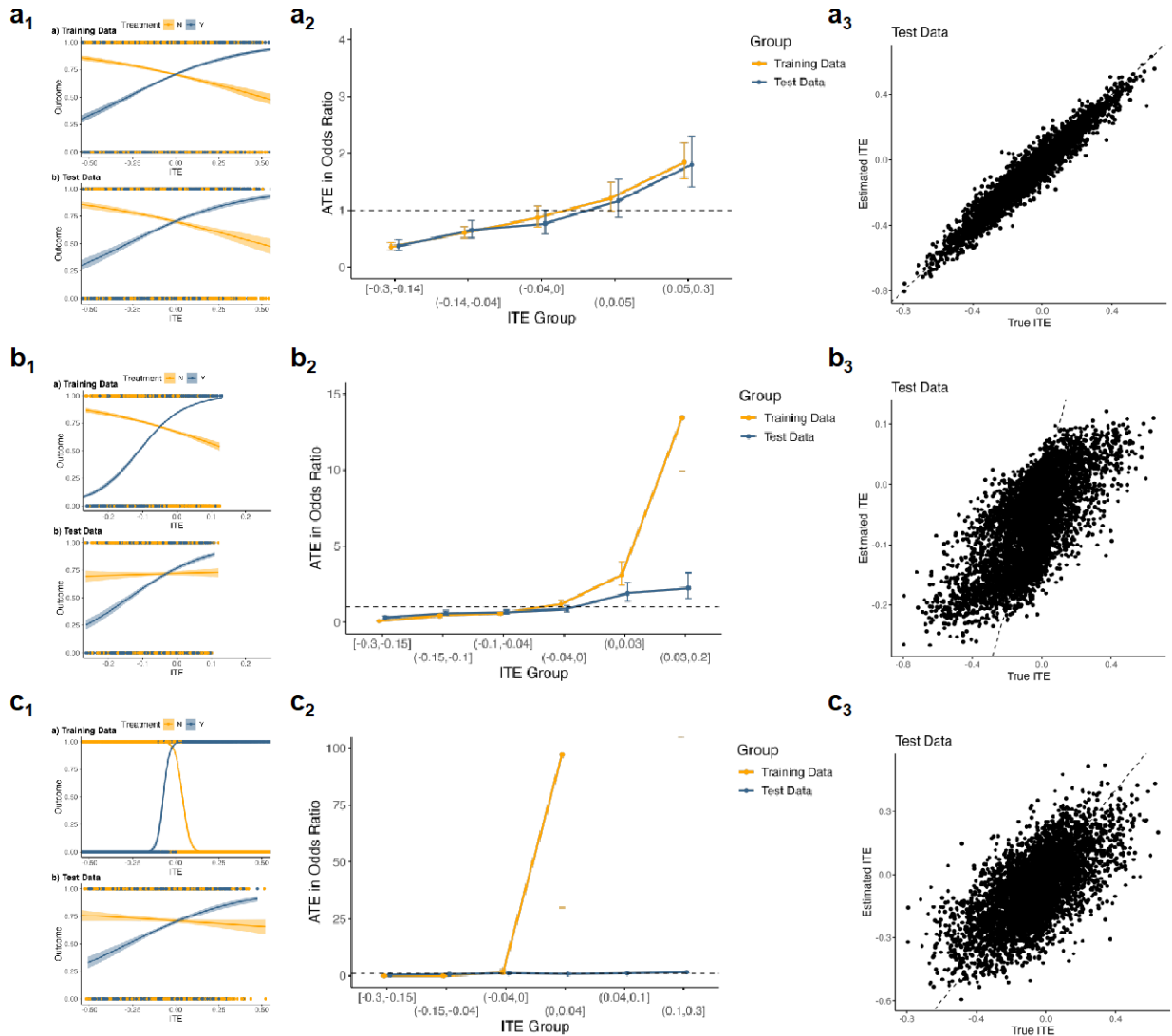

Figure S7.3.1. Simulation study: comparative analysis of causal machine learning-based individualized treatment effects. $a_1$-$a_3$, T learner Logistic Regression. $b_1$-$b_3$, Causal Forest. $c_1$-$c_3$, T learner Random Forest. $a_1,b_1,c_1$, Comparing patient outcomes against estimated ITE values between training data and test data. $a_2,b_2,c_2$, Line plots depict ATE in odds ratio within different ITE subgroups and provide the confidence intervals at 95% level. A horizontal dashed line at 1.0 means null treatment effects. $a_3,b_3,c_3$, Dot plots of estimated ITEs against the true ITEs in the test data. The black dashed line indicates ideal prediction when estimated ITEs equal to true ITEs. Abbreviations: ITE: individualized treatment effect, ATE: average treatment effect.

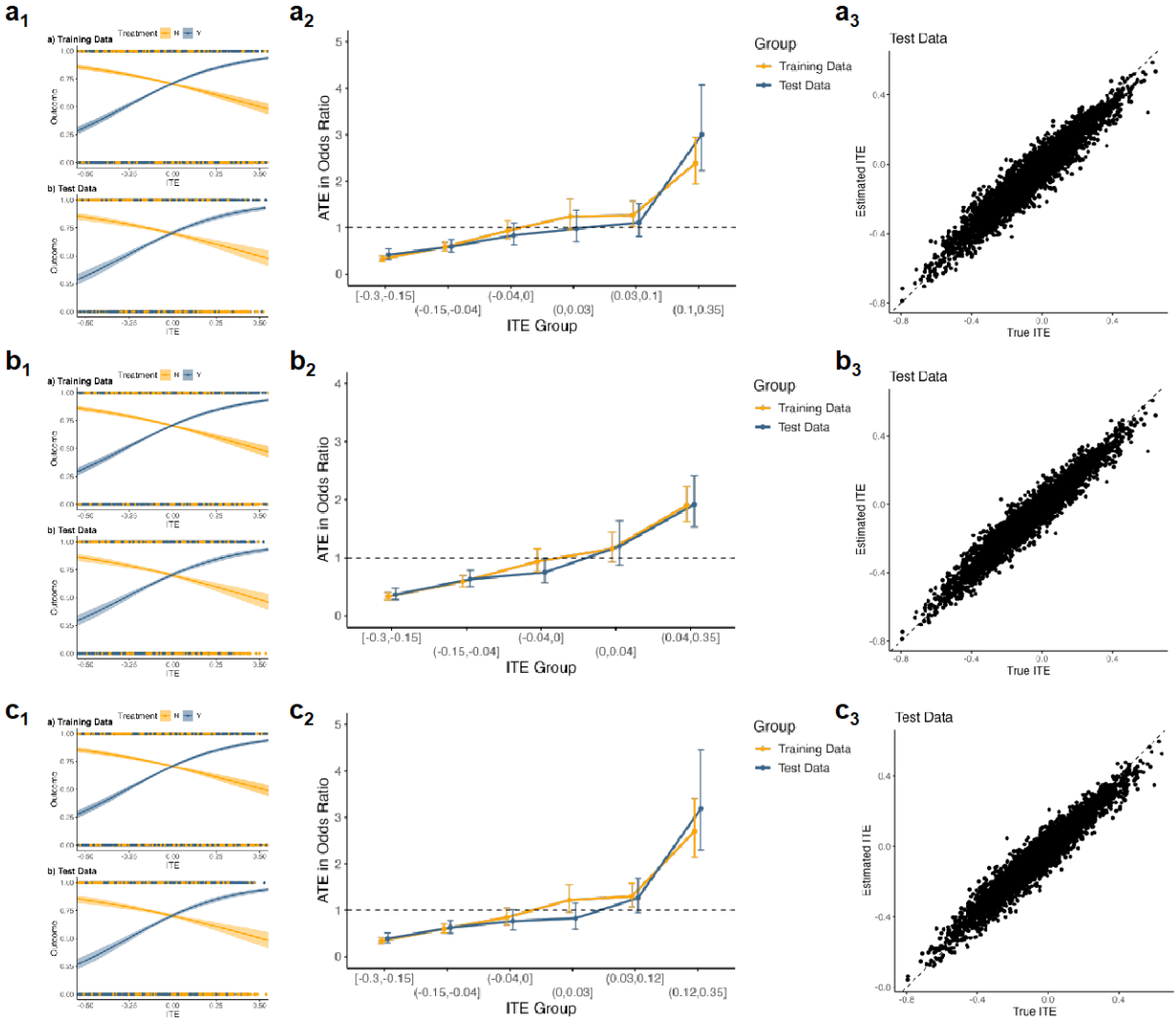

Figure S7.3.2. Simulation study: comparative analysis of causal machine learning-based individualized treatment effects. **a₁-a₃**, T learner Support Vector Machine. **b₁-b₃**, T learner Penalized Logistic Regression. **c₁-c₃**, S learner Penalized Logistic Regression. **a₁,b₁,c₁**, Comparing patient outcomes against estimated ITE values between training data and test data. **a₂,b₂,c₂**, Line plots depict ATE in odds ratio within different ITE subgroups and provide the confidence intervals at 95% level. A horizontal dashed line at 1.0 means null treatment effects. **a₃,b₃,c₃**, Dot plots of estimated ITEs against the true ITEs in the test data. The black dashed line indicates ideal prediction when estimated ITEs equal to true ITEs. Abbreviations: ITE: individualized treatment effect, ATE: average treatment effect.